\documentclass[journal]{IEEEtran}
\ifCLASSINFOpdf
\else
\fi
\usepackage{dsfont}
\usepackage{amsfonts}
\usepackage{amsmath}
\usepackage{amssymb}
\usepackage{amsthm}
\usepackage{mathrsfs}
\usepackage{graphicx}
\usepackage{subeqnarray}
\usepackage{cases}
\usepackage{graphicx,epstopdf}
\usepackage{booktabs}
\usepackage{tabularx}
\usepackage{array}
\usepackage{diagbox} 
\usepackage{multirow}
\usepackage{cite}
\usepackage{subfigure}
\usepackage{stfloats}

\newtheorem{proposition}{P{\scriptsize ROPOSITION}}[section]
\newtheorem{theorem}[proposition]{T{\scriptsize HEOREM}}
\newtheorem{remark}{R{\scriptsize  EMARK}}[section]

\newtheorem{algorithm}[proposition]{A{\scriptsize LGORITHM}}
\newcommand{\tabincell}[2]{\begin{tabular}{@{}#1@{}}#2\end{tabular}}
\newcounter{mytempeqncnt}

\begin{document}
%\title{ A Blind Color Image Watermarking Scheme Using Fast Structure-Preserving Quaternion SVD Algorithm}
\title{Efficient Robust Watermarking Based on Quaternion Singular Value Decomposition and Coefficient Pair Selection}


\author{Yong~Chen, Zhi-Gang~Jia,
        Ya-Xin~Peng, Yan~Peng% <-this % stops a space
\thanks{Y~Chen is the School of Mechatronic Engineering and Automation, Shanghai University, and Research Institute of USV Engineering, Shanghai 200444, P. R. China (Email: imagecy@shu.edu.cn)}% <-this % stops a space

\thanks{Y-X~Peng is with the Department of Mathematics, School of Science, Shanghai University, Shanghai 200444, P. R. China (Email: yaxin.peng@shu.edu.cn)}% <-this % stops a space

\thanks{Y~Peng is the School of Mechatronic Engineering and Automation, Shanghai University, and Research Institute of USV Engineering, Shanghai 200444, P. R. China (Email: pengyan@shu.edu.cn)}% <-this % stops a space

\thanks{* Corresponding author: Z-G~Jia is the School of Mathematics and Statistics, Jiangsu Normal University, and Research Institute of Mathematical Science, Jiangsu 221116, P. R. China (Email: zhgjia@jsnu.edu.cn)}% <-this % stops a space
}

\maketitle

\begin{abstract}
Quaternion singular value decomposition (QSVD) is a robust technique of digital watermarking which can extract high quality watermarks from watermarked images with low distortion. In this paper, QSVD technique is further investigated and an efficient robust watermarking scheme is proposed. The improved algebraic structure-preserving method is proposed to handle the problem of ``explosion of complexity'' occurred in the conventional QSVD design. Secret information is transmitted blindly by incorporating in QSVD two new strategies, namely, coefficient pair selection and adaptive embedding. Unlike conventional QSVD which embeds watermarks in a single imaginary unit, we propose to adaptively embed the watermark into the optimal hiding position using the Normalized Cross-Correlation (NC) method. This avoids the selection of coefficient pair with less correlation, and thus, it reduces embedding impact by decreasing the maximum modification of coefficient values. In this way, compared with conventional QSVD, the proposed watermarking strategy avoids more modifications to a single color image layer and a better visual quality of the watermarked image is observed. Meanwhile, adaptive QSVD resists some common geometric attacks, and it improves the robustness of conventional QSVD. With these improvements, our method outperforms conventional QSVD. Its superiority over other state-of-the-art methods is also demonstrated experimentally.

\end{abstract}


\IEEEpeerreviewmaketitle

\section{Introduction}
\IEEEPARstart{T}{oday} in the era of big data, an increasingly serious problem is the illegal copying, modification, and forgery of digital media. Digital watermarking is considered as a powerful technology to protect intellectual property theft, which involves the insertion of specific information (i.e., a watermark) into multimedia documents as a form of identification \cite{TR07}. Digital watermarks can be embedded in noise-tolerant multimedia, including image, audio, or video. Therefore, image watermarks can be classified as robust and fragile. Robust watermarking is meant to be resilient to modification attempts, whereas fragile watermarking makes the embedded information sensitive to modification. Data hiding can be reversible or irreversible, where the former is used to ensure exact recovery of the original host image after extracting the embedded information \cite{ZWWX19}. Different reversible watermarking methods have been proposed to achieve high data hiding capacity with low distortion, such as lossless compression \cite{CSTS05}, difference expansion (DE) \cite{LSZLT12,DC14,DC15}, histogram shifting (HS)\cite{LLYZ13,WYWW18,JYZL19} and prediction-error expansion (PEE) \cite{LYZ11,OLZN15,OLZN13,C12}. In addition, such as the patchwork theory \cite{ZPP10} and integer wavelet transform (IWT) \cite{AGLTDL12}, they can also successfully resist JPEG compression. For irreversible watermarking methods, some of them involve embedding and extracting the watermark using feature invariant coefficient, and combining the watermark bits to establish association rules by slightly modifying the host image pixels \cite{SNZPL13,SNZL13,FLCZ18}. Although these schemes are irreversible, the correlation of their feature invariant coefficients is robust enough to resist various possible attacks and ensure that the watermarked image has high invisibility without affecting the normal use of the host image.

The SVD-based methods \cite{SNZPL13,SNZL13,CS17,MAS08,BL13,FLCZ18,LWZZ16} and its variants \cite{BM05,MY15} accomplish robust embedding by slightly modifying the singular values or singular vectors of the factorized matrix in non-overlapping blocks. Due to the stability of SVD, that is, small numerical changes of singular values or the correlation of feature invariant elements in singular vectors after different attacks, these methods were reported to be robust to most attacks. Chang \emph{et al.}\cite{CS17} proposed a model that exploits the features of watermarks to establish association rules and embeds the rules into a host image. Their model combined with SVD technology can resist different image processing attacks. The experimental results also indirectly reflect that SVD technology is a powerful tool for robust watermarking. Bao \emph{et al.}\cite{BL13} proposed an image-adaptive watermarking scheme by applying a simple quantization-index-modulation process on the singular value of wavelet subband. The scheme is robust against the JPEG compression. Mohammad \emph{et al.} \cite{MAS08} proposed an improved SVD-based watermarking method by directly using the reference watermark. This method successfully solves the problem of false positive detection and is extremely robust against geometric distortion attacks. Note that in the watermarking scheme implemented by modifying the singular value, since the singular value is closely related to the pixels of the entire image, the watermarked images all show poor visual quality after being reconstructed. Su \emph{et al.} \cite{SNZPL13,SNZL13} proposed a color image watermarking scheme by using the correlation of feature invariant elements in the first column of the factorized matrix $U$ of SVD. Experimental results show that their algorithm is robust against some common attacks with low distortion. However, these watermarking schemes are designed to individually mark each image layer, which ignores the correlation of the three image layers (R,G,B) and the synchronization of the watermark embedding. Recently, Li \emph{et al.}\cite{LWZZ16} rearranged the three color layers of the color image block into a column matrix $M$. The watermarking scheme exploits the SVD of the matrix $M$ to embed and extract the watermark in a holistic manner. Liu \emph{et al.}\cite{FLCZ18} proposed a blind watermarking scheme based on QSVD. The scheme uses the traditional QSVD algorithm to embed the watermark into the optimal coefficient pair of the imaginary part $\mathbf{i}$ of the quaternion matrix $U$. Experimental results show that their algorithm has superior performance in terms of robustness and invisibility. Among those methods, QSVD has attracted considerable attention since the method has the potential to well exploit the integration of quaternion and the feature invariant coefficient pairs of SVD in natural images. So that we focus on developing a new QSVD-based approach of watermarking with making an improvement on the QSVD algorithm and introducing a novel adaptive embedding technique.

Quaternion were introduced by Hamilton in 1843 \cite{WR11}. The literature on quaternion matrices have extensive applications in many research fields, consisting of quantum mechanics \cite{DJD21}, field theory \cite{DJSD20} and image processing \cite{NS12,JNS19,JNW19}. In particular, the quaternion matrix applied to the image representation, which integrates well the three image layers of the color image. It not only allows the three image layers to be processed synchronously but also avoids loss of correlation between image layers. Hosny \emph{et al.} \cite{HD18} proposed a robust color image watermarking method based on quaternion Legendre-Fourier moments (QLFMs) to resist geometric attacks and common signal processing attacks. Experimental results show that this method has superior performance in terms of visual imperceptibility and robustness when resisting various attacks. Later, they developed a resilient color image watermarking method using accurate quaternion radial substituted Chebyshev moments (QRSCMs) \cite{HD19}. The high-precision, fast, and stable calculation of QRSCMs makes the designed watermarking scheme have better performance. Based on accurate polar harmonic Fourier moments (PHFMs) and chaotic mapping, Ma \emph{et al.} \cite{MCWLWS20} proposed a novel watermarking algorithm for copyright protection, which has strong robustness against geometric attacks by relying on the geometric invariance of accurate PHFMs. These watermarking methods are feasible and have room for improvement in reducing the computational complexity and increasing the watermark capacity.

The SVD \cite{GC20, GC96} is one of the most powerful tools in matrix computations, The SVD of a quaternion matrix was theoretically derived in 1997 by Zhang \cite{ZF57}. On the basis of theory, the SVD calculation of the quaternion matrix cannot be realized quickly and effectively, especially the large-scale matrix. In 2005, the quaternion toolkit developed by Sangwine \emph{et al.}\cite{SN} facilitated the research of workers and was widely used by most scholars. However, we clearly recognize that its computational complexity is quite expensive. Recently, Jia \emph{et al.}\cite{JWL24} proposed a real structure-preserving algorithm of Givens based transformation for solving the right eigenvalue problem of Hermitian quaternion matrices, using the special structures and properties of real counterpart of Hermitian quaternion matrices. Afterwards, Jia \emph{et al.}'s method \cite{JWL24} is developed by Li \emph{et al.}\cite{LWZZ67} to calculate the SVD of the quaternion matrix, where the algorithm costs about one-third CPU time of that by running ${\tt svd}$ of Quaternion toolkit. The advantage of real structure-preserving algorithm is that the real arithmetic number is greatly reduced but the assignment numbers is large. Therefore, Li \emph{et al.} \cite{LWZZ96} further proposed a structure-preserving algorithm of Householder based transformation to calculate the SVD of the quaternion matrix, which takes about half CPU time of that by running the Givens based algorithm. It effectively reduces the assignment number, but the spatial structure is not fully utilized in the bidiagonal reduction process for quaternion matrices. With the rapid development of these disciplines, high-performance algorithms for solving these problem are still greatly in need.

The mechanism of QSVD-based embedding can be described as follows. Firstly, a color host image is represented by a pure quaternion matrix. Secondly, the pure quaternion matrix is divided into non-overlapping blocks and the QSVD algorithm is performed to obtain its $U$ matrix at each block that needs to be embedded. Finally, the watermark bits are embedded or extracted according to some feature invariance in the quaternion matrix $U$. Though the QSVD-based watermarking scheme works well, there is still room for improvement. Actually, focusing on the highly correlated coefficient pairs can fully exploit the local data, which is very helpful to enhance watermark performance.

In this paper, the QSVD technique is further improved and an efficient robust watermarking scheme is presented. The main contributions of this brief are as follows:
\begin{itemize}
  \item Efficient algorithm: Unlike the conventional QSVD algorithm, we propose an efficient real structure-preserving QSVD algorithm combined with the quaternion Householder transformation and the generalized Givens transformation. It focuses on the optimization of the spatial structure and the improvement of replacing the quaternion Householder transformation with the generalized Givens transformation once the two quaternions are calculated in the bidiagonal reduction of quaternion matrix.
      In fact, it is also verified in the numerical experiments that our structure-preserving algorithm only costs about half of the CPU time of the algorithms proposed in \cite{LWZZ67} and \cite{LWZZ96}.
  \item Coefficient pairs selection: As an intermediate step of watermarking scheme, according to some characteristics of quaternion matrix $U$, such as magnitude and sign, we propose to calculate the highly correlated coefficient pairs for watermark embedding using the Normalized Cross-Correlation (NC) method. Compared with the statistical methods in the conventional QSVD, an optimal feature-invariant position is obtained, which is more robust to some common geometric attacks.
  \item Adaptive embedding: Conventional QSVD embeds a bit into one of three imaginary units ($\mathbf{i,j,k}$), which causes more modification in a single image layer. Based on the determination of the coefficient pairs, we adaptively select the optimal hidden position in the three coefficient pairs of the three imaginary units. In this way, a watermarked image with better visual quality is obtained compared to conventional QSVD.
\end{itemize}
With these improvements, our method performs better than the conventional QSVD-based methods. It can extract high quality watermark with low distortion.

The rest of this paper is organized as follows. In Section \ref{sec1}, we present some properties of quaternion matrices and the real counterparts. After that, we present an efficient structure-preserving algorithm for computing the SVD of quaternion matrices in Section \ref{sec2}. In Section \ref{sec3}, the advantages of coefficient pairs selection and adaptive embedding are explained. In Section \ref{sec4}, based on the proposed QSVD algorithm, the proposed watermark embedding and extraction schemes are described in details. In Section \ref{sec5}, our algorithm is evaluated in comparison with the latest structure-preserving algorithm, and our watermarking scheme is evaluated in comparison with the conventional QSVD and some other state-of-the-art works. Concluding remarks are given in the last section.

\section{Preliminaries}\label{sec1}
In this section we present some basic results for quaternion matrices and their real counterparts.

We always use the following notation. Let the superscripts $^T$ and $^H$ denote the transpose and the conjugate transpose, respectively. $\mathbb{R}$ and $\mathbb{C}$ are the sets of the real and complex numbers, respectively. We denote by $\mathbb{H}$ the set of quaternion:
$$\mathbb{H}=\{q = a + b\mathbf{i} + c\mathbf{j} + d\mathbf{k}, \quad a, b, c, d\in \mathbb{R}\},$$
where three imaginary units $\mathbf{i,j, k}$ satisfy
$$\mathbf{i}^2 = \mathbf{j}^2 = \mathbf{k}^2 = \mathbf{ijk}=-1.$$
When $a=0$, $q$ is called pure quaternion. For $q = a + b\mathbf{i} + c\mathbf{j} + d\mathbf{k}\in \mathbb{H}$, the conjugate of $q$ is $\bar{q}=a-b\mathbf{i}-c\mathbf{j}-d\mathbf{k}$,  and $|q|^{2}=q\bar{q}=a^{2}+b^{2}+c^{2}+d^{2}$. If $q\ne 0$, then the inverse $q^{-1} = \overline{q}/|q|^2$ is unique.
Let $\mathbb{H}^{m \times n}$ and $\mathbb{H}^{n}$ be the collections of all $m\times n$ matrices with entries in $\mathbb{H}$ and $n$-dimensional vectors, respectively. For any quaternion matrix  $Q=Q_{0}+Q_{1}\mathbf{i}+Q_{2}\mathbf{j}+Q_{3}\mathbf{k} \in \mathbb{H}^{m\times n}$, $Q^{T}=Q_{0}^T+Q_{1}^T\mathbf{i}+Q_{2}^T\mathbf{j}+Q_{3}^T\mathbf{k}$, and  $Q^{H}=\bar{Q}^{T}=Q_{0}^T-Q_{1}^T\mathbf{i}-Q_{2}^T\mathbf{j}-Q_{3}^T\mathbf{k}$.

In \cite{SN06}, Sangwine proposed to encode the three image layers of an RGB image on the three imaginary units of a pure quaternion matrix, that is,
\begin{equation*}
Q(x,y)=R(x,y)\mathbf{i}+G(x,y)\mathbf{j}+B(x,y)\mathbf{k},
\end{equation*}
where $R(x,y)$, $G(x,y)$ and $B(x,y)$ are the red, green and blue values of the pixel $(x,y)$, respectively. Obviously, the advantage of the quaternion matrix used to investigate many color image problems is that quaternion can treat three image layers holistically without neglecting the correlation and synchronicity between the image layers, and its disadvantage is computational complexity.

In the following text, we always apply the real counterpart of $Q$ defined in \cite{JWL24} to deal with the computational complexity of quaternion, denoted by
\begin{equation}\label{eq1}
   Q^{R}\equiv\left[
              \begin{smallmatrix}
                ~~Q_0&~~Q_2&~~Q_1&~~Q_3 \\
                -Q_2&~~Q_0&~~Q_3&-Q_1 \\
                -Q_1&-Q_3&~~Q_0&~~Q_2 \\
                -Q_3&~~Q_1&-Q_2&~~Q_0 \\
              \end{smallmatrix}
            \right],
\end{equation}
and their relationship is equivalent. The algebraic structure of $Q^R$ is called JRS-symmetry in \cite{JWL24}, which is implicitly preserved in the algebraic structure-preserving algorithms in \cite{JWL24,LWZZ67,LWZZ96,JWZC18}. Thus, for the convenience of loading, transformation, extraction and computation, $Q^{R}$ is simplified to a block column or a block row as follows:
\begin{equation}\begin{split}\label{eq2}
   Q_{c}^{R}=&\left[\begin{smallmatrix}
                Q_0^T&-Q_2^T & -Q_1^T & -Q_3^T
                \end{smallmatrix}
            \right]^T, \\
             Q_{r}^{R}=&\left[\begin{smallmatrix}
                Q_0& Q_2& Q_1& Q_3
              \end{smallmatrix}
            \right],\end{split}
\end{equation}
which avoids the dimensional expansion problem caused by the real counterpart. In this way, the computation workload, computational time and storage spaces are greatly reduced by using real operation instead of quaternion operation.
\section{Improved quaternion singular value decomposition}\label{sec2}

In this section, we present an improved version of QSVD.

\begin{theorem}[QSVD \cite{ZF57}]\label{th1}
Let $Q\in\mathbb{H}^{m\times n}$. Then there exist two unitary quaternion matrices $U\in \mathbb{H}^{m\times m}$ and $V\in \mathbb{H}^{n\times n}$ such that
$$Q=U \Sigma V^{H},$$
where $\Sigma={\rm diag}(\sigma_{1},\sigma_{2},\cdots,\sigma_{\ell})$, $\sigma_{1}\geq \sigma_{2}\geq\cdots\geq \sigma_{\ell}\geq 0$ are positive singular values of $Q$, and $\ell=min(m,n).$
\end{theorem}

Firstly, we recall two important quaternion unitary transformations: generalized Givens transformation \cite{JWZC18} and Householder transformations \cite{JWL24,LWZZ96,GBM89}.
\begin{algorithm} [generalized Givens transformation \cite{JWZC18}]\label{al1}
{\rm Let $x=[x_1 ~x_2]^T \in \mathbb{H}^2$ be given with $x_2\neq 0.$ Then there exists a generalized Givens matrix \begin{equation*}
\mathscr{G}=\left[
\begin{array}{cc}
q_{11} & q_{12} \\
q_{21} & q_{22} \\
\end{array}
\right],
\end{equation*}
such that $\mathscr{G}^{H}x=[\|x\|_2 ~ 0]^T.$ A choice of $\mathscr{G}$ is
\begin{equation*}
  q_{11}=\frac{x_1}{\|x\|_2}, q_{21}=\frac{x_2}{\|x\|_2}.
\end{equation*}
In order to ensure stability, the selection problem of $q_{12}$, $q_{22}$ will be discussed in the following two cases.
\begin{description}
  \item[(a)] {\rm if}~ $|x_1|\leq |x_2|$, $$q_{12}=|q_{21}|, \quad q_{22}=-|q_{12}|q_{21}^{-H}q_{11}^H;$$
  \item[(b)] {\rm if}~ $|x_1|> |x_2|$, $$q_{22}=|q_{11}|, \quad q_{12}=-|q_{11}|q_{11}^{-H}q_{21}^H.$$
\end{description}
}
\end{algorithm}

\begin{algorithm}[Householder transformation \cite{JWL24,LWZZ96,GBM89}] \label{al2}
{\rm Given two different quaternion vectors $x=[x_1,\cdots,x_n]^T,~y=[y_1,\cdots,y_n]^T\in\mathbb{H}^n$ with $\|x\|=\|y\|$ and $y^{H}x\in\mathbb{R}$,
there exists a quaternion Householder matrix defined by $\mathscr{H}=I-2uu^{H},$
where $u=\frac{y-x}{\|y-x\|}$, such that $\mathscr{H}y=x$.
For any real vector $v\in\mathbb{R}^n$ with  $\|v\|=1$,
\begin{itemize}
\item when $x=\alpha v$  with $\alpha\in\mathbb{H}$ and $|\alpha|=\|y\|$,
$\mathscr{H}_1:=I-2uu^{H},$
where $u=\frac{y-x}{\|y-x\|},~(\text{proposed in {\rm\cite{GBM89}}} )$
\item when $x=\|y\|v$,
 $\mathscr{H}_2:=(I-2uu^T)G,$
 where $u=\frac{Gy-x}{\|Gy-x\|}$,
$G={\rm diag}(g_1,g_2,\dots,g_n),$
$$g_\ell=
\begin{cases}
\frac{\overline{y_\ell}}{|y_\ell|},\quad y_\ell \neq0, \\
1,~\text{otherwise},
\end{cases}
$$
(proposed in {\rm\cite{JWL24}})
\item  and
when $x=\|y\|v$, %
 $\mathscr{H}_3:=G\mathscr{H}_1,$
where $G={\rm diag}(g_1,g_2,\dots,g_n),$
$$g_\ell=
\begin{cases}
\frac{\overline{z_{\ell}}}{|z_{\ell}|},\quad z_{\ell}\neq 0, \\
1,~\text{otherwise}
\end{cases}
\text{with}~z=\mathscr{H}_1y.
$$
(proposed in {\rm\cite{LWZZ96}})
\end{itemize}
}
\end{algorithm}
Then, we proposes an efficient QSVD algorithm based on $\mathscr{G}$ in Algorithm \ref{al1} and $\mathscr{H}_{3}$ in Algorithm \ref{al2}.

Let $Q$ be an $m \times n$ quaternion matrix with $ m \geq n$. The bidiagonal reduction proceeds by alternately pre- and postmultiplying $Q$ by Householder transformations $\mathscr{H}_{3}$ and/or generalized Givens transformations $\mathscr{G}$. The reduction of the bidiagonalization is shown in Fig.~\ref{fig1}. For the first step, we determine a quaternion Householder transformation $\mathscr{H}_{3}^{l}$ to eliminate all hatted quaternions in the first column of $Q$ being zero except for the first hatted element. Then another transformation $\mathscr{H}_{3}^{r}$ is determined to eliminate all hatted quaternions in first row of $Q$ being zero except for the first hatted element. Next, the reduction proceeds recursively on the matrix $Q(2:m,2:n)$. Once the two quaternions are rotated in the last two rows and columns of the matrix, we replace $\mathscr{H}_{3}$ with $\mathscr{G}$ to reduce the expensive computation.

In applying the rotations, however, we must distinguish two cases. In the first case, when QSVD of a large-scale quaternion matrix is executed, the efficiency gain of rotating two quaternions by using $\mathscr{G}$ instead of $\mathscr{H}_{3}$ is negligible. However, our proposed efficient Algorithm \ref{al3}, shown in Fig.~\ref{fig2}, optimizes the data structure of the algorithm in \cite{LWZZ96} to avoid the redundant data being transmitted and calculated, and is compared with $\mathscr{H}_{2}$ based and $\mathscr{H}_{3}$ based QSVD algorithm for CPU time and precision in Example \ref{sec5}.1. In the second case, if we wish to calculate the QSVD of a $4\times 4$ quaternion matrix, the reduced computational amount of replacing the $\mathscr{H}_{3}$ with $\mathscr{G}$ is huge, see Remark \ref{rem1} for the explanation.
\begin{figure*}[!t]
\normalsize
\setcounter{mytempeqncnt}{\value{equation}}
\setcounter{equation}{5}
\begin{equation*}
\left[
\begin{smallmatrix}
\widehat{\times}  & \times & \times & \times \\
\widehat{\times } & \times & \times & \times \\
\widehat{\times}  & \times & \times & \times \\
\widehat{\times } & \times & \times & \times \\
\end{smallmatrix}
\right]\overset{\mathscr{H}_{3}^{l}}{\Longrightarrow}
\left[
\begin{smallmatrix}
r & \widehat{\times}   & \widehat{\times } & \widehat{\times}  \\
0 & \times   & \times  & \times  \\
0 & \times   & \times  & \times\\
0 & \times   & \times  & \times\\
\end{smallmatrix}
\right]\overset{\mathscr{H}_{3}^{r}}{\Longrightarrow}
\left[
\begin{smallmatrix}
r & r  & 0 & 0 \\
0 & \widehat{\times } & \times & \times \\
0 & \widehat{\times}  & \times & \times \\
0 & \widehat{\times } & \times & \times\\
\end{smallmatrix}
\right]\overset{\mathscr{H}_{3}^{l}}{\Longrightarrow}
\left[
\begin{smallmatrix}
r & r & 0 & 0 \\
0 & r & \widehat{\times} & \widehat{\times} \\
0 & 0 & \times & \times \\
0 & 0 & \times & \times \\
\end{smallmatrix}
\right]\overset{\mathscr{G}^{r}}{\Longrightarrow}
\left[
\begin{smallmatrix}
r & r & 0 & 0 \\
0 & r & r & 0 \\
0 & 0 & \widehat{\times} & \times \\
0 & 0 & \widehat{\times} & \times \\
\end{smallmatrix}
\right]\overset{\mathscr{G}^{l}}{\Longrightarrow}
\left[
\begin{smallmatrix}
r & r & 0 & 0 \\
0 & r & r & 0 \\
0 & 0 & r & \widehat{\times} \\
0 & 0 & 0 & \times \\
\end{smallmatrix}
\right]\overset{G^{r}}{\Longrightarrow}
\left[
\begin{smallmatrix}
r & r & 0 & 0 \\
0 & r & r & 0 \\
0 & 0 & r & r\\
0 & 0 & 0 & \widehat{\times} \\
\end{smallmatrix}
\right]\overset{G^{l}}{\Longrightarrow}
\left[
\begin{smallmatrix}
r & r & 0 & 0 \\
0 & r & r & 0 \\
0 & 0 & r & r \\
0 & 0 & 0 & r \\
\end{smallmatrix}
\right].
\end{equation*}

\setcounter{equation}{\value{mytempeqncnt}}
\caption{The bidiagonal reduction process of $4\times 4$ quaternion matrices. Here, the letter $r$ denotes some real numbers and $\widehat{\times}$ represents the element to be processed, the superscripts $^l$ and $^r$ denote the alternately pre- and post multiplication, respectively.}\label{fig1}
\hrulefill
\vspace*{4pt}
\end{figure*}

\begin{figure*}[!t]
\normalsize
\setcounter{mytempeqncnt}{\value{equation}}
\setcounter{equation}{5}
\begin{algorithm} [QSVD based on $\mathscr{H}_{3}$ and $\mathscr{G}$ ]\label{al3}
{\rm Given a quaternion matrix $Q=Q_0+Q_1\mathbf{i}+Q_2\mathbf{j}+Q_3\mathbf{k}\in \mathbb{H}^{m\times n}$, where $Q_{k} \in \mathbb{R}^{m\times n}$, $k=0,1,2,3$, this algorithm presents two orthogonal matrices $U \in \mathbb{H}^{m\times m}$, $V\in \mathbb{H}^{n\times n}$ and a diagonal matrix $\Sigma \in \mathbb{R}^{m\times n}$. Without loss of generality, we suppose $m\geq n$}.
\end{algorithm}
Two auxiliary functions: $in(p,q)_{n}: =[p:q,n+p:n+q,2n+p:2n+q,3n+p:3n+q]$, $id(p)_{n}:=[p,n+p,2n+p,3n+p],$  for any positive integers $p$ and $q$. ${\tt eye, zeros}$ and ${\tt svd}$ are functions of MATLAB.\\
\textbf{\rm \textbf{Function}:} $[U,\Sigma,V]={\tt QSVD}(Q_0,Q_1,Q_2,Q_3)$

$C:=Q^{R}_{c}=[Q^{T}_0,-Q^{T}_2,-Q^{T}_1,-Q^{T}_3]^{T}, U=[{\tt eye}(m),{\tt zeros}(m,3m)], V=[{\tt eye}(n);{\tt zeros}(3n,n)]$.

1. for $s=1:n-2$

2. \quad $C(in(s,m)_{m},s:n)=\mathscr{H}^{s}_{3}\ast C(in(s,m)_{m},s:n)$;\quad $U(:,in(s,m)_{m})=U(:,in(s,m)_{m})\ast \mathscr{H}^{s}_{3}$;

4. \quad $T=[C(s:m,s+1:n),-C(m+s:2m,s+1:n),-C(2m+s:3m,s+1:n),-C(3m+s:4m,s+1:n)]$;

5. \quad if $s<n-2$

6. \quad \quad $T=T\ast \mathscr{H}^{s}_{3}$;\quad $V(in(s+1,n)_{n},:)=\mathscr{H}^{s}_{3}\ast V(in(s+1,n)_{n},:)$;

8. \quad else

9. \quad \quad $T=T\ast \mathscr{G}^{s}$;\quad $V(in(s+1,n)_{n},:)=(\mathscr{G}^{s})^{T}\ast V(in(s+1,n)_{n},:)$;

10.\quad end

11.\quad $C(in(s,m)_{m},s+1:n)=[T(:,1:n-s);-T(:,n-s+1:2(n-s));$
\begin{center}
$-T(:,2(n-s)+1:3(n-s));-T(:,3(n-s)+1:4(n-s))]$;
\end{center}
12. end

13. $s=n-1$;

14. if $m=n$

15.\quad $C(in(s,n)_{n},s:n)=(\mathscr{G}^{s})^{T}\ast C(in(s,n)_{n},s:n)$;\quad $U(:,in(s,n)_{n})=U(:,in(s,n)_{n})\ast \mathscr{G}^{s}$;

16. else

17.\quad $C(in(s,m)_{m},s:n)=\mathscr{H}^{s}_{3}\ast C(in(s,m)_{m},s:n)$;\quad $U(:,in(s,m)_{m})=U(:,in(s,m)_{m})\ast \mathscr{H}^{s}_{3}$;

18. end

19. $T=[C(s:m,n),-C(m+s:2m,n),-C(2m+s:3m,n),-C(3m+s:4m,n)]$;

20. $T=T\ast G^{s}$

21. $C(in(s,m)_{m},n)=[T(:,1);-T(:,2);-T(:,3);-T(:,4)]$;\quad $V(id(n)_{n},:)=(G^{s})^{T}\ast V(id(n)_{n},:)$;

22. $s=n$;

23. if $m=n$

24.\quad $C(id(s)_{m},s:n)=(G^{s})^{T}\ast C(id(s)_{m},s:n)$;\quad $U(:,id(s)_{m})=U(:,id(s)_{m})\ast G^{s}$;

25. else

26.\quad $C(in(s,m)_{m},s:n)=\mathscr{H}^{s}_{3}\ast C(in(s,m)_{m},s:n)$;\quad $U(:,in(s,m)_{m})=U(:,in(s,m)_{m})\ast \mathscr{H}^{s}_{3}$;

27. end

28. $B=C(1:m, 1:n)$; $[u_{b},d_{b},v_{b}^{T}]={\tt svd}(B);$

29.$$\begin{smallmatrix}&U_{0}&=&U^{T}(1:m,1:m)\ast u_{b};&  &U_{1}&=&-U^{T}(1:m,2m+1:3m)\ast u_{b};& &U_{2}&=&-U^{T}(1:m,m+1:2m)\ast u_{b};&  &U_{3}&=&-U^{T}(1:m,3m+1:4m)\ast u_{b};&\\
   &V_{0}&=&V(1:n,1:n)\ast v_{b};&  &V_{1}&=&-V(2n+1:3n,1:n)\ast v_{b};& &V_{2}&=&-V(n+1:2n,1:n)\ast v_{b};&  &V_{3}&=&-V(3n+1:4n,1:n)\ast v_{b};&\end{smallmatrix}$$
30. $\Sigma=d_{b}$; $U=U_{0}+U_{1}\mathbf{i}+U_{2}\mathbf{j}+U_{3}\mathbf{k}$; $V=V_{0}+V_{1}\mathbf{i}+V_{2}\mathbf{j}+V_{3}\mathbf{k}.$

Where $\mathscr{H}^{s}_{3}$, $\mathscr{G}^{s}$ and $G^{s}$ are transformations performed by the current loop step $s$.
\setcounter{equation}{\value{mytempeqncnt}}
\caption{The QSVD reduction process of quaternion matrices.}\label{fig2}
\hrulefill
\vspace*{4pt}
\end{figure*}

\begin{remark}\label{rem1}
{\rm In Table \ref{tab1}, we present the comparison on real arithmetic numbers and assignment numbers between the quaternion generalized Givens transformation and the quaternion Householder transformation when two quaternions are rotated, where assignment numbers refer to call subroutines or perform matrix operations. The $\mathscr{H}_{3}$ is replaced by $\mathscr{G}$, which reduces 4 assignment numbers and 41 real flops. From Fig. \ref{fig1}, the QSVD of a $4\times 4$ quaternion matrix replaces $\mathscr{H}_{3}$ twice with $\mathscr{G}$, so that approximately 274 real flops are reduced. A color image of size $512\times 512$ is divided into $128\times 128$ image blocks of size $4\times 4$. The proposed QSVD applied to all image blocks will have approximately $128\times 128\times 274$ expensive real flops reduced. Algorithm \ref{al3} takes about $64(mn^2-n^3/3)$ flops for the bidiagonalization for $Q^{R}$ of an $m\times n$ quaternionic matrix $Q$ without generating unitary matrices.}
\end{remark}

\begin{table*}[!htbp]
\caption{Computation amounts and assignment numbers for quaternion Transformations.}\label{tab1}
\centering
\begin{tabular}{|p{2cm}<{\centering}|p{2cm}<{\centering}|p{2cm}<{\centering}|p{2cm}<{\centering}|p{2cm}<{\centering}|}
\hline
 Methods & \multicolumn{2}{p{4cm}<{\centering}|}{Generate $\mathscr{H}_{3}(\mathscr{G})$} & \multicolumn{2}{p{4cm}<{\centering}|}{Transformation $\mathscr{H}_{3}(\mathscr{G})x, x\in \mathbb{H}^{2}$}\\
\cline{2-5}
     &Assignment&Real flops&Assignment&Real flops\\
\hline
$\mathscr{H}_{3}$     & 11 & 46 & 4 & 184  \\
\hline
$\mathscr{G}$   & 9 & 69 & 2 & 120  \\
\hline
\end{tabular}
\end{table*}

\section{Coefficient pairs selection and adaptive embedding}\label{sec3}
\subsection{Coefficient pair selection}
We first give an example to illustrate the advantage of coefficient pair selection. Suppose that a $4\times 4$ quaternion matrix $Q_{k}$ is one of the blocks of the host image, and has the QSVD as
\begin{equation*} Q_{k}= \left[                               \begin{smallmatrix}
                                 q_{11} & q_{12} & q_{13}& q_{14} \\
                                 q_{21} & q_{22} & q_{23}& q_{24} \\
                                 q_{31} & q_{32} & q_{33}& q_{34} \\
                                 q_{41} & q_{42} & q_{43}& q_{44} \\
                                \end{smallmatrix}
                              \right]:=U\Sigma V^{H}
                              \end{equation*}\\
 \begin{equation}\label{eq3}
                              =\left[
                                \begin{smallmatrix}
                                 u_{11} & u_{12} & u_{13} & u_{14}\\
                                 u_{21} & u_{22} & u_{23} & u_{24}\\
                                 u_{31} & u_{32} & u_{33} & u_{34}\\
                                 u_{41} & u_{42} & u_{43} & u_{44}\\
                                \end{smallmatrix}
                              \right]
                              \left[
                                \begin{smallmatrix}
                             \sigma_{1} & 0           & 0 & 0 \\
                                    0    & \sigma_{2} & 0 & 0 \\
                                    0    & 0      & \sigma_{3} & 0 \\
                                    0    & 0           & 0 & \sigma_{4}\\
                                \end{smallmatrix}
                              \right]
                              \left[
                                \begin{smallmatrix}
                                 v_{11} & v_{12} & v_{13} & v_{14}  \\
                                 v_{21} & v_{22} & v_{23} & v_{24} \\
                                 v_{31} & v_{32} & v_{33} & v_{34} \\
                                 v_{41} & v_{42} & v_{43} & v_{44} \\
                                \end{smallmatrix}
\right].\end{equation}

It is noted that the component $U$ has an interesting property, i.e., all elements of the three imaginary parts $(\mathbf{i,j,k})$ of the first column of the $U$ component have the same symbol respectively and their values are very close. For example, a pure quaternion matrix $Q_{k}=Q_{1}\mathbf{i}+Q_{2}\mathbf{j}+Q_{3}\mathbf{k} \in \mathbb{H}^{4 \times 4}$ obtained from a digital image block, where
\begin{equation*}
Q_1=\left[
\begin{smallmatrix}
0.3930 &0.5392 &0.4243 &0.2593 \\
0.1695 &0.2347 &0.9128 &0.0087  \\
0.5745 &0.6261 &0.8279 &0.5718 \\
0.7796 &0.5523 &0.2507 &0.6526
\end{smallmatrix}
\right],
\end{equation*}
\begin{equation*}
Q_2=\left[
\begin{smallmatrix}
0.6489 &0.4875 &0.8106 &0.8273 \\
0.1580 &0.9323 &0.7149 &0.3924 \\
0.3141 &0.9348 &0.6791 &0.7853 \\
0.0619 &0.6796 &0.2172 &0.2155
\end{smallmatrix}
\right],
\end{equation*}
\begin{equation*}
Q_3=\left[
\begin{smallmatrix}
0.4207 &0.7816 &0.2996 &0.2786  \\
0.3970 &0.5307 &0.3980 &0.9020  \\
0.5799 &0.6404 &0.7491 &0.1425  \\
0.2203 &0.2465 &0.9459 &0.4230
\end{smallmatrix}
\right],
\end{equation*}
by performing Algorithm \ref{al3}, we have
\begin{equation*}
\begin{split}
U&=\left[
\begin{smallmatrix}
 -0.0056 & ~~0.0006 &  -0.0972  & ~~0.0076\\
~~0.0091 & ~~0.0909 &  -0.1524  &  -0.0395\\
 -0.0342 &  -0.1173 & ~~0.1112  &  -0.0320\\
 -0.0141 &  -0.1553 & ~~0.1215  & ~~0.1281\\
\end{smallmatrix}
\right]\\
&+\left[
\begin{smallmatrix}
  -0.2826 &  -0.0989  &  -0.0042  & ~~0.1016 \\
  -0.2657 &  -0.5240  & ~~0.2734  &  -0.0360 \\
  -0.4256 & ~~0.0661  &  -0.0610  &  -0.2312 \\
  -0.3224 & ~~0.5914  &  -0.2827  & ~~0.2786 \\
\end{smallmatrix}
\right]\mathbf{i}\\
&+\left[
\begin{smallmatrix}
   -0.2802 & ~~0.2037 &  ~~0.5790 &  ~~0.3170 \\
   -0.2323 &  -0.3537 &   -0.4099 &  ~~0.0873 \\
   -0.2624 & ~~0.0558 &  ~~0.2277 &   -0.4115 \\
   -0.0732 &  -0.0974 &   -0.3526 &  ~~0.1343 \\
\end{smallmatrix}
\right]\mathbf{j}\\
&+\left[
\begin{smallmatrix}
   -0.2731 &  ~~0.2883 & ~~0.0993  &  -0.4089 \\
   -0.3252 &   -0.0483 &  -0.2767  &  -0.0796 \\
   -0.3249 &   -0.1236 & ~~0.0824  & ~~0.5552 \\
   -0.2654 &   -0.1882 &  -0.0963  &  -0.2468 \\
\end{smallmatrix}
\right]\mathbf{k}.
\end{split}
\end{equation*}

As can be seen from $U$ matrix, the symbol and magnitude relationship of the elements of the three imaginary units of $u_{11}, u_{21}, u_{31}$ and $u_{41}$ can be used for robust watermarking.

In order to select the optimal embedding coefficient pair in $u_{11}, u_{21}, u_{31}$ and $u_{41}$, for six standard $512\times 512$ sized color images Baboon, F-16, House, Lena, Lostlake and Monolake \cite{UoG}, we adopt the following procedure to experiments.
\begin{itemize}
  \item The test image of size $M\times N$ is divided into image blocks of size $4\times 4$.
  \item Each image block is calculated by QSVD to obtain a $U$ matrix.
  \item Matrix $U^{\mathbf{i}}_{x}$ is composed of the coefficients of the imaginary part i of $u_{x 1}$ obtained for each image block, and $U^{\mathbf{j}}_{x}$ and $U^{\mathbf{k}}_{x}$ are also the same, where the dimensions of $U^{\mathbf{i}}_{x}, U^{\mathbf{j}}_{x}$ and $U^{\mathbf{k}}_{x}$ are $m\times n,$ $m=M/4,$ $n=N/4,$ and $x\in \{1,2,3,4\}$, respectively.
  \item Calculate the correlation between two matrices $U^{\mathbf{i}}_{x}$ and $U^{\mathbf{i}}_{y}$ (between $U^{\mathbf{j}}_{x}$ and $U^{\mathbf{j}}_{y}$, between $U^{\mathbf{k}}_{x}$ and $U^{\mathbf{k}}_{y}$) using Normalized Cross-Correlation (NC), where $y\in \{1,2,3,4\}, x\neq y$.
\end{itemize}
As can be seen from the Table \ref{tab2}, the average value of the imaginary part $\mathbf{i}$ of the NC$(u_{21},u_{31})$ is 0.9924, the average value of $\mathbf{j}$ is 0.9913, and the average value of $\mathbf{k}$ is 0.9924, which shows that the three coefficient pairs of the three imaginary parts of $u_{21}$ and $u_{31}$ are respectively the closest elements for many standard images. Therefore, it is noted that there exists a strong correlation between the coefficients of the three imaginary units $(\mathbf{i, j, k})$ of $u_{21}$ and $u_{31}$, respectively. This property can be explored for image watermarking.

\begin{table*}[!htbp]
\centering
\caption{The NC values of different elements in first column of $U$ component after QSVD.}\label{tab2}
\begin{tabular}{c|ccccccc}
\toprule  
\multicolumn{2}{c}{\quad Image} & NC($u_{11},u_{21}$) & NC($u_{11},u_{31}$) & NC($u_{11},u_{41}$) & NC($u_{21},u_{31}$) & NC($u_{21},u_{41}$) & NC($u_{31},u_{41}$)\\
\midrule  
\multirow{4}*{$\mathbf{i}$}
& Baboon  & 0.9856 & 0.9775 & 0.9703 & 0.9866 & 0.9727 & 0.9814\\
& F-16    & 0.9968 & 0.9922 & 0.9840 & 0.9978 & 0.9897 & 0.9945\\
& House   & 0.9942 & 0.9860 & 0.9802 & 0.9953 & 0.9877 & 0.9957\\
& Lena    & 0.9993 & 0.9978 & 0.9962 & 0.9993 & 0.9980 & 0.9994\\
& Lostlake& 0.9884 & 0.9762 & 0.9696 & 0.9886 & 0.9765 & 0.9885\\
& Monolake& 0.9873 & 0.9772 & 0.9707 & 0.9866 & 0.9769 & 0.9860\\
\cline{1-2}  
\multicolumn{2}{c}{\quad Average} & 0.9920 & 0.9845 & 0.9785 & \textbf{0.9924} & 0.9836 & 0.9909\\
\midrule 
\multirow{4}*{$\mathbf{j}$}
& Baboon  & 0.9772 & 0.9645 & 0.9559 & 0.9786 & 0.9590 & 0.9726\\
& F-16    & 0.9966 & 0.9908 & 0.9832 & 0.9972 & 0.9897 & 0.9958\\
& House   & 0.9946 & 0.9867 & 0.9759 & 0.9953 & 0.9834 & 0.9910\\
& Lena    & 0.9962 & 0.9899 & 0.9830 & 0.9966 & 0.9906 & 0.9967\\
& Lostlake& 0.9872 & 0.9734 & 0.9661 & 0.9877 & 0.9741 & 0.9874\\
& Monolake& 0.9926 & 0.9864 & 0.9827 & 0.9922 & 0.9864 & 0.9920\\
\cline{1-2}  
\multicolumn{2}{c}{\quad Average} & 0.9907 & 0.9820 & 0.9745 & \textbf{0.9913} & 0.9805 & 0.9893\\
\midrule  
\multirow{4}*{$\mathbf{k}$}
& Baboon  & 0.9772 & 0.9616 & 0.9509 & 0.9779 & 0.9560 & 0.9734\\
& F-16    & 0.9986 & 0.9964 & 0.9915 & 0.9987 & 0.9937 & 0.9961\\
& House   & 0.9946 & 0.9859 & 0.9724 & 0.9956 & 0.9817 & 0.9900\\
& Lena    & 0.9978 & 0.9947 & 0.9916 & 0.9979 & 0.9948 & 0.9979\\
& Lostlake& 0.9905 & 0.9814 & 0.9764 & 0.9908 & 0.9813 & 0.9905\\
& Monolake& 0.9941 & 0.9896 & 0.9871 & 0.9937 & 0.9897 & 0.9937\\
\cline{1-2}   
\multicolumn{2}{c}{\quad Average} & 0.9921 & 0.9849 & 0.9783 & \textbf{0.9924} & 0.9829 & 0.9903\\
\bottomrule 
\multicolumn{2}{c}{Total average} & 0.9916 & 0.9838 & 0.9771 & \textbf{0.9920} & 0.9823 & 0.9902\\
\bottomrule 
\end{tabular}
\end{table*}

\subsection{Adaptive embedding }

The three coefficient pairs of the three imaginary units between $u_{21}$ and $u_{31}$ have high correlation respectively. Usually, an imaginary unit is specified in three imaginary units to select its coefficient pair as the optimal for watermark embedding, such as the coefficient pair of the imaginary unit $\mathbf{i}$ is specified in Liu \emph{et al.}'s method \cite{FLCZ18}. In this paper, one of the three imaginary units is adaptively selected according to different images and different image blocks. From the longitudinal data in Table \ref{tab2}, the optimal embedding position of the watermark in the two images of Baboon and Lena is the imaginary part $\mathbf{i}$ of NC($u_{21}, u_{31}$), and their values are 0.9866 and 0.9993, respectively. For images F-16, House, Lostlake and Monolake, the optimal embedding position is the imaginary part $\mathbf{k}$ of NC($u_{21}, u_{31}$), and their values are 0.9987, 0.9956, 0.9908 and 0.9937, respectively. Moreover, the host image being tested is divided into a number of image blocks, which adds to the importance of adaptively selecting the imaginary unit and its coefficient pair. Therefore, the optimal coefficient pair are adaptively selected for watermark embedding according to different images and different image blocks, which avoids the selection of coefficient pair with low correlation, and thus, it reduces embedding impact by decreasing the maximum modification of coefficient values.

According to the selection of the optimal coefficient pair described above, $u_{21}$ and $u_{31}$ are modified with $\Delta_{1}$ and $\Delta_{2}$ respectively to hide the watermark information and the details will be shown in Section \ref{sec4}. The values of the corresponding modifications $\Delta_{1}$ and $\Delta_{2}$ can be divided into the following cases:
\begin{itemize}
  \item Imaginary unit $\mathbf{i}$:
      $\Delta_{1}=\delta_{11}\mathbf{i}, \Delta_{2}=\delta_{21}\mathbf{i}$,
  \item Imaginary unit $\mathbf{j}$:
      $\Delta_{1}=\delta_{12}\mathbf{j}, \Delta_{2}=\delta_{22}\mathbf{j}$,
  \item Imaginary unit $\mathbf{k}$:
      $\Delta_{1}=\delta_{13}\mathbf{k}, \Delta_{2}=\delta_{23}\mathbf{k}$,
\end{itemize}
where $\delta_{ij}\in \mathbb{R}, i\in \{1,2\}, j\in \{1,2,3\}.$ Assuming that the modifications are specified $\Delta_{1}=\delta_{11}\mathbf{i}, \Delta_{2}=\delta_{21}\mathbf{i}$\cite{FLCZ18}, then $u_{21}$ is modified to $u^{\star}_{21}=\alpha_{0}+(\alpha_{1}+\delta_{11})\mathbf{i}+\alpha_{2}\mathbf{j}+\alpha_{3}\mathbf{k}$ ($\alpha_{0,1,2,3}\in \mathbb{R}$), and $u_{31}$ is modified to $u^{\star}_{31}=\beta_{0}+(\beta_{1}+\delta_{21})\mathbf{i}+\beta_{2}\mathbf{j}+\beta_{3}\mathbf{k}$ ($\beta_{0,1,2,3}\in \mathbb{R}$). Let $v_{11}=\gamma_{0}+\gamma_{1}\mathbf{i}+\gamma_{2}\mathbf{j}+\gamma_{3}\mathbf{k}$ ($\gamma_{0,1,2,3}\in \mathbb{R}$), then the element $q_{21}$ in Eq.\eqref{eq3} is reconstructed by
\begin{equation*}
\begin{split}
q_{21}&=\sigma_{1}u^{\star}_{21}v_{11}+\sigma_{2}u_{22}v_{21}+\sigma_{3}u_{23}v_{31}+\sigma_{4}u_{24}v_{41}\\
&=\sigma_{1}(\alpha_{0}\gamma_{0}-(\alpha_{1}+\delta_{11})\gamma_{1}-\alpha_{2}\gamma_{2}-\alpha_{3}\gamma_{3})\\
&+\sigma_{1}[(\alpha_{0}\gamma_{1}+(\alpha_{1}+\delta_{11})\gamma_{0}+\alpha_{2}\gamma_{3}-\alpha_{3}\gamma_{2})]\mathbf{i}\\
&+\sigma_{1}[(\alpha_{0}\gamma_{2}-(\alpha_{1}+\delta_{11})\gamma_{3}+\alpha_{2}\gamma_{0}+\alpha_{3}\gamma_{1})]\mathbf{j}\\
&+\sigma_{1}[(\alpha_{0}\gamma_{3}+(\alpha_{1}+\delta_{11})\gamma_{2}-\alpha_{2}\gamma_{1}-\alpha_{3}\gamma_{0})]\mathbf{k}\\
&+\sigma_{2}u_{22}v_{21}+\sigma_{3}u_{23}v_{31}+\sigma_{4}u_{24}v_{41},
\end{split}
\end{equation*}
and the other reconstructed elements $q_{ij}$($i\in \{2,3\},j\in \{1,2,3,4\}$) are also the same. We find that the modification $\Delta_{1}$ in coefficient of the imaginary unit $\mathbf{j}$ is always being subtracted, which causes the image layer G to be modified more than the other two image layers R and B. Adaptive embedding avoids this situation and obtains a better PSNR as shown in Fig. \ref{PSNR} of Experiment \ref{sec5}.2. Finally, we remark that the proposed method can achieve better performance by fully exploiting the local correlation.

\section{A Novel Color Image Watermarking Scheme based on QSVD}\label{sec4}

\subsection{Embedding and extracting procedure}

Suppose the original host color image represented by the pure quaternion matrix $Q \in \mathbb{H}^{m \times n}$, and the original binary watermark image $W\in \mathbb{R}^{e\times f}$.
\begin{algorithm}[Watermark embedding process]
\qquad

{\rm  Step 1. %Obtaining the watermark sequence.
Converting the original binary watermark image into a vector sequence for indexing and embedding. The watermark sequence $[w_{1},w_{2},\cdots, w_{k}]$ are obtained and $w_{k} \in \{1, 0\}$.

Step 2. %Obtaining the embedding block.
The host image is divided into non-overlapping image blocks $Q_{k}\in \mathbb{H}^{4\times 4}$ in a holistic way. In order to improve the watermarking robustness against cropping attack of the proposed method, the image blocks are selected by the random coordinate sequence based on the private key {\tt Ka}.

Step 3. Performing Algorithm \ref{al3} on each block $Q_{k}$ as Eq.~\eqref{eq4} to obtain the matrices $U_{k}, V_{k}$ and $\Sigma_{k}$,
\begin{equation}\label{eq4}
[U_{k},\Sigma_{k},V_{k}]={\tt QSVD}(Q_{k}).
\end{equation}

Step 4. Changing the relationship of the optimal coefficient pair between $(u_{21})_{\mathbf{i/j/k}}$ and $(u_{31})_{\mathbf{i/j/k}}$ according to the watermark information. If the embedded binary watermark bit is 1, the value of $(u_{21}-u_{31})_{\mathbf{i/j/k}}$ should be negative and its magnitude is greater than $T$. If the embedded binary watermark bit is 0, the value of $(u_{21}-u_{31})_{\mathbf{i/j/k}}$ should be positive and its magnitude is greater than a threshold $T$. When these two conditions are violated, the elements of $(u_{21})_{\mathbf{i,j,k}}$ and $(u_{31})_{\mathbf{i,j,k}}$ should be modified as $(u^{\star}_{21})_{\mathbf{i,j,k}}$ and $(u^{\star}_{31})_{\mathbf{i,j,k}}$, respectively, based on the following rules in Eqs.\eqref{eq5}-\eqref{eq6} to embed the watermark $w_{k}$ and obtain the modified $U_{k}^{\star}$,
\begin{equation}\label{eq5}
 {\rm if}~ w_{k}=1,
   \begin{cases}
   (u^{\star}_{21})_{\mathbf{i/j/k}}={\rm sign}((u_{21})_{\mathbf{i/j/k}})\times(u_{avg}+\frac{T}{2})\\
   (u^{\star}_{31})_{\mathbf{i/j/k}}={\rm sign}((u_{31})_{\mathbf{i/j/k}})\times(u_{avg}-\frac{T}{2})
   \end{cases}\\
\end{equation}
\begin{equation}\label{eq6}
 {\rm if}~ w_{k}=0,
   \begin{cases}
   (u^{\star}_{21})_{\mathbf{i/j/k}}={\rm sign}((u_{21})_{\mathbf{i/j/k}})\times(u_{avg}-\frac{T}{2})\\
   (u^{\star}_{31})_{\mathbf{i/j/k}}={\rm sign}((u_{31})_{\mathbf{i/j/k}})\times(u_{avg}+\frac{T}{2})
   \end{cases}\\
\end{equation}
where $w_{k}$ is the watermark bit, $(x)_{\mathbf{i/j/k}}$ denotes an coefficient of the imaginary part $\mathbf{i}$ or $\mathbf{j}$ or $\mathbf{k}$ of the quaternion $x$, sign($x$) is the sign of $x$, $abs(x)$ is the absolute value of $x$ and $u_{avg}=(abs((u_{21})_{\mathbf{i/j/k}})+abs((u_{31})_{\mathbf{i/j/k}}))/2$.

Step 5. Obtaining the watermarked image block by
\begin{equation}\label{eq7}
Q^{\star}_{k}=U^{\star}_{k}\Sigma_{k}V_{k}.
\end{equation}

Step 6. Repeating Steps 3-5 with the private key {\tt Ka} until all watermark information is embedded in the host image.

Step 7. Recombine all $4\times4$ watermarked blocks $Q^{\star}_{k}$ to obtain the watermarked image $Q^{\star}$.

The simplified embedding mechanisms can be summarized in Fig.~\ref{liuchengtu}.
}
\end{algorithm}
Our method is a blind watermarking algorithm, that is, the watermark information can be extracted without the original host image and the original watermark information. The detailed steps of the watermark extraction are explained as follows.

\begin{figure}
\centering
\includegraphics[width=8cm,height=5cm]{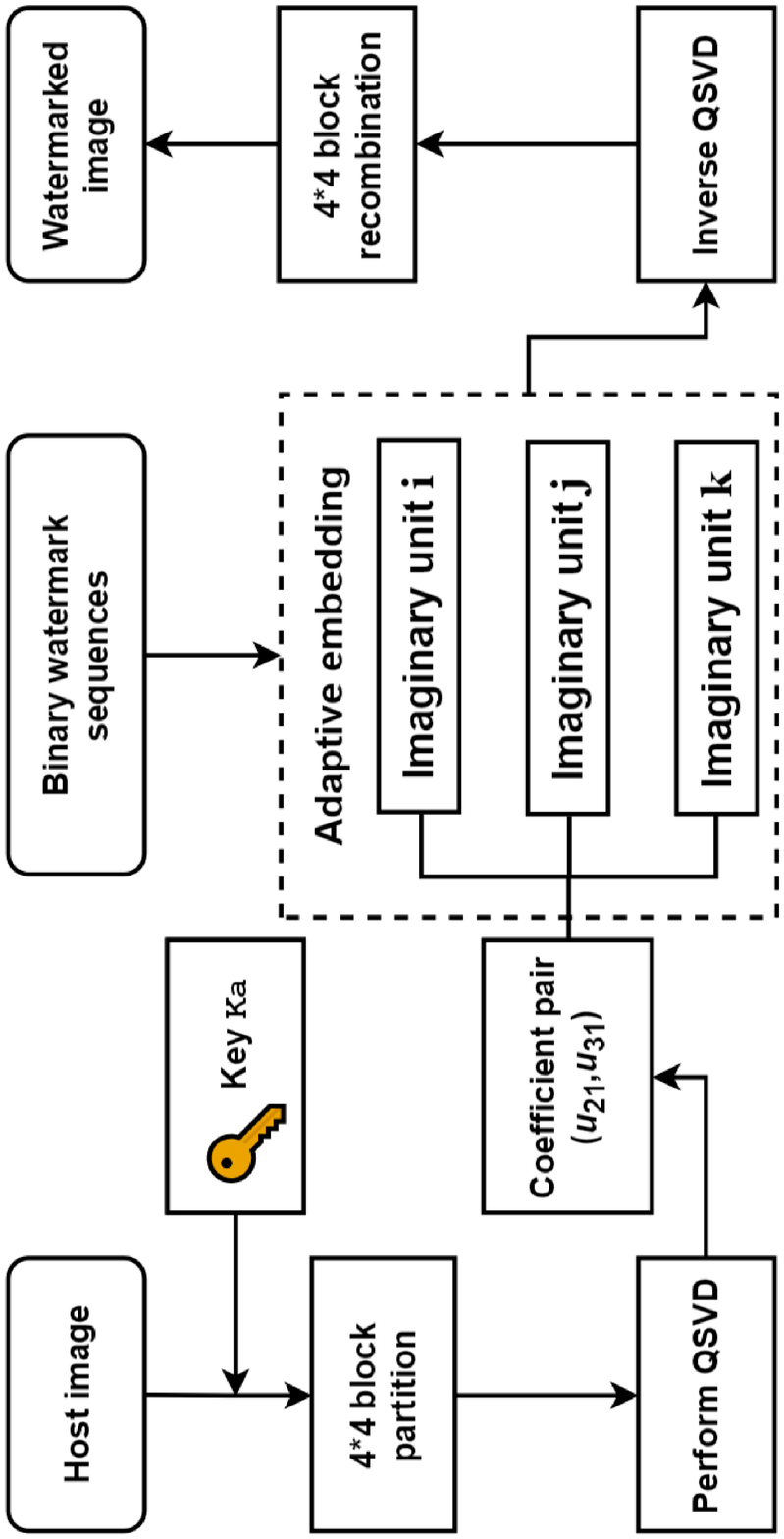}
\caption{Embedding mechanism of the proposed method.}\label{liuchengtu}
\end{figure}

\begin{algorithm}[Watermark extraction process]
\qquad

{\rm Step 1. Obtaining the watermarked image. The watermarked image $Q^{\star}=Q^{\star}_{0}+Q^{\star}_{1}\mathbf{i}+Q^{\star}_{2}\mathbf{j}+Q^{\star}_{3}\mathbf{k}$ is not a pure quaternion matrix. We set the real part $Q^{\star}_{0}$ of $Q^{\star}$ to the same size zero matrix, then assume $Q^{\star}=Q^{\star}_{1}\mathbf{i}+Q^{\star}_{2}\mathbf{j}+Q^{\star}_{3}\mathbf{k}$ as a suspicious image with embedded watermark information.

Step 2. Obtaining the watermarked image block. The watermarked image $Q^{\star}$ is partitioned into non-overlapping watermarked blocks with size of $4\times 4$ pixels. Then, using the random coordinate sequence based on the private key {\tt Ka} to select the watermarked block $Q^{\star}_{k}$.

Step 3. Applying Algorithm \ref{al3} to the watermarked block $Q^{\star}_{k}$ and obtaining the matrices $U^{\star}_{k}, \Sigma^{\star}_{k}, V^{\star}_{k}$.

Step 4. Using Eqs.~\eqref{eq8}-\eqref{eq10} to extract the watermark information $w_{k}^{\star}$

\begin{equation}\label{eq8}
\begin{cases}\begin{smallmatrix}
{\rm diff_{\mathbf{i}}}=(u^{\star}_{21}-u^{\star}_{31})_{\mathbf{i}},  \\
{\rm diff_{\mathbf{j}}}=(u^{\star}_{21}-u^{\star}_{31})_{\mathbf{j}},  \\
{\rm diff_{\mathbf{k}}}=(u^{\star}_{21}-u^{\star}_{31})_{\mathbf{k}},
\end{smallmatrix}\end{cases}
\end{equation}
\begin{equation}\label{eq9}
{\rm Diff_{min}=min(abs(diff_{\mathbf{i}}), abs(diff_{\mathbf{j}}), abs(diff_{\mathbf{k}})),}
\end{equation}
\begin{equation}\label{eq10}
w_{k}^{\star}=
\begin{cases}
1, \quad {\rm if~ Diff_{min}} \geq 0, \\
0, \quad {\rm if~ Diff_{min}} <0.
\end{cases}
\end{equation}

Step 5. Repeating Steps 3-4 by the private key {\tt Ka} until all watermarked blocks have been performed. Extracted binary sequence $w^{\star}$ is reconverted to the matrix of original size.

Step 6. Reconstructing the final extracted watermark $W^{\star}$ from the matrix of extracted watermark information.
}
\end{algorithm}

\subsection{Evaluation criteria}

Generally, the performance of the watermarking methods is investigated by measuring their invisibility, robustness, capacity, etc. For the imperceptible feature, the traditional method PSNR, as shown in Eq.~\eqref{eq11}, is used to measure the similarity between the original color image $Q \in\mathbb{H}^{M\times N}$ and the watermarked image $Q^{\star}\in \mathbb{H}^{M\times N}$.
\begin{equation}\label{eq11}
{\rm PSNR}=10 \times lg\frac{3\times M\times N\times (1)^{2}}{\parallel Q-Q^{\star}\parallel_{F}^{2}}dB
\end{equation}
where $1$ represents the maximum value of a pixel in a color channel, and $M$, $N$ denote the height and width of an image respectively.
$\parallel Q-Q^{\star}\parallel_{F}^{2}=trace((Q-Q^{\star})^{H}(Q-Q^{\star}))$$=\parallel (Q-Q^{\star})^{R}_{c}\parallel_{F}^{2}.$
In general, a larger PSNR indicates the watermarked image more closely resembles the original host image, which means that the watermark more imperceptible.

In addition, method BER, as denoted in Eq.~\eqref{eq12}, is used to measure the dissimilarity between the extracted watermark and the original watermark. Generally, a lower BER reveals that the extracted watermark resembles the original watermark more closely.

\begin{equation}\label{eq12}
{\rm BER}=\frac{b}{e \times f}
\end{equation}
where $b$ is the number of incorrectly detected bits, $e \times f$ is the size of the original watermark.

\section{Numerical experiments}\label{sec5}
In this section we present two numerical examples. In \textbf{Example \ref{sec5}.1}, we compare Algorithm \ref{al3} with two other latest algorithms: one is the performance of the $\mathscr{H}_{2}$-based algorithm \cite{LWZZ67}, another is the implementation of the $\mathscr{H}_{3}$-based algorithm \cite{LWZZ96}. All these experiments are performed on matrix $Q_0,Q_1,Q_2,Q_3 \in \mathbb{R}^{m\times n}$ with $m \geq n$, obtained using the function \texttt{rand}. Therefore, by probability one, quaternion matrices $Q$ are of full column ranks. In \textbf{Example \ref{sec5}.2}, we use Algorithm \ref{al3} to embed and extract watermark on the color image represented by pure quaternion matrix, and display its evaluation results in comparison with the state-of-the-art methods \cite{FLCZ18} and \cite{LWZZ16} regarding invisibility and robustness. All these computations are performed on an Intel(R) Core(TM) $64\times 64$ Corei7-8750H @ 2.20GHz/8.00GB computer using MATLAB R2017b.

\textbf{\rm \textbf{Example \ref{sec5}.1}:} Given a quaternion matrix $Q=Q_0+Q_{1}\mathbf{i}+Q_{2}\mathbf{j}+Q_{3}\mathbf{k} \in \mathbb{H}^{ak \times bk}$, where $Q_p$, ($p$=0,1,2,3) are all random real matrices, $a$ and $b$ are two given positive integers. $a$, $b$ and $k$ determine the dimension of $Q$.

\begin{figure}[h]
  \centering
    \includegraphics[width=8cm,height=5cm]{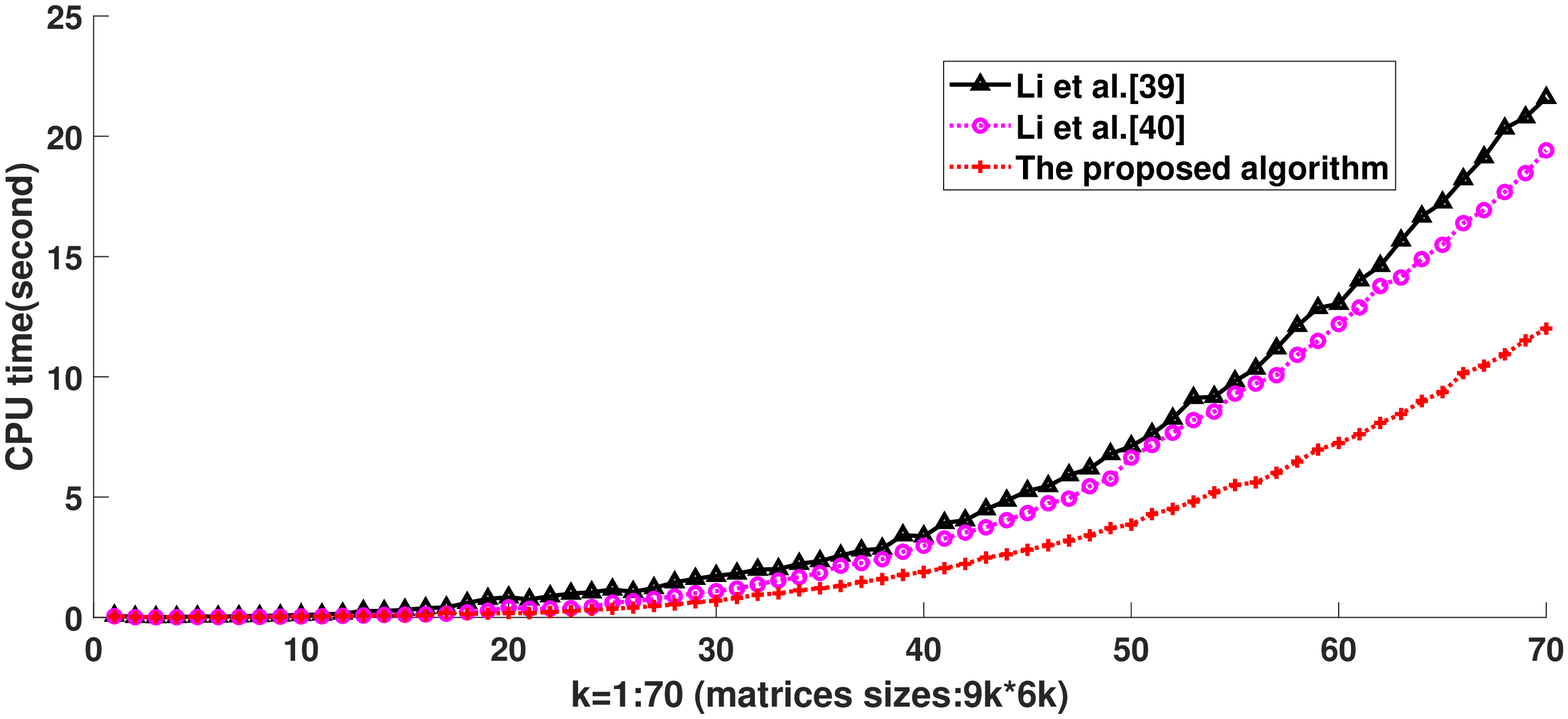}\\
    \includegraphics[width=8cm,height=5cm]{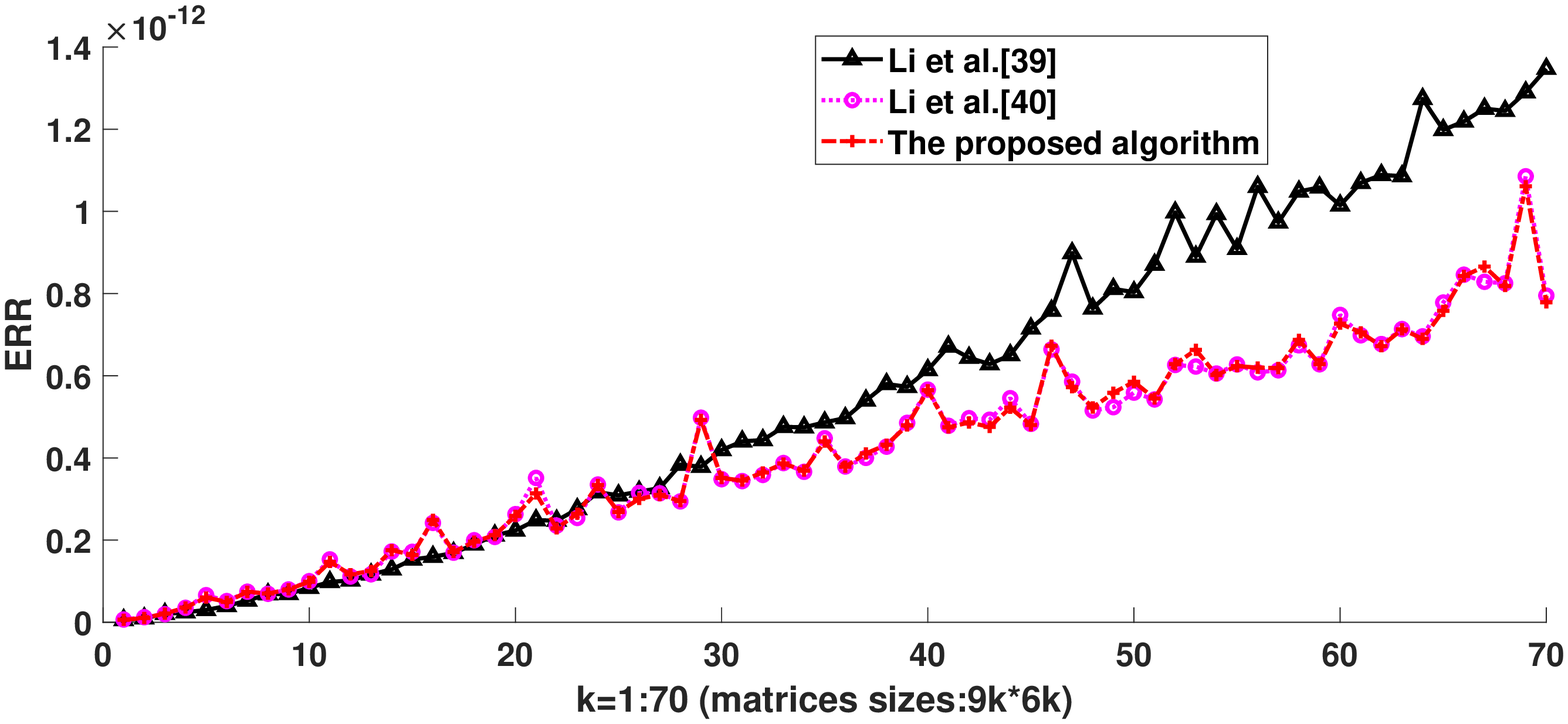}
  \caption{\scriptsize The CPU times (seconds) and the absolute residuals of QSVD with generating $U$ and $V$.}\label{uv}

\end{figure}

For $a=9$, $b=6$, $k=1:70$ and $m=ak, n=bk$, we compare the CPU times and accuracy of three real structure-preserving algorithms of QSVD with generating unitary transformation matrix $U$ and $V$. From Fig.~\ref{uv}, we observe that the CPU time of the proposed algorithm is the smallest, and it costs about half CPU time of that by running the algorithms of \cite{LWZZ67} and \cite{LWZZ96}. From the absolute residual curves of Fig.~\ref{uv}, we find that the three algorithms are very stable, located in a high-precision interval. Among them, the proposed algorithm and the algorithm in \cite{LWZZ96} are more stable than the algorithm in \cite{LWZZ67}. From the above discussion, we see that our real structure-preserving algorithm is efficient and convenient. Notice that in Fig.~\ref{uv}, we do not compare CPU times of one which adopts the function \texttt{svd} in quaternion toolkit, or one that directly uses the built-in function \texttt{svd} on the representation $Q^{R}$ of $Q$, because Li \emph{et al.}\cite{LWZZ67} have already shown that their algorithm is superior to these two methods.

\begin{figure*}
  \centering
  \includegraphics[width=18cm, height=48mm]{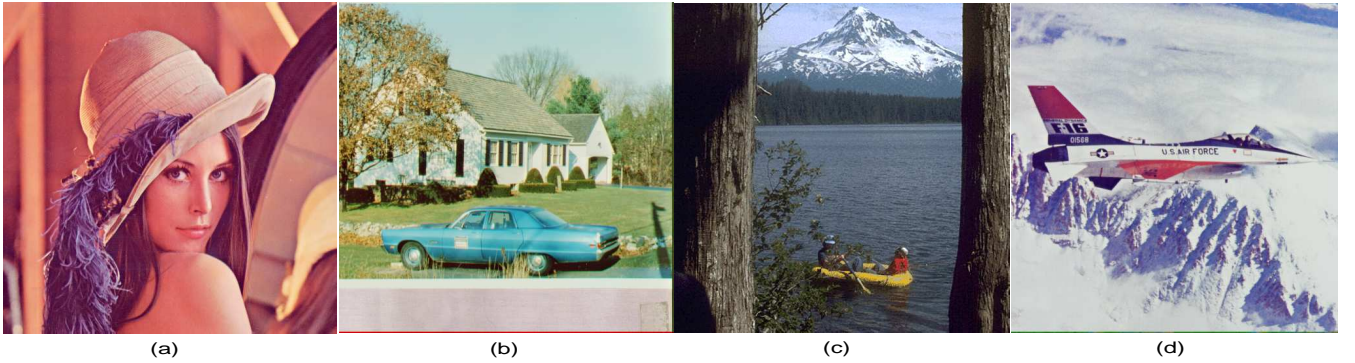}
  \caption{The original host image: \textbf{(a)} Lena, \textbf{(b)} House, \textbf{(c)} Lostlake,\textbf{(d)} F-16.}\label{Hostimage}
\end{figure*}

\begin{figure}
  \centering
  \includegraphics[width=4cm, height=33mm]{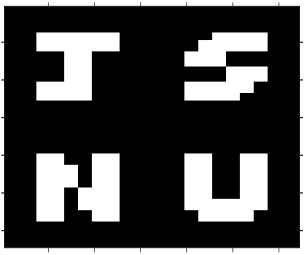}
  \caption{The original watermark image.}\label{watermark}
\end{figure}

\textbf{\rm \textbf{Example \ref{sec5}.2}:} Four standard $512\times 512$ sized color images in the CVG-UGR image database, i.e., Lena, House, Lostlake and F-16 are used as the host images \cite{UoG} as shown in Fig.~\ref{Hostimage}, while a binary Logo of size $64 \times 64$ are used as original watermark as shown in Fig.~\ref{watermark}. The proposed method is compared with the existing state-of-the-art methods \cite{FLCZ18} and \cite{LWZZ16} for evaluating invisibility, robustness, capacity, security, and efficiency based on the same embedding rate of $\frac{1}{64}$ bits per pixel (bpp).

In order to ensure the invisibility of the watermark and the quality of the extracted watermark, the criterion for all the experiments is to first ensure that the PSNR value is greater than 40 dB, and then ensure that the BER value of the extracted watermark is less than 0.3. Therefore, the threshold $T$ of this paper is set in the appropriate interval $[0.002, 0.04]$ to verify the effectiveness of the proposed method.

\textbf{1. The watermark invisibility analysis}.

\begin{figure*}
\centering
\includegraphics[width=8cm,height=5cm]{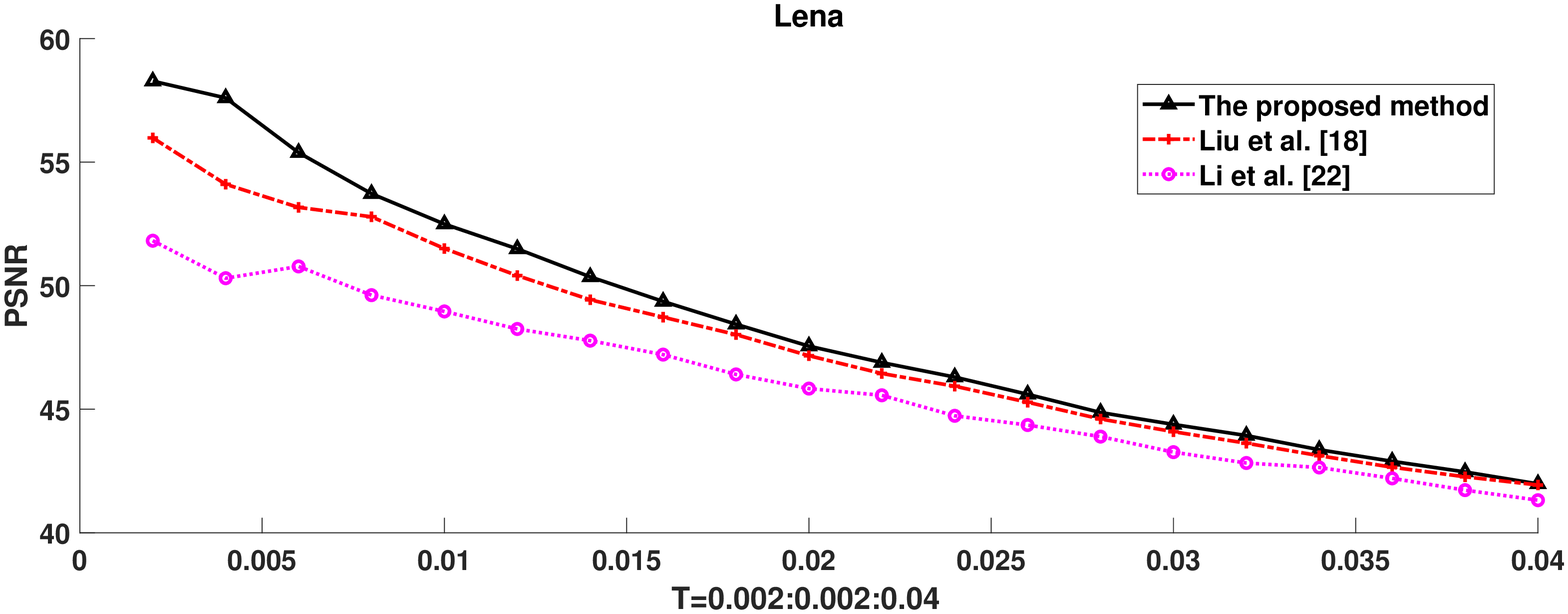}
\includegraphics[width=8cm,height=5cm]{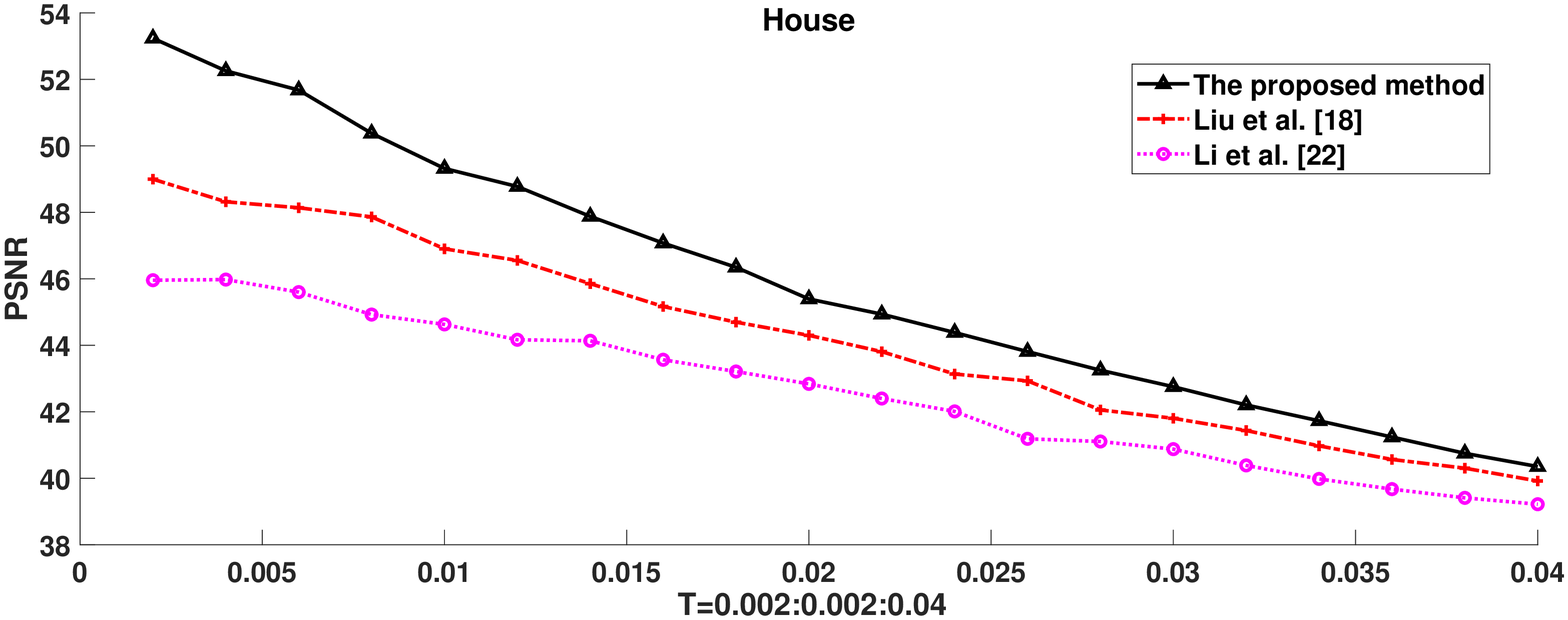}
\includegraphics[width=8cm,height=5cm]{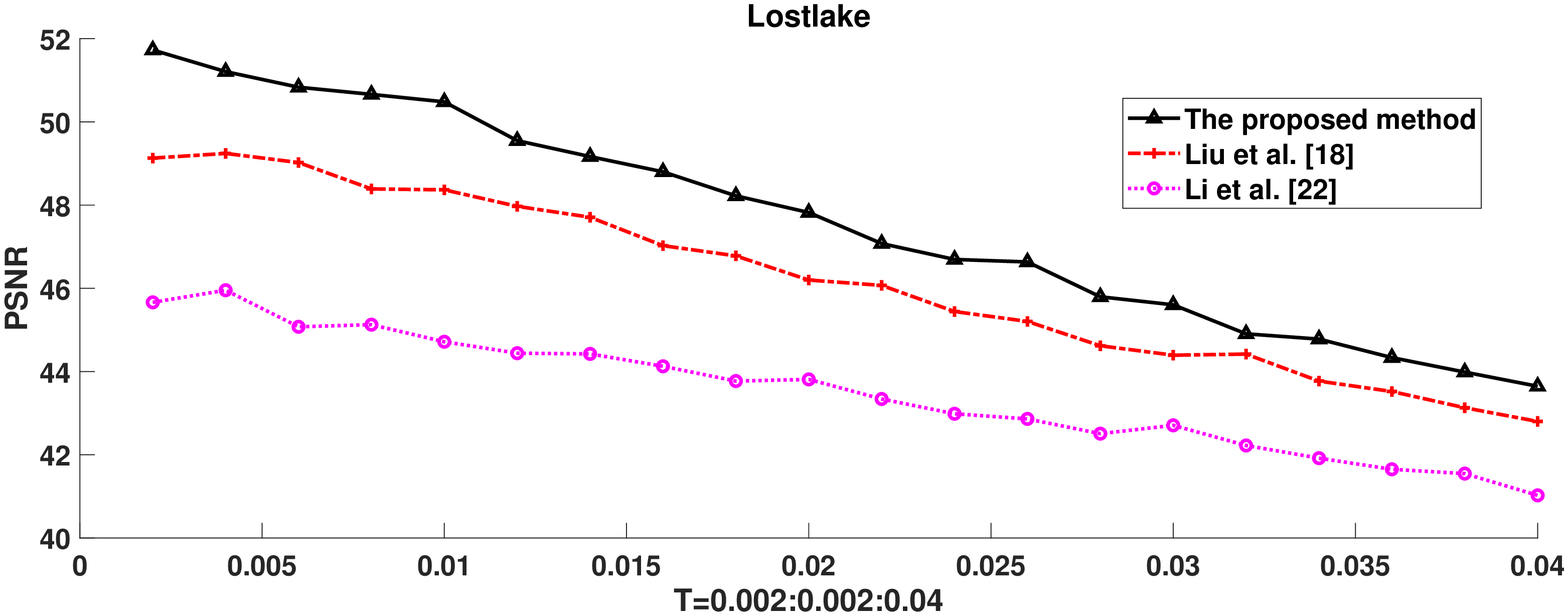}
\includegraphics[width=8cm,height=5cm]{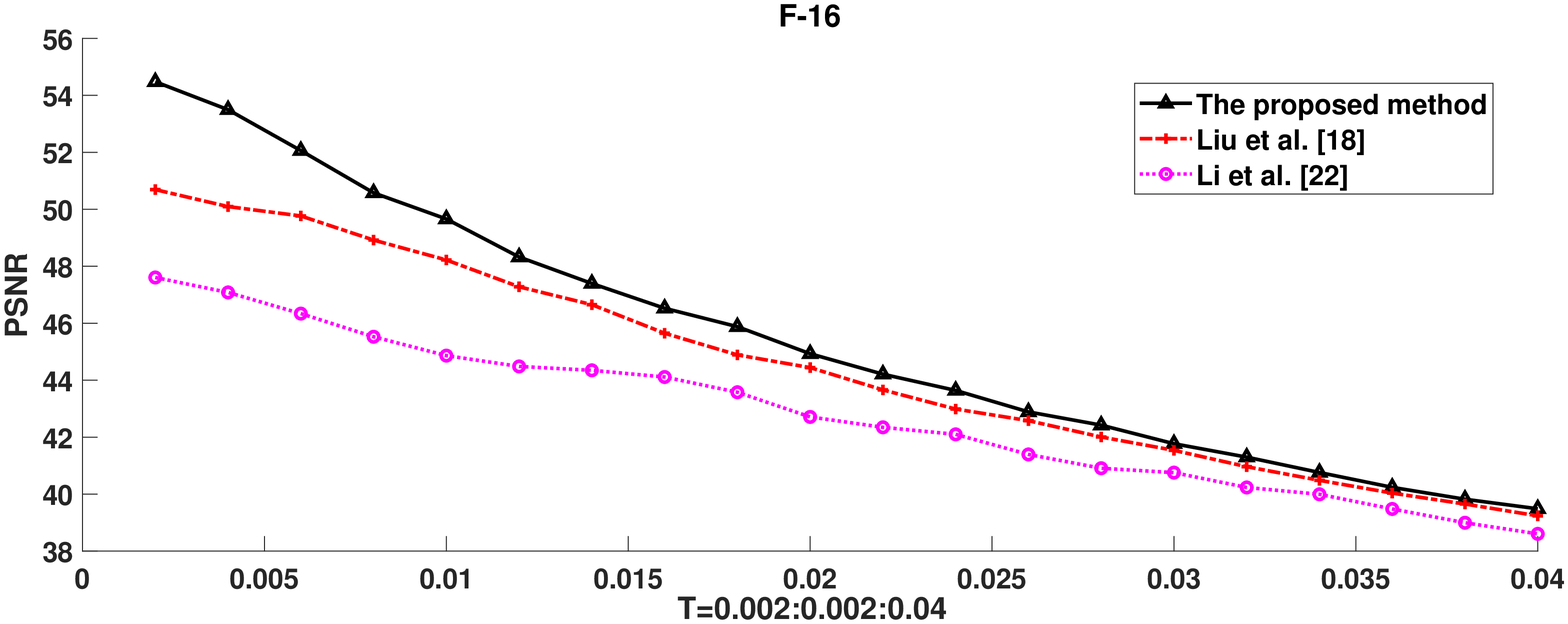}
\caption{Invisibility comparison between our method and two methods of Liu \emph{et al.}\cite{FLCZ18} and Li \emph{et al.}\cite{LWZZ16} }\label{PSNR}
\end{figure*}

In order to evaluate the invisibility of the watermark, we embed the watermark of Fig.~\ref{watermark} in all host images, and then compare it with two recent works of Liu \emph{et al.}\cite{FLCZ18} and Li \emph{et al.}\cite{LWZZ16}. For the three methods, we vary the threshold $T$ from 0.002 to 0.04 with step size 0.002. Fig.~\ref{PSNR} plots their PSNR values and trends, which demonstrates that the better visual quality of watermarked images can be obtained by the proposed method.

\textbf{2. The watermark robustness analysis}

Regarding robustness, different strengths of several different attacks are commonly performed on the watermarked color images. These attacks include: JPEG compression (compression ratio is 20\%, 40\%, 60\%, respectively), motion blur (mask is (4, 4), (6, 6), (9, 9), respectively), cropping (ratio is 10\%, 30\%, 50\%, respectively), scaling (ratio is 0.5, 2, 4, respectively), speckle noise (variance is 0.05), salt \& pepper noise (density is 0.05).

\begin{figure*}[!htbp]
\centering
\subfigure[JPEG40]{
\includegraphics[width=8cm,height=5cm]{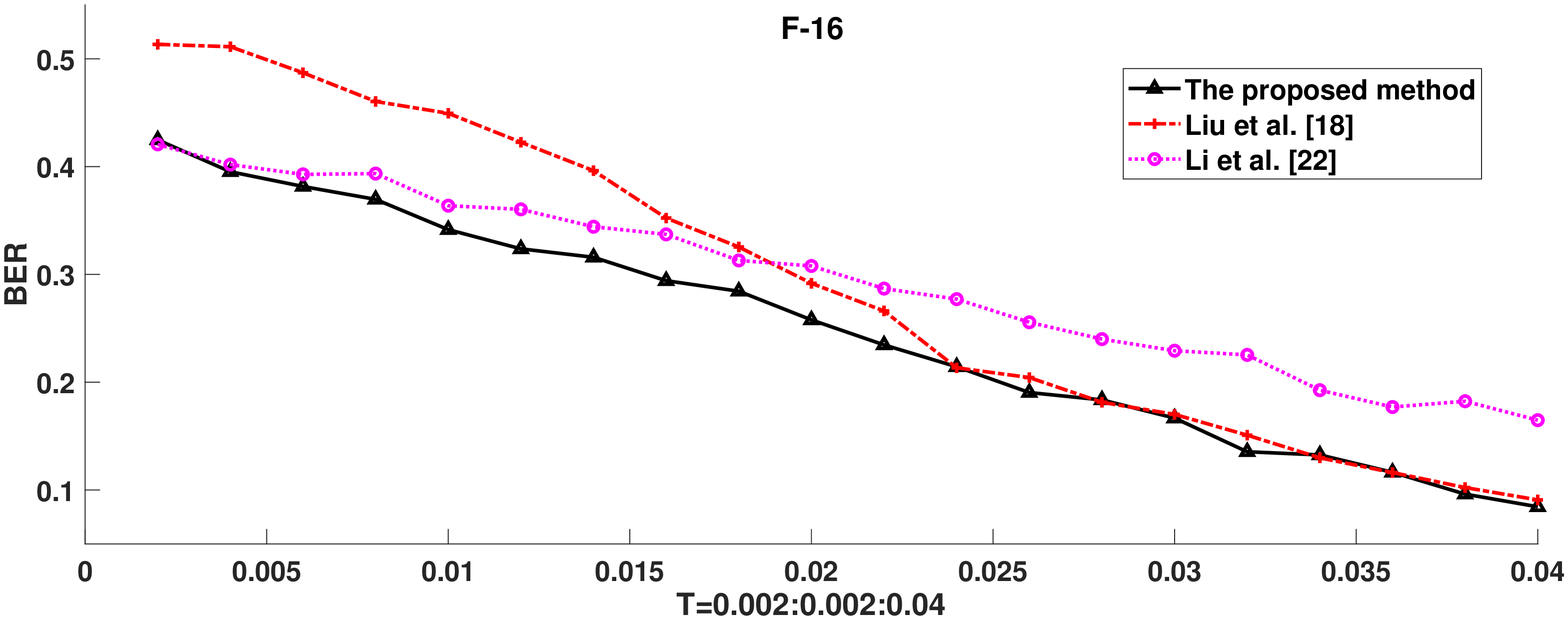}
}
\quad
\subfigure[JPEG60]{
\includegraphics[width=8cm,height=5cm]{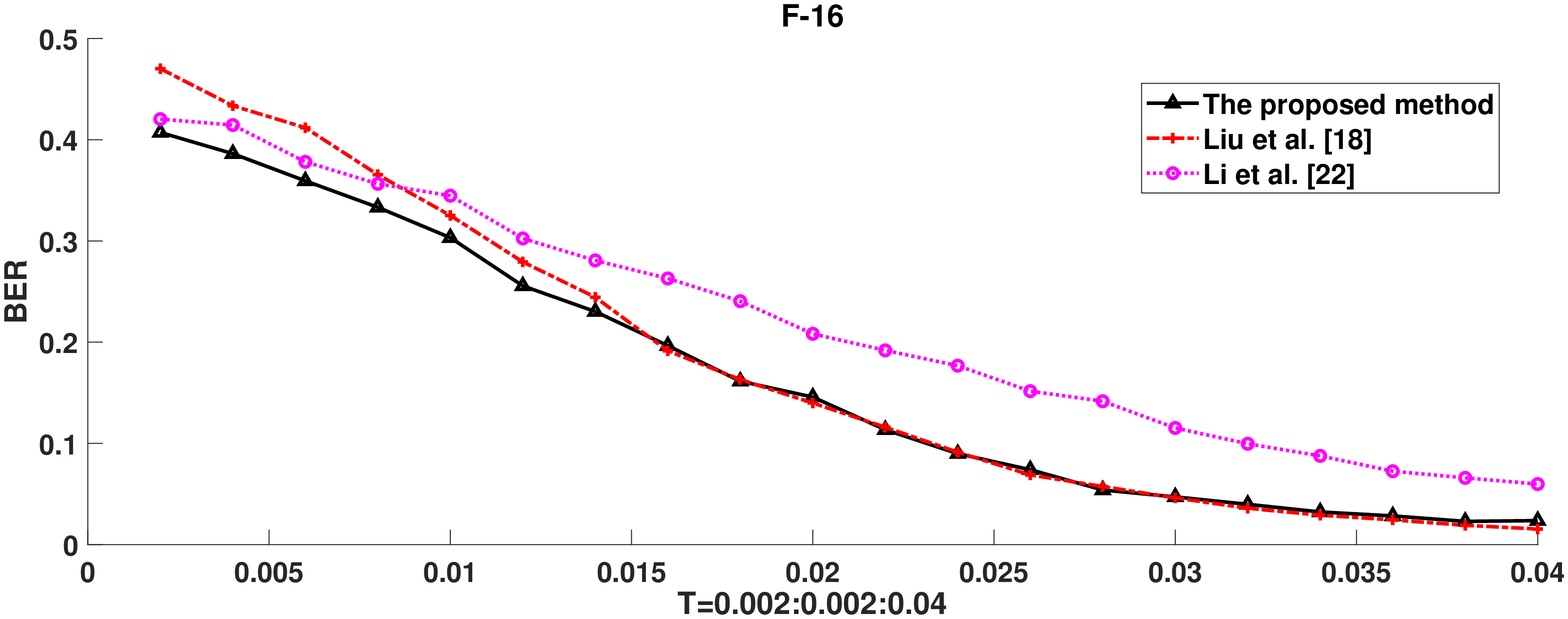}
}
\quad
\subfigure[Motion(4,4)]{
\includegraphics[width=8cm,height=5cm]{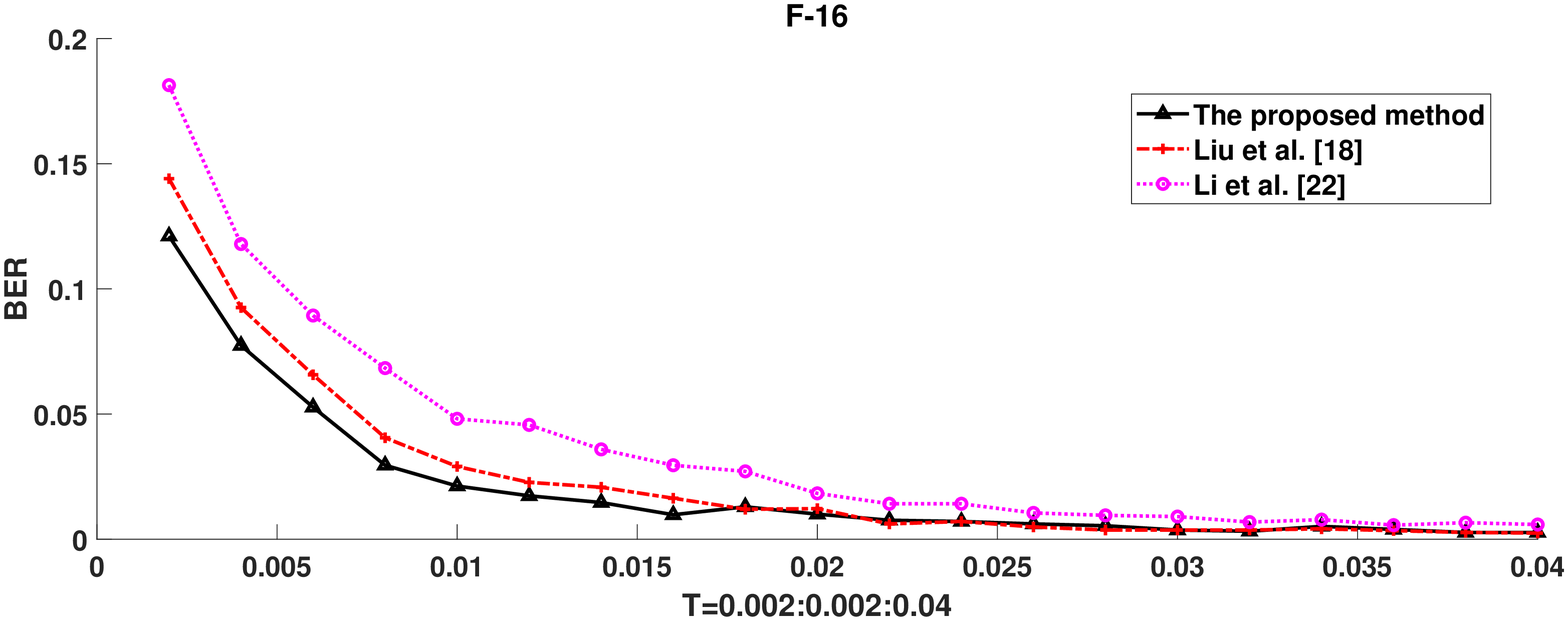}
}
\quad
\subfigure[Motion(9,9)]{
\includegraphics[width=8cm,height=5cm]{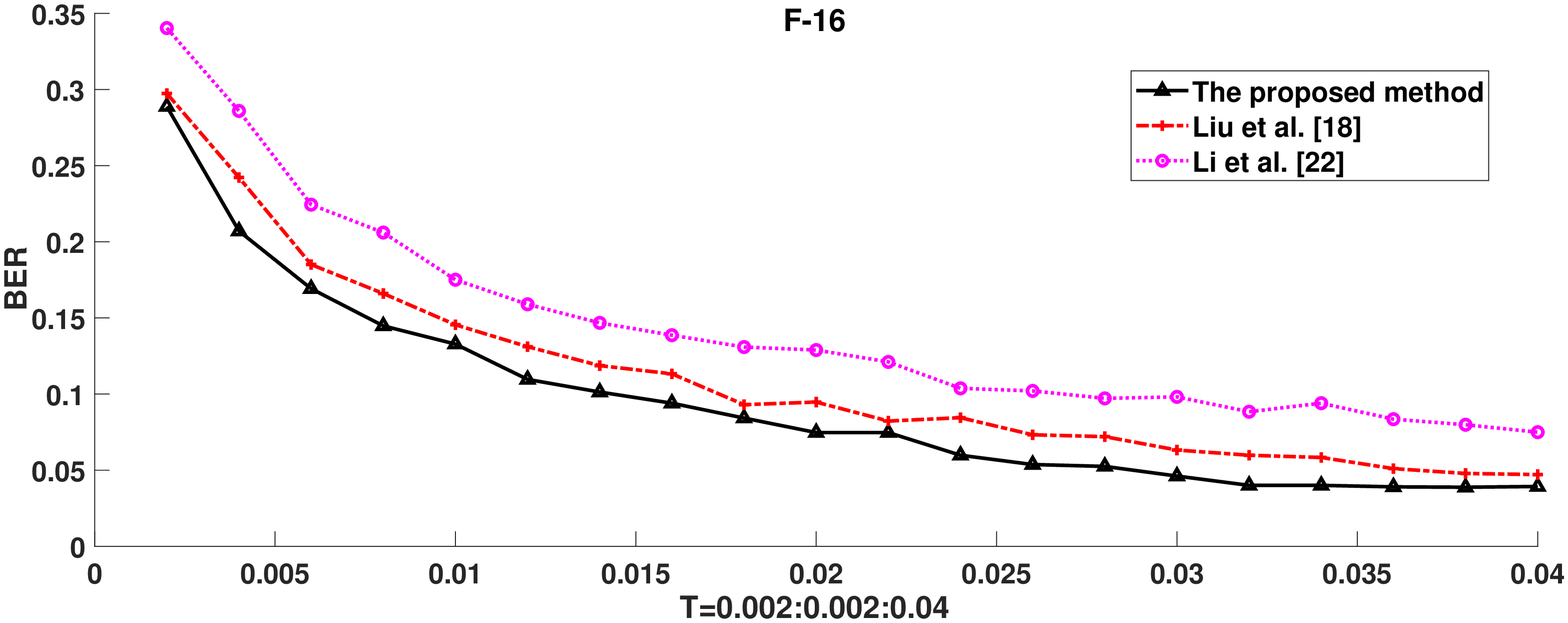}
}
\quad
\subfigure[Scaling0.5]{
\includegraphics[width=8cm,height=5cm]{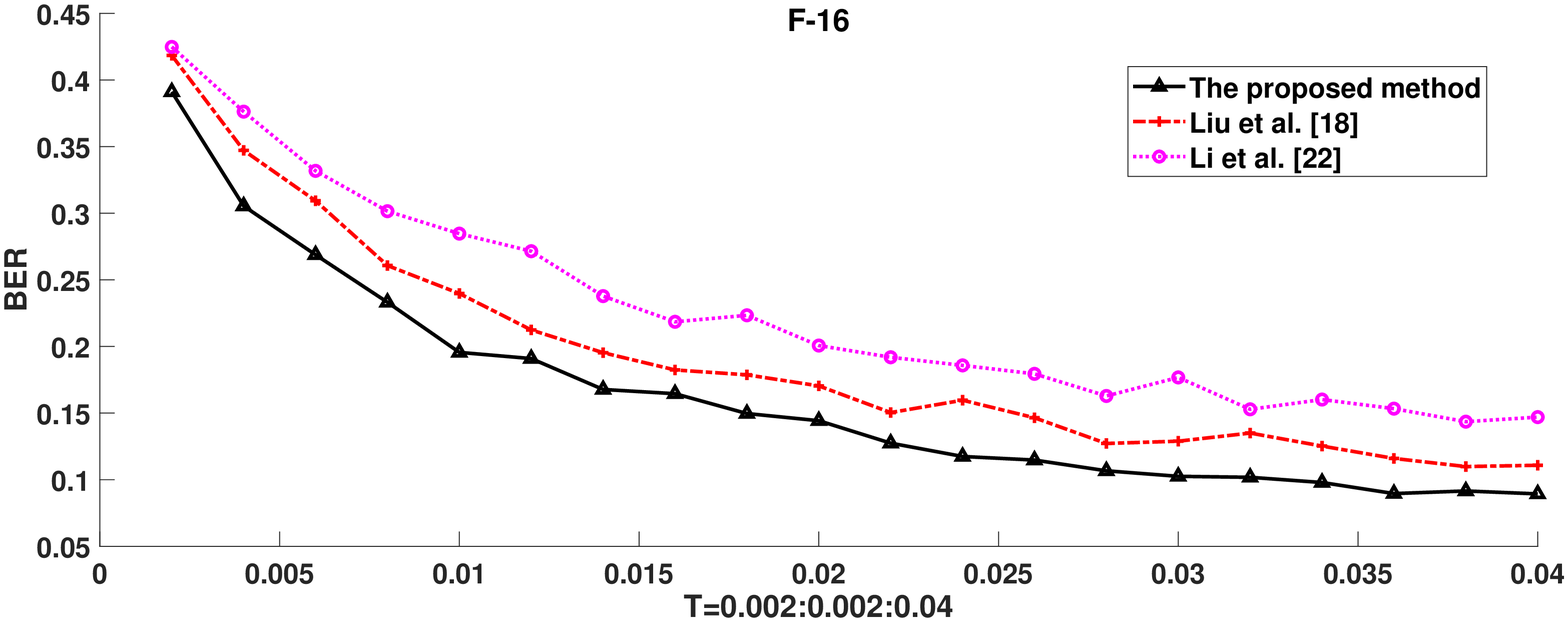}
}
\quad
\subfigure[Scaling2]{
\includegraphics[width=8cm,height=5cm]{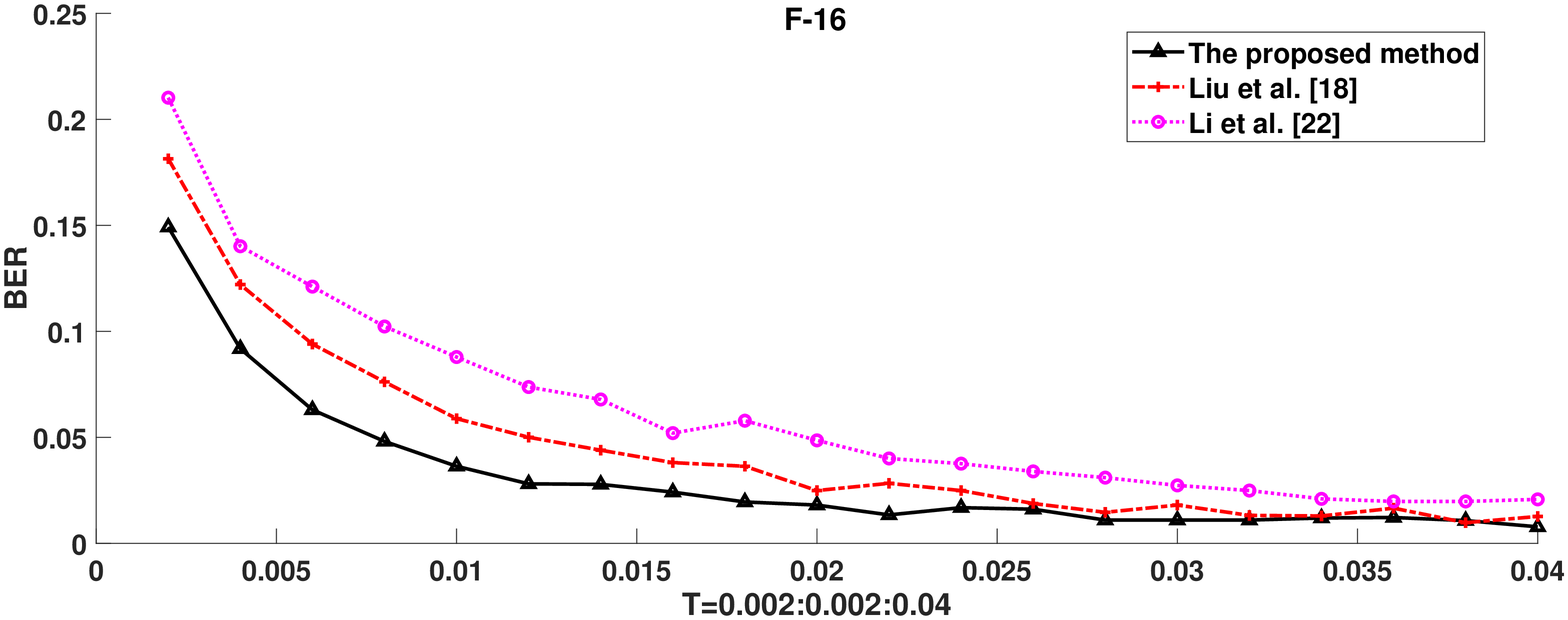}
}
\quad
\subfigure[Cropping10\%]{
\includegraphics[width=8cm,height=5cm]{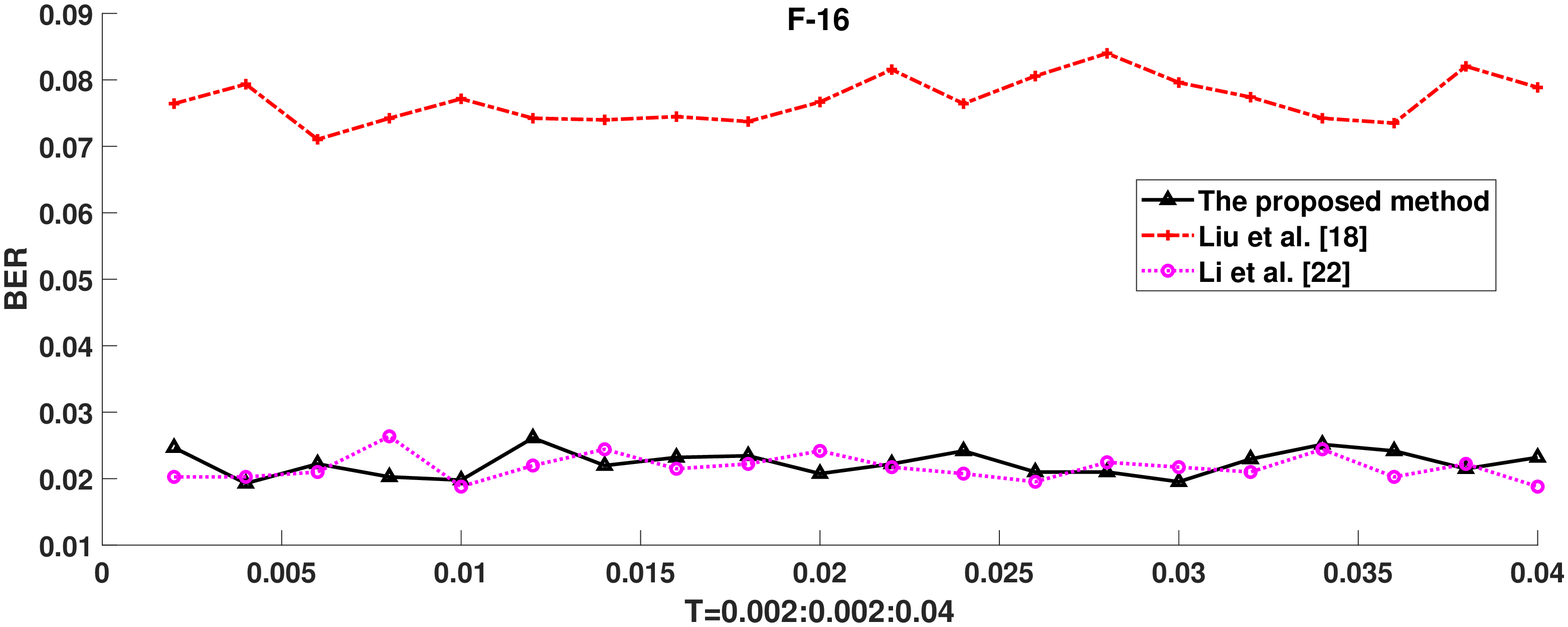}
}
\quad
\subfigure[Cropping50\%]{
\includegraphics[width=8cm,height=5cm]{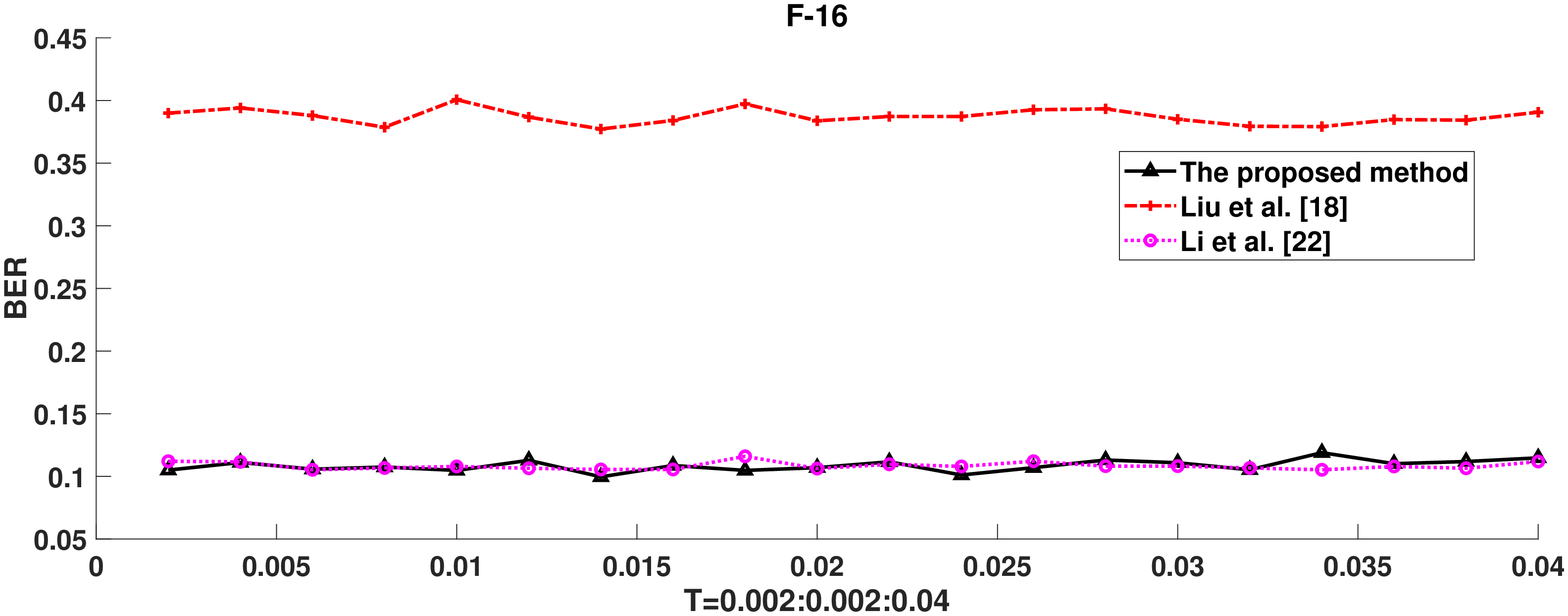}
}
\caption{Robustness comparison between our method and two methods of Liu \emph{et al.}\cite{FLCZ18} and Li \emph{et al.}\cite{LWZZ16} on watermarked image F-16 with different attacks.}\label{FBER_1}
\end{figure*}

\begin{figure*}[!htbp]
\centering
\subfigure[JPEG 20]{
\includegraphics[width=8cm,height=5cm]{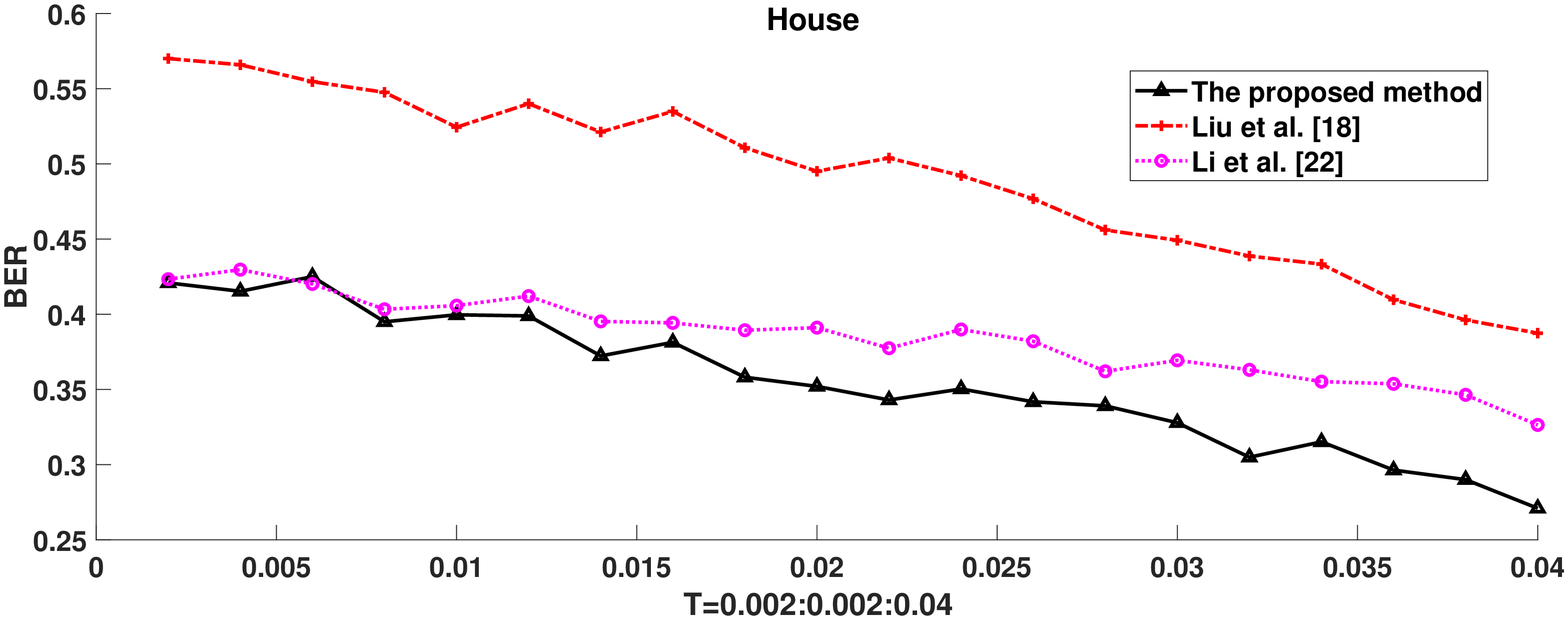}
}
\quad
\subfigure[Motion (6,6)]{
\includegraphics[width=8cm,height=5cm]{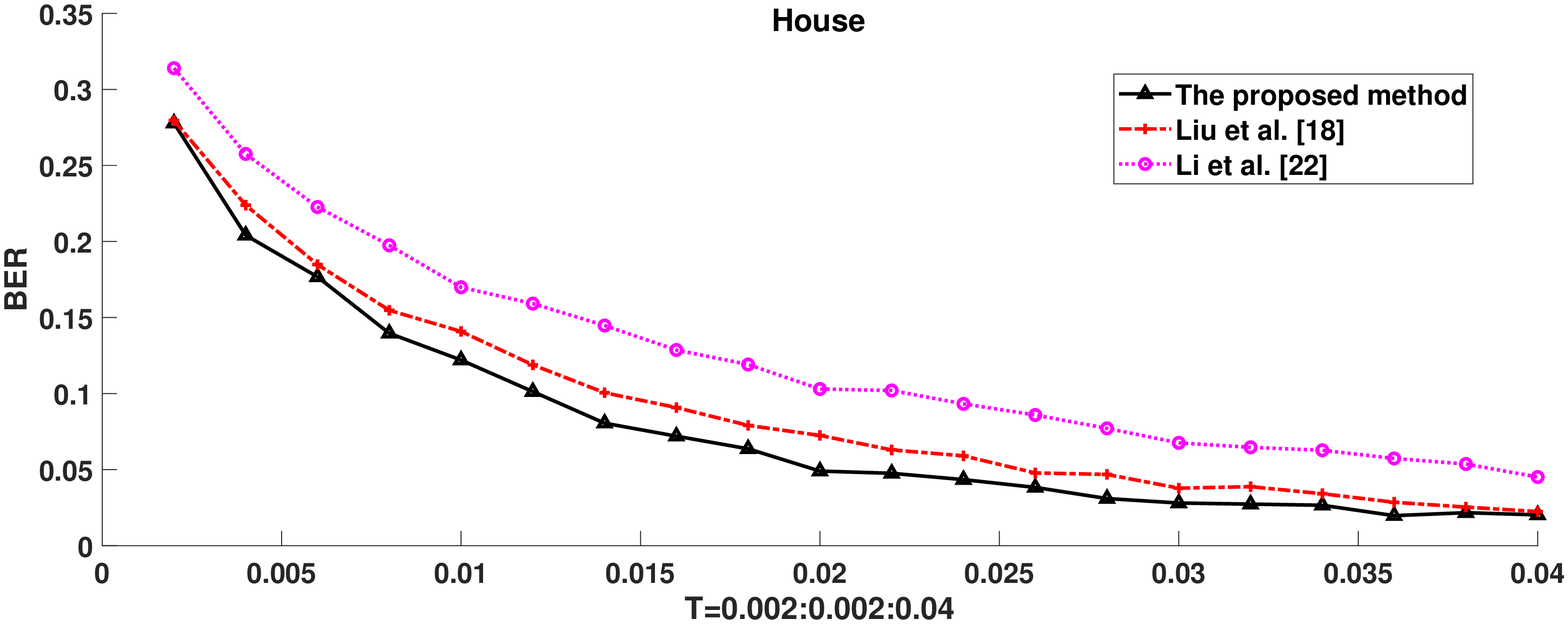}
}
\quad
\subfigure[Scaling 4]{
\includegraphics[width=8cm,height=5cm]{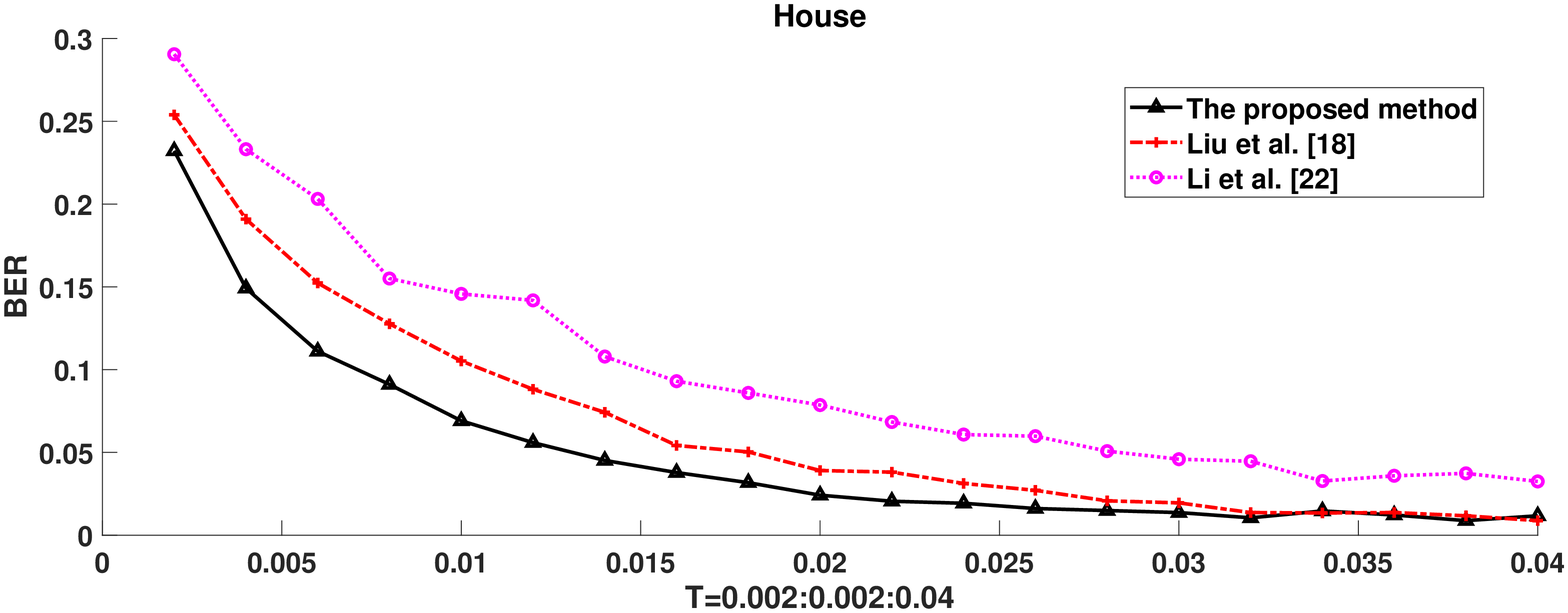}
}
\quad
\subfigure[Cropping 30\%]{
\includegraphics[width=8cm,height=5cm]{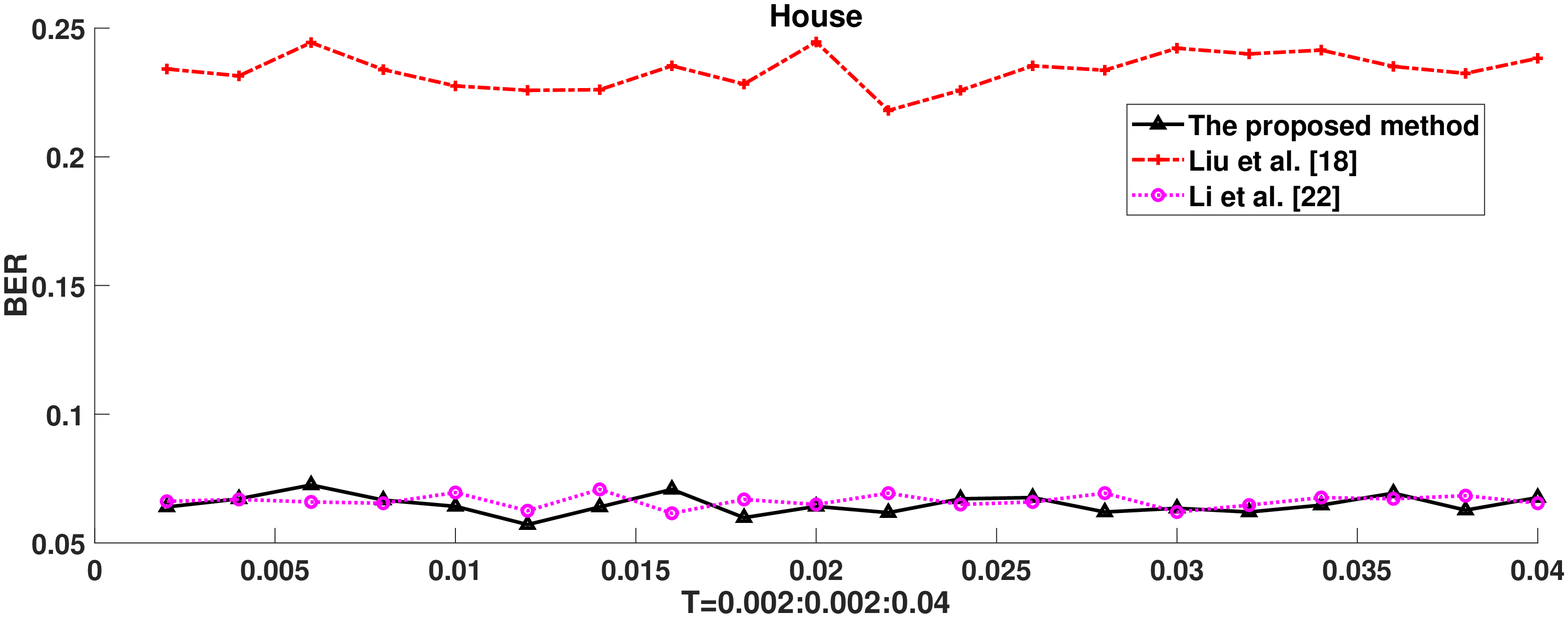}
}
\quad
\subfigure[Speckle 0.05]{
\includegraphics[width=8cm,height=5cm]{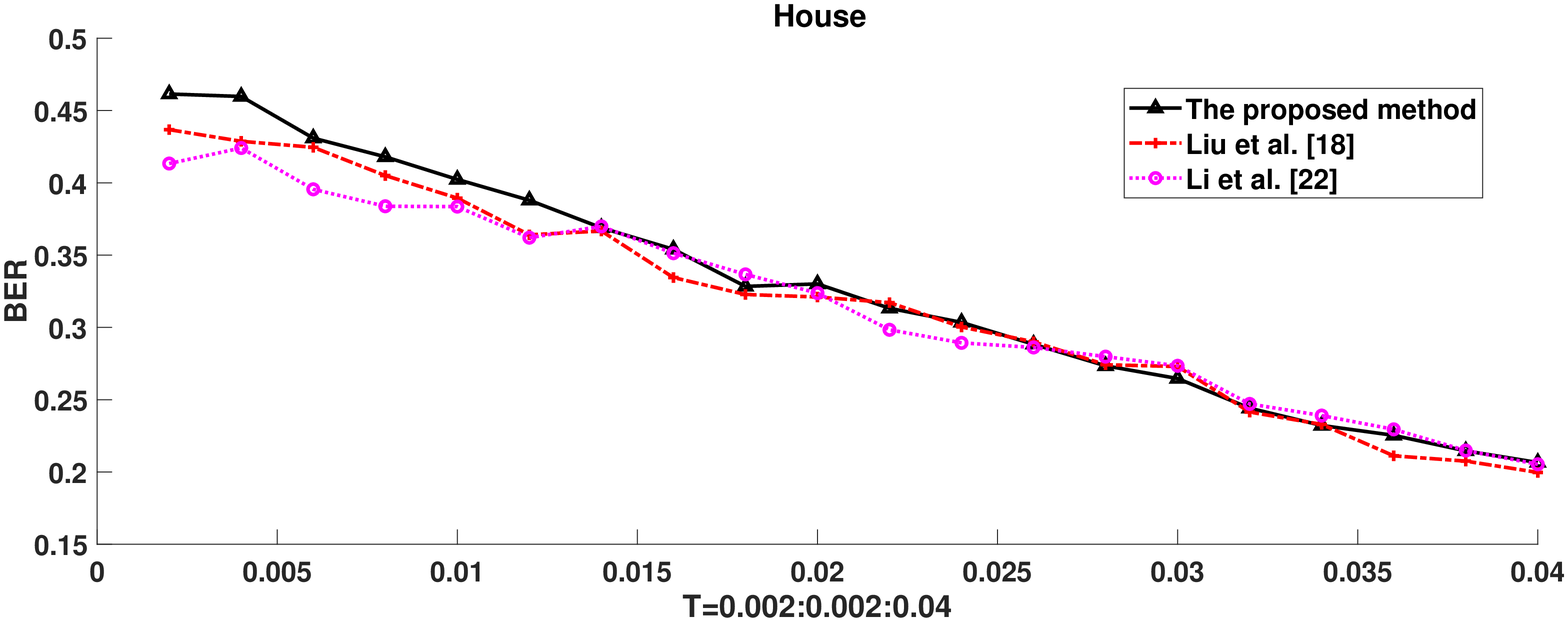}
}
\quad
\subfigure[Salt \& pepper 0.05]{
\includegraphics[width=8cm,height=5cm]{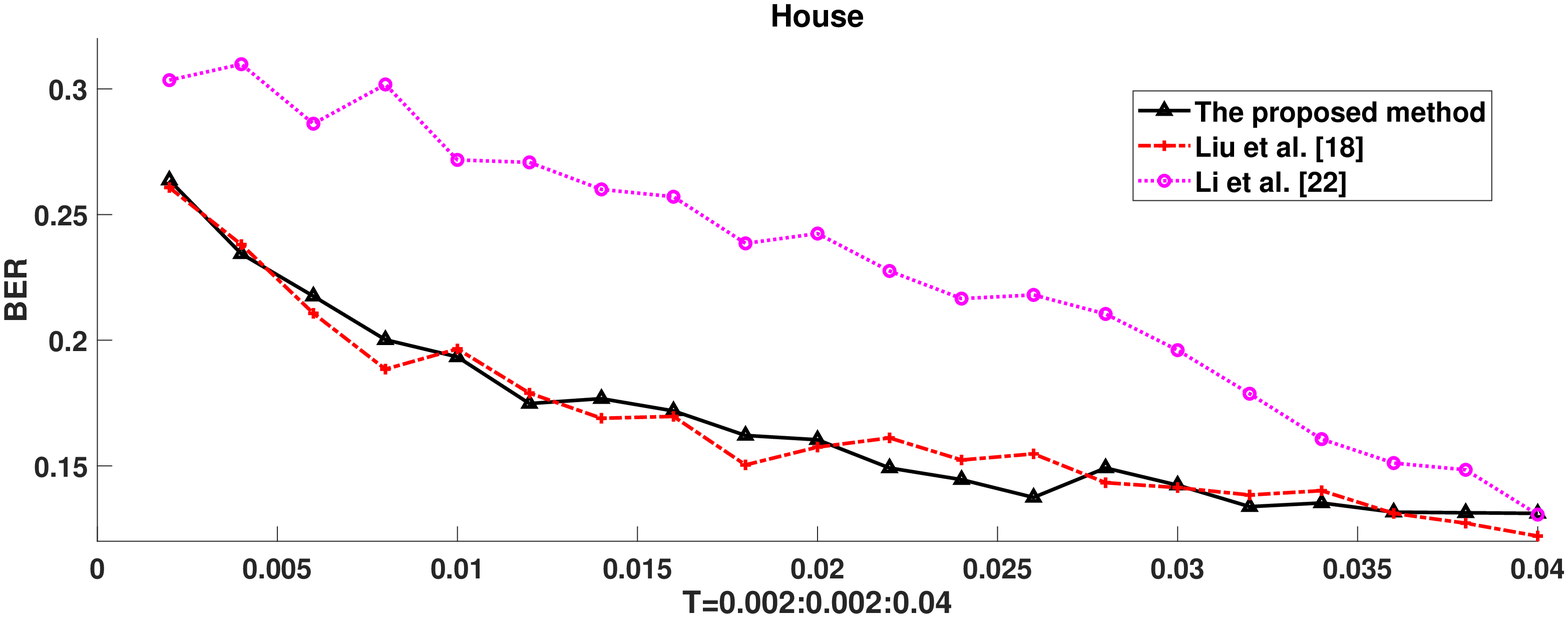}
}
\caption{Robustness comparison between our method and two methods of Liu \emph{et al.}\cite{FLCZ18} and Li \emph{et al.}\cite{LWZZ16} on watermarked image House with different attacks.}\label{FBER_2}
\end{figure*}

The comparison results of the three watermarking schemes are shown in Figs.~\ref{FBER_1}-\ref{FBER_2} and Table \ref{tab3}. Figs.~\ref{FBER_1}-\ref{FBER_2} plots the variation of the BER values of the three methods tested with images F-16 and House within the set threshold interval, where the image House is tested and verified as a different image. Table \ref{tab3} visually shows the watermarks extracted at the threshold $T=0.035$ by the three methods tested with all the images in Fig.~\ref{Hostimage}. From Figs.~\ref{FBER_1}-\ref{FBER_2} and Table \ref{tab3}, the proposed method yields a superior performance than the other two methods. For the JPEG compression case, the BER values of the proposed method are basically the smallest among the three methods within the threshold interval. Especially, when the compression factor is 20\% (Fig.~\ref{FBER_2} (a)), the proposed method exhibits much stronger robustness. For the motion blur and the scaling cases, the proposed method is strictly more robust than the other two methods. For the cropping case, the BER values of the proposed method are comparable with those of Li \emph{et al.}'s method, but are much better than those of Liu \emph{et al.}'s method. These reflect that the proposed QSVD method with the highly correlated coefficient pairs can obtain stronger robustness. For the speckle noise case, the BER values of the three methods are close in the threshold interval $T$. For the salt \& pepper noise case, the BER values of the proposed method are comparable with those of Liu \emph{et al.}'s method, but are much better than those of Li \emph{et al.}'s method. In addition, we can also find that the robustness is gradually enhanced with the increase of threshold $T$. From Fig.~\ref{PSNR}, if the PSNR values of the three methods are the same, the proposed method can choose a larger threshold $T$ to achieve higher robustness. Overall, the proposed QSVD method is more robust than the other two methods.

To further highlight the robustness of the proposed method, we calculate the mean of the BER values obtained by different strengths of several different attacks using the data in Table \ref{tab3}. They are 0.1404 (method \cite{FLCZ18}), 0.1204 (method \cite{LWZZ16}) and 0.0930 (proposed method). From the mean results, we can also conclude that the proposed method is more resistant to different image processing attacks than the other two methods.

\begin{table*}[!htb]
\caption{Comparison of the robustness (BER) between the proposed method and methods \cite{FLCZ18}, \cite{LWZZ16} when the threshold $T=0.035$.}\label{tab3}
\centering

\begin{tabular}{c|c|ccc|ccc}
\toprule  
\multicolumn{2}{c}{\quad Watermarked image} &\multicolumn{3}{c}{\quad F-16  } & \multicolumn{3}{c}{\quad Lena  }\\
\midrule  

\multicolumn{2}{c}{\quad Algorithms} & Method \cite{FLCZ18} & Method \cite{LWZZ16} & The proposed & Method \cite{FLCZ18} & Method \cite{LWZZ16} & The proposed \\
\midrule  
 \multirow{16}*{\tabincell{c}{Attack \\type}}
& JPEG 40 &
\begin{minipage}{0.07\textwidth} \includegraphics[width=12mm, height=10mm]{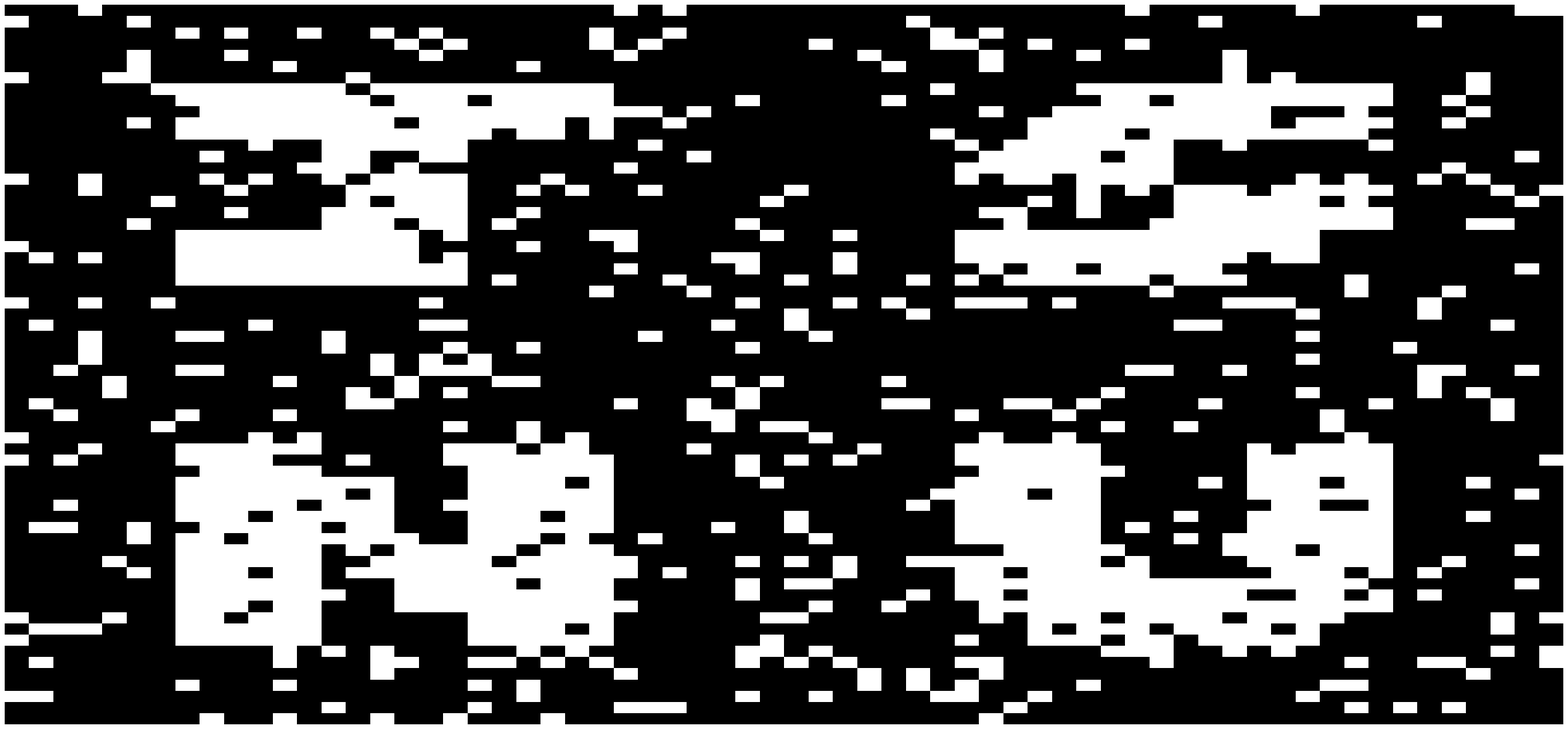} \end{minipage} &
\begin{minipage}{0.07\textwidth} \includegraphics[width=12mm, height=10mm]{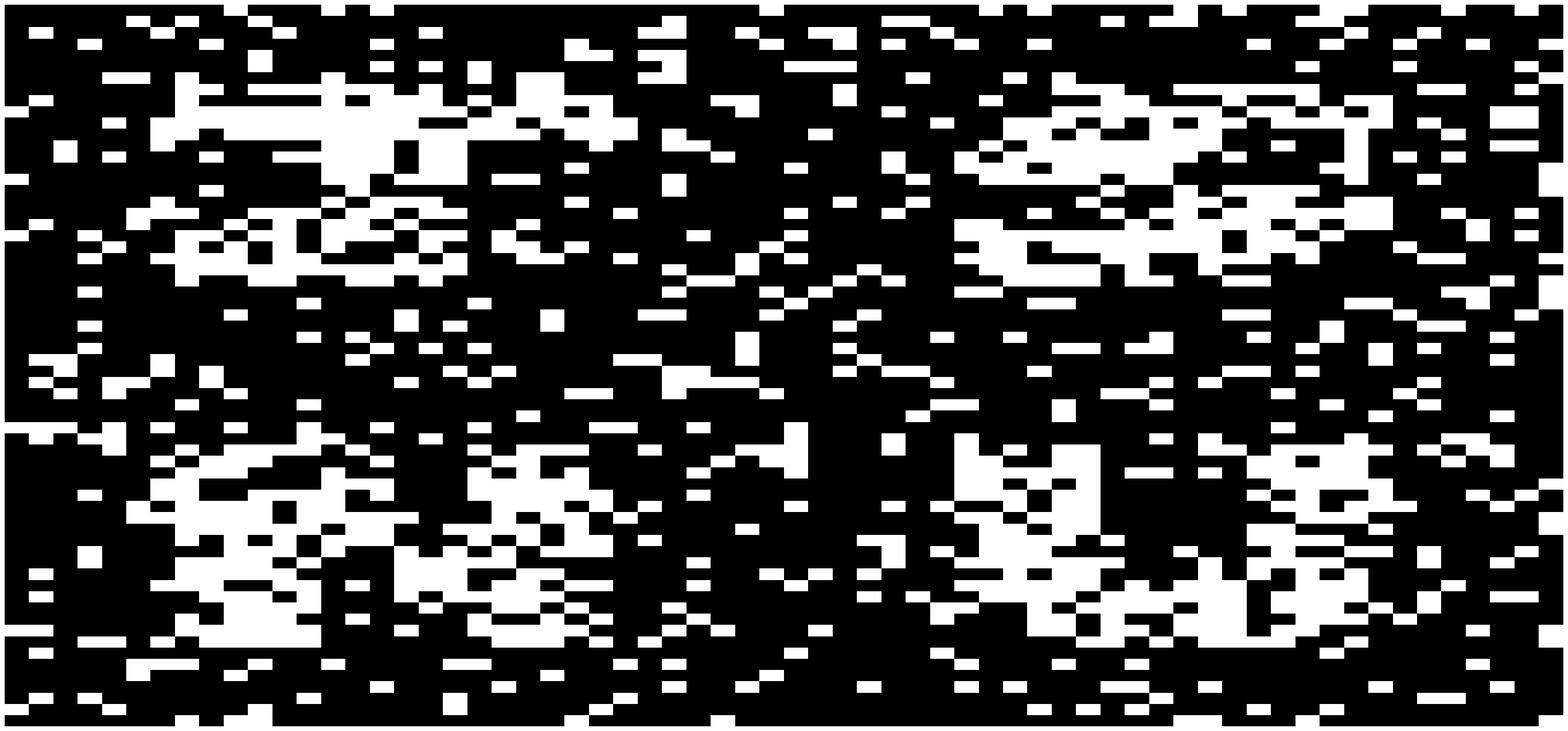} \end{minipage} &
\begin{minipage}{0.07\textwidth} \includegraphics[width=12mm, height=10mm]{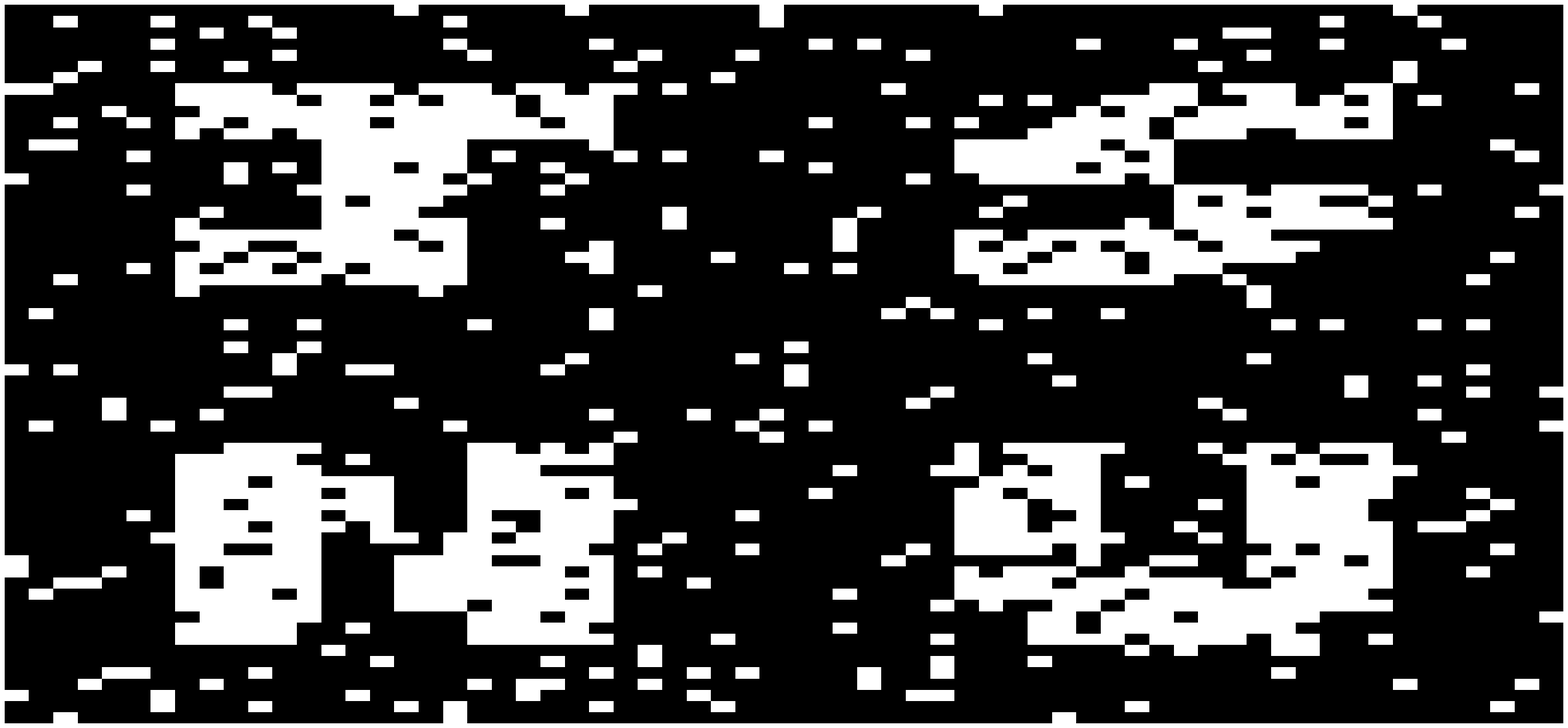} \end{minipage}&
\begin{minipage}{0.07\textwidth} \includegraphics[width=12mm, height=10mm]{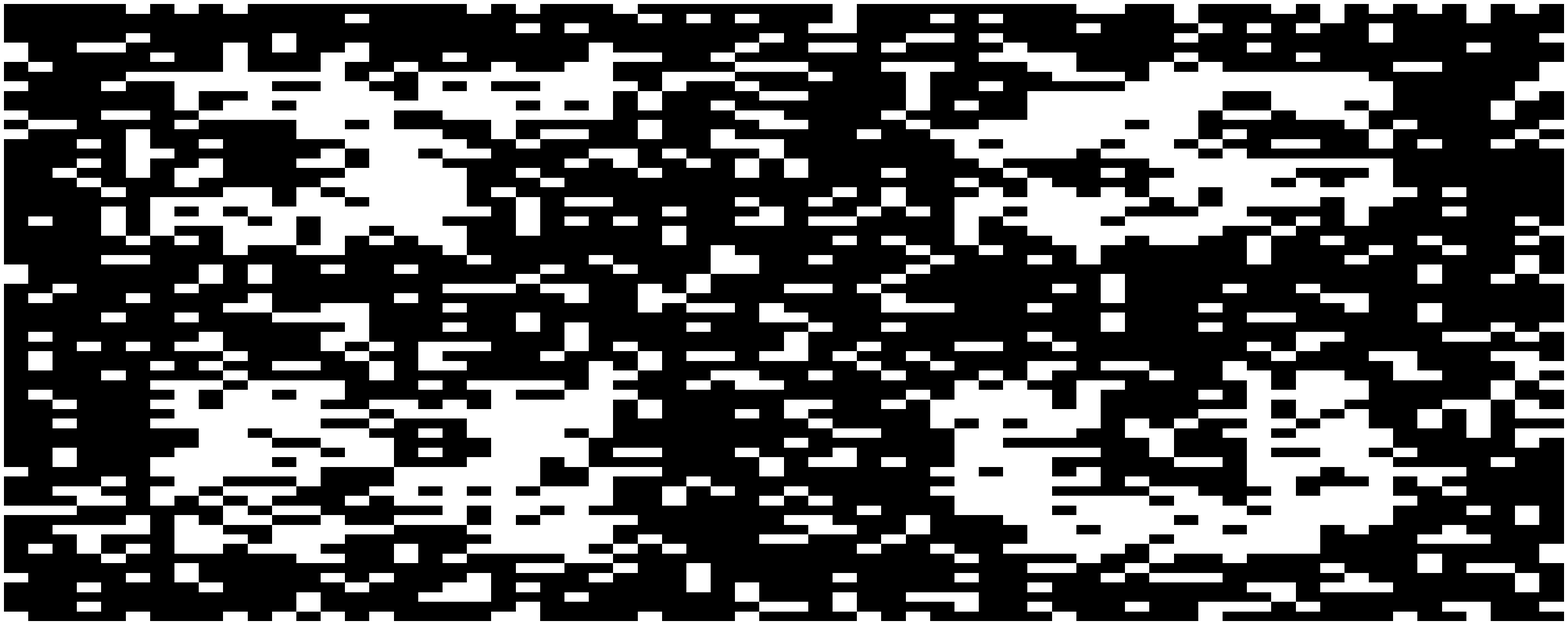} \end{minipage} &
\begin{minipage}{0.07\textwidth} \includegraphics[width=12mm, height=10mm]{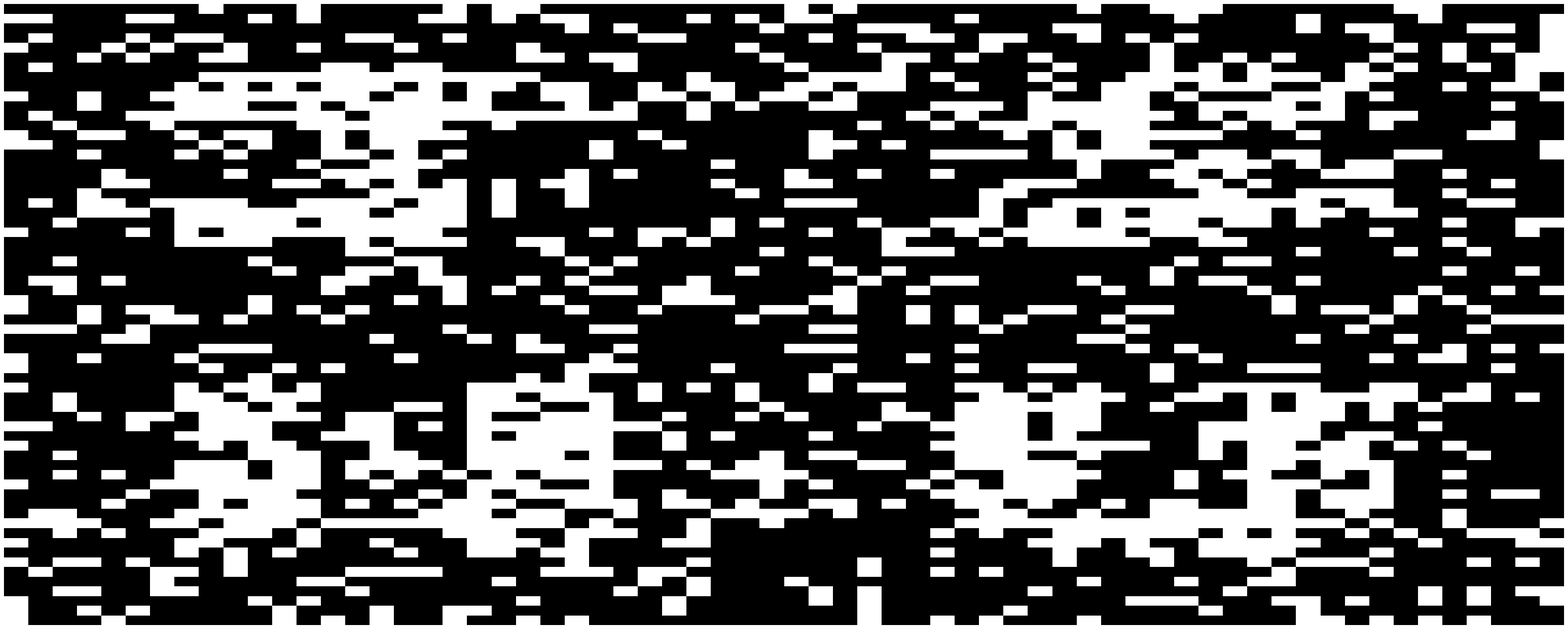} \end{minipage} &
\begin{minipage}{0.07\textwidth} \includegraphics[width=12mm, height=10mm]{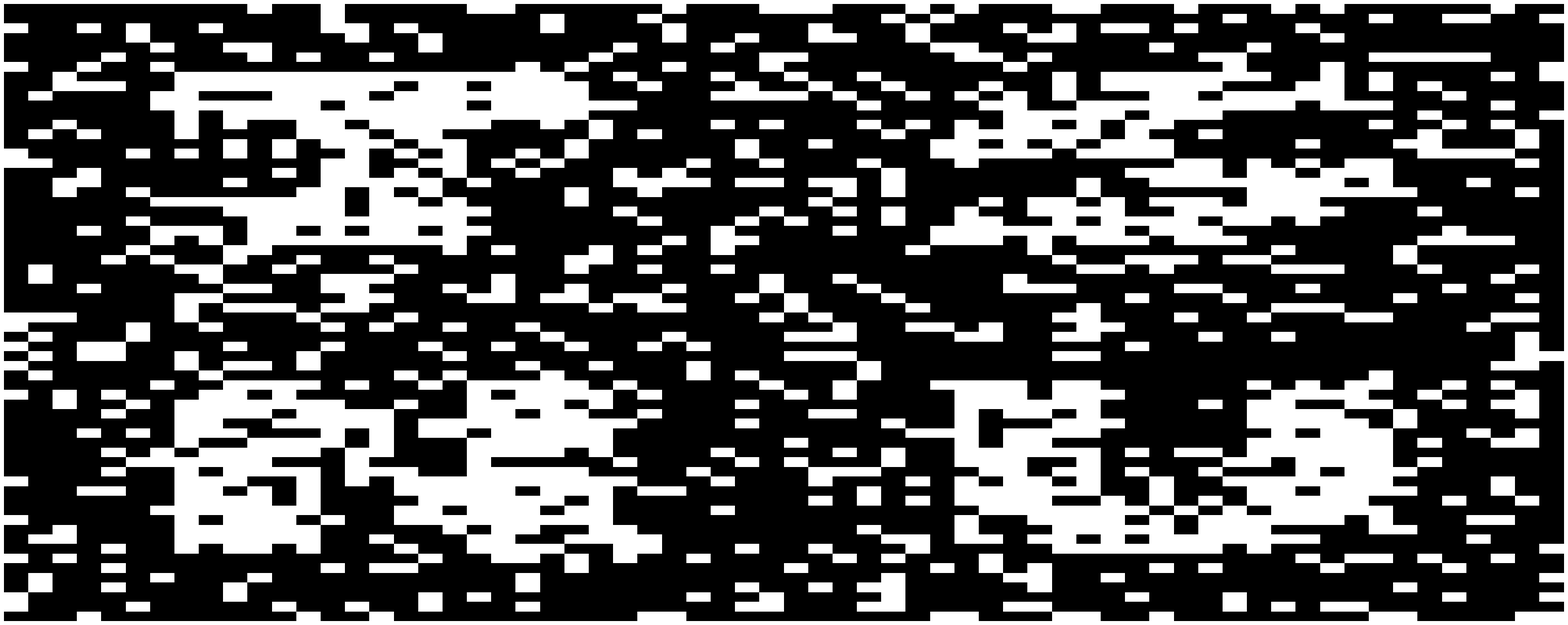} \end{minipage}\\
\cline{2-8}
& BER & 0.1245 & 0.1890 & 0.1208 & 0.2246 & 0.2419 & 0.2156\\
\cline{2-8}
&JPEG 60 &
\begin{minipage}{0.07\textwidth} \includegraphics[width=12mm, height=10mm]{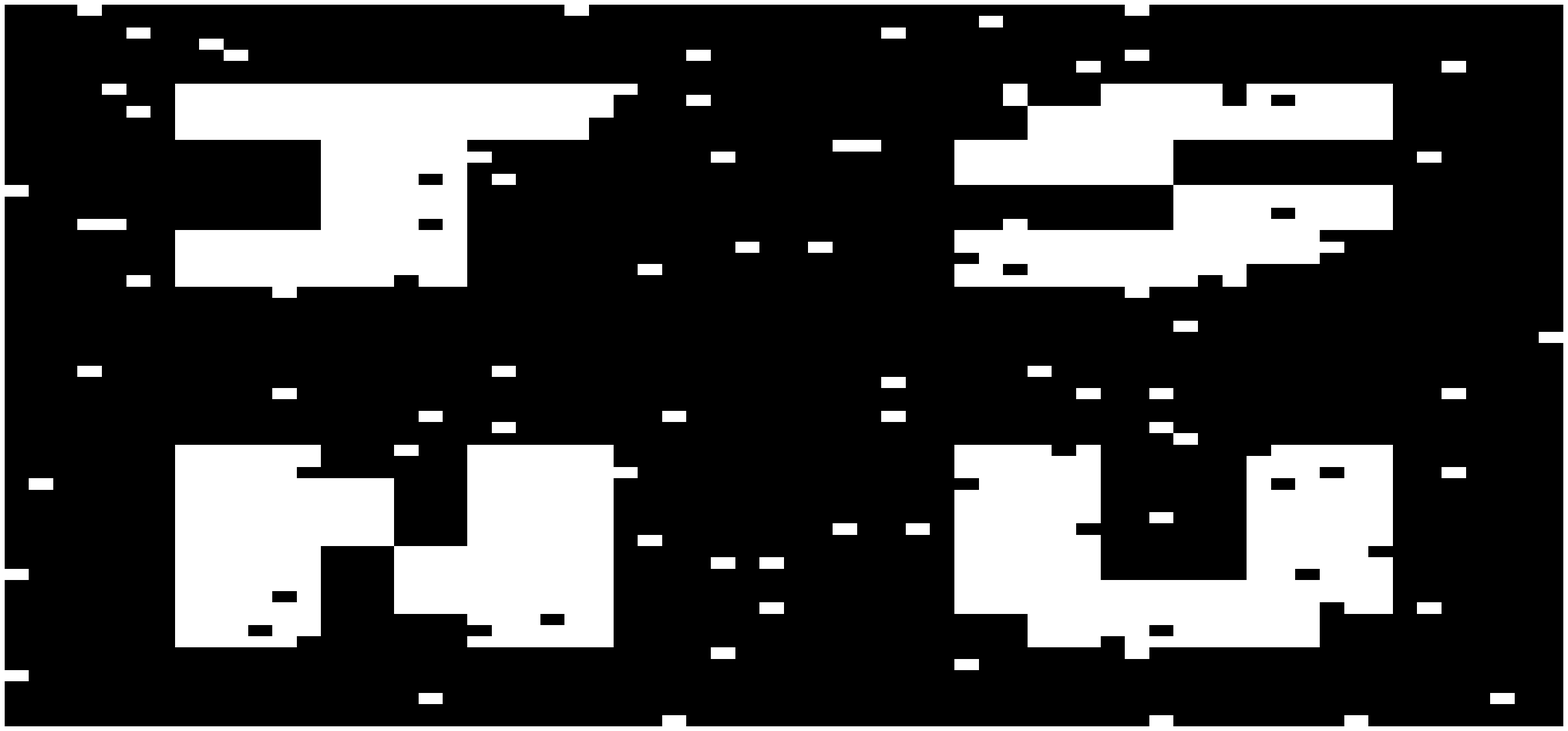} \end{minipage}  &
\begin{minipage}{0.07\textwidth} \includegraphics[width=12mm, height=10mm]{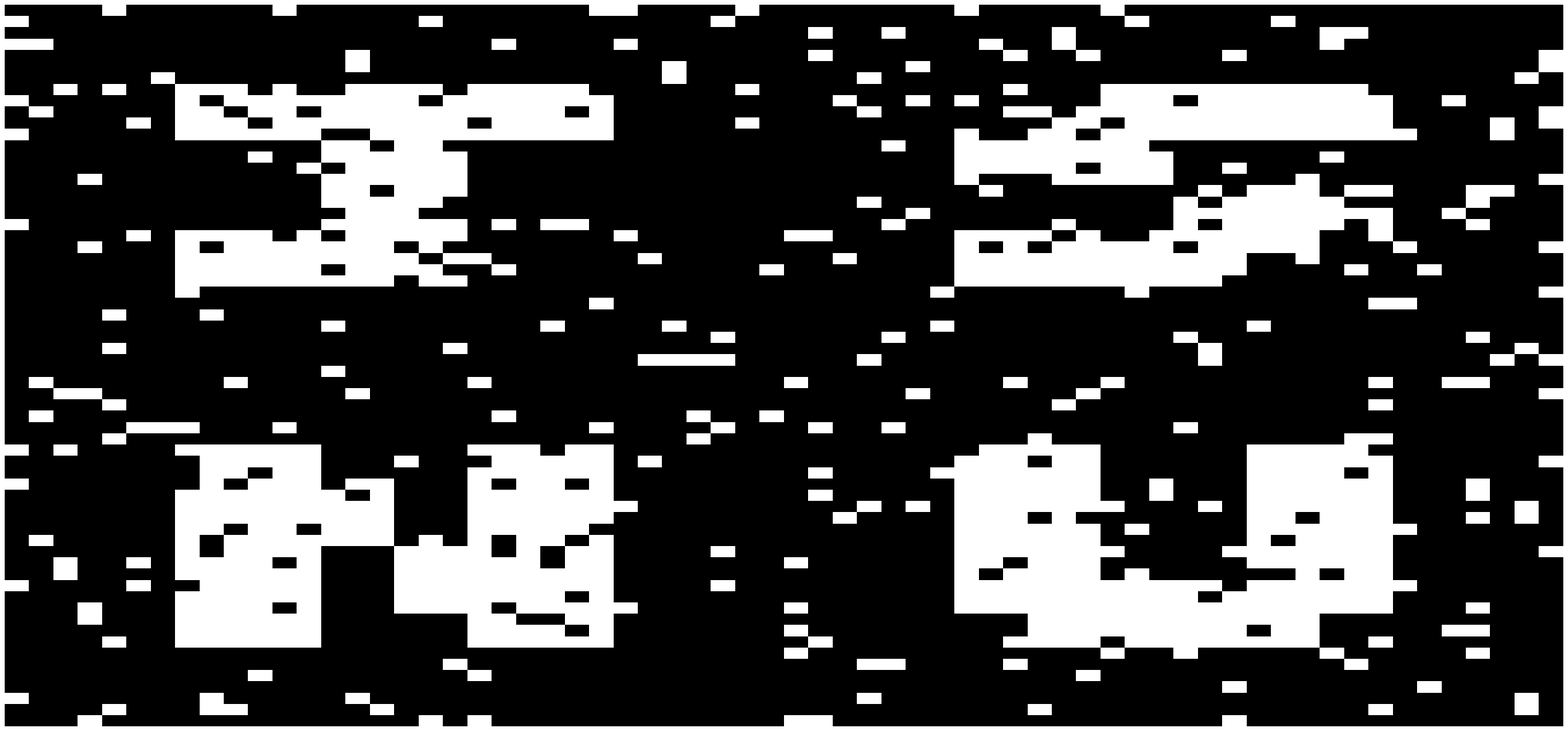} \end{minipage} &
\begin{minipage}{0.07\textwidth} \includegraphics[width=12mm, height=10mm]{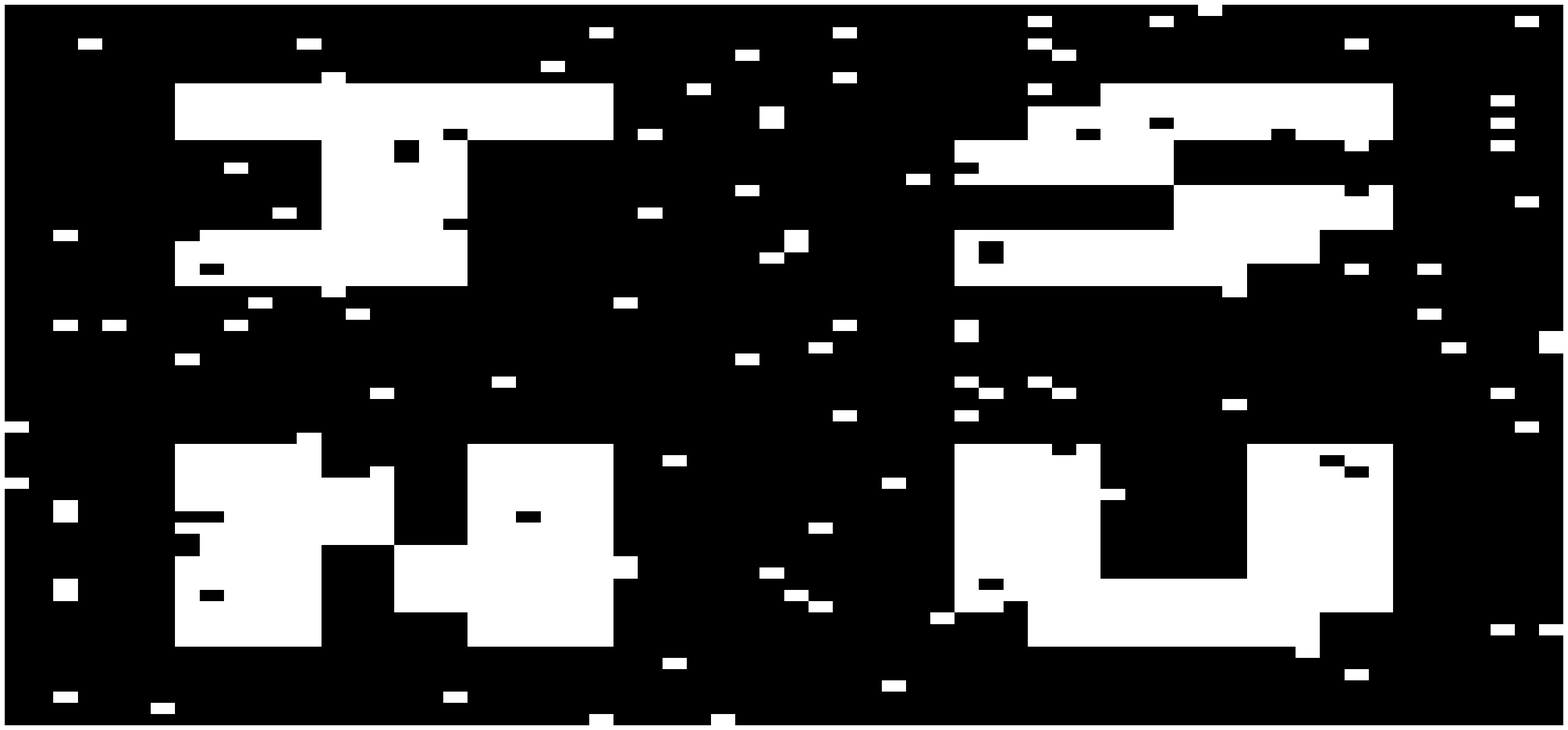} \end{minipage}&
\begin{minipage}{0.07\textwidth} \includegraphics[width=12mm, height=10mm]{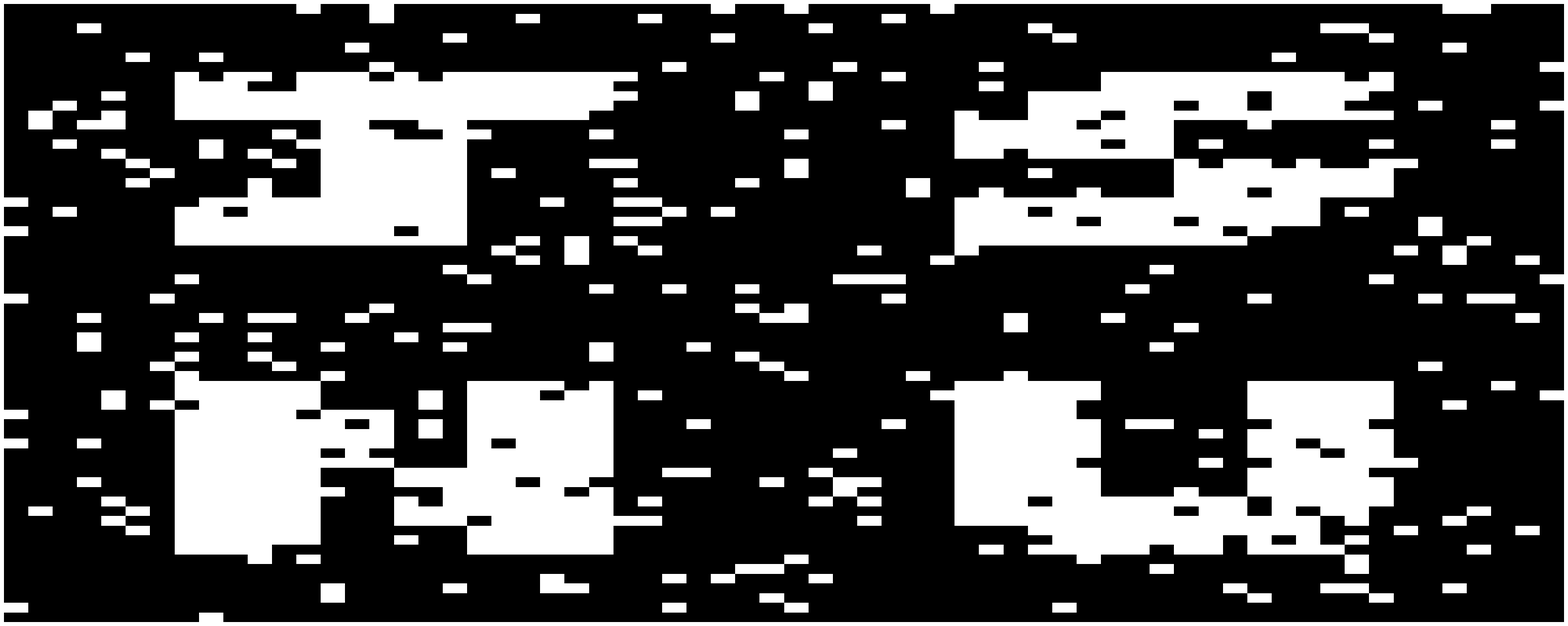} \end{minipage}  &
\begin{minipage}{0.07\textwidth} \includegraphics[width=12mm, height=10mm]{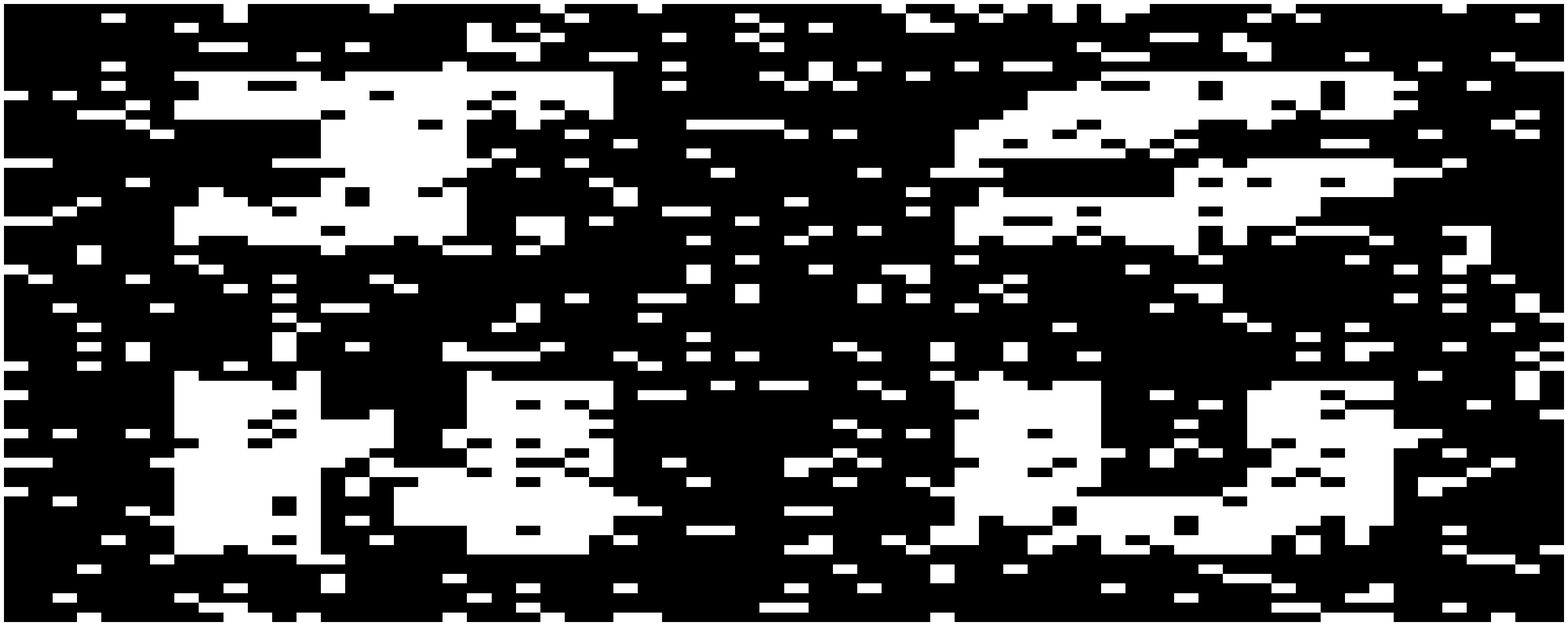} \end{minipage} &
\begin{minipage}{0.07\textwidth} \includegraphics[width=12mm, height=10mm]{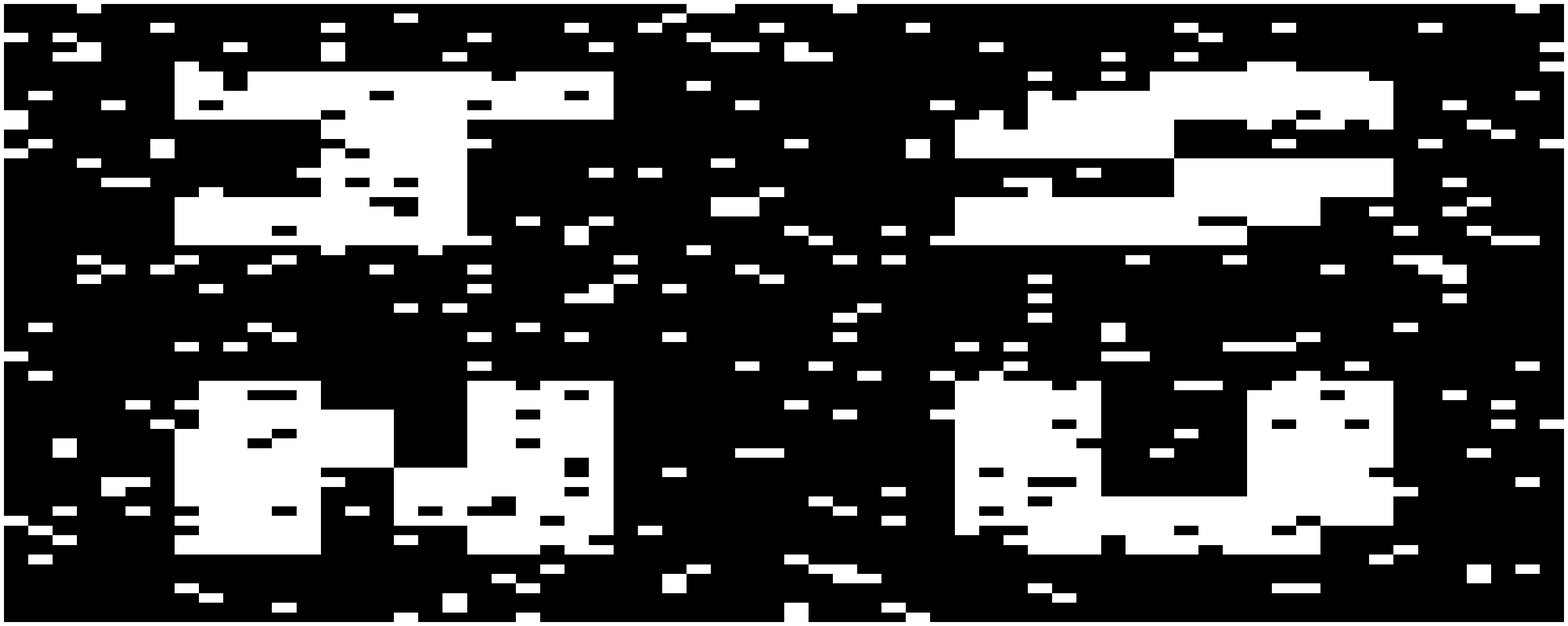} \end{minipage}\\
\cline{2-8}
& BER & 0.0271 & 0.0820 & 0.0266 & 0.0806 & 0.1299 & 0.0774\\
\cline{2-8}
&\tabincell{c}{Motion\\blur(4,4)} &
\begin{minipage}{0.07\textwidth} \includegraphics[width=12mm, height=10mm]{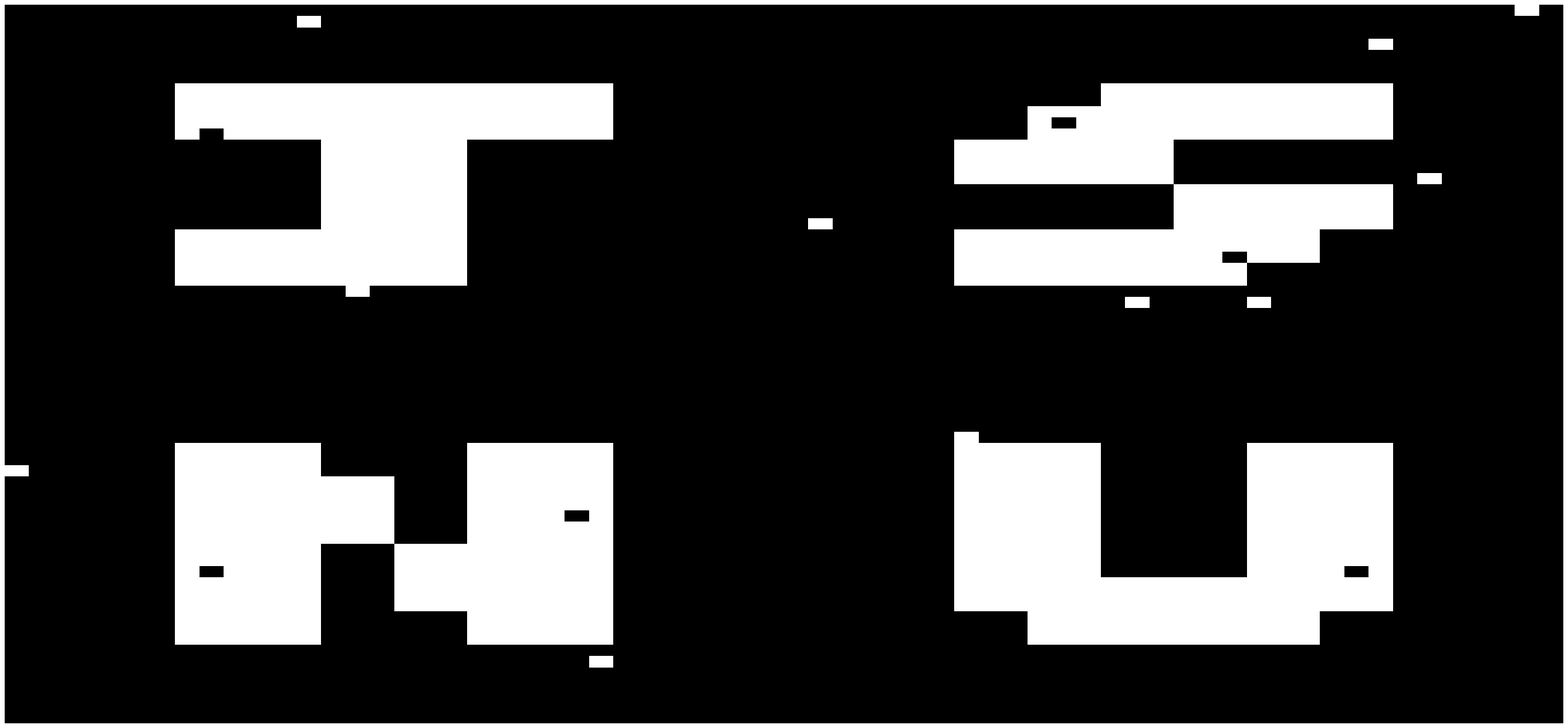} \end{minipage} &
\begin{minipage}{0.07\textwidth} \includegraphics[width=12mm, height=10mm]{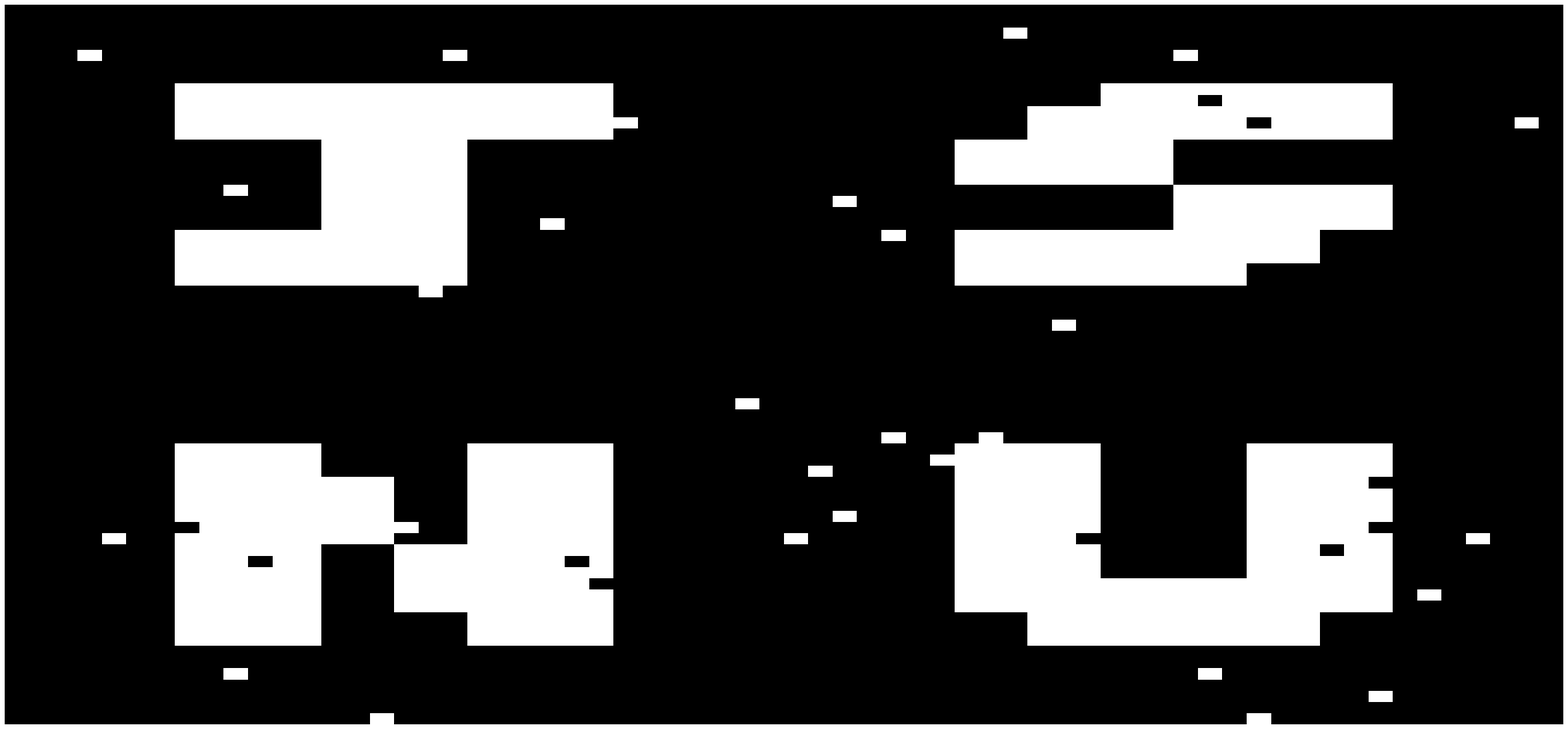} \end{minipage} &
\begin{minipage}{0.07\textwidth} \includegraphics[width=12mm, height=10mm]{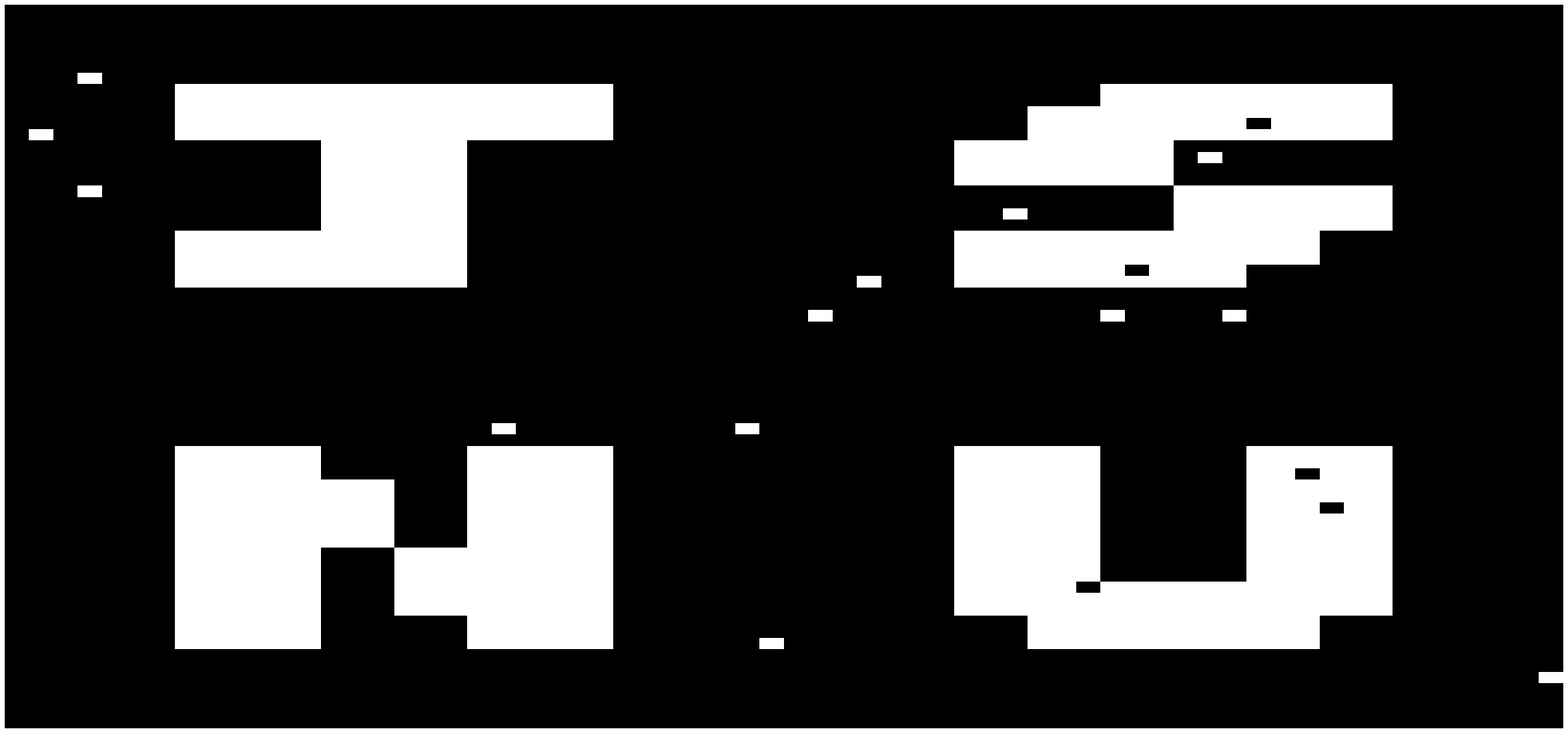} \end{minipage}&
\begin{minipage}{0.07\textwidth} \includegraphics[width=12mm, height=10mm]{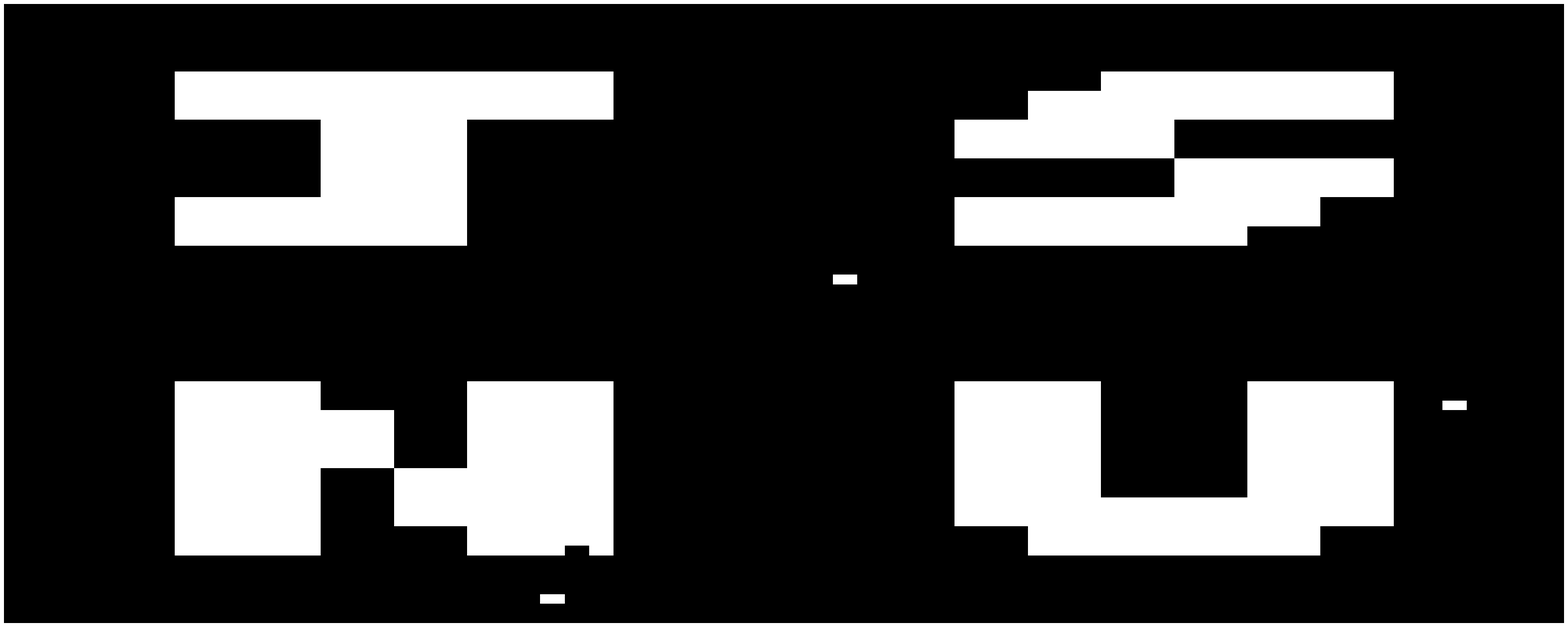} \end{minipage} &
\begin{minipage}{0.07\textwidth} \includegraphics[width=12mm, height=10mm]{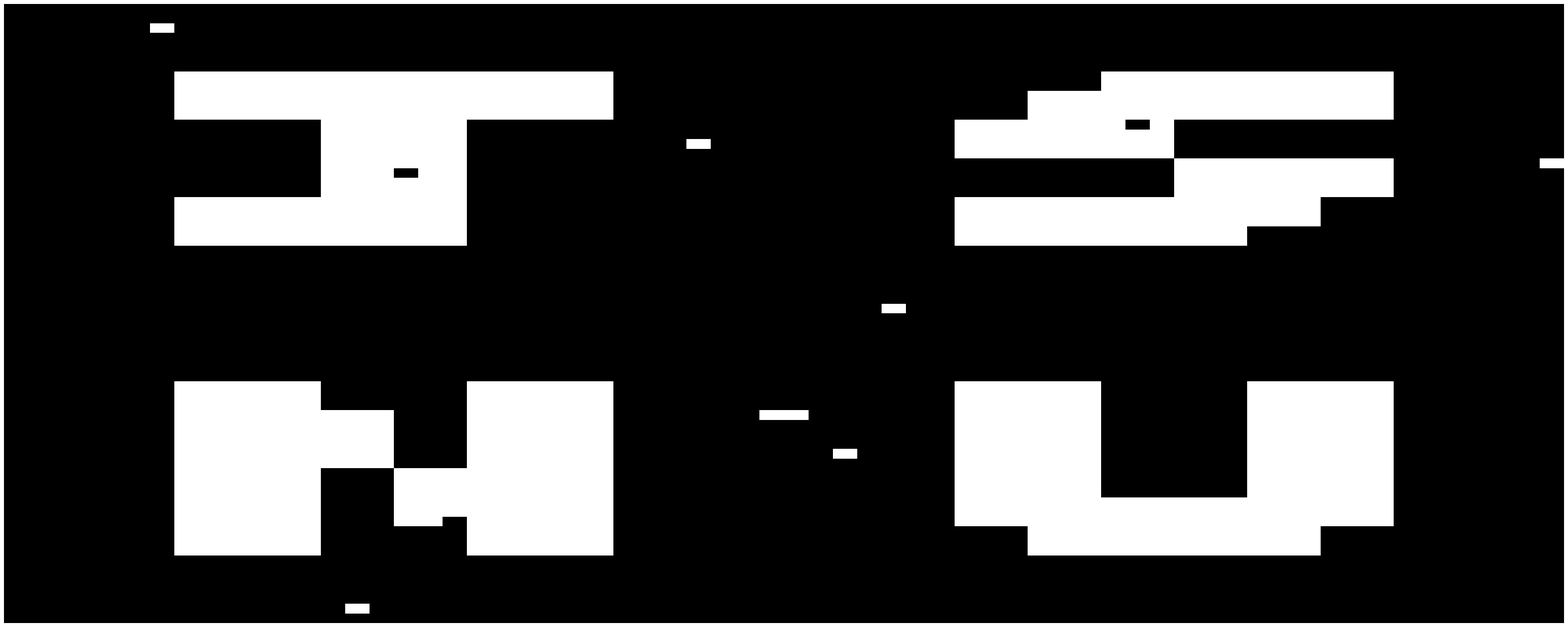} \end{minipage} &
\begin{minipage}{0.07\textwidth} \includegraphics[width=12mm, height=10mm]{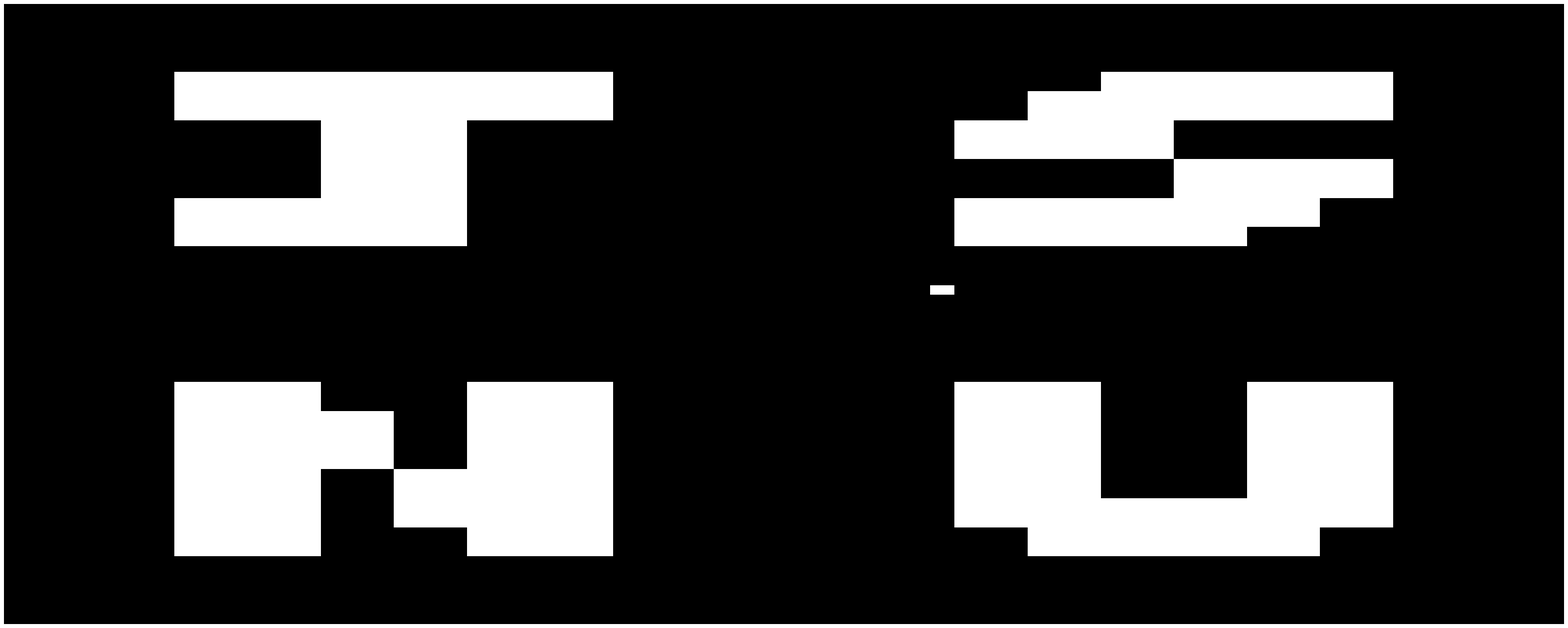} \end{minipage}\\
\cline{2-8}
& BER & 0.0049 & 0.0081 & 0.0032 & 0.0009 & 0.0027 & 0.0002\\
\cline{2-8}
&\tabincell{c}{Motion\\blur(9,9)}&
\begin{minipage}{0.07\textwidth} \includegraphics[width=12mm, height=10mm]{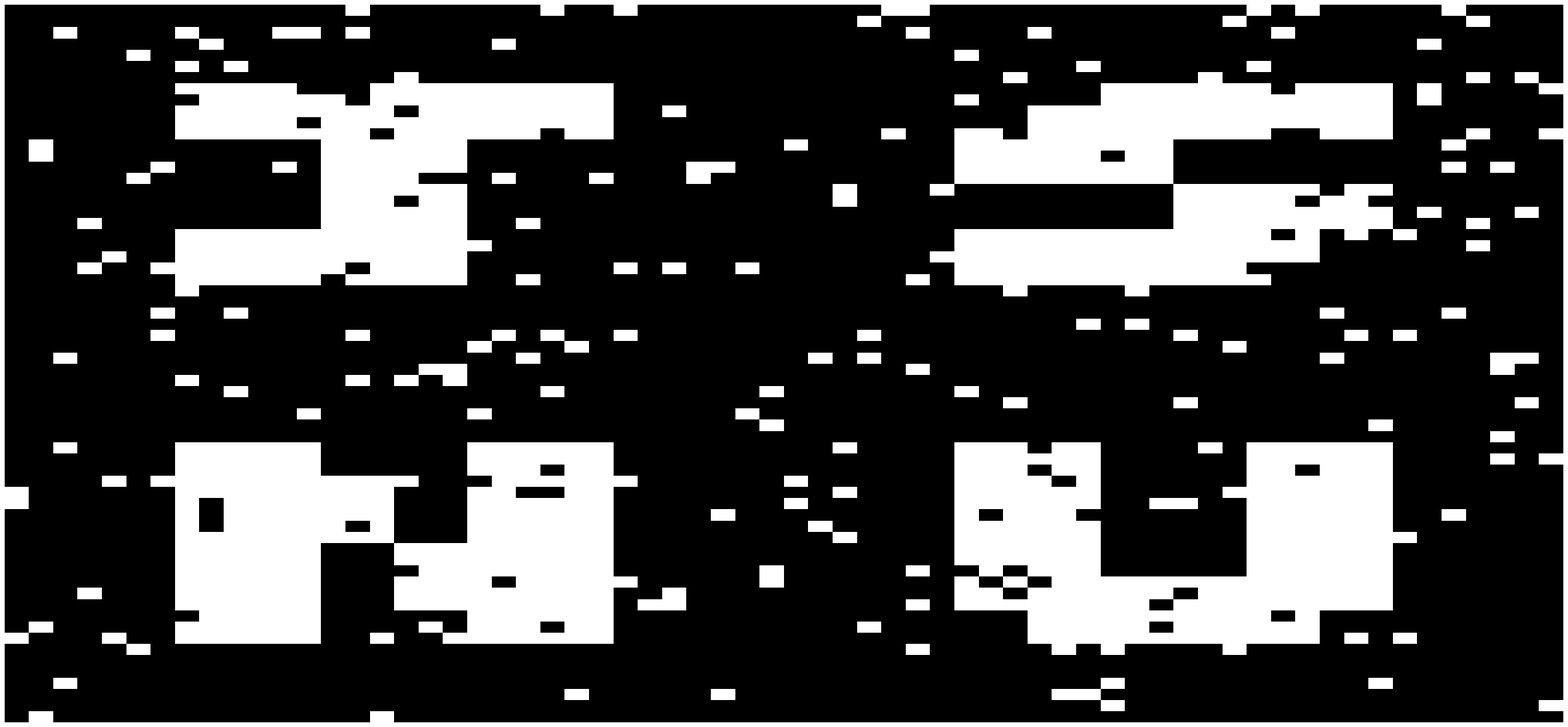} \end{minipage} &
\begin{minipage}{0.07\textwidth} \includegraphics[width=12mm, height=10mm]{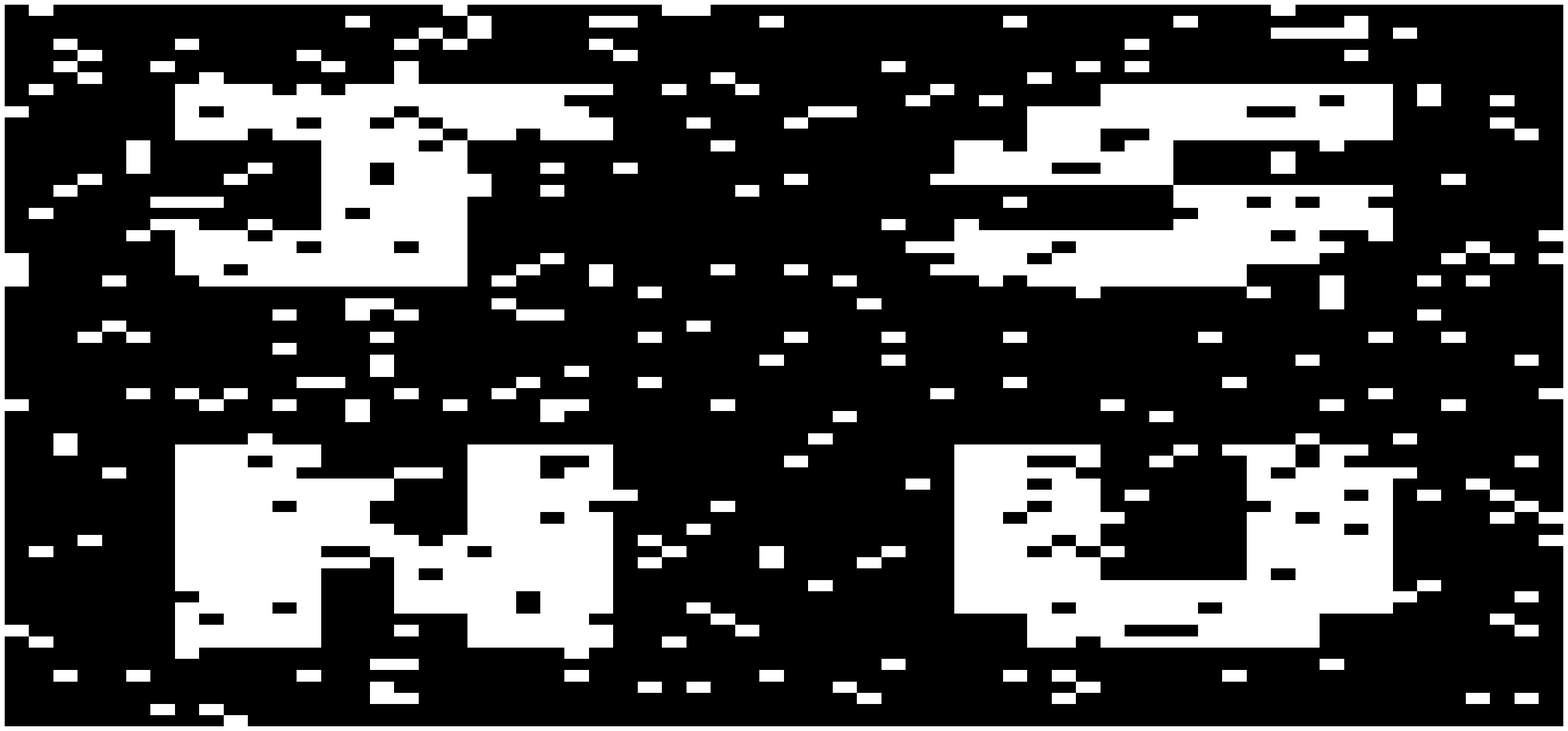} \end{minipage} &
\begin{minipage}{0.07\textwidth} \includegraphics[width=12mm, height=10mm]{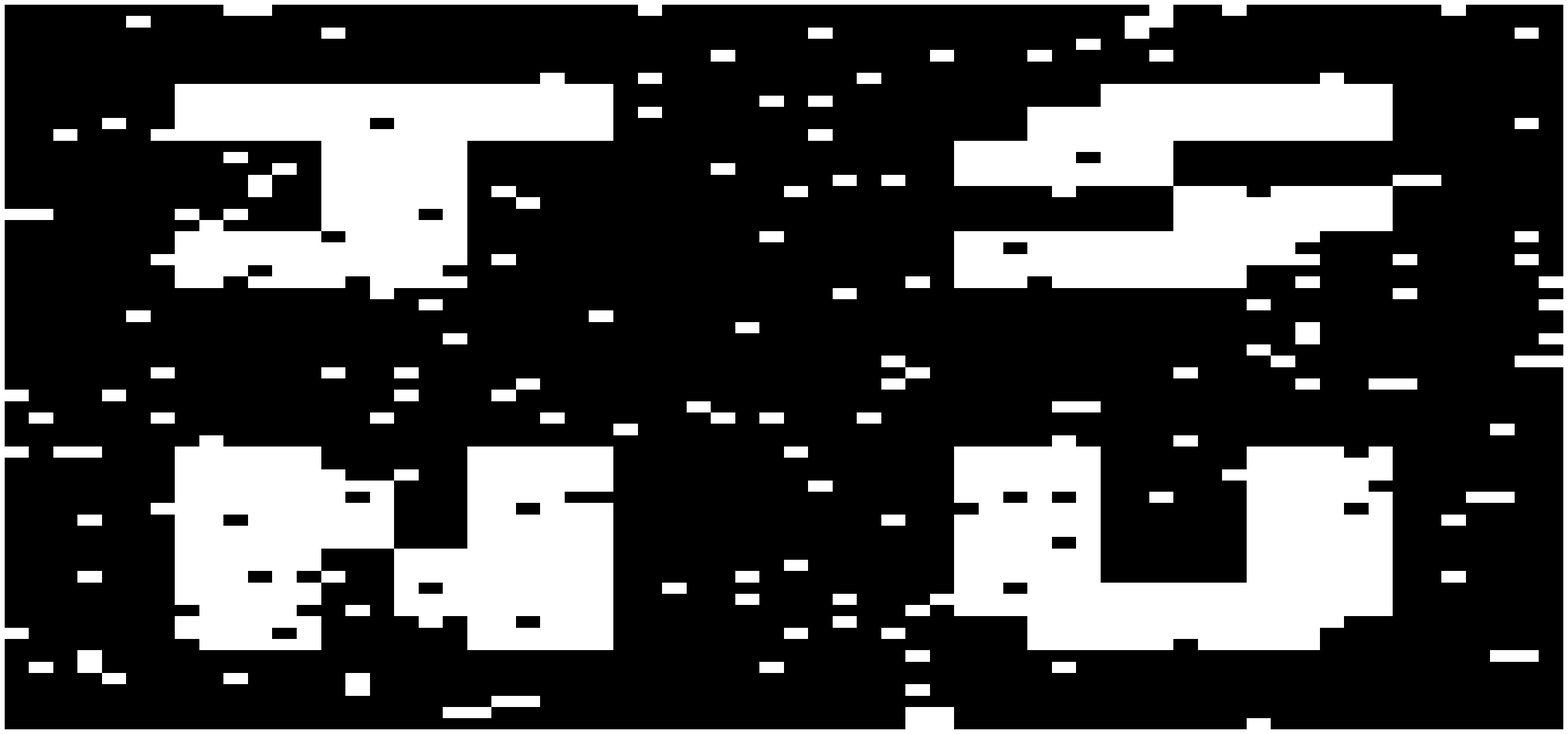} \end{minipage}&
\begin{minipage}{0.07\textwidth} \includegraphics[width=12mm, height=10mm]{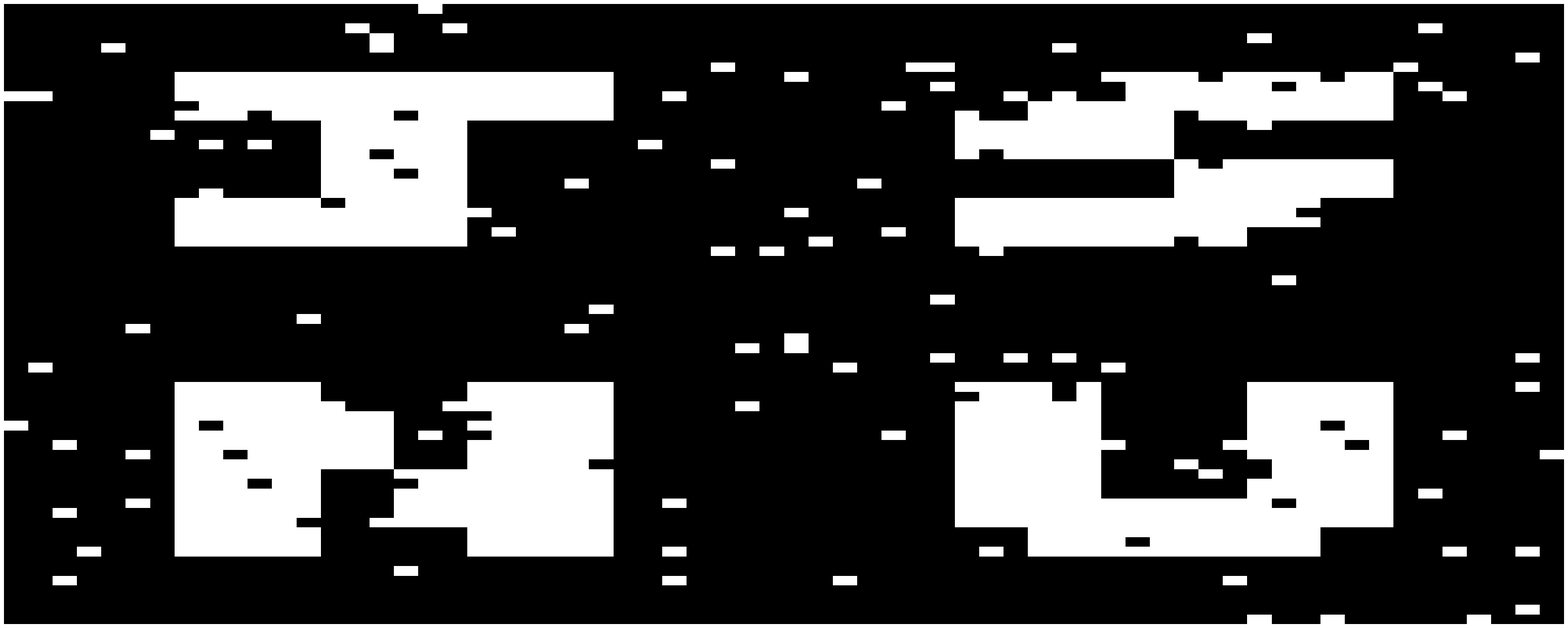} \end{minipage} &
\begin{minipage}{0.07\textwidth} \includegraphics[width=12mm, height=10mm]{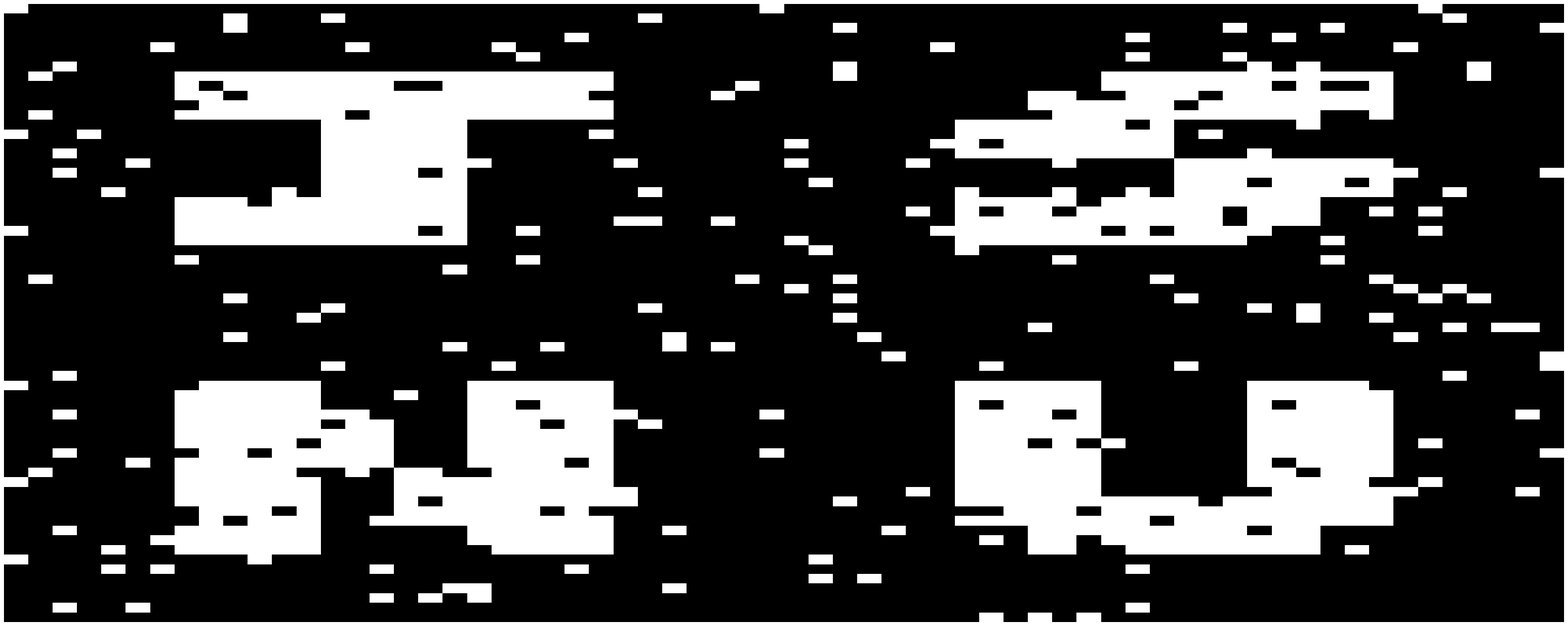} \end{minipage} &
\begin{minipage}{0.07\textwidth} \includegraphics[width=12mm, height=10mm]{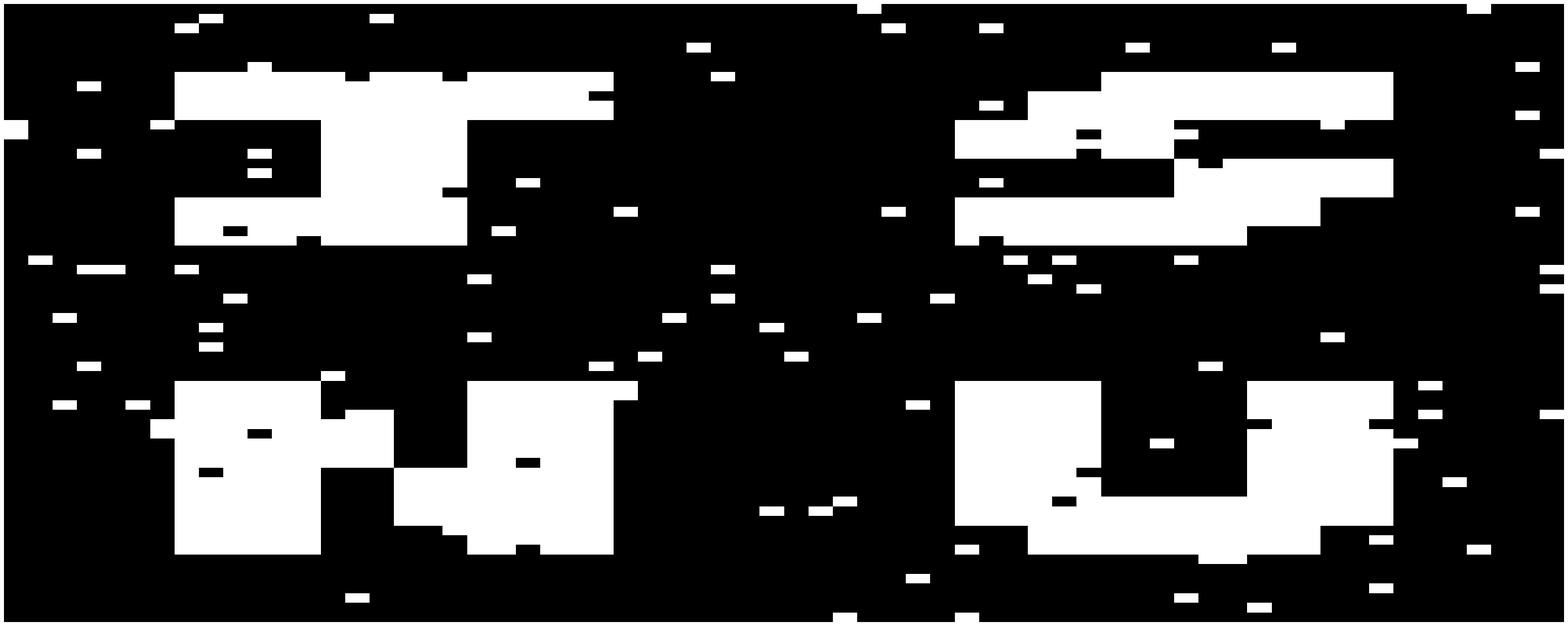} \end{minipage}\\
\cline{2-8}
& BER & 0.0562 & 0.0725 & 0.0427 & 0.0308 & 0.0598 & 0.0266\\
\cline{2-8}
&Scaling-0.5 &
\begin{minipage}{0.07\textwidth} \includegraphics[width=12mm, height=10mm]{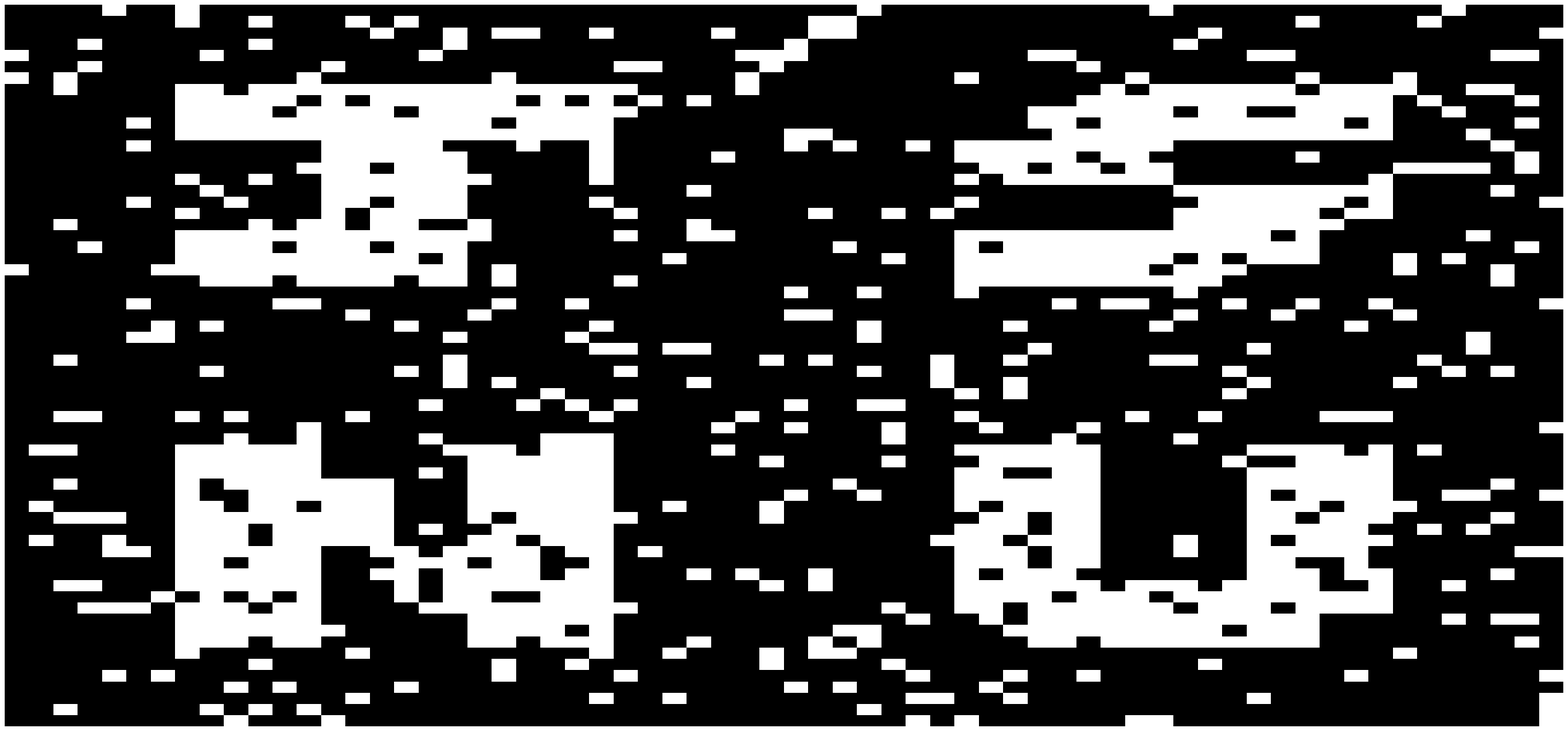} \end{minipage} &
\begin{minipage}{0.07\textwidth} \includegraphics[width=12mm, height=10mm]{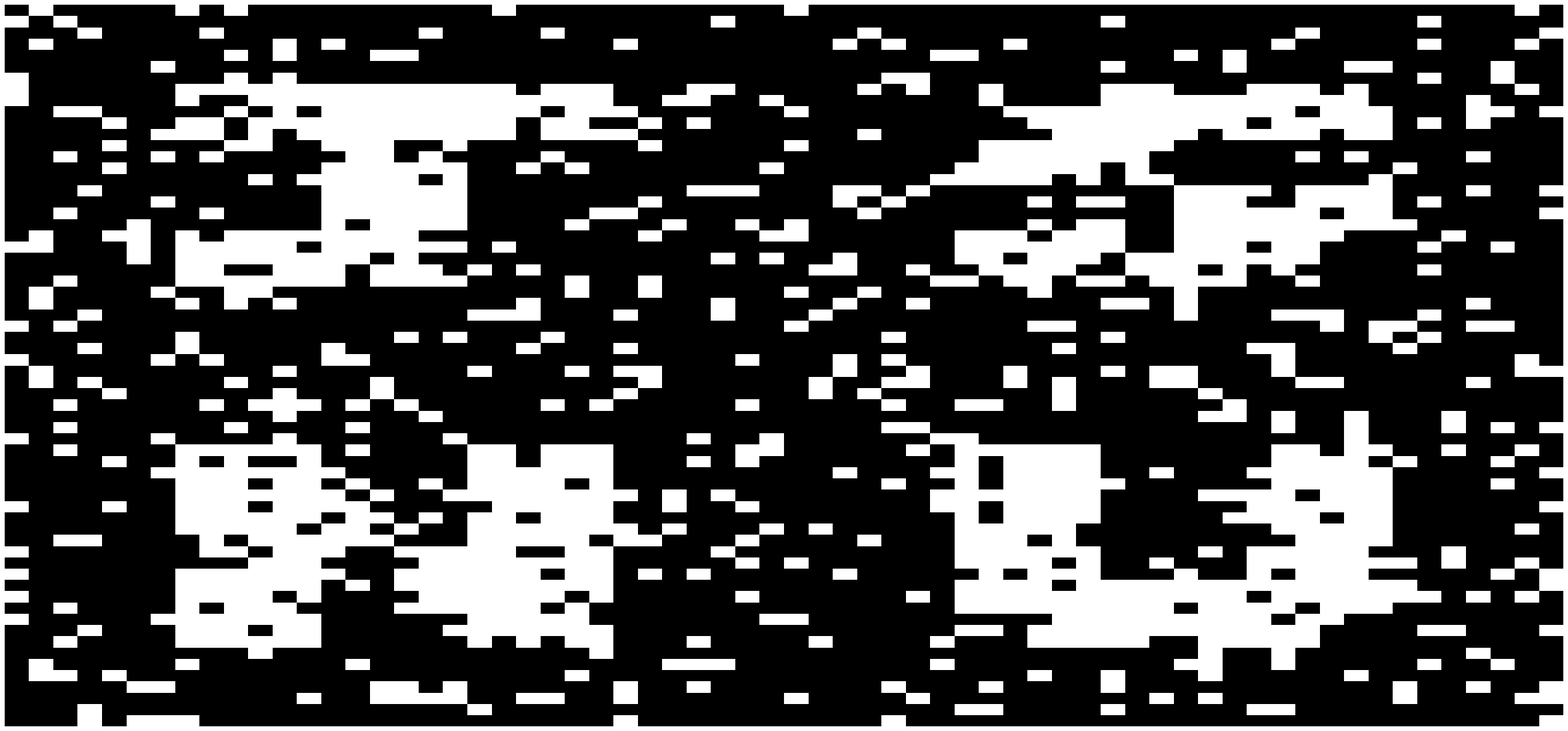} \end{minipage} &
\begin{minipage}{0.07\textwidth} \includegraphics[width=12mm, height=10mm]{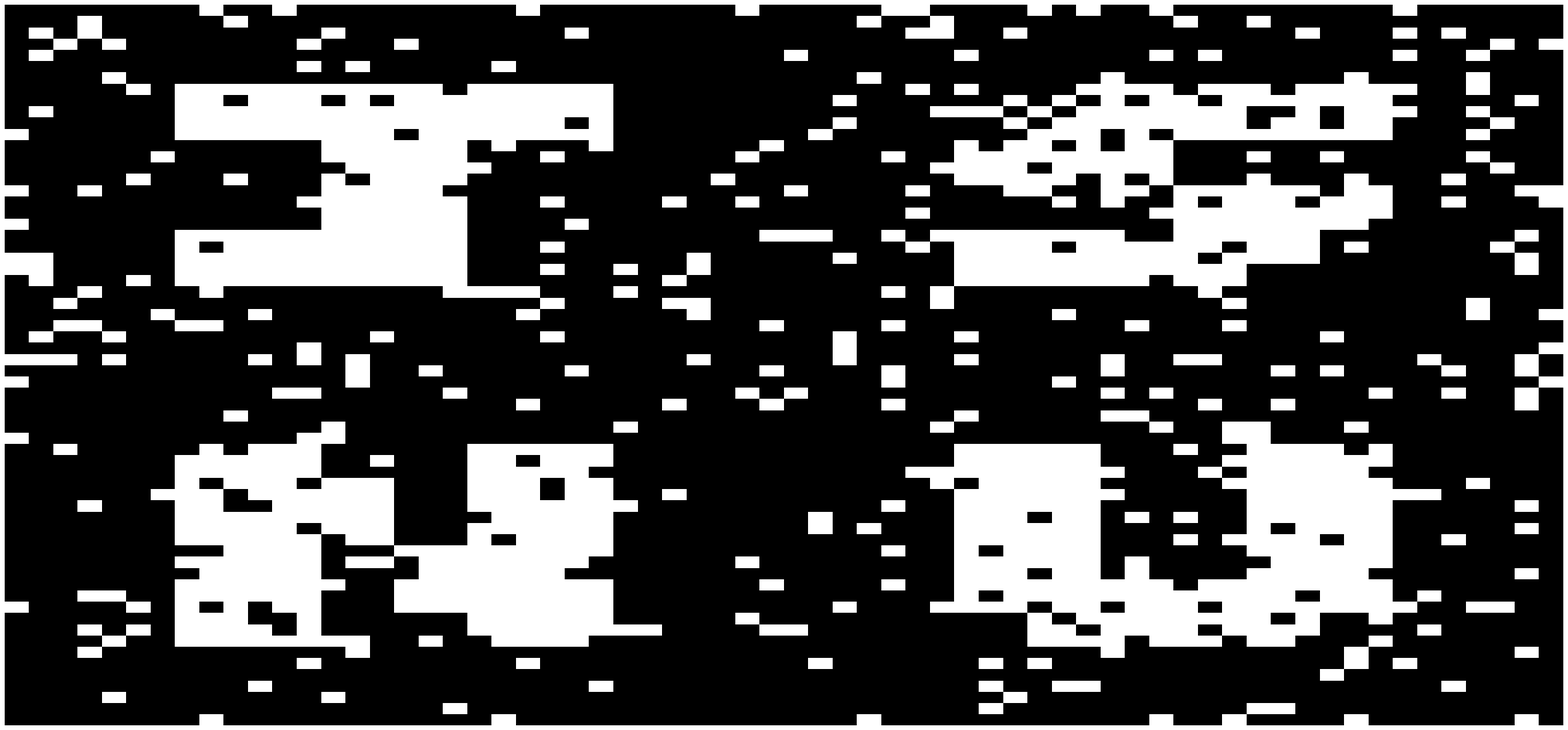} \end{minipage}&
\begin{minipage}{0.07\textwidth} \includegraphics[width=12mm, height=10mm]{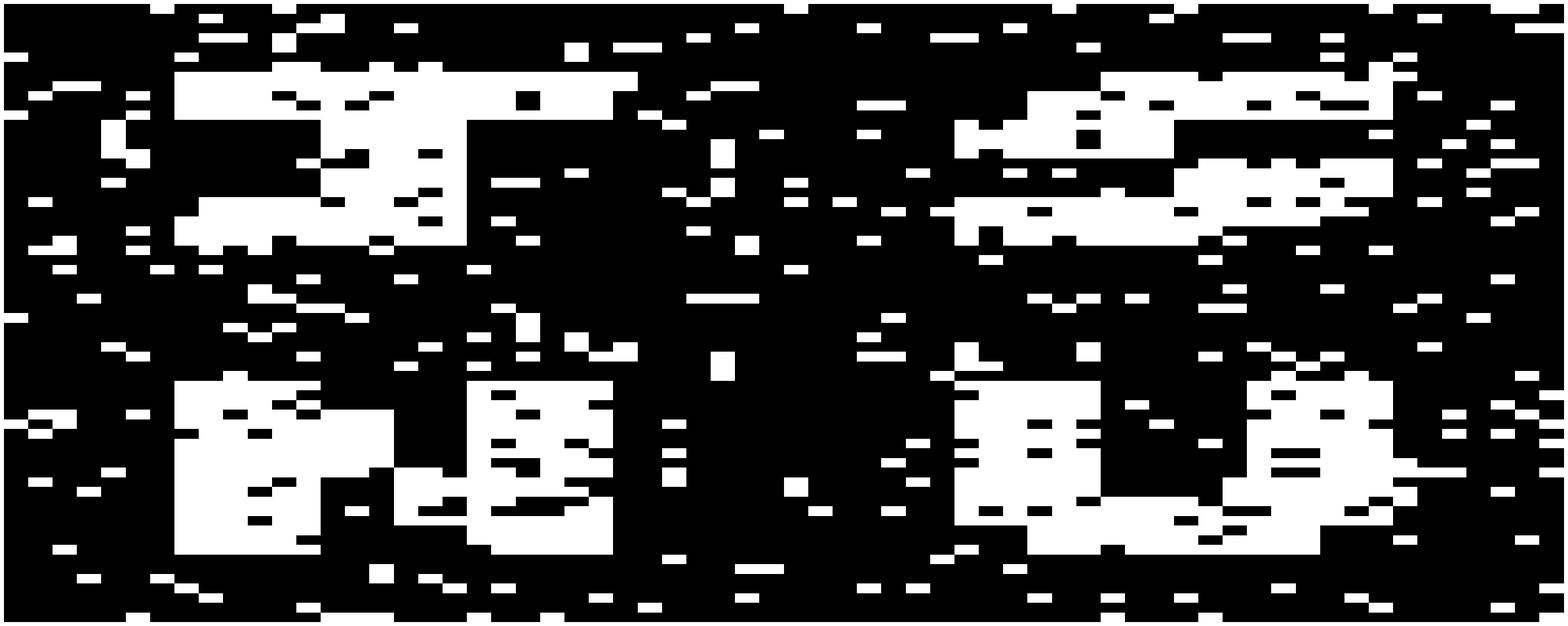} \end{minipage} &
\begin{minipage}{0.07\textwidth} \includegraphics[width=12mm, height=10mm]{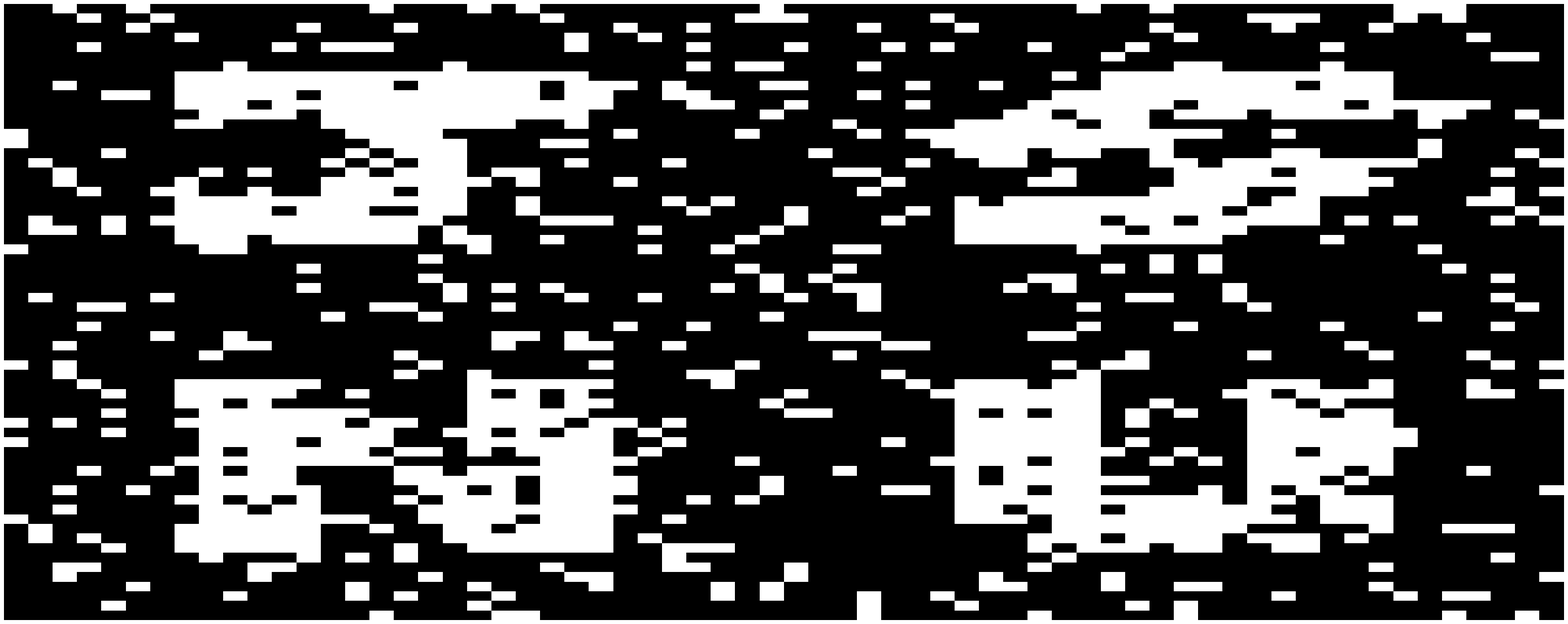} \end{minipage} &
\begin{minipage}{0.07\textwidth} \includegraphics[width=12mm, height=10mm]{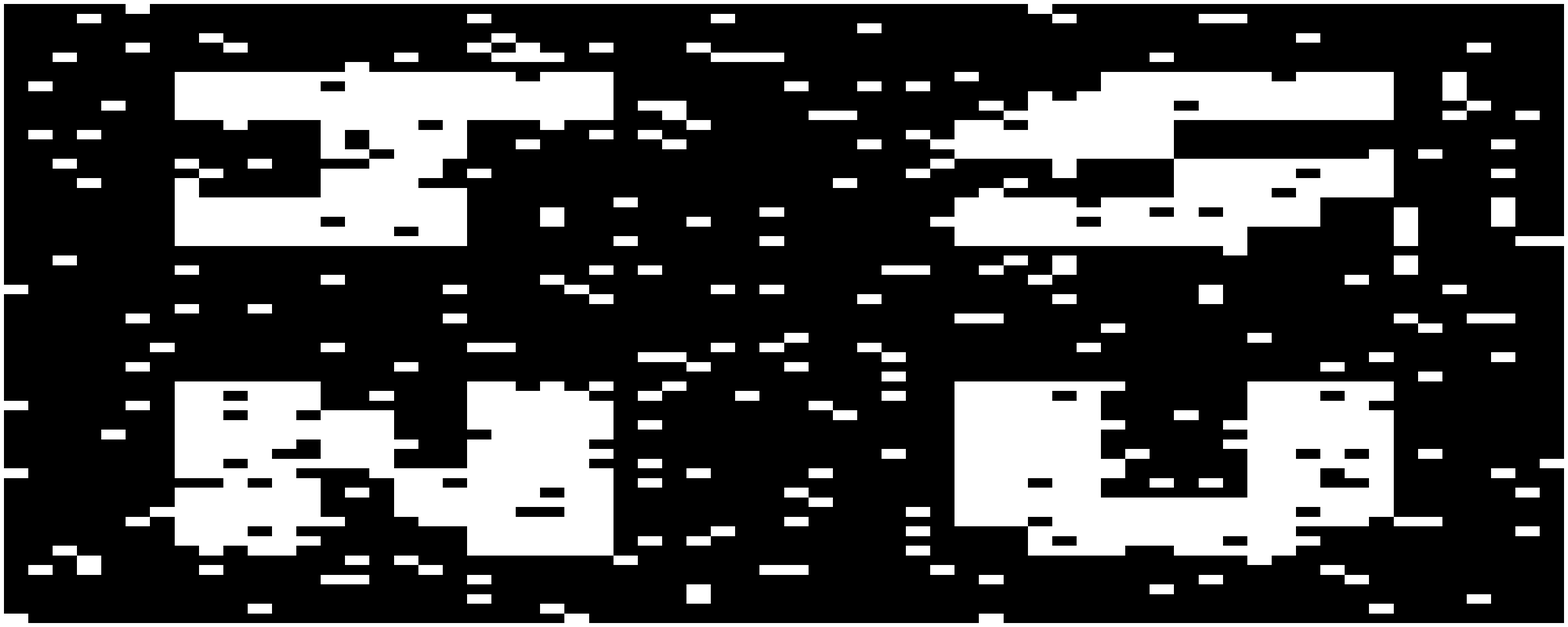} \end{minipage}\\
\cline{2-8}
& BER & 0.1267 & 0.1470 & 0.0996 & 0.0947 &0.1418 & 0.0801\\
\cline{2-8}
&Scaling-2&
\begin{minipage}{0.07\textwidth} \includegraphics[width=12mm, height=10mm]{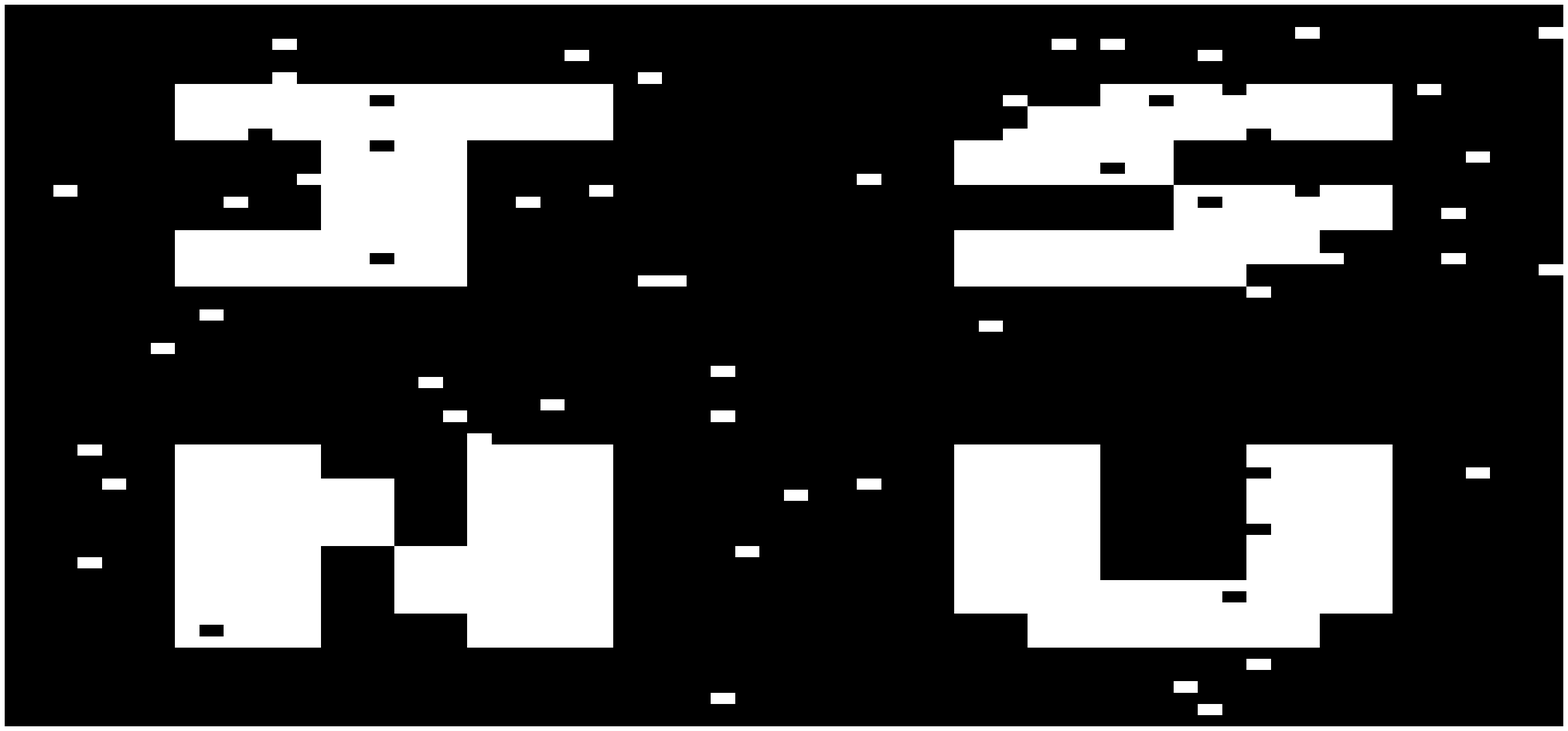} \end{minipage} &
\begin{minipage}{0.07\textwidth} \includegraphics[width=12mm, height=10mm]{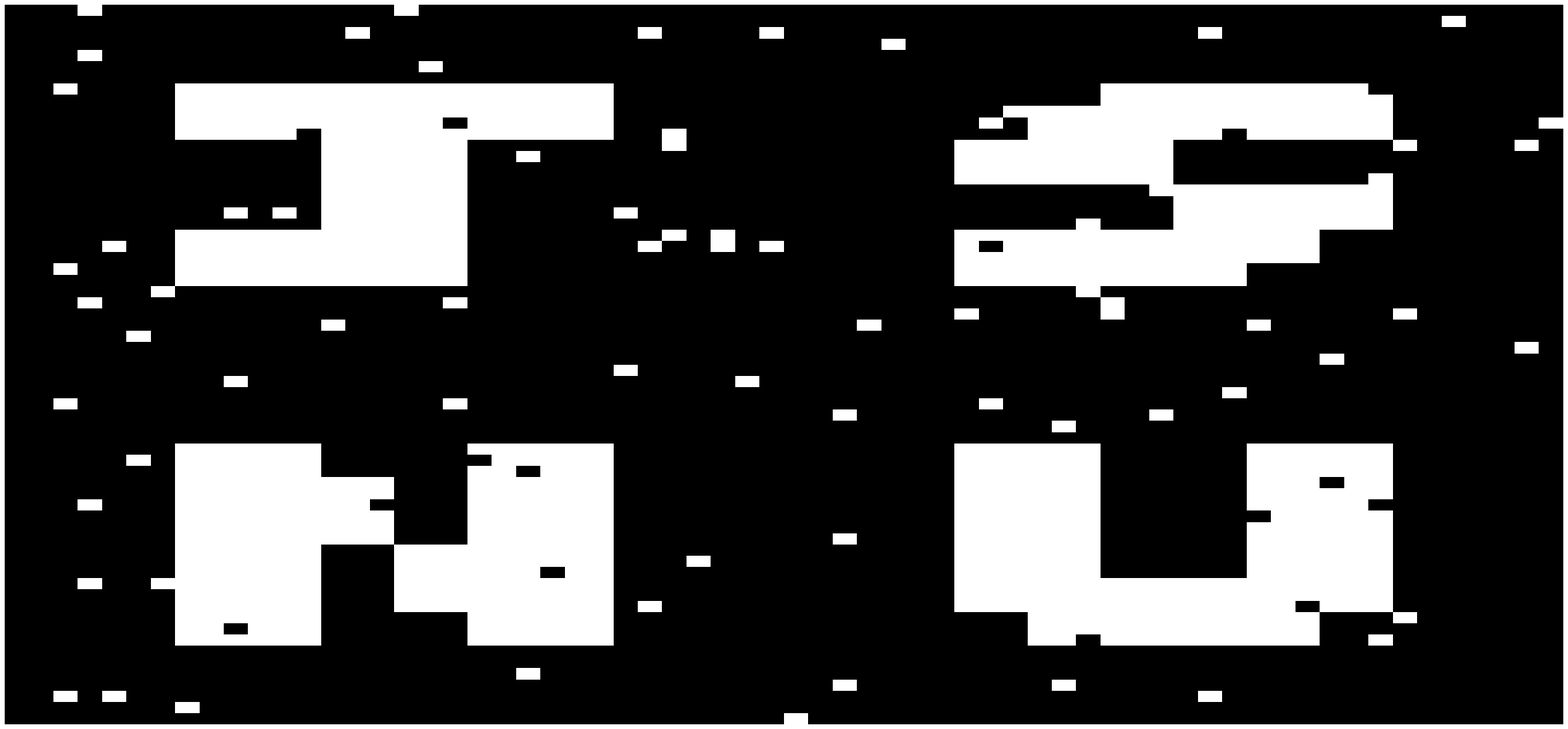} \end{minipage} &
\begin{minipage}{0.07\textwidth} \includegraphics[width=12mm, height=10mm]{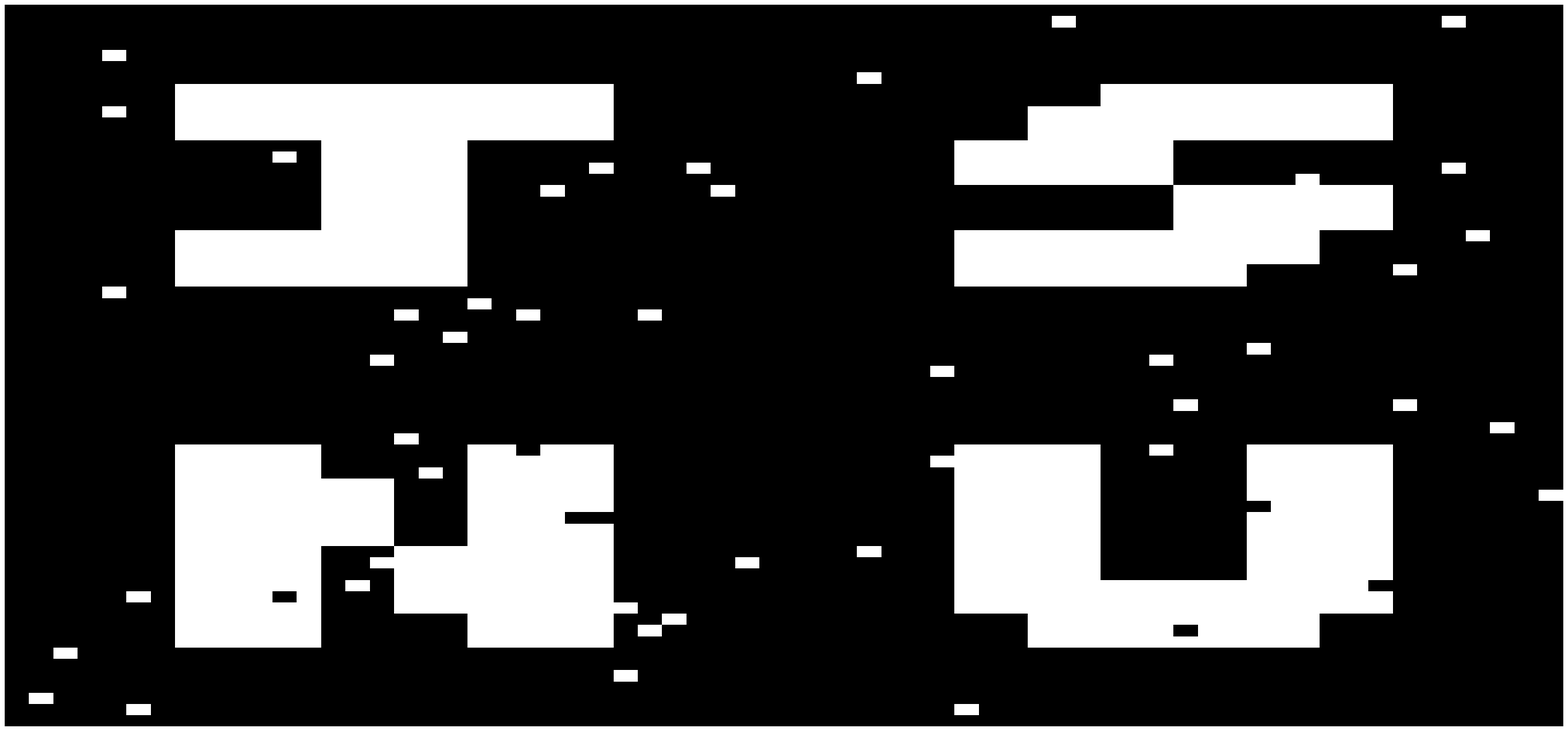} \end{minipage}&
\begin{minipage}{0.07\textwidth} \includegraphics[width=12mm, height=10mm]{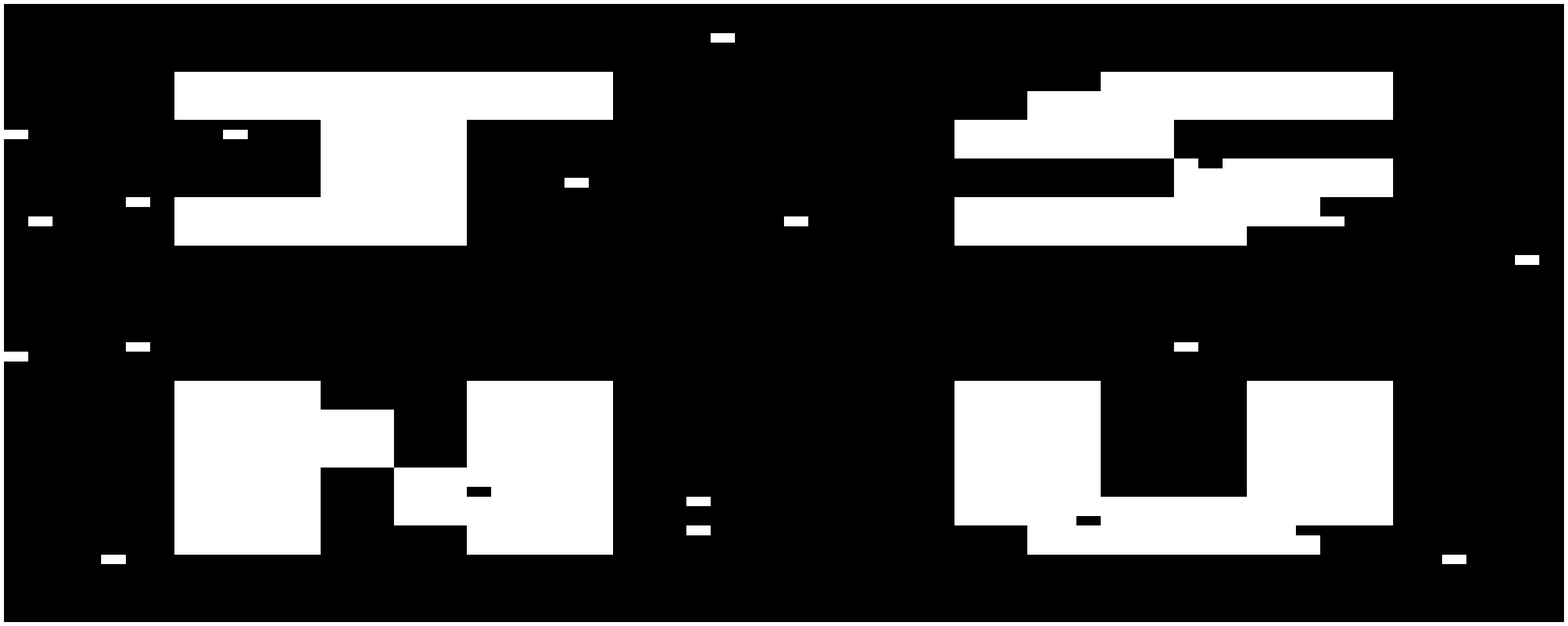} \end{minipage} &
\begin{minipage}{0.07\textwidth} \includegraphics[width=12mm, height=10mm]{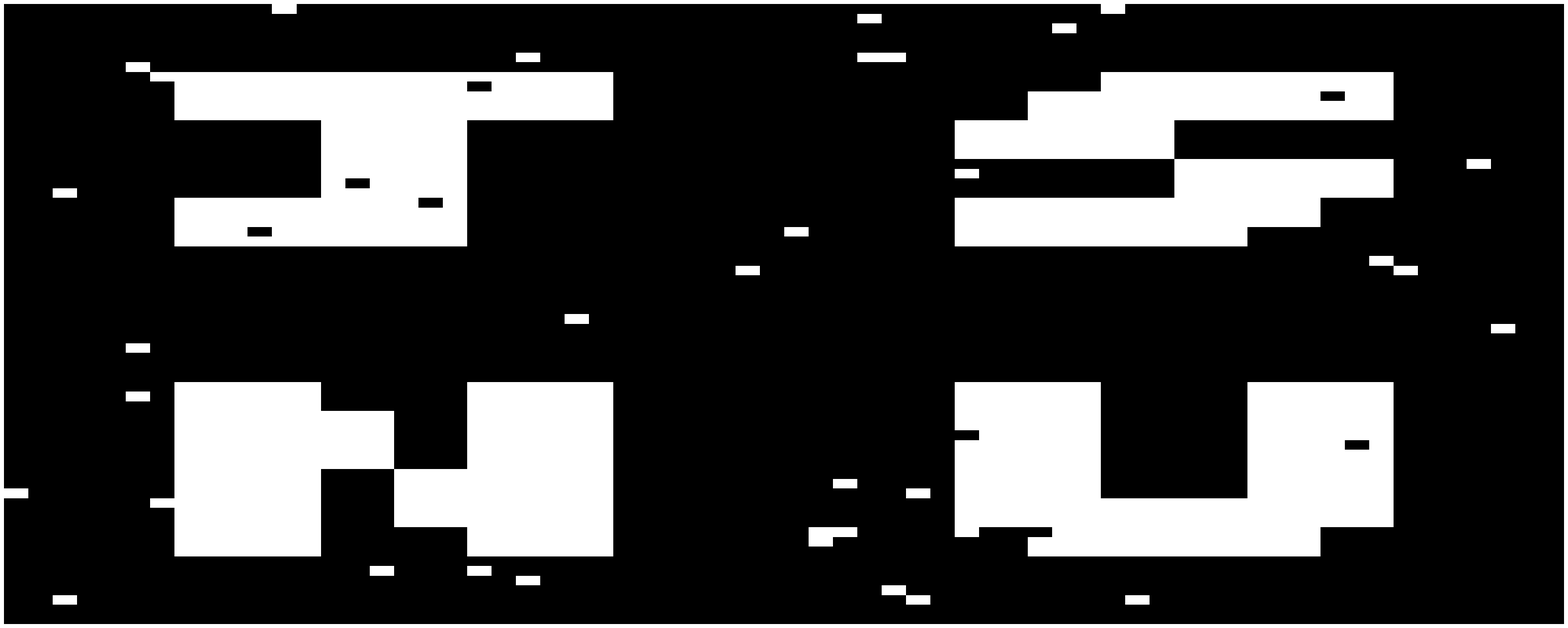} \end{minipage} &
\begin{minipage}{0.07\textwidth} \includegraphics[width=12mm, height=10mm]{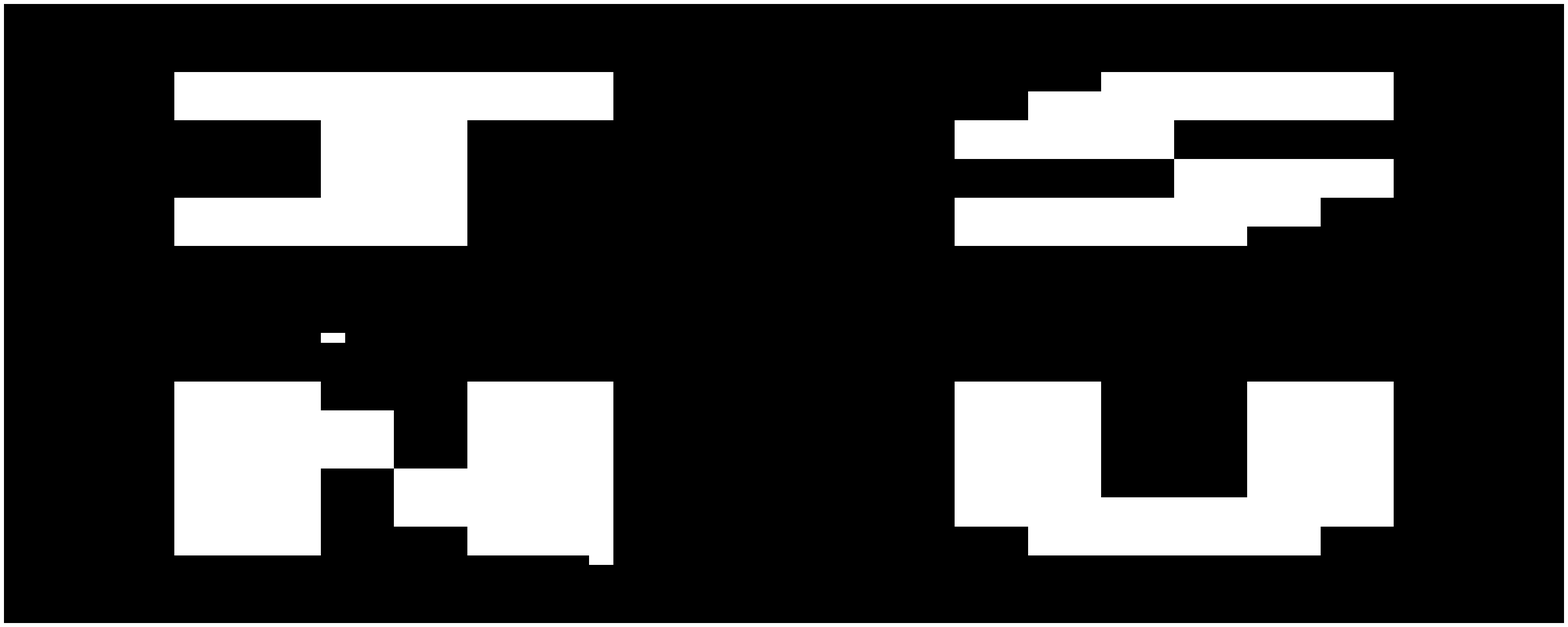} \end{minipage}\\
\cline{2-8}
& BER & 0.0105 & 0.0205 & 0.0090 & 0.0049 & 0.0105 & 0.0005\\
\cline{2-8}
&Cropping 10\%&
\begin{minipage}{0.07\textwidth} \includegraphics[width=12mm, height=10mm]{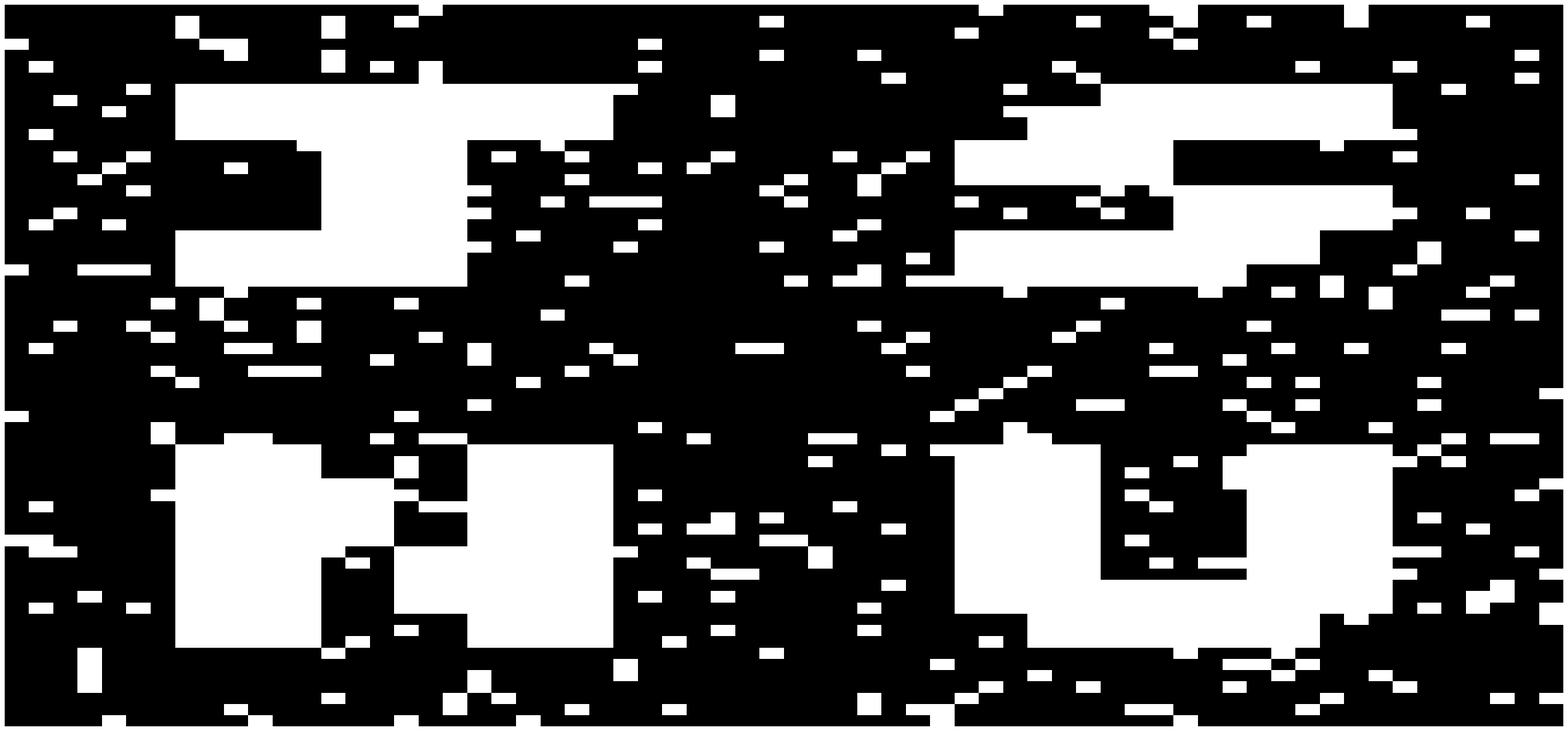} \end{minipage} &
\begin{minipage}{0.07\textwidth} \includegraphics[width=12mm, height=10mm]{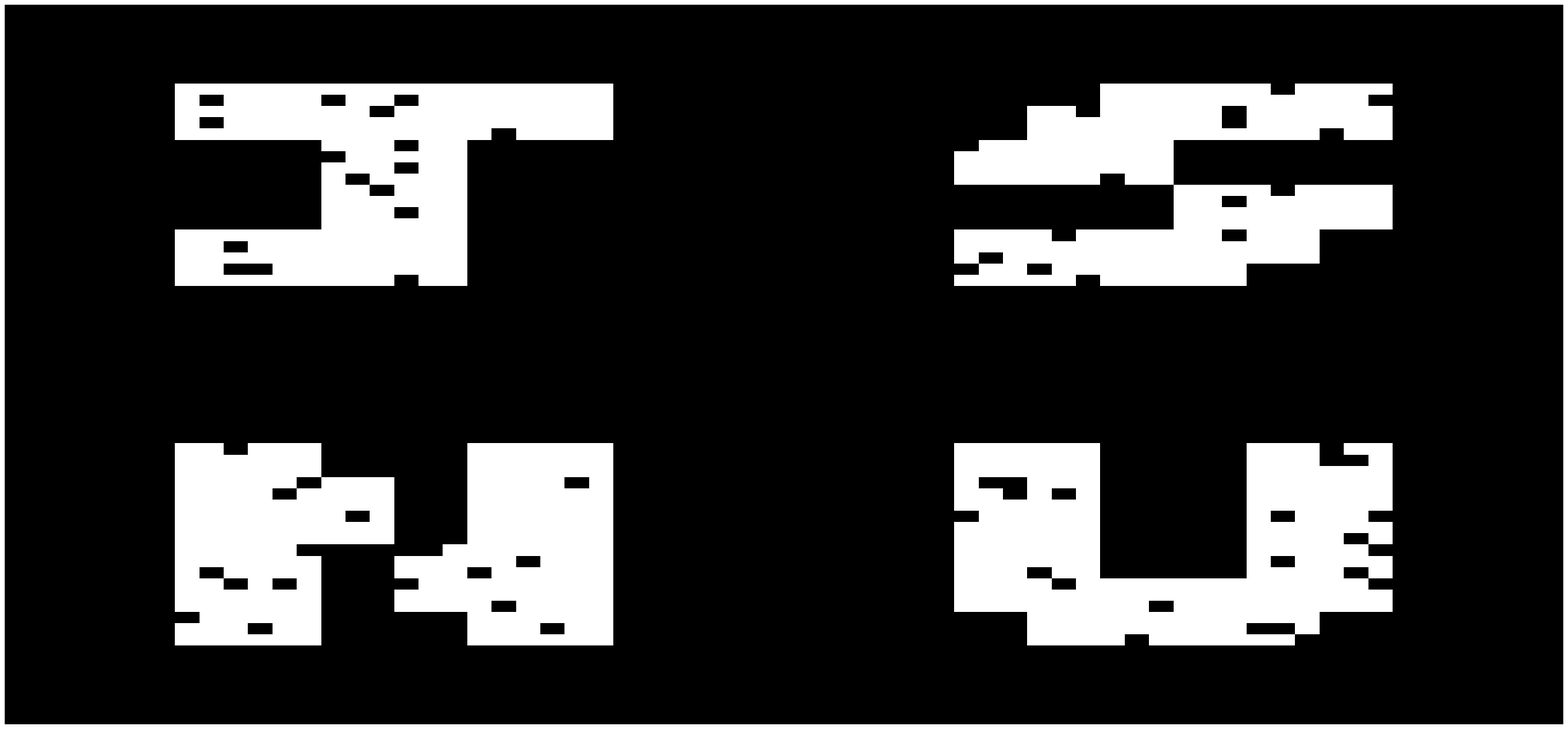} \end{minipage} &
\begin{minipage}{0.07\textwidth} \includegraphics[width=12mm, height=10mm]{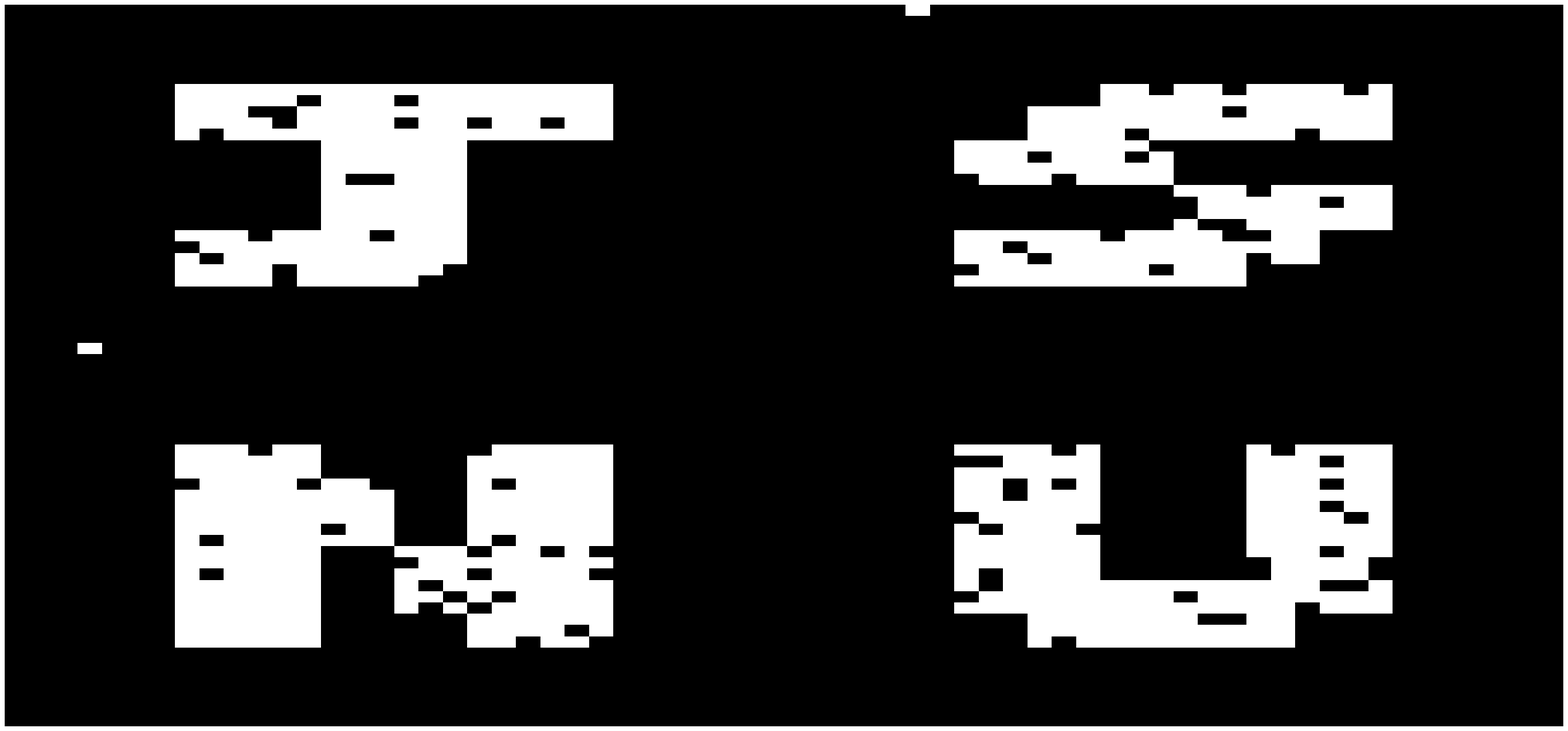} \end{minipage}&
\begin{minipage}{0.07\textwidth} \includegraphics[width=12mm, height=10mm]{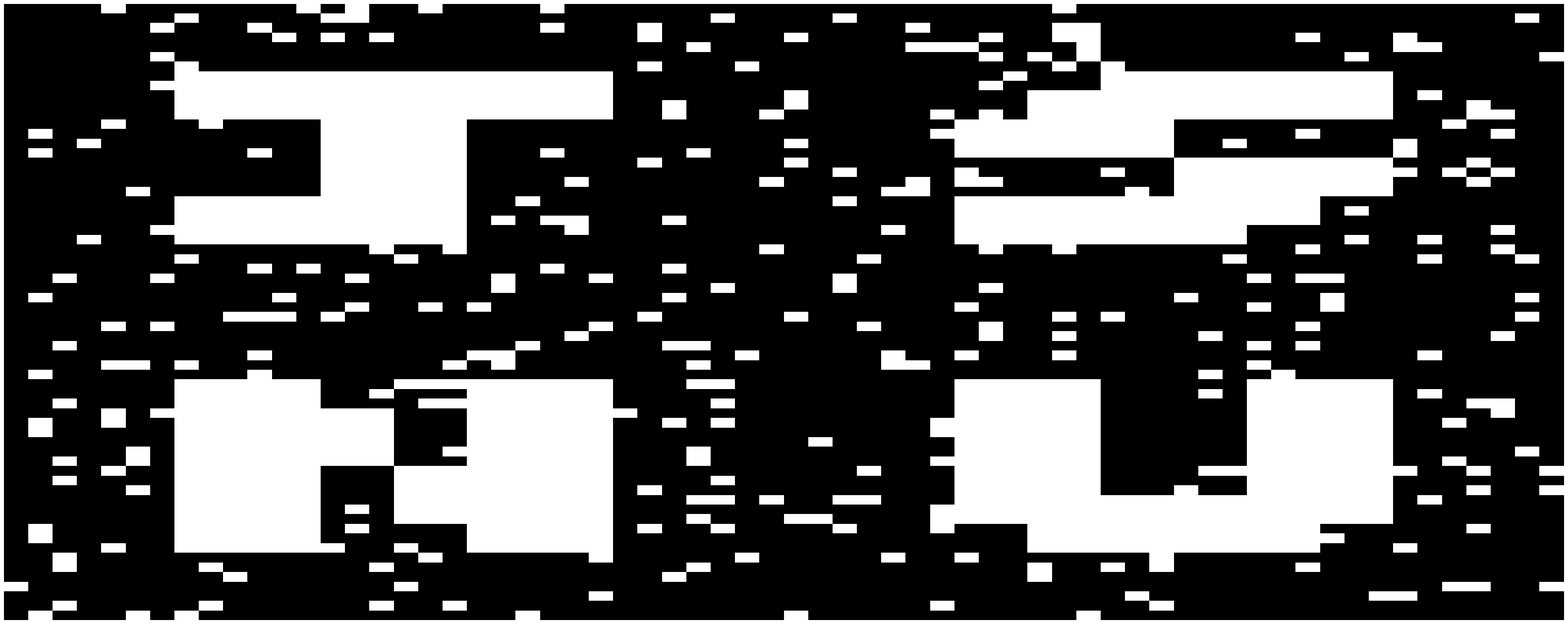} \end{minipage} &
\begin{minipage}{0.07\textwidth} \includegraphics[width=12mm, height=10mm]{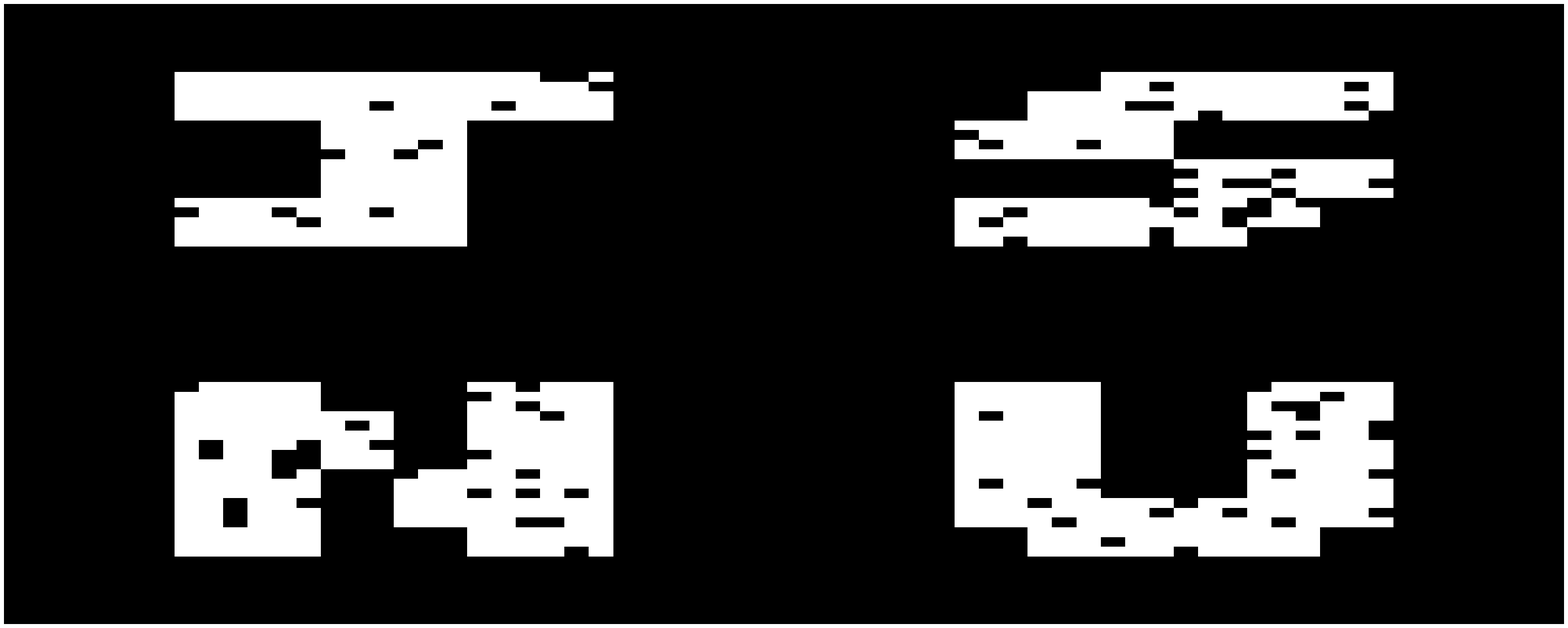} \end{minipage} &
\begin{minipage}{0.07\textwidth} \includegraphics[width=12mm, height=10mm]{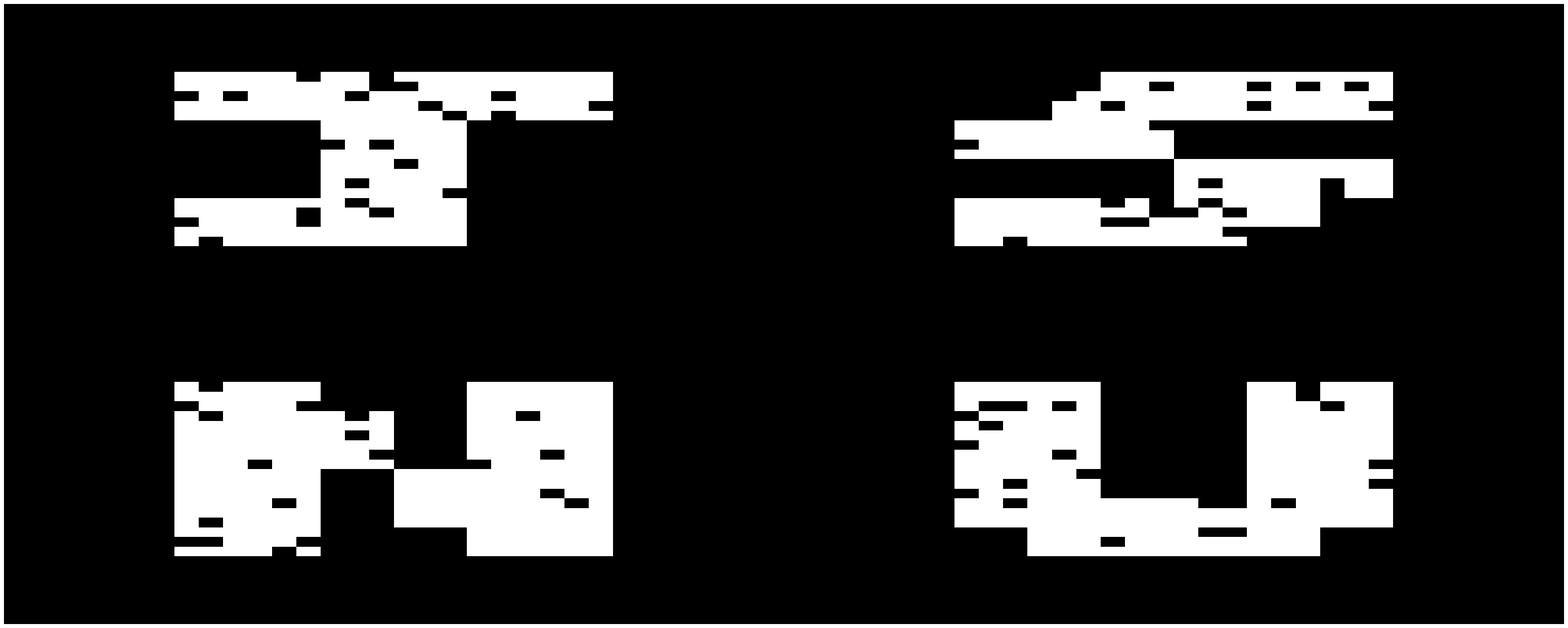} \end{minipage}\\
\cline{2-8}
& BER & 0.0723 & 0.0210 & 0.0200 & 0.0764 & 0.0227 & 0.0220\\
\cline{2-8}
&Cropping 50\% &
\begin{minipage}{0.07\textwidth} \includegraphics[width=12mm, height=10mm]{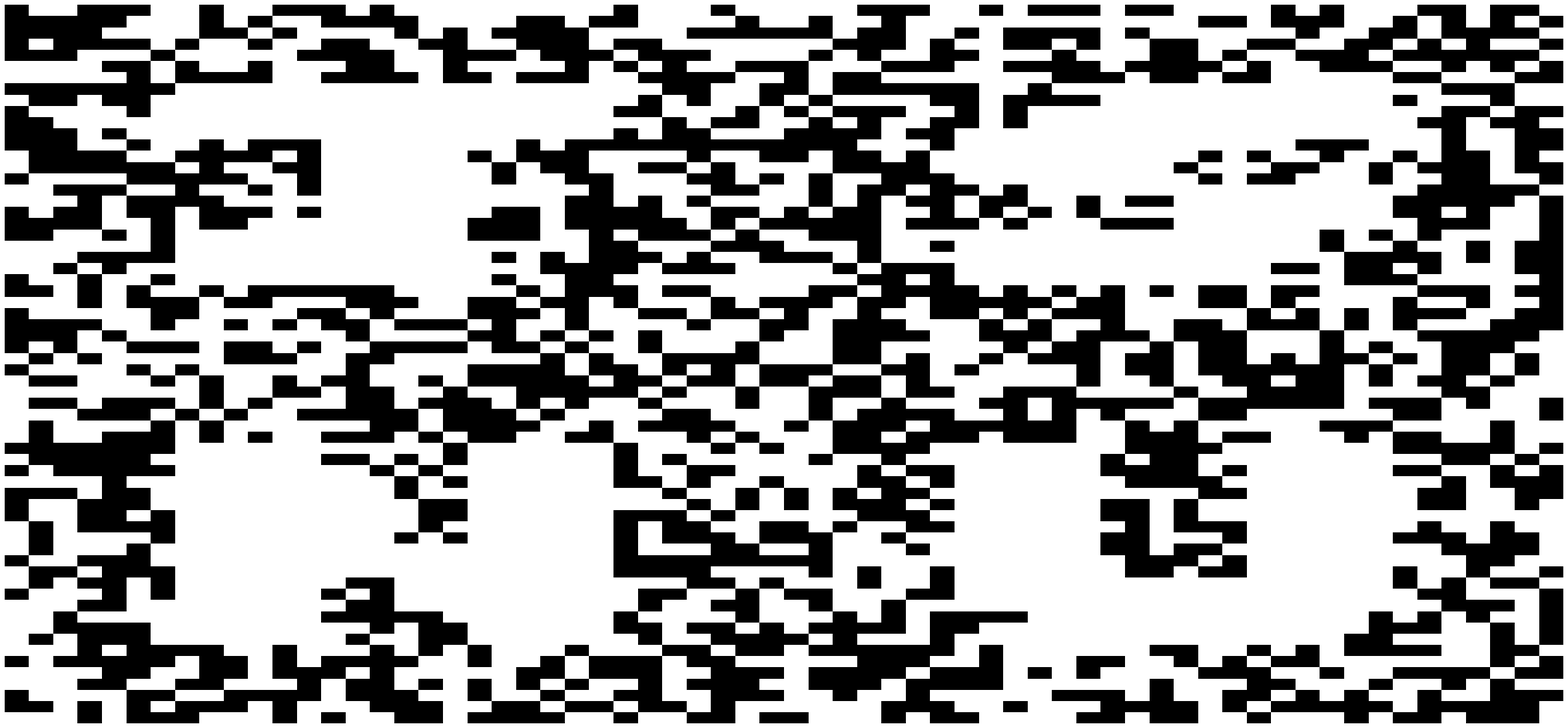} \end{minipage} &
\begin{minipage}{0.07\textwidth} \includegraphics[width=12mm, height=10mm]{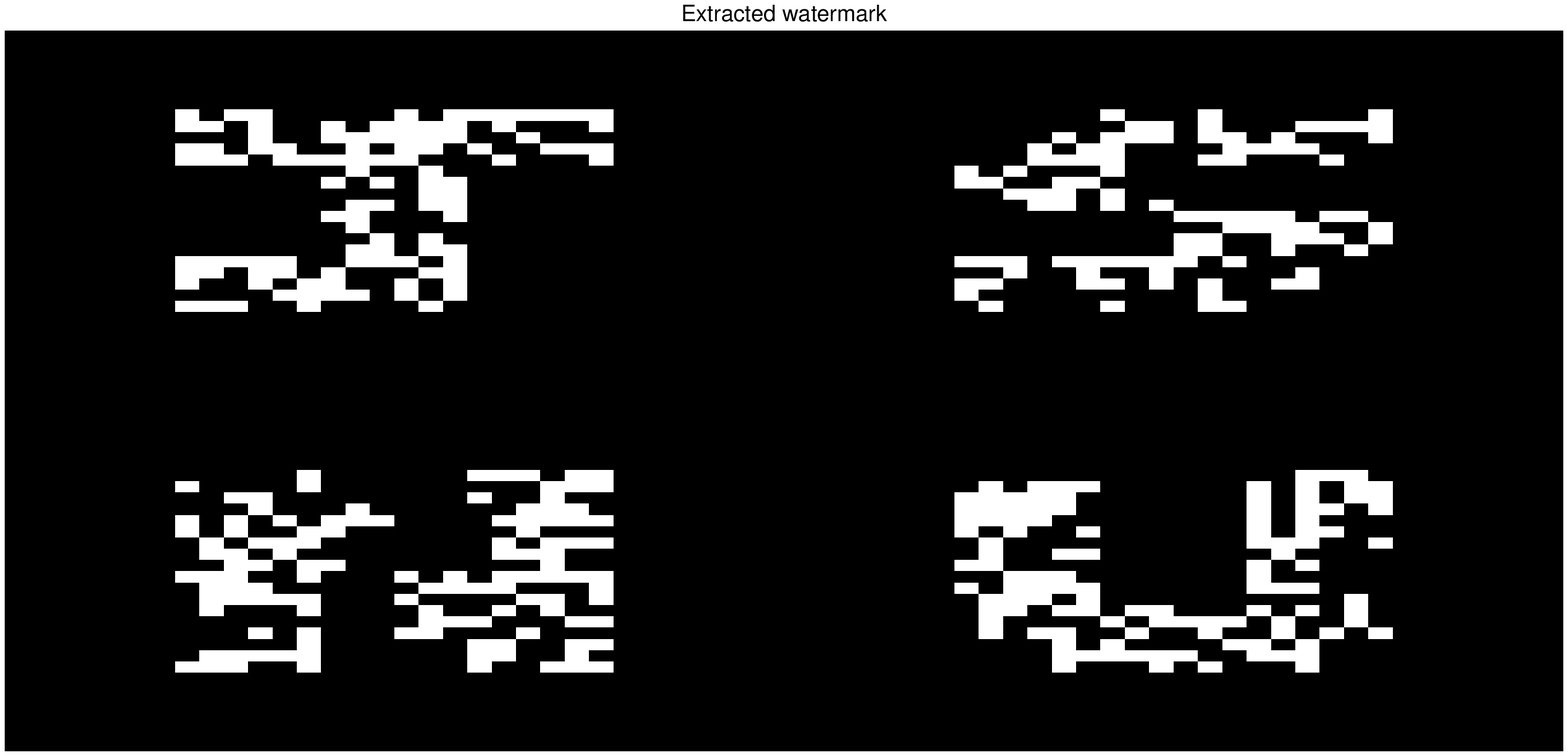} \end{minipage} &
\begin{minipage}{0.07\textwidth} \includegraphics[width=12mm, height=10mm]{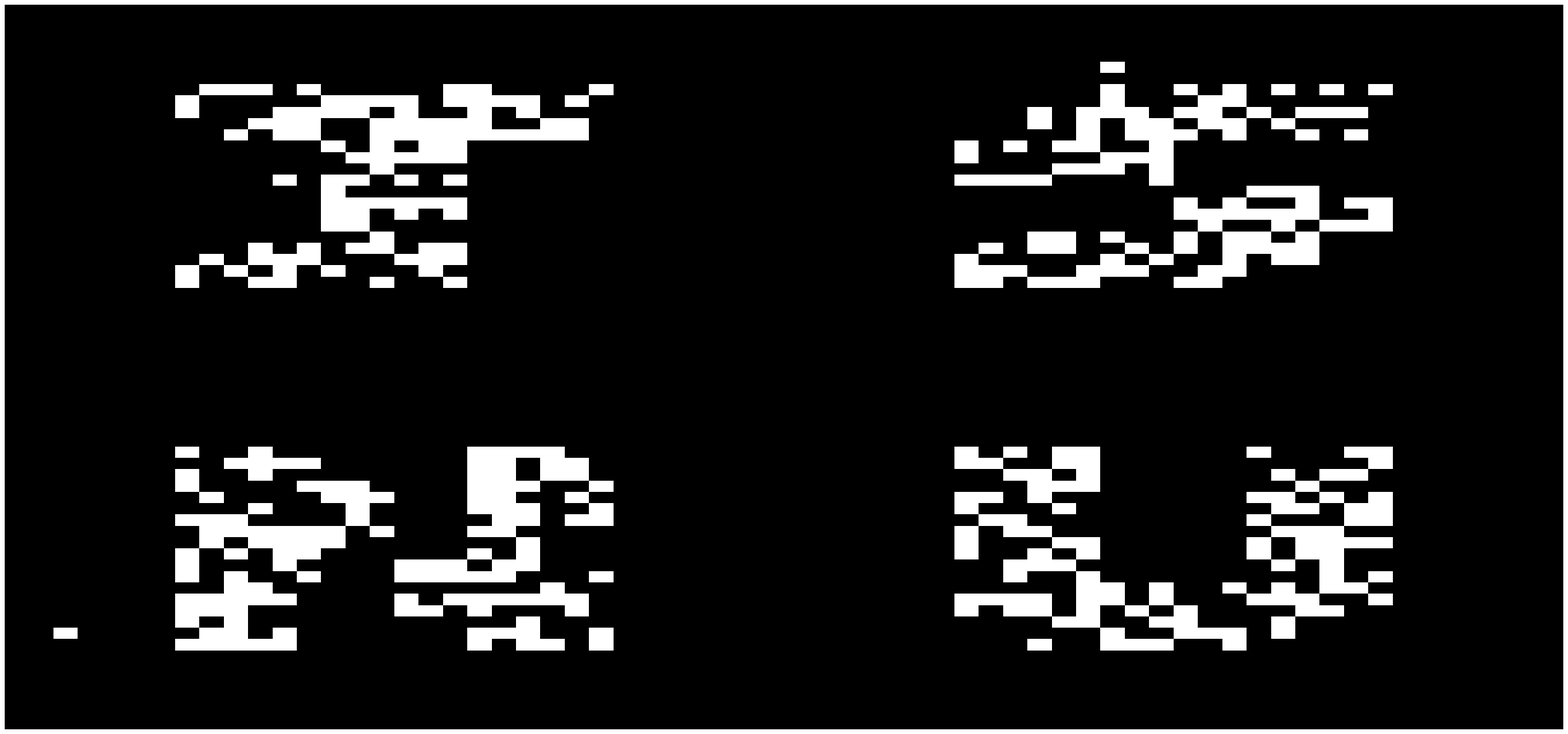} \end{minipage}&
\begin{minipage}{0.07\textwidth} \includegraphics[width=12mm, height=10mm]{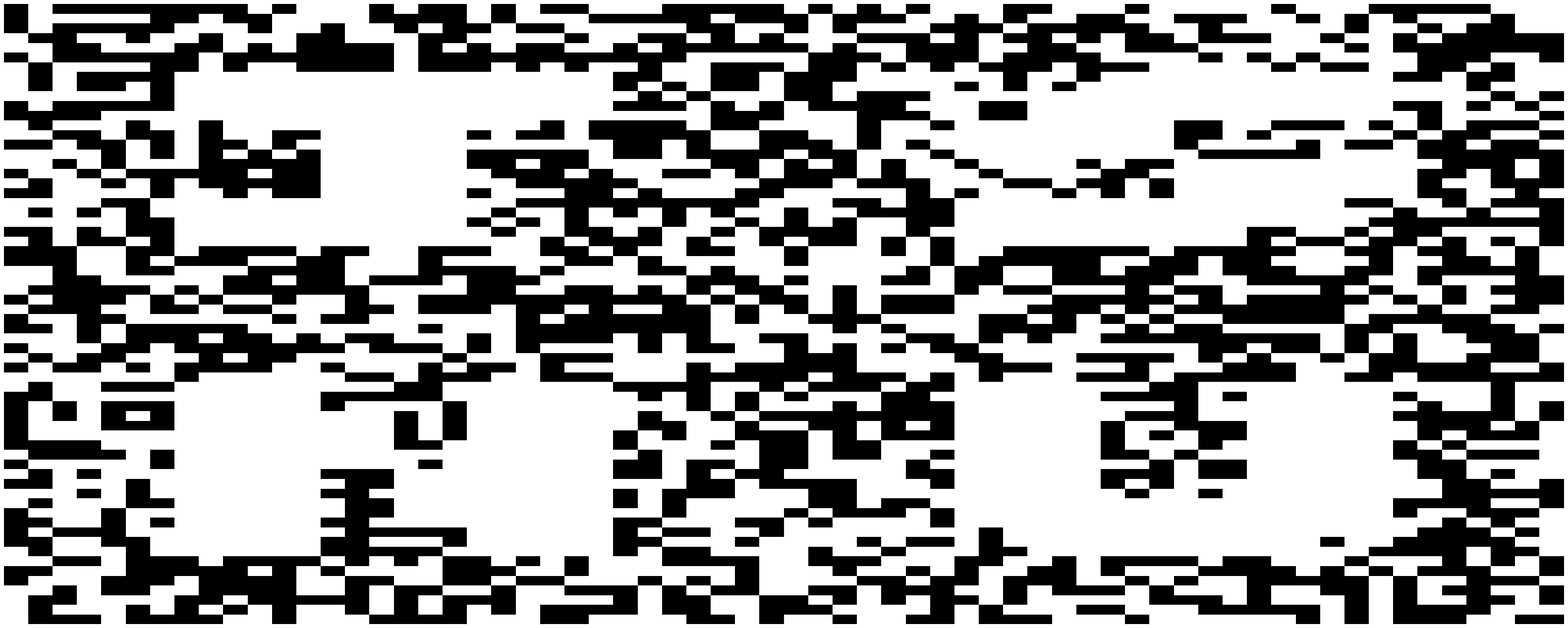} \end{minipage} &
\begin{minipage}{0.07\textwidth} \includegraphics[width=12mm, height=10mm]{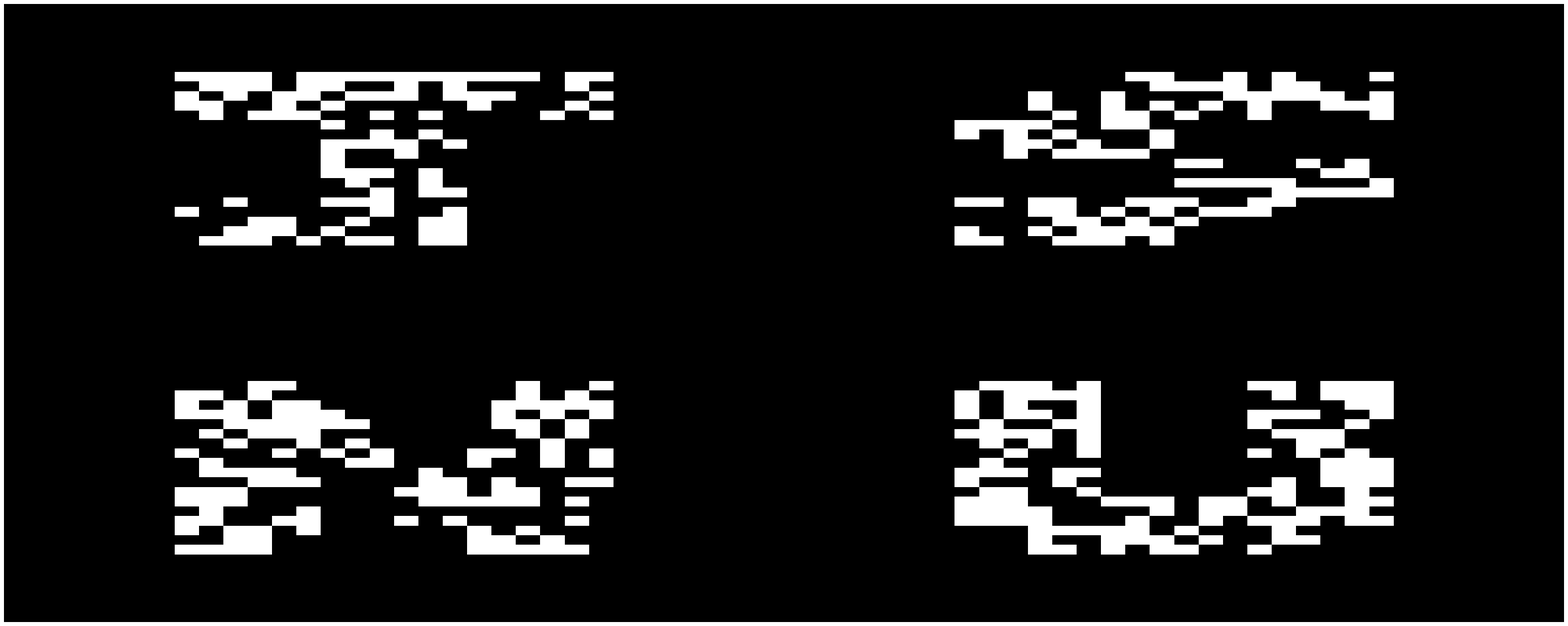} \end{minipage} &
\begin{minipage}{0.07\textwidth} \includegraphics[width=12mm, height=10mm]{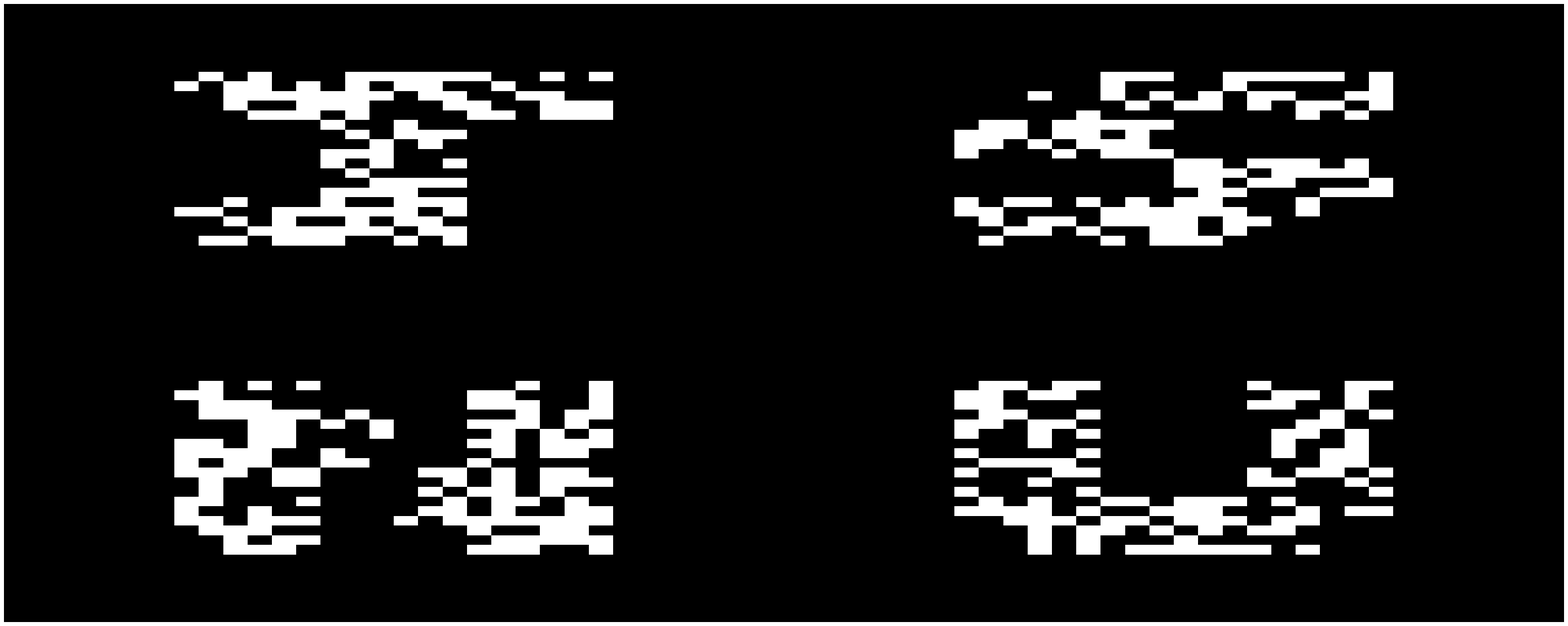} \end{minipage}\\
\cline{2-8}
& BER & 0.3916 & 0.1084 & 0.1052 & 0.3811 & 0.1096 & 0.1050\\
\midrule  
\multicolumn{2}{c}{\quad Watermarked image} &\multicolumn{3}{c}{\quad  House } & \multicolumn{3}{c}{\quad Lostlake }\\
\midrule  
\multirow{19}*{\tabincell{c}{Attack \\type}}
&JPEG20 &
\begin{minipage}{0.07\textwidth} \includegraphics[width=12mm, height=10mm]{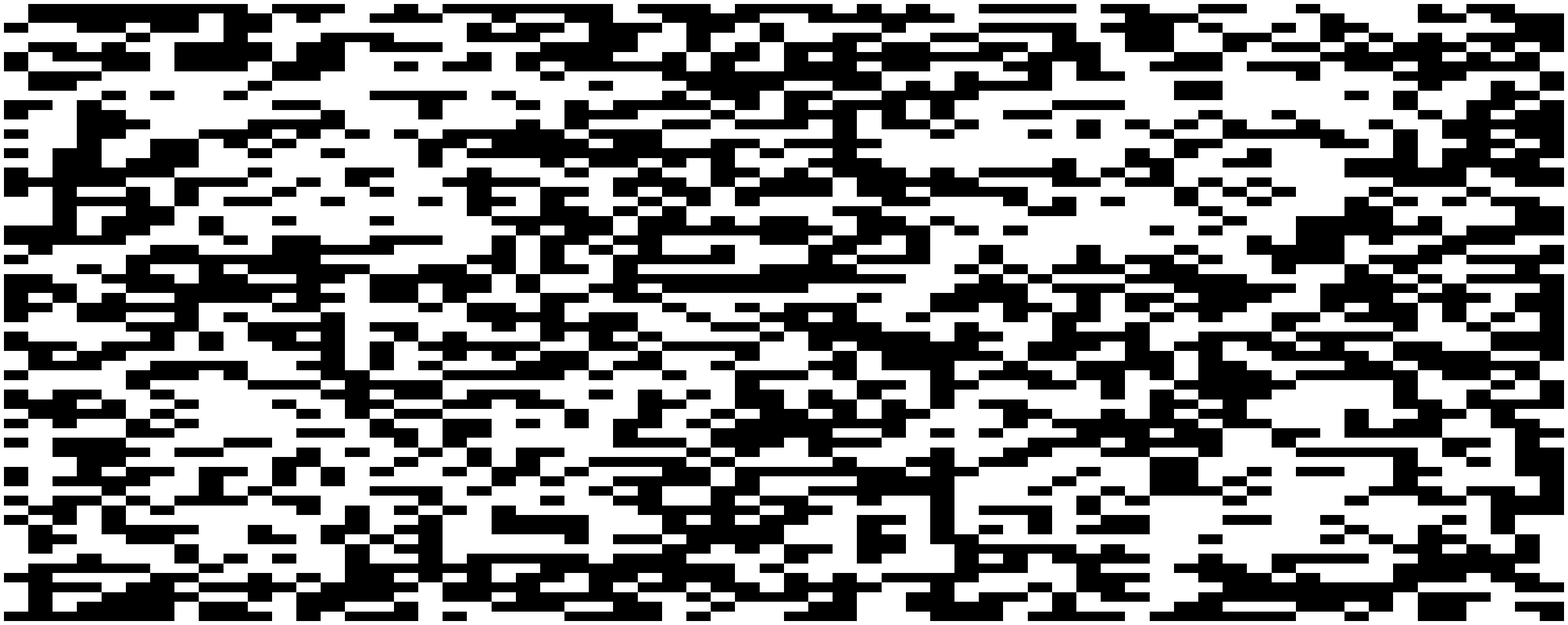} \end{minipage} &
\begin{minipage}{0.07\textwidth} \includegraphics[width=12mm, height=10mm]{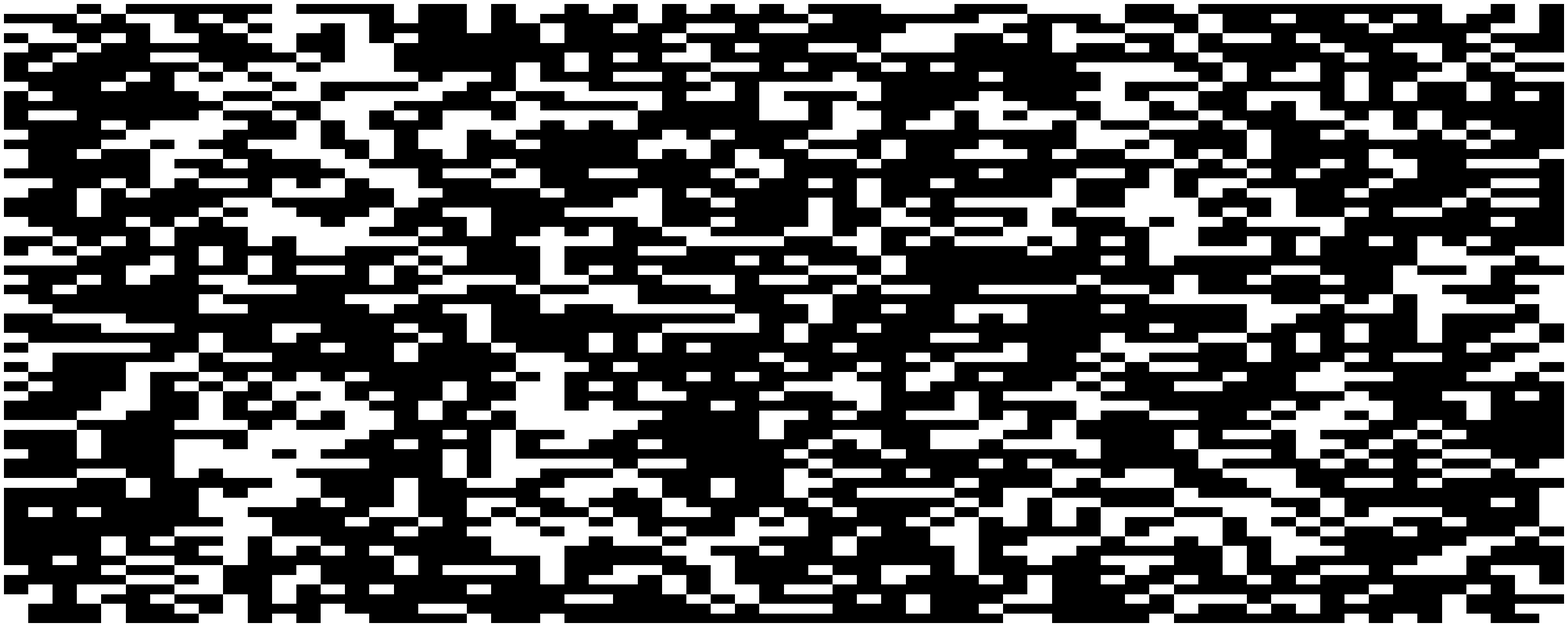} \end{minipage} &
\begin{minipage}{0.07\textwidth} \includegraphics[width=12mm, height=10mm]{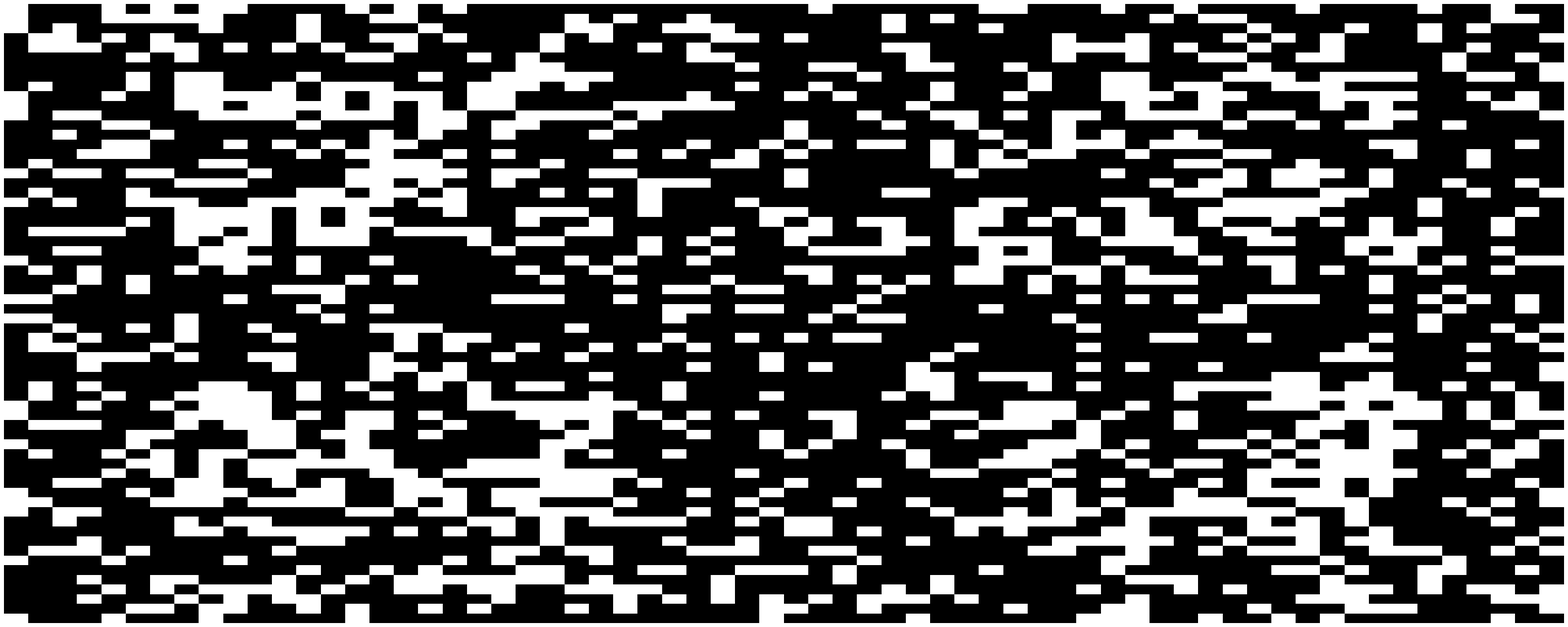} \end{minipage}&
\begin{minipage}{0.07\textwidth} \includegraphics[width=12mm, height=10mm]{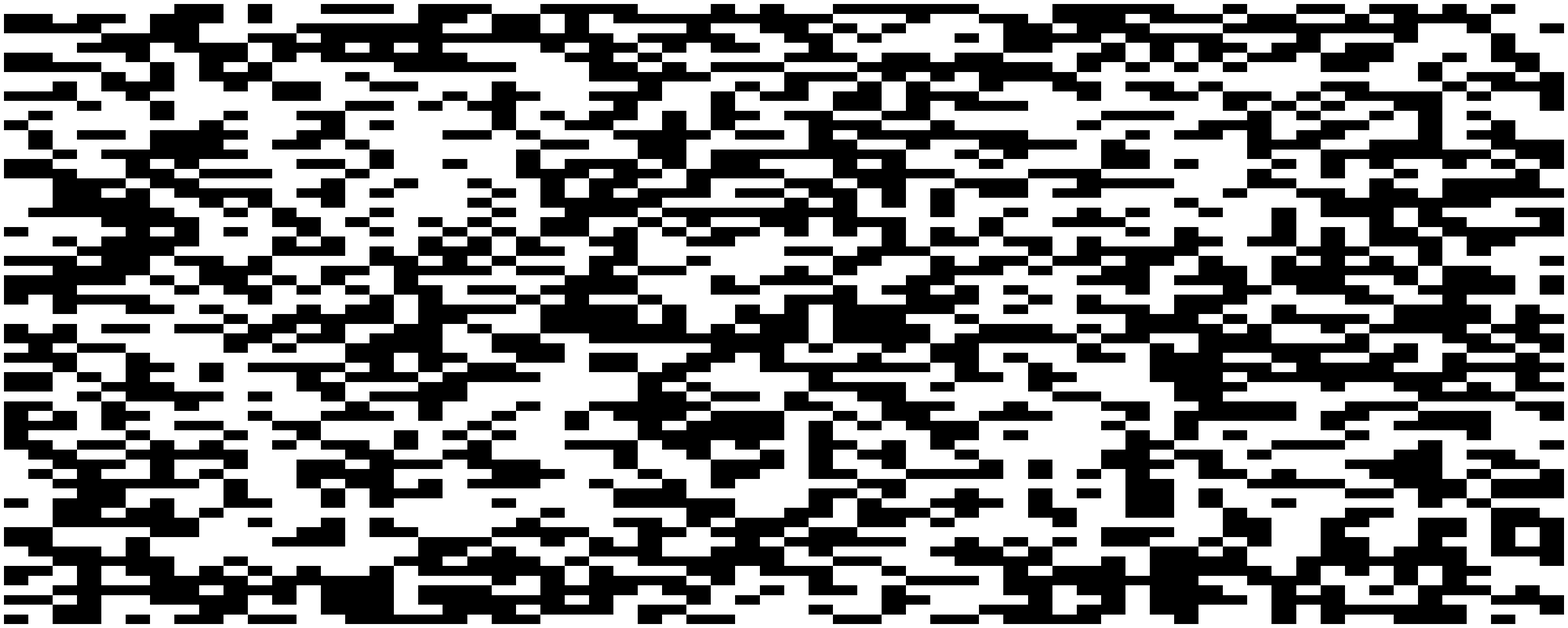} \end{minipage} &
\begin{minipage}{0.07\textwidth} \includegraphics[width=12mm, height=10mm]{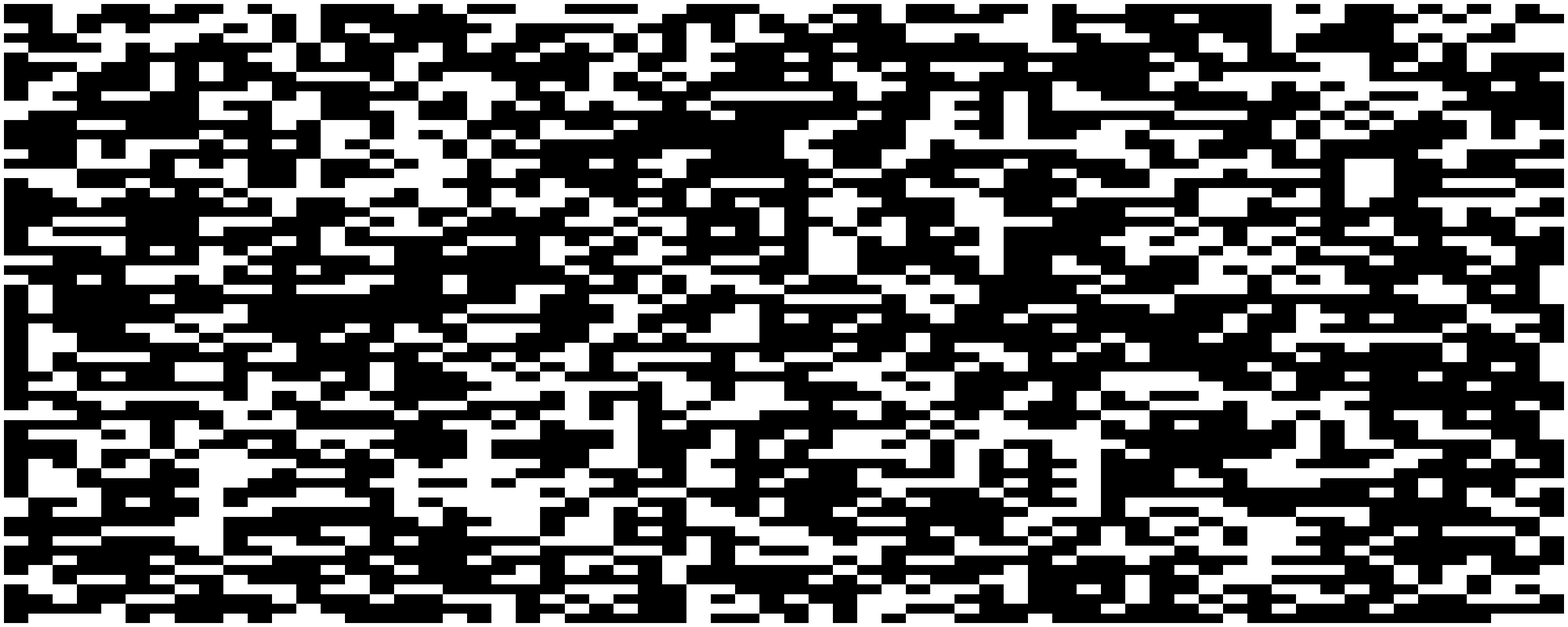} \end{minipage} &
\begin{minipage}{0.07\textwidth} \includegraphics[width=12mm, height=10mm]{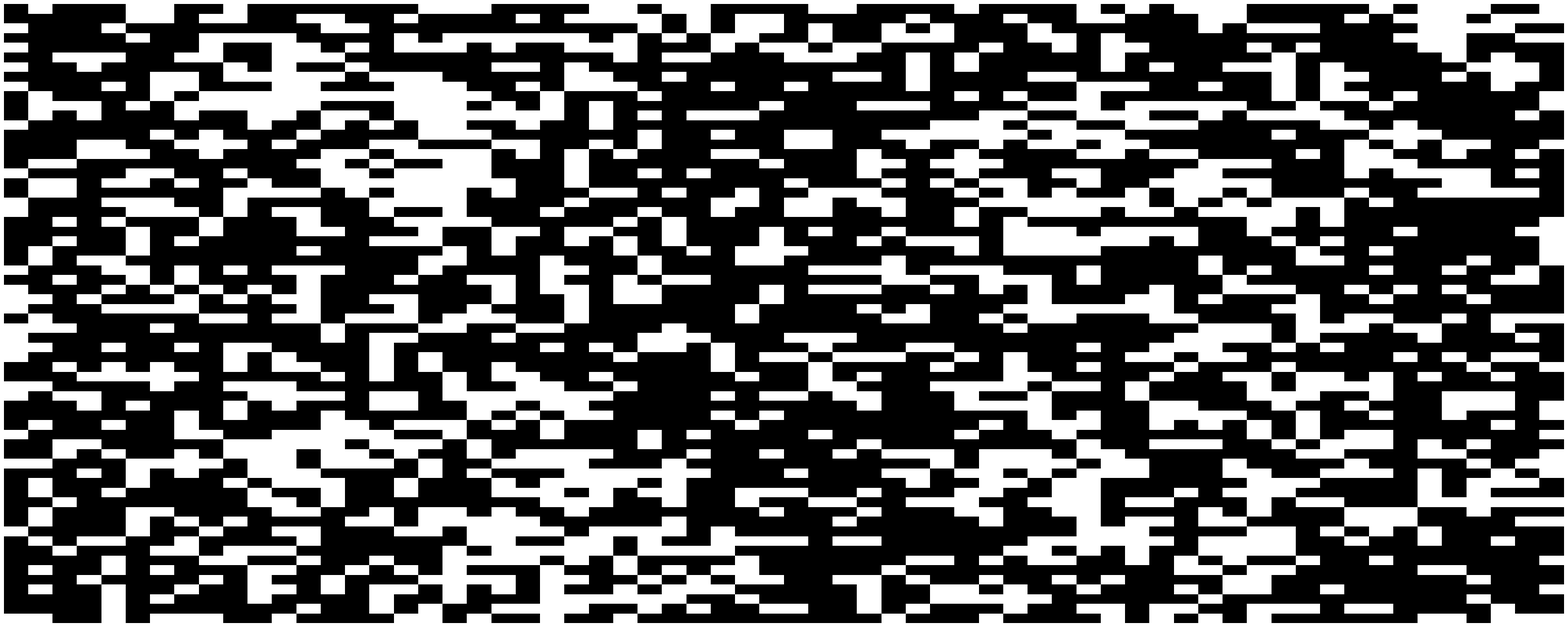} \end{minipage}\\
\cline{2-8}
& BER & 0.4209 & 0.3518 & 0.3042 & 0.4397 & 0.3892 & 0.3601\\
\cline{2-8}
&Motion blur (6,6) &
\begin{minipage}{0.07\textwidth} \includegraphics[width=12mm, height=10mm]{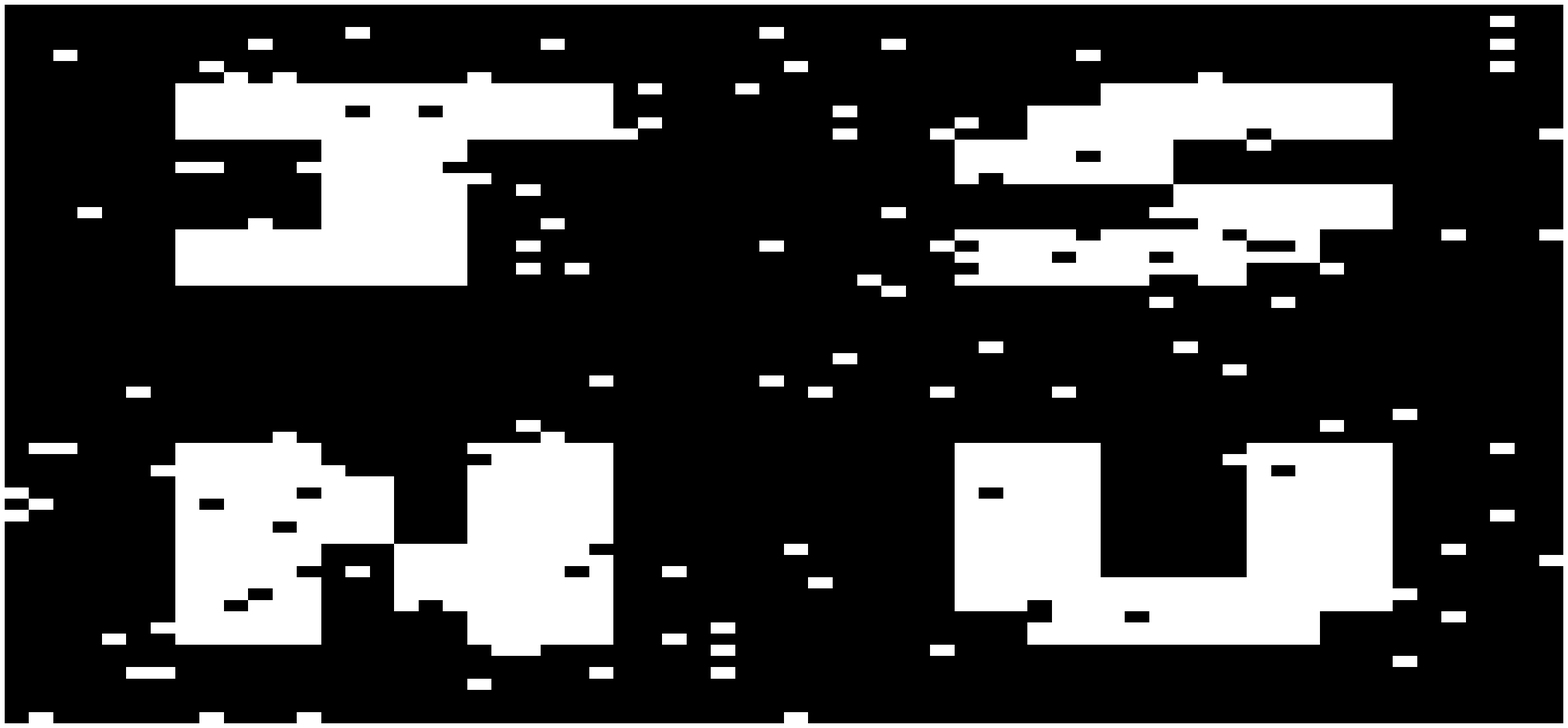} \end{minipage} &
\begin{minipage}{0.07\textwidth} \includegraphics[width=12mm, height=10mm]{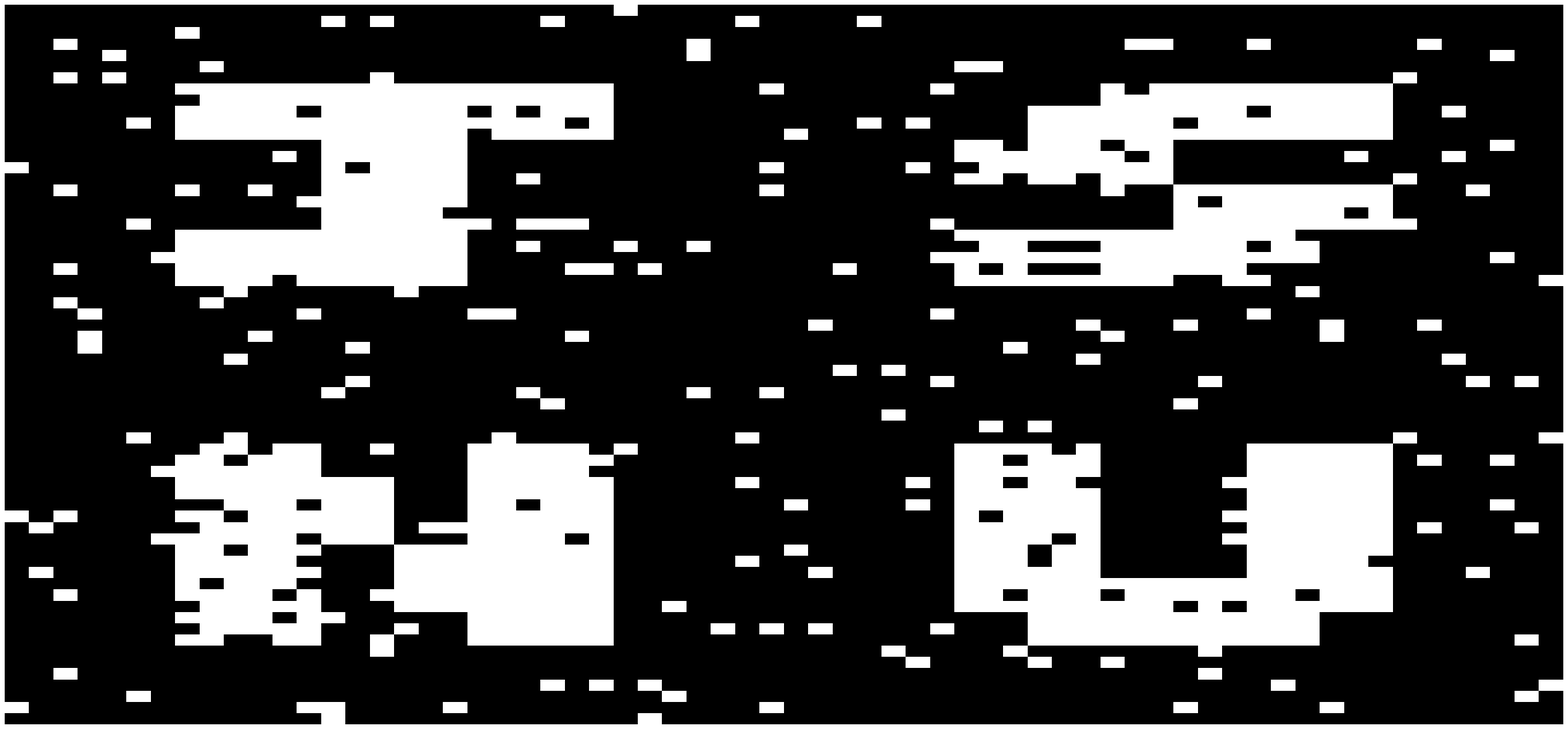} \end{minipage} &
\begin{minipage}{0.07\textwidth} \includegraphics[width=12mm, height=10mm]{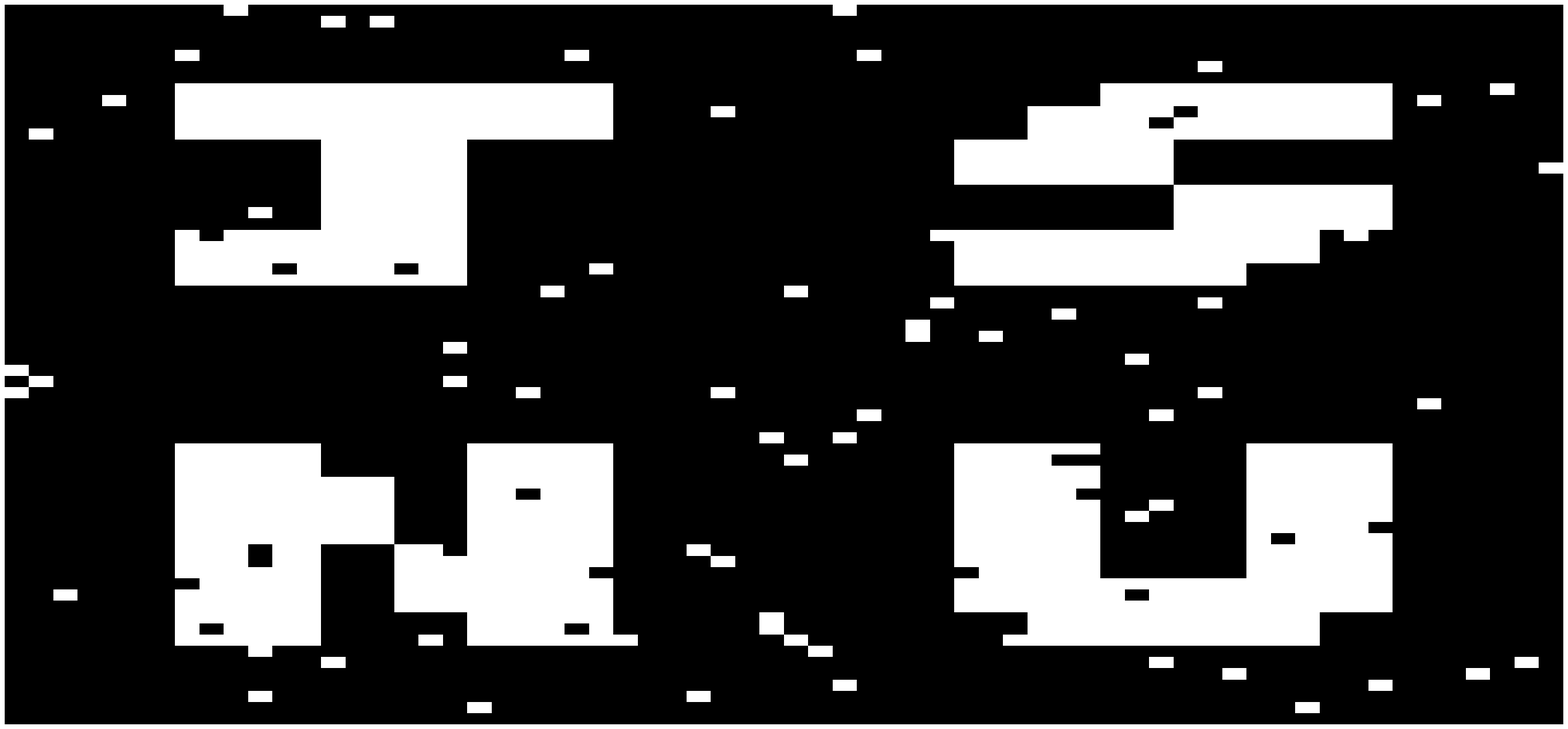} \end{minipage}&
\begin{minipage}{0.07\textwidth} \includegraphics[width=12mm, height=10mm]{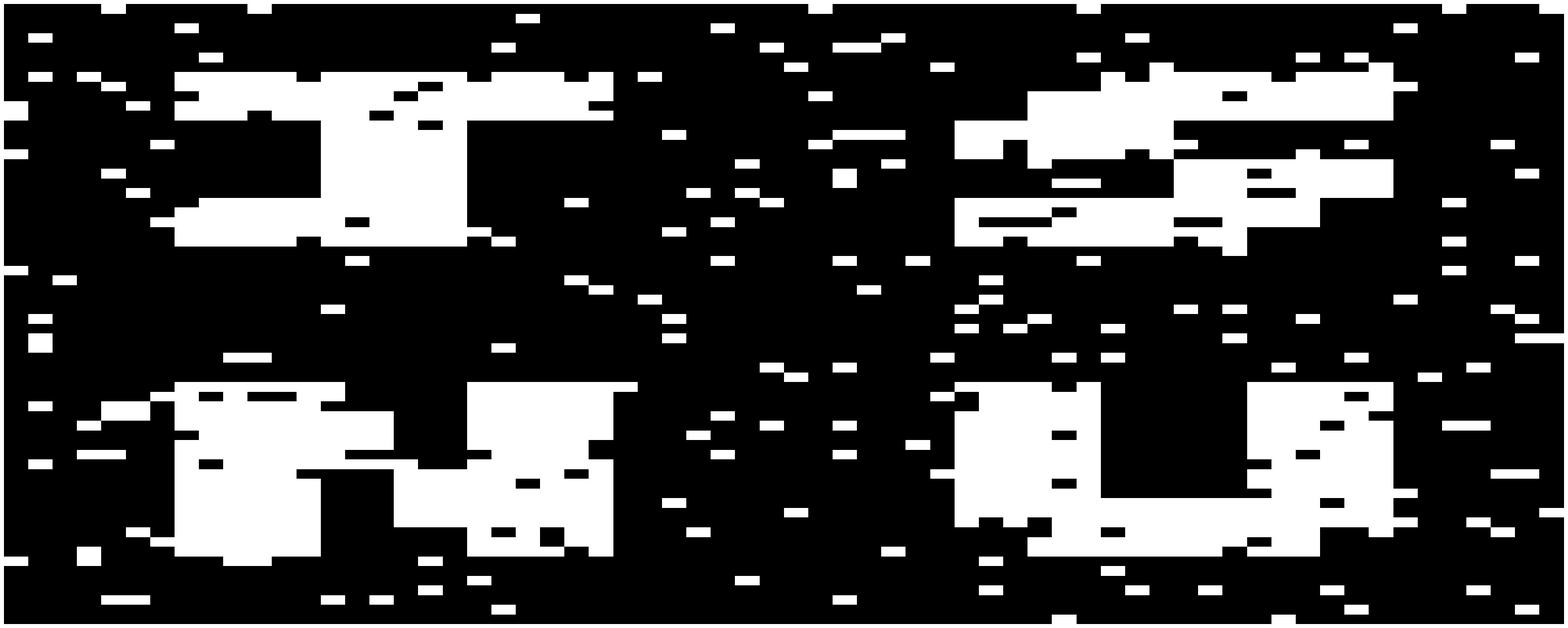} \end{minipage} &
\begin{minipage}{0.07\textwidth} \includegraphics[width=12mm, height=10mm]{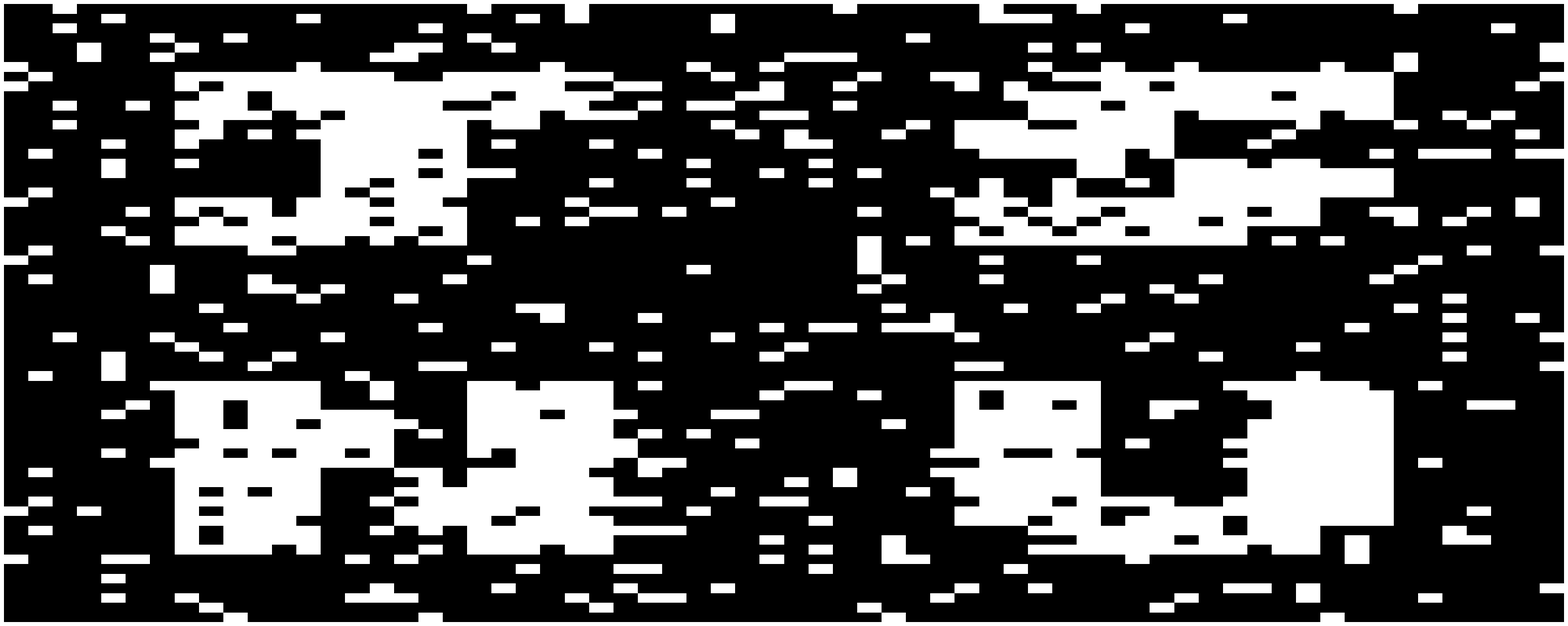} \end{minipage} &
\begin{minipage}{0.07\textwidth} \includegraphics[width=12mm, height=10mm]{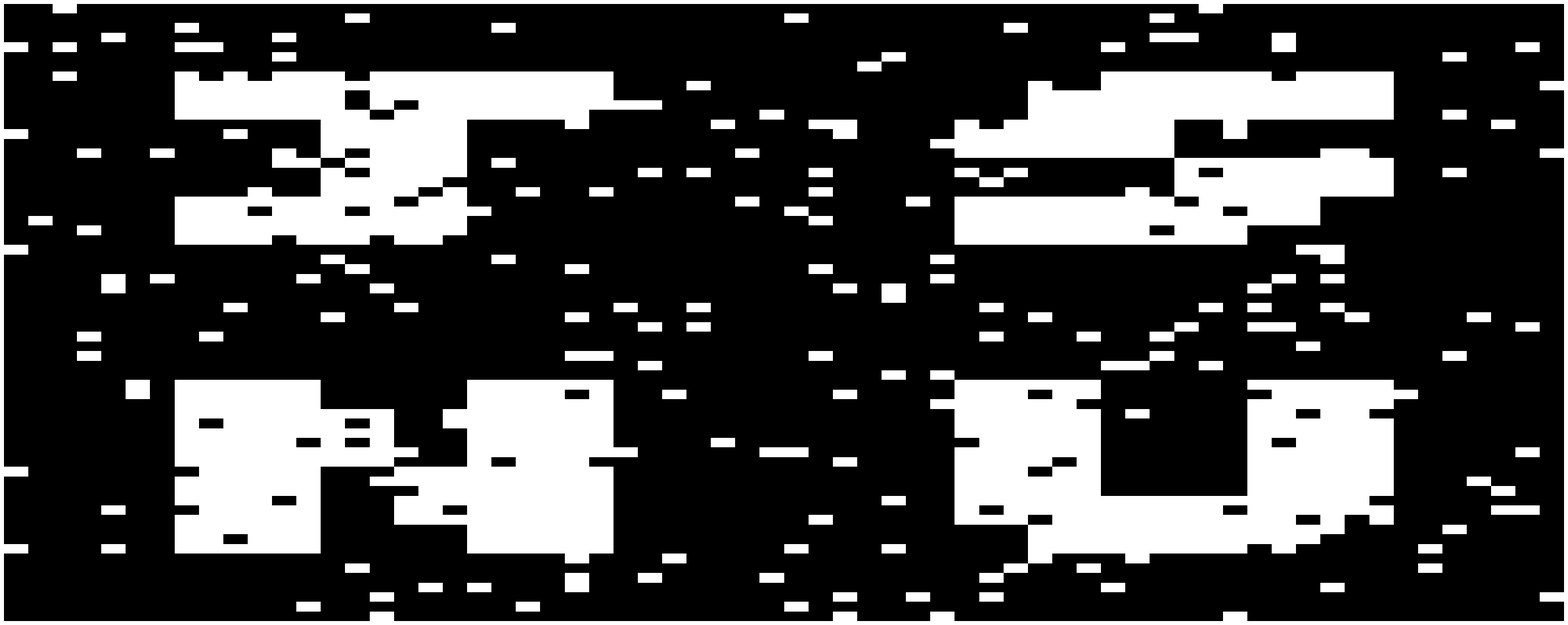} \end{minipage}\\
\cline{2-8}
& BER & 0.0320 & 0.0603 & 0.0208 & 0.0601 & 0.1160 & 0.0603\\
\cline{2-8}
&Scaling-4 &
\begin{minipage}{0.07\textwidth} \includegraphics[width=12mm, height=10mm]{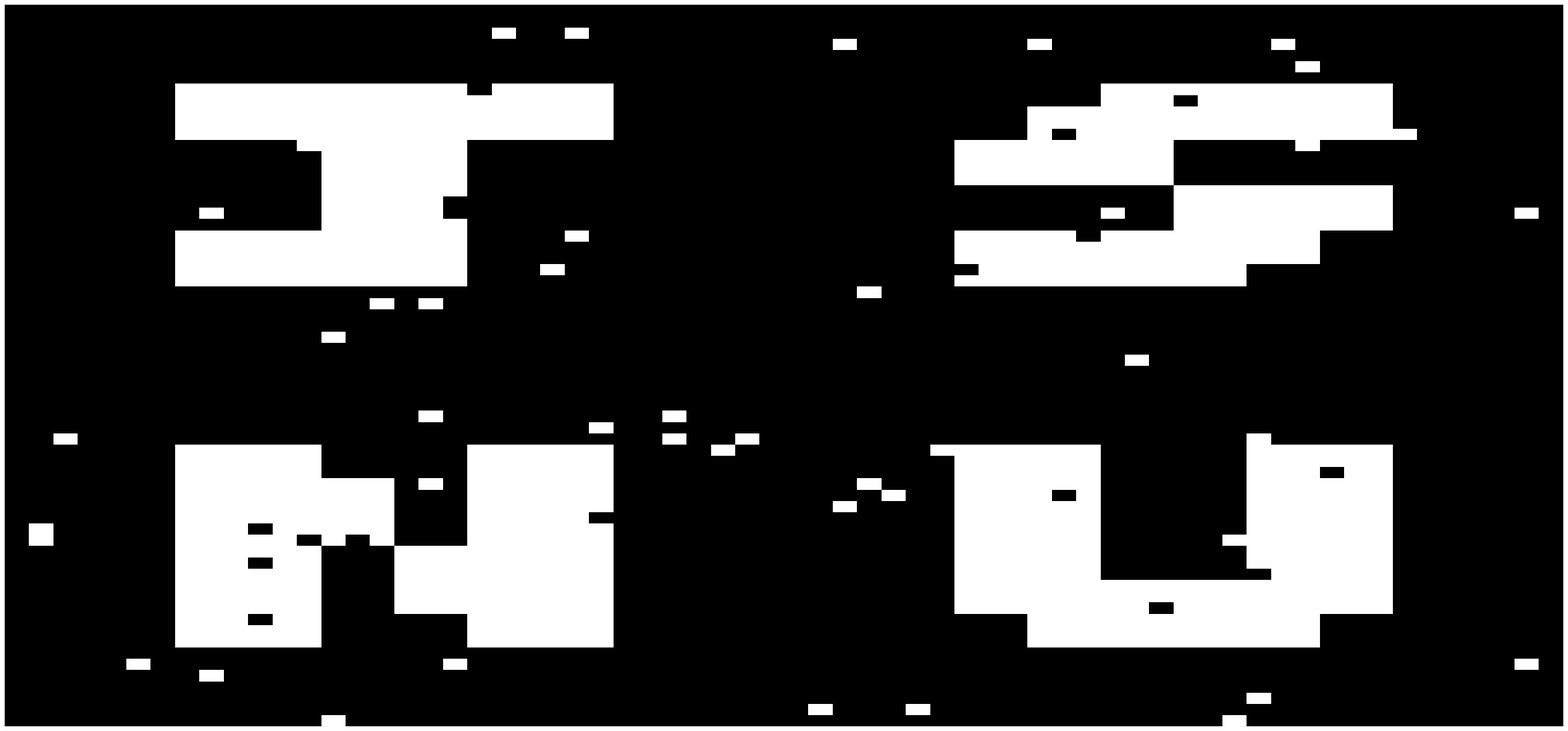} \end{minipage} &
\begin{minipage}{0.07\textwidth} \includegraphics[width=12mm, height=10mm]{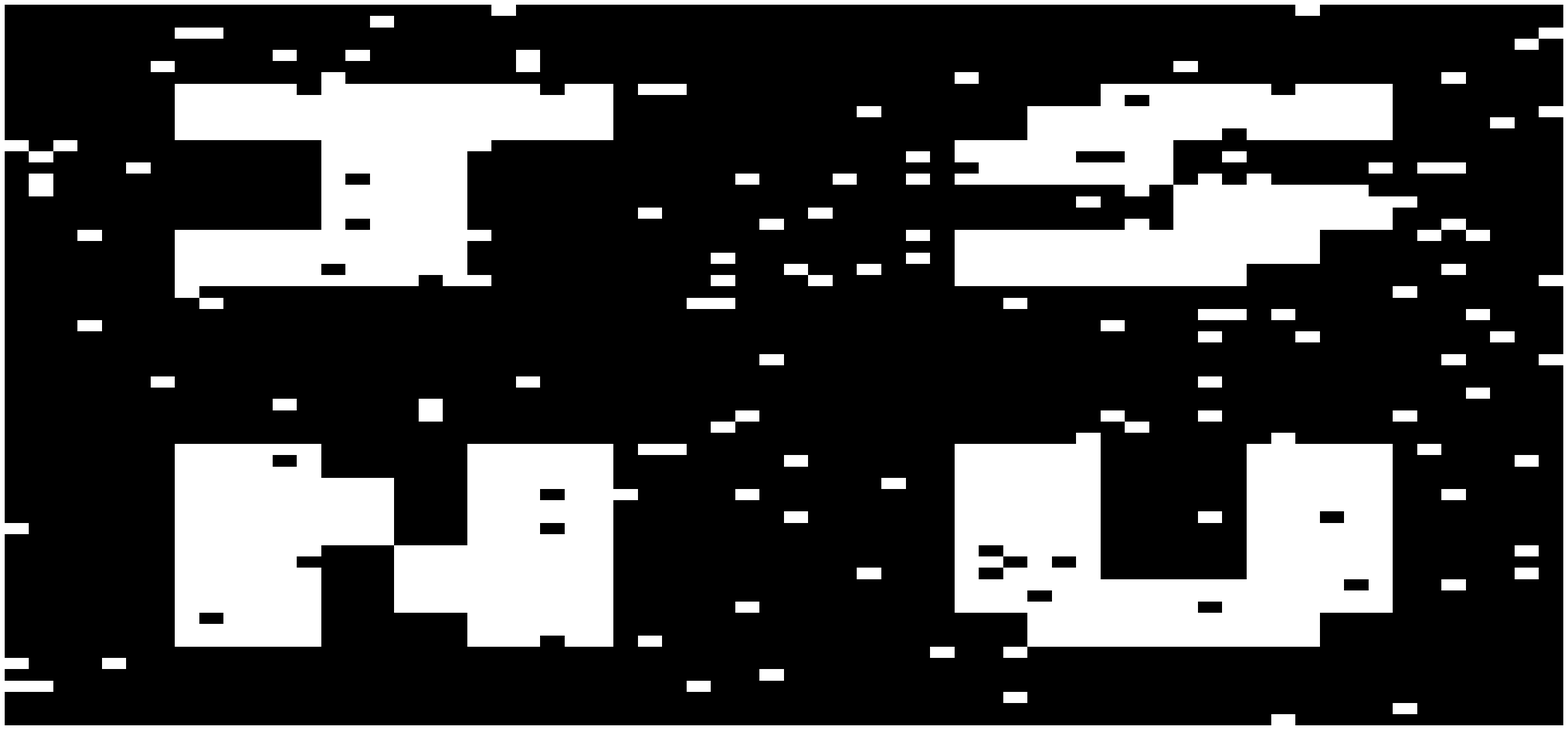} \end{minipage} &
\begin{minipage}{0.07\textwidth} \includegraphics[width=12mm, height=10mm]{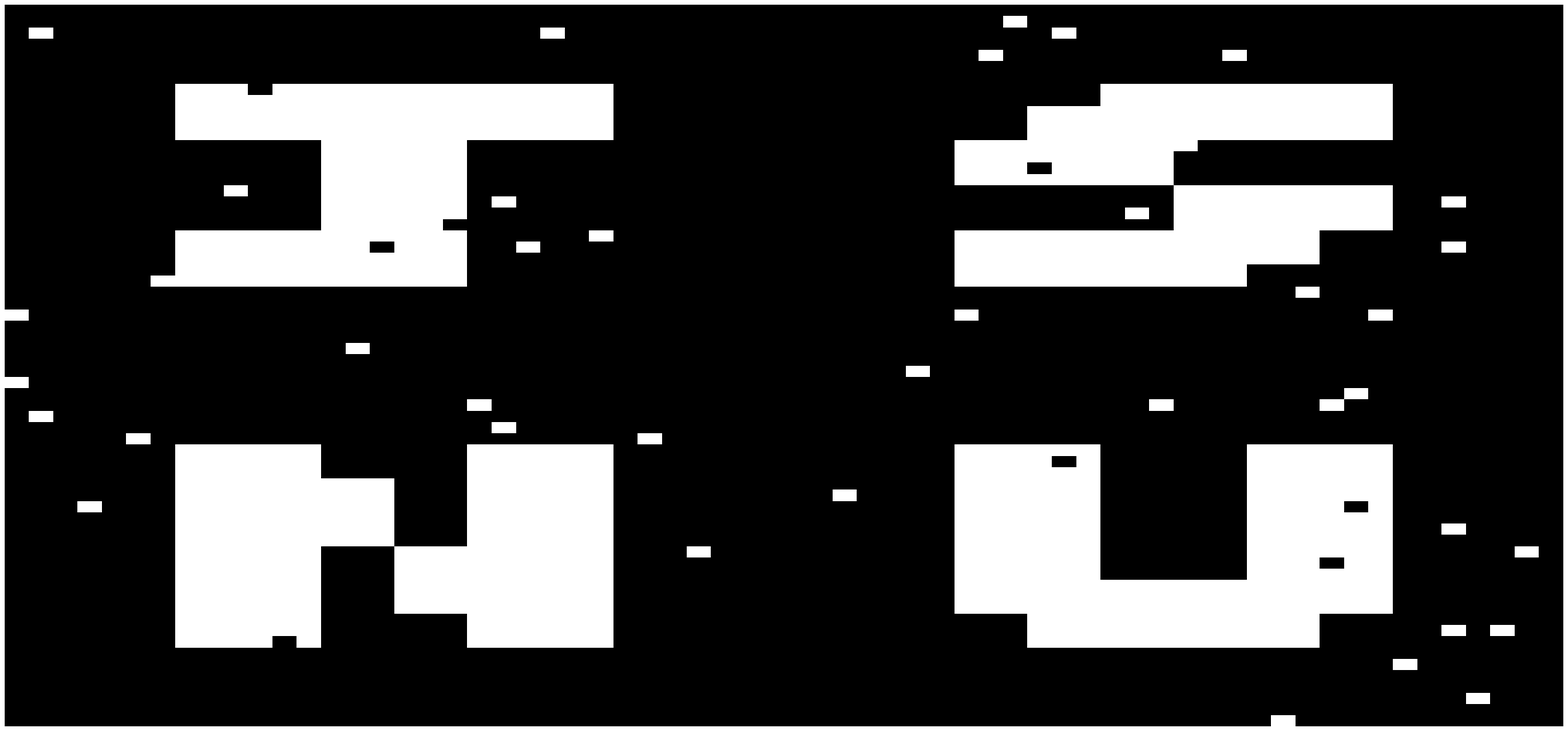} \end{minipage}&
\begin{minipage}{0.07\textwidth} \includegraphics[width=12mm, height=10mm]{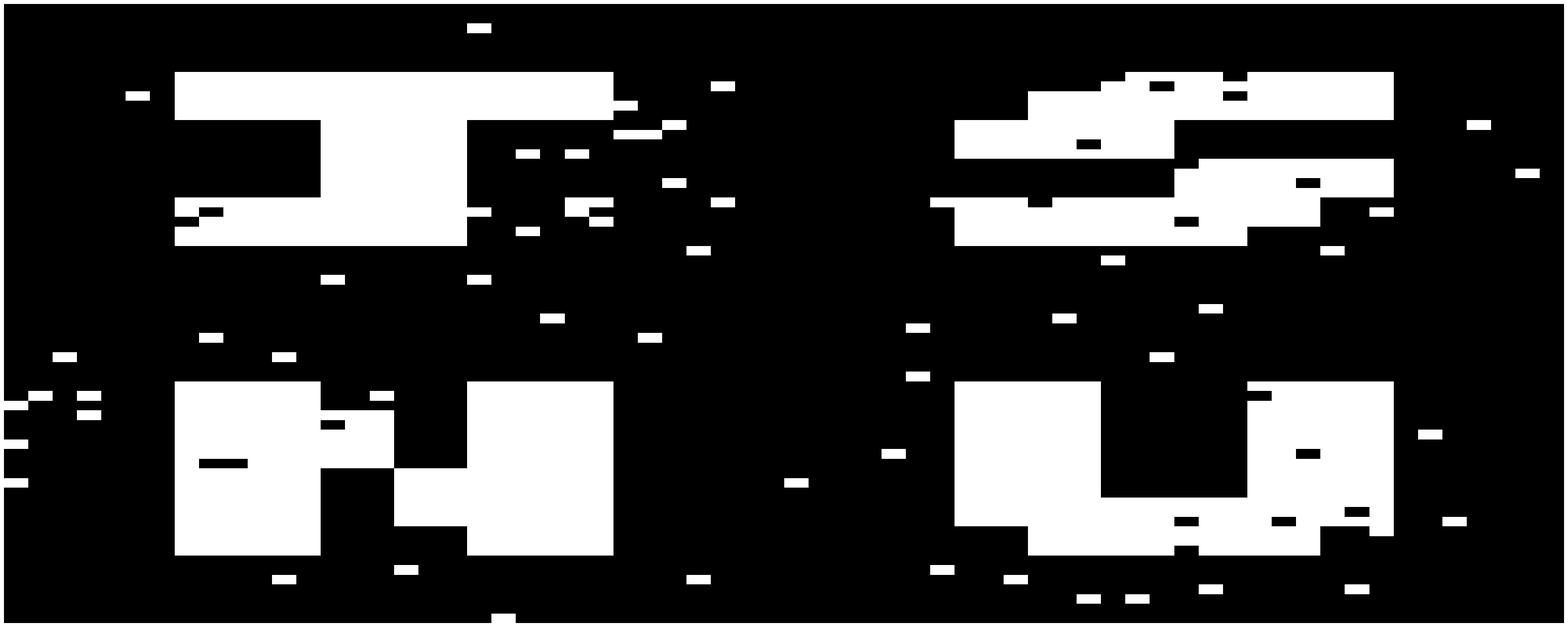} \end{minipage} &
\begin{minipage}{0.07\textwidth} \includegraphics[width=12mm, height=10mm]{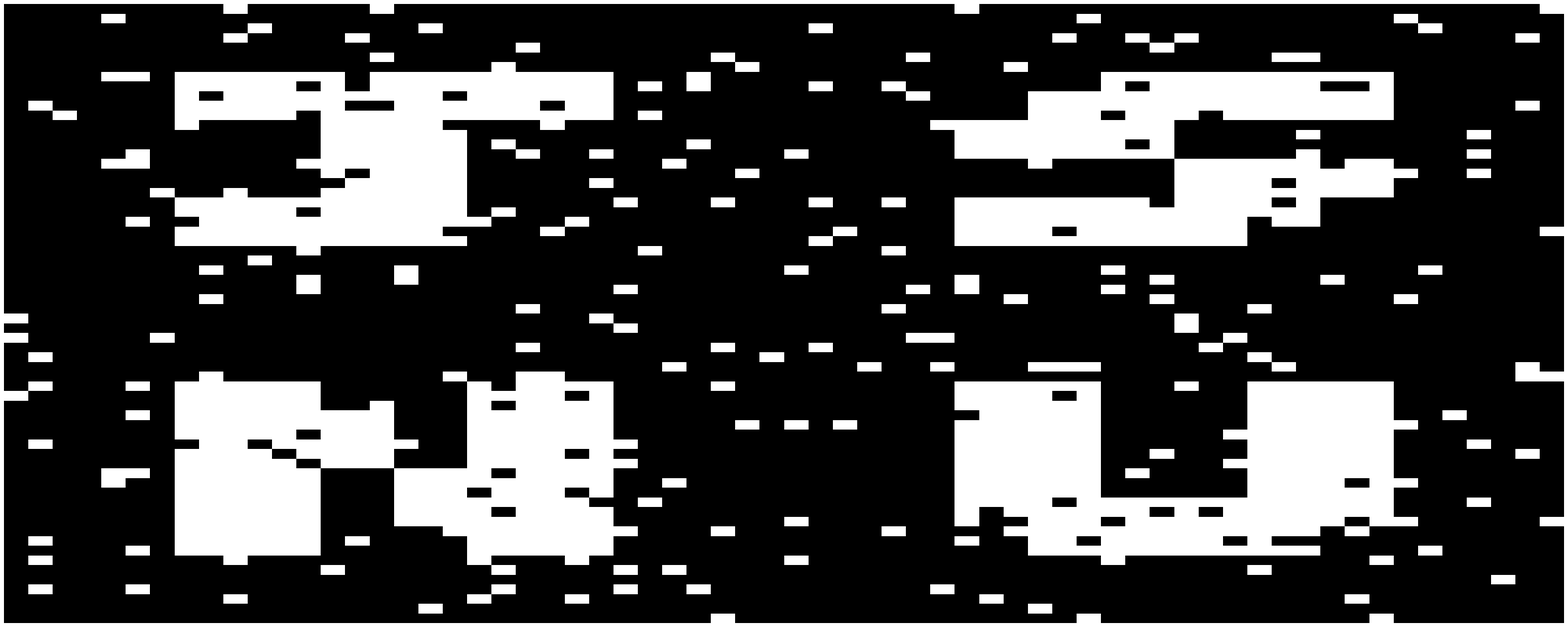} \end{minipage} &
\begin{minipage}{0.07\textwidth} \includegraphics[width=12mm, height=10mm]{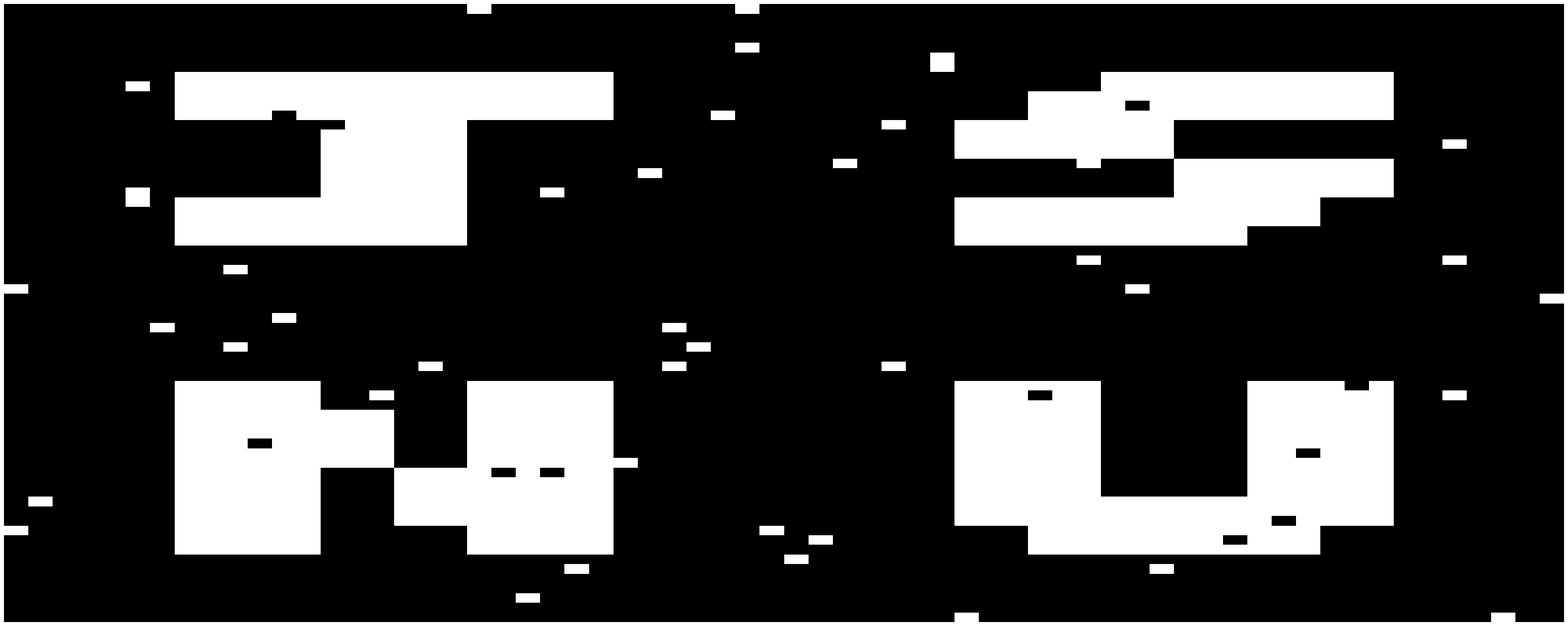} \end{minipage}\\
\cline{2-8}
& BER & 0.0149 & 0.0364 & 0.0117 & 0.0190 & 0.0635 & 0.0129\\
\cline{2-8}
&Cropping 30\%&
\begin{minipage}{0.07\textwidth} \includegraphics[width=12mm, height=10mm]{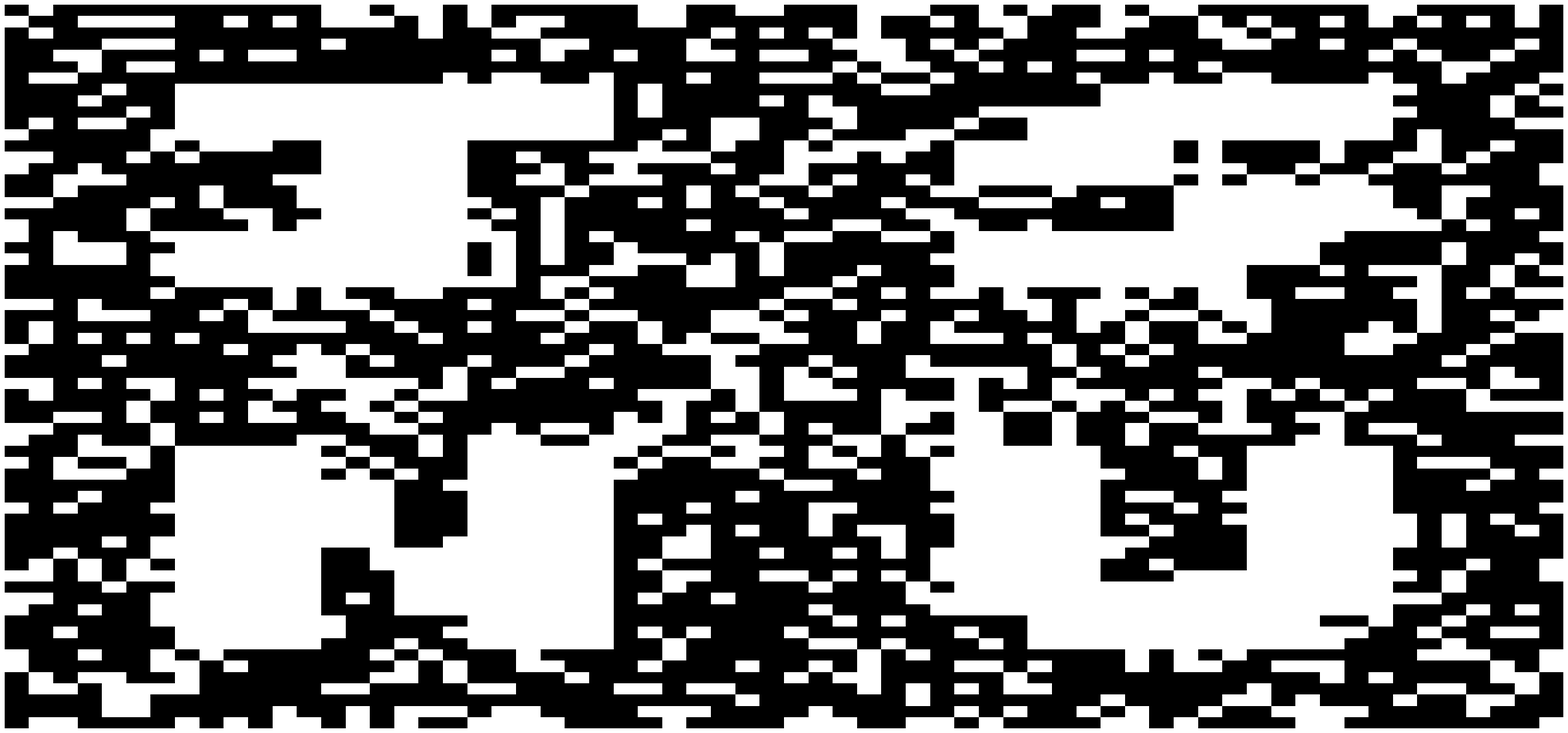} \end{minipage}  &
\begin{minipage}{0.07\textwidth} \includegraphics[width=12mm, height=10mm]{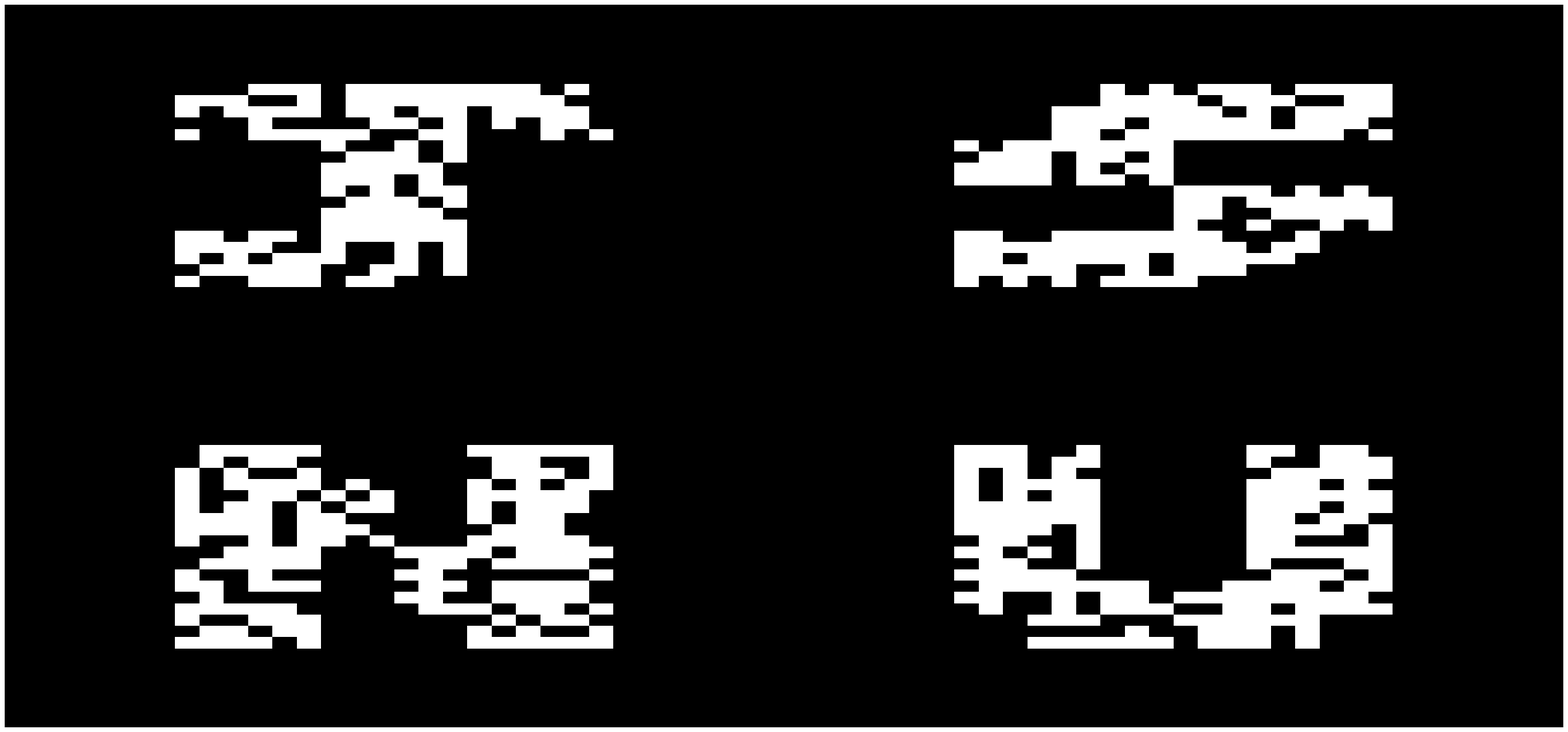} \end{minipage} &
\begin{minipage}{0.07\textwidth} \includegraphics[width=12mm, height=10mm]{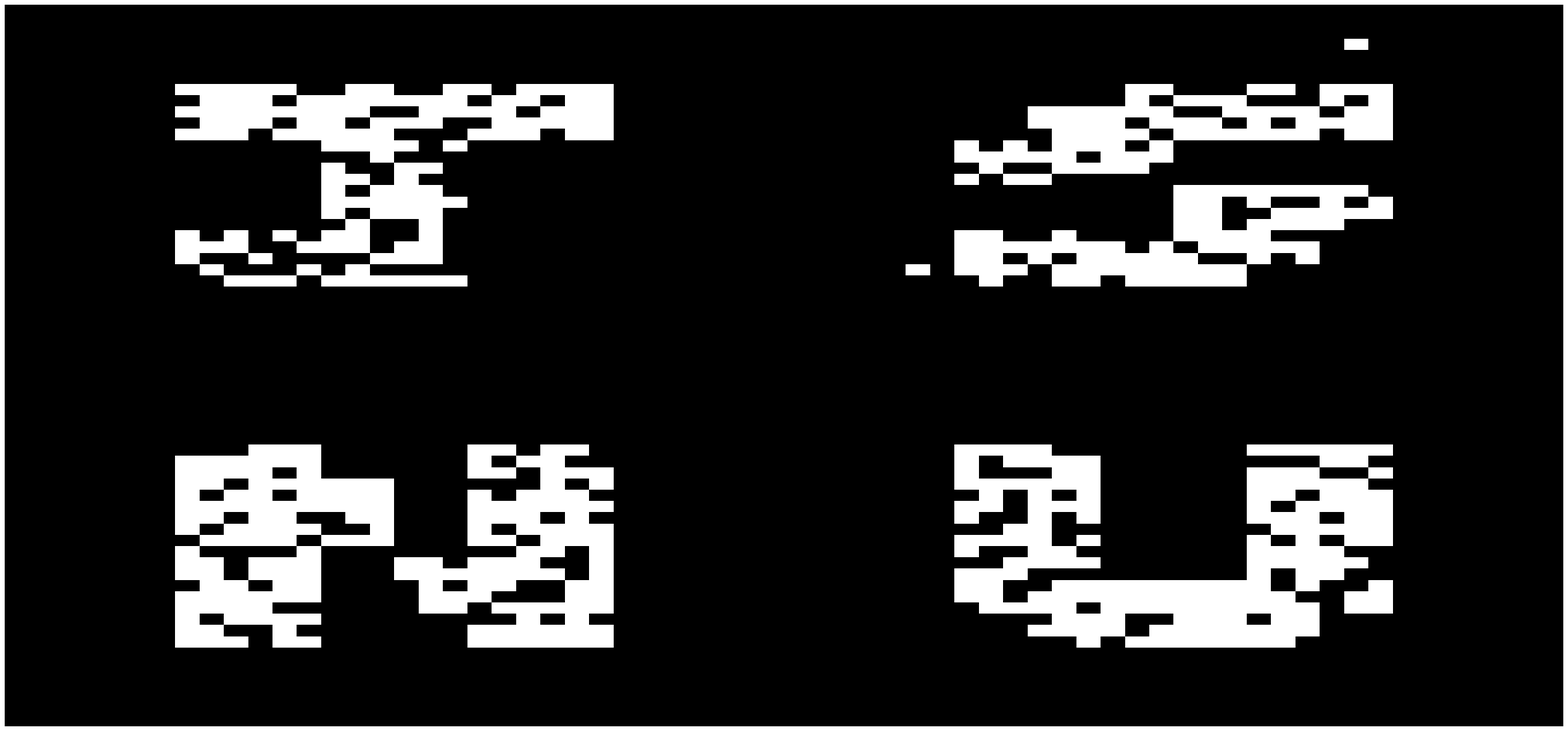} \end{minipage} &
\begin{minipage}{0.07\textwidth} \includegraphics[width=12mm, height=10mm]{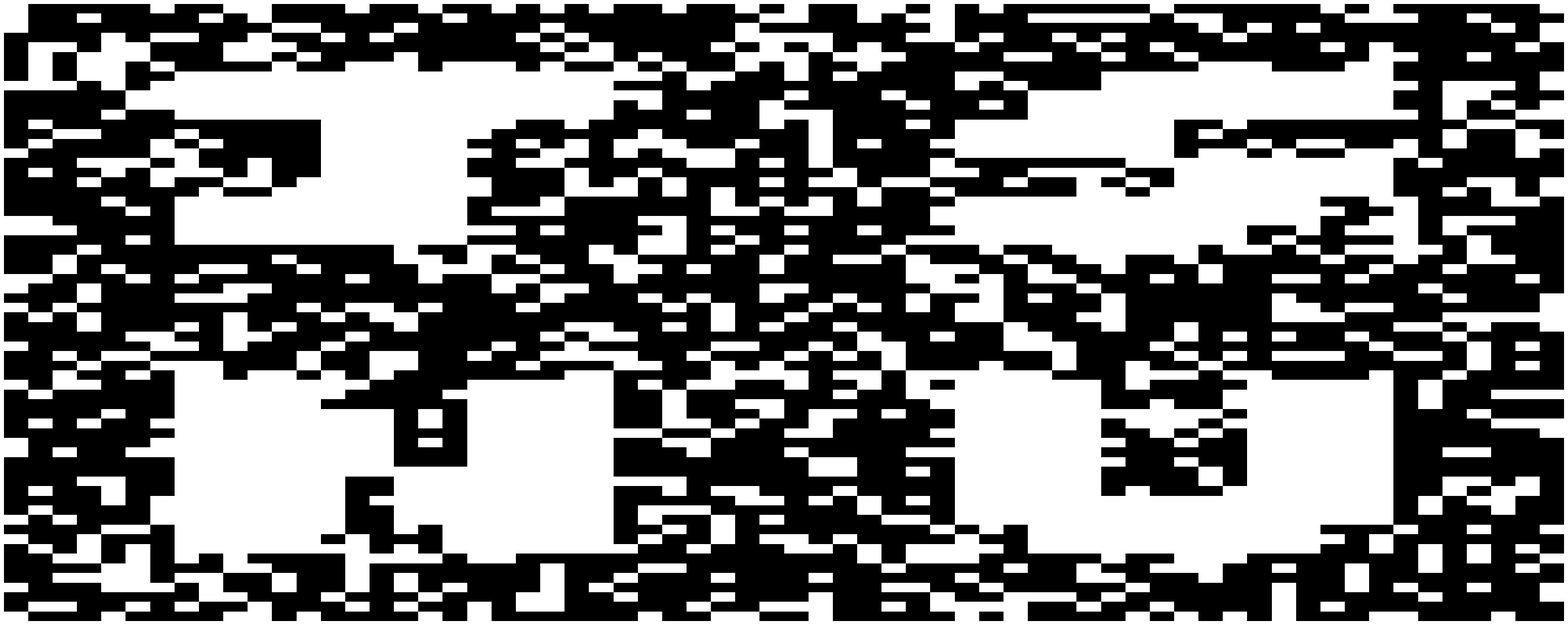} \end{minipage}  &
\begin{minipage}{0.07\textwidth} \includegraphics[width=12mm, height=10mm]{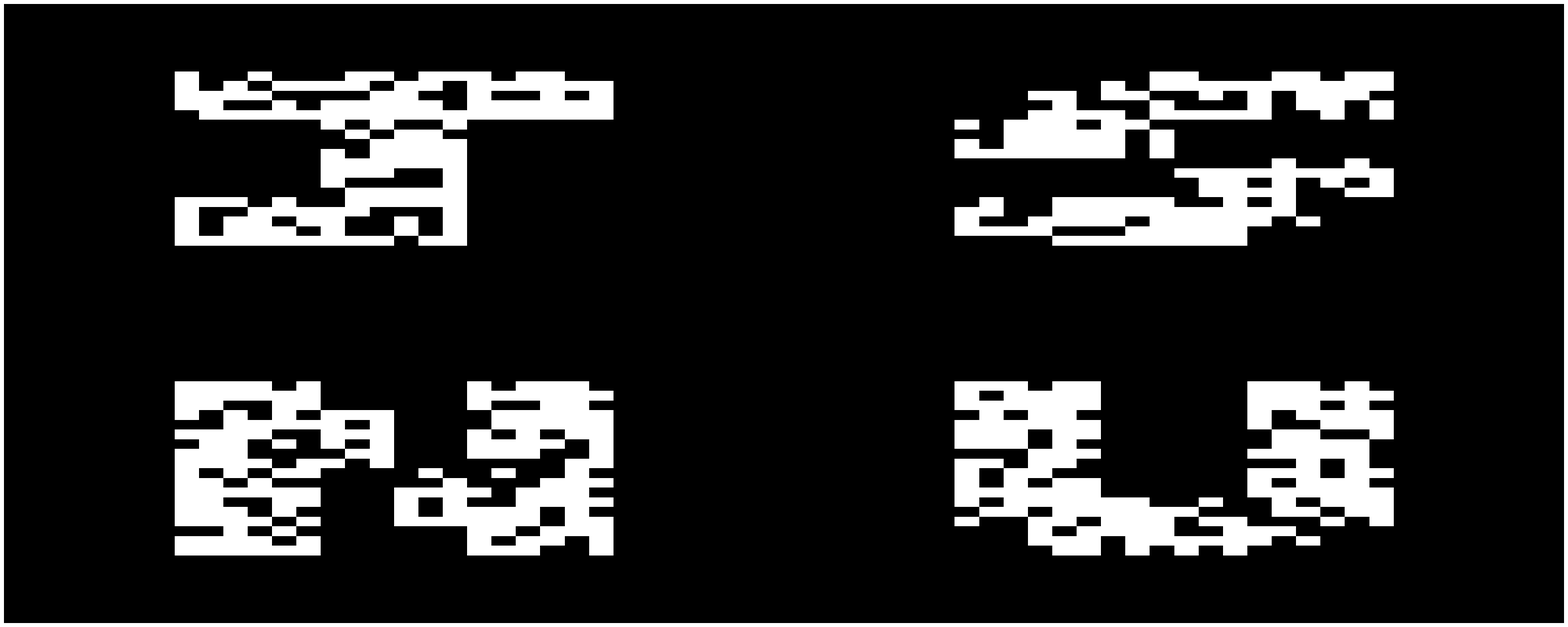} \end{minipage} &
\begin{minipage}{0.07\textwidth} \includegraphics[width=12mm, height=10mm]{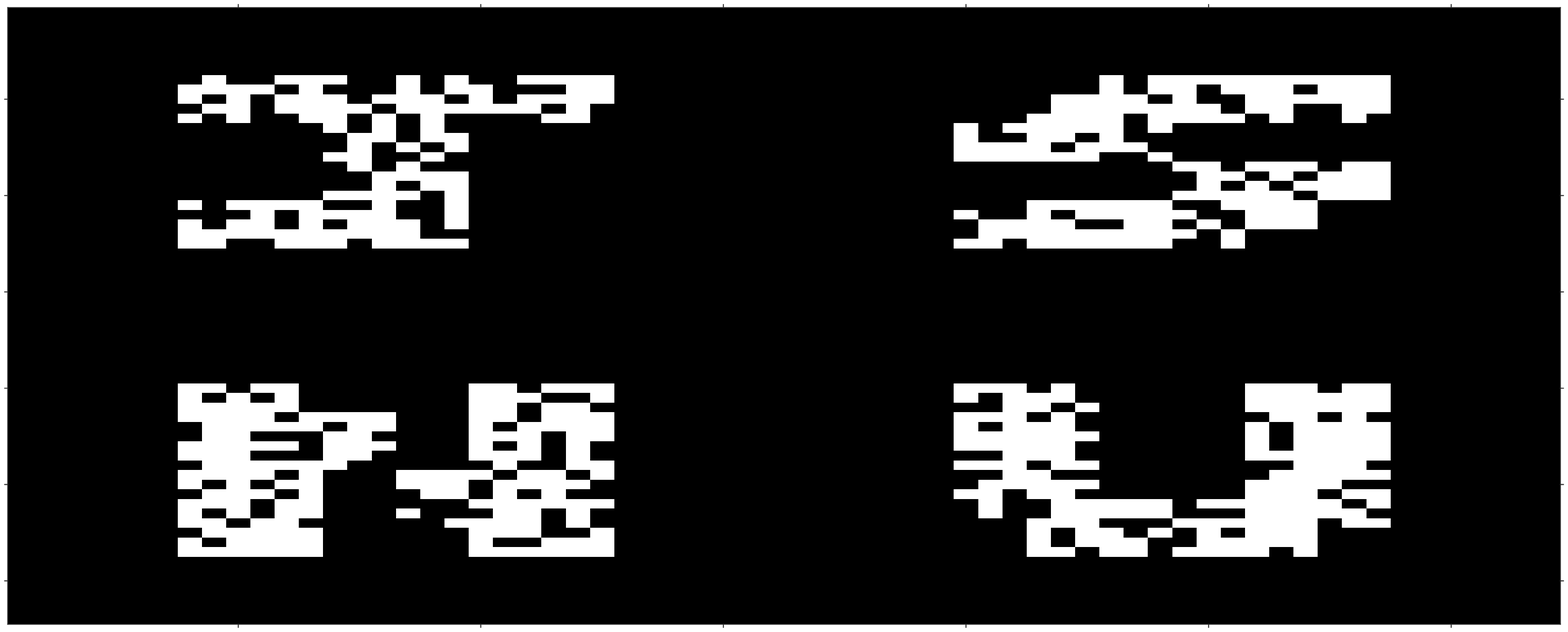} \end{minipage} \\
\cline{2-8}
& BER & 0.2368 & 0.0674 & 0.0664 & 0.2390 & 0.0684 & 0.0671\\
\cline{2-8}
& Speckle 0.05 &
\begin{minipage}{0.07\textwidth} \includegraphics[width=12mm, height=10mm]{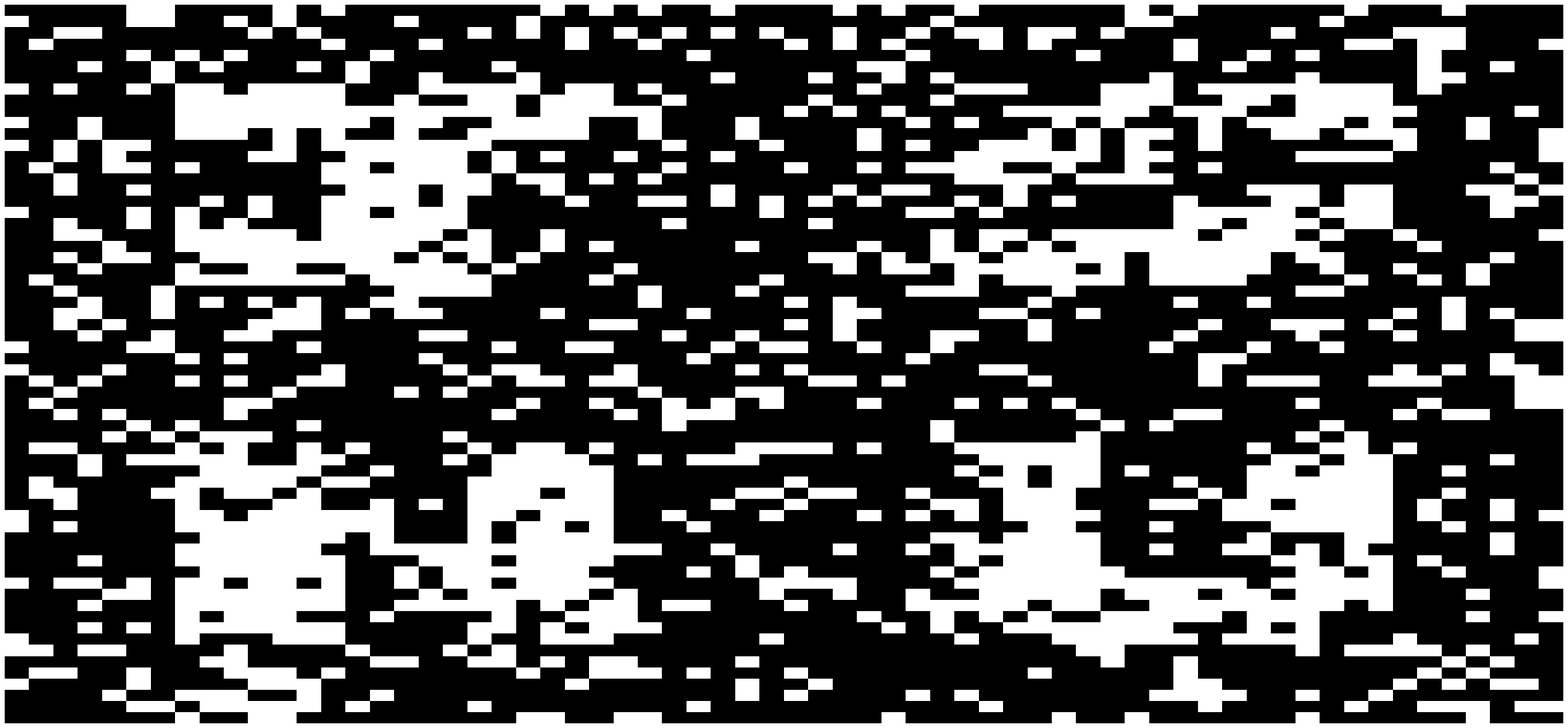} \end{minipage} &
\begin{minipage}{0.07\textwidth} \includegraphics[width=12mm, height=10mm]{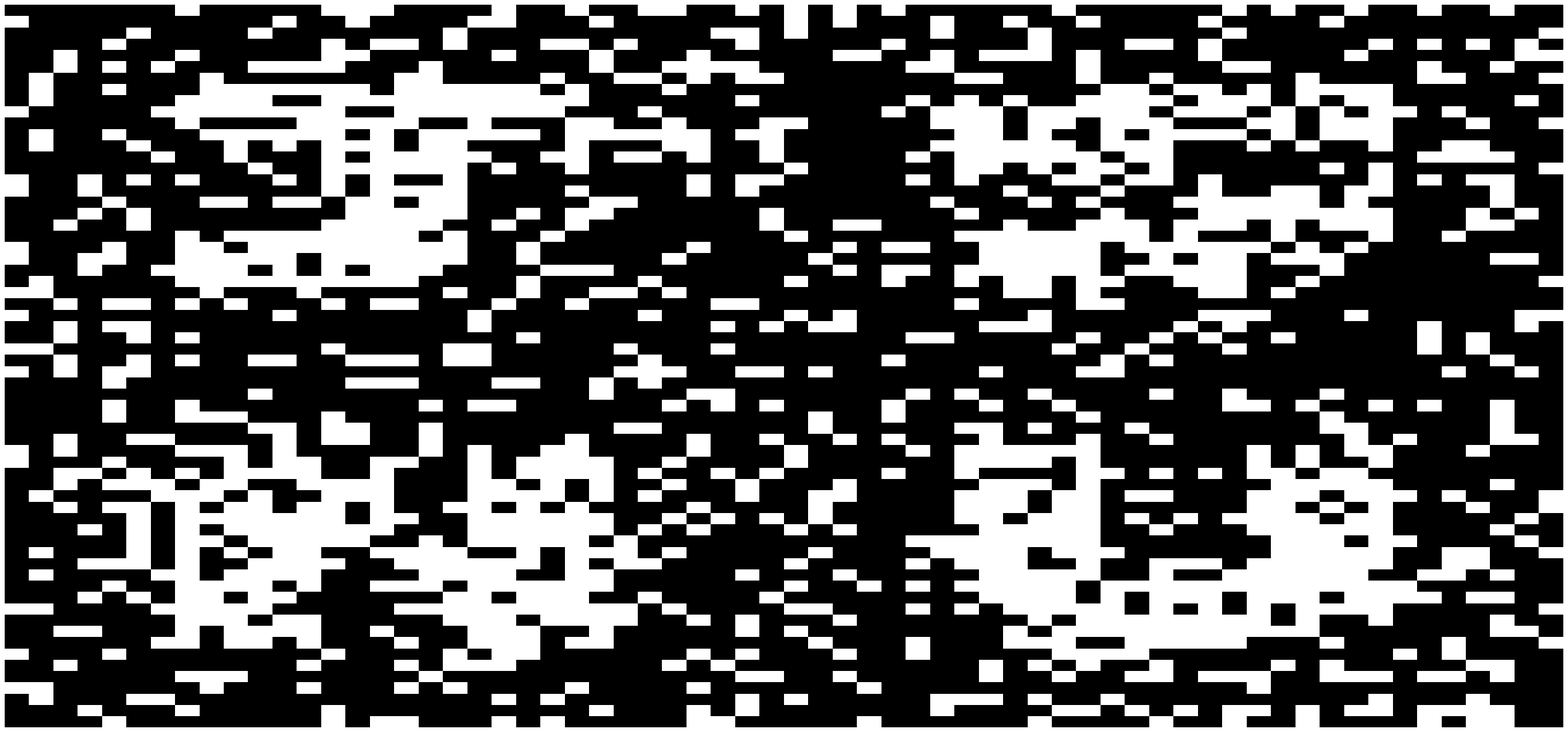} \end{minipage} &
\begin{minipage}{0.07\textwidth} \includegraphics[width=12mm, height=10mm]{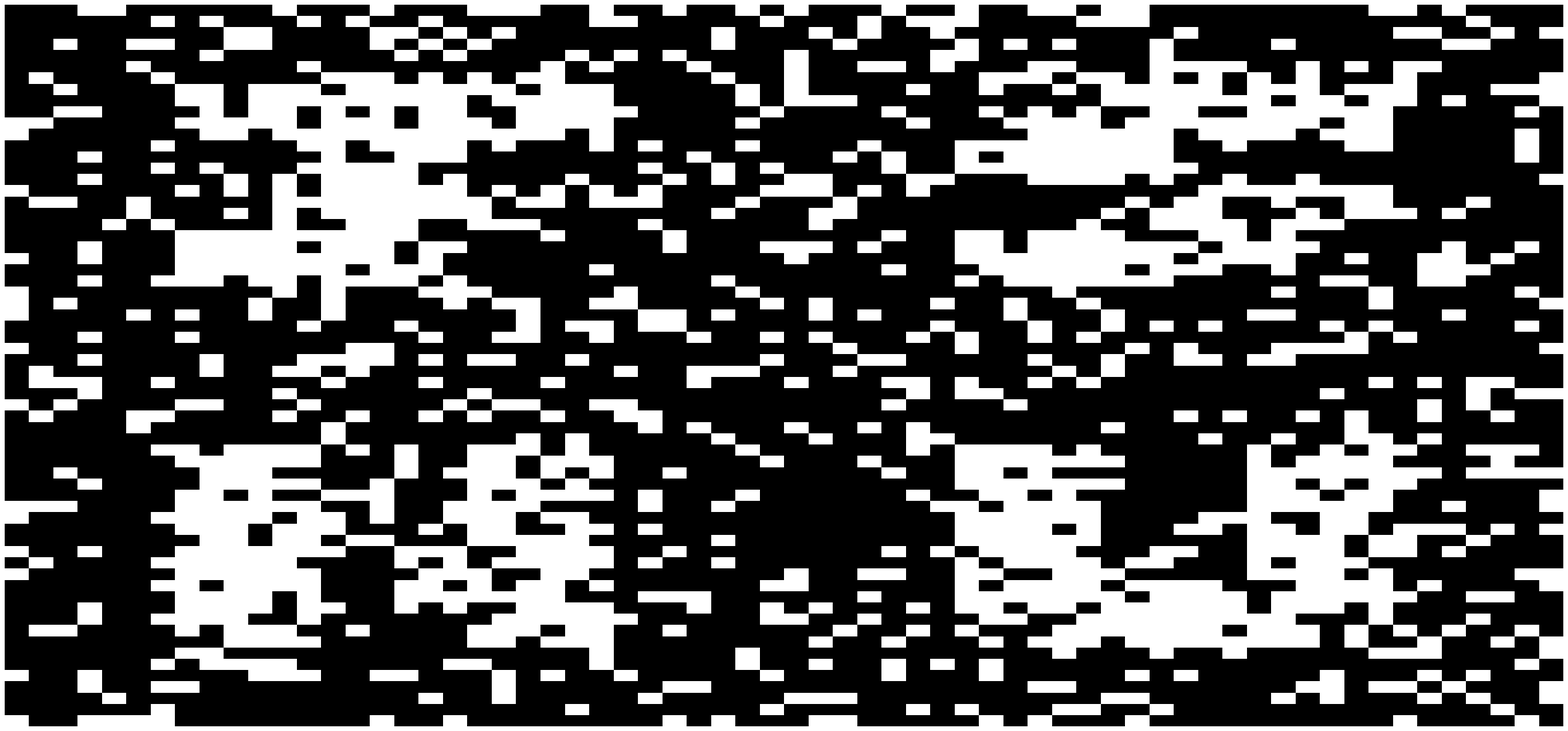} \end{minipage}&
\begin{minipage}{0.07\textwidth} \includegraphics[width=12mm, height=10mm]{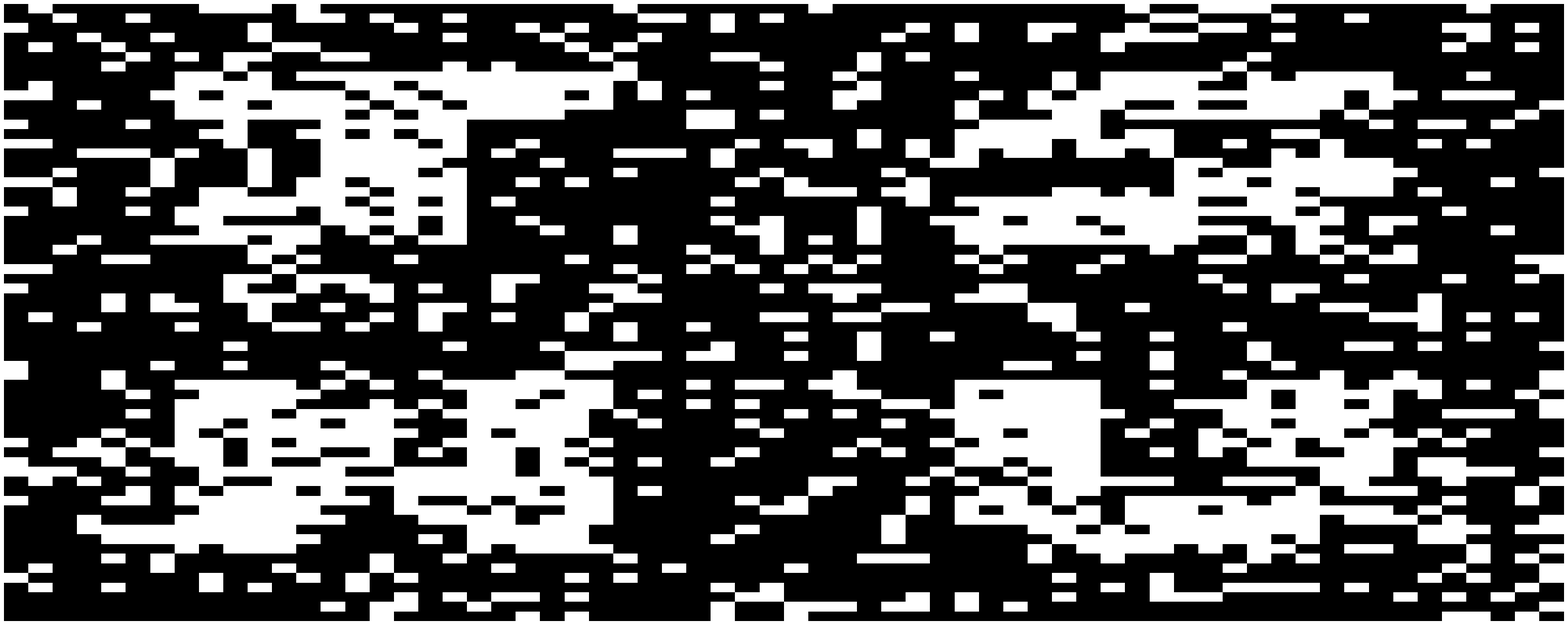} \end{minipage} &
\begin{minipage}{0.07\textwidth} \includegraphics[width=12mm, height=10mm]{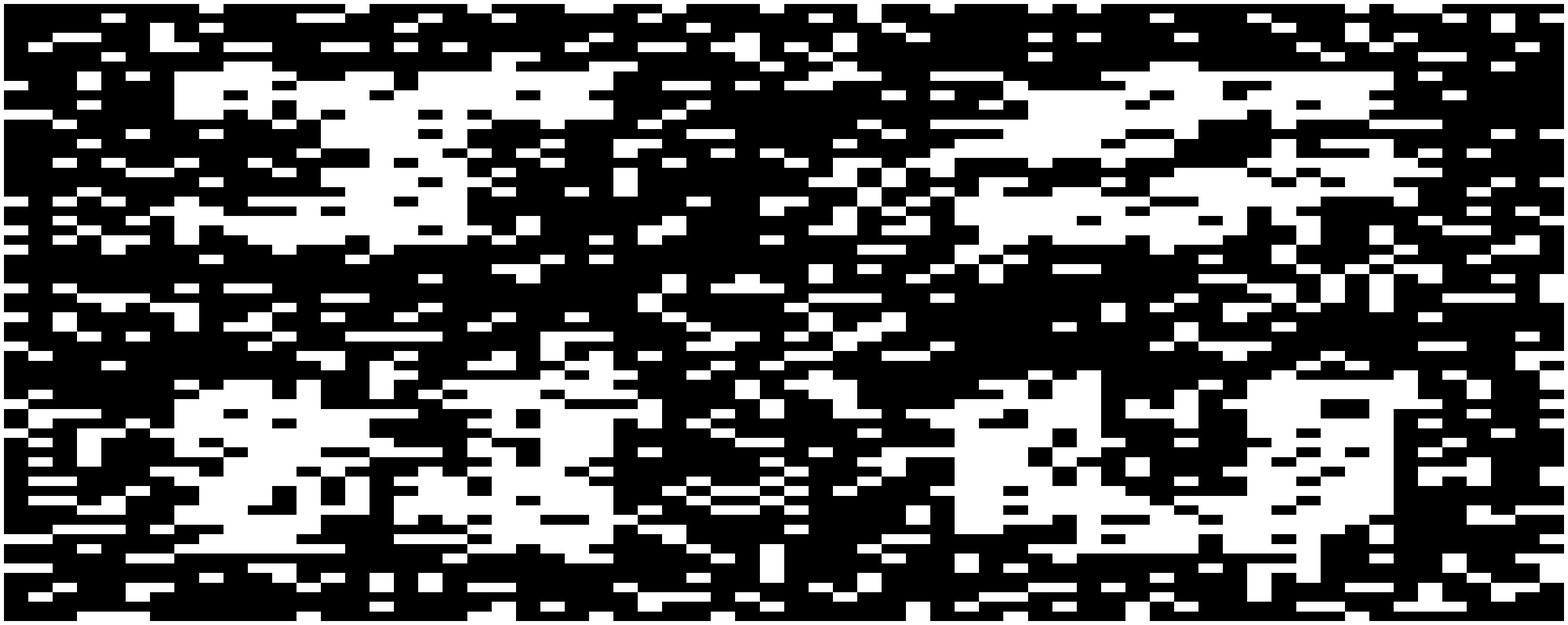} \end{minipage} &
\begin{minipage}{0.07\textwidth} \includegraphics[width=12mm, height=10mm]{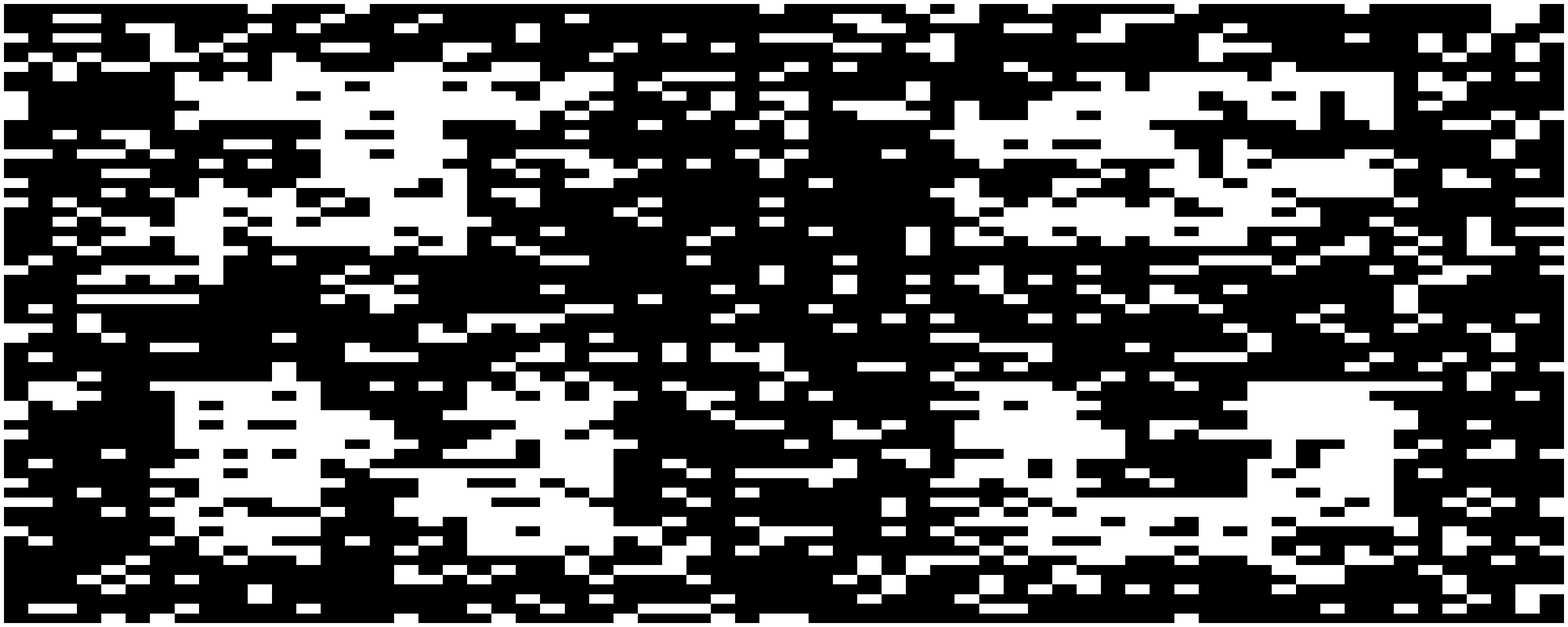} \end{minipage}\\
\cline{2-8}
& BER & 0.2158 & 0.2319 & 0.2214 & 0.2048 & 0.2129 & 0.2146\\
\cline{2-8}
&Salt\&pepper 0.05 &
\begin{minipage}{0.07\textwidth} \includegraphics[width=12mm, height=10mm]{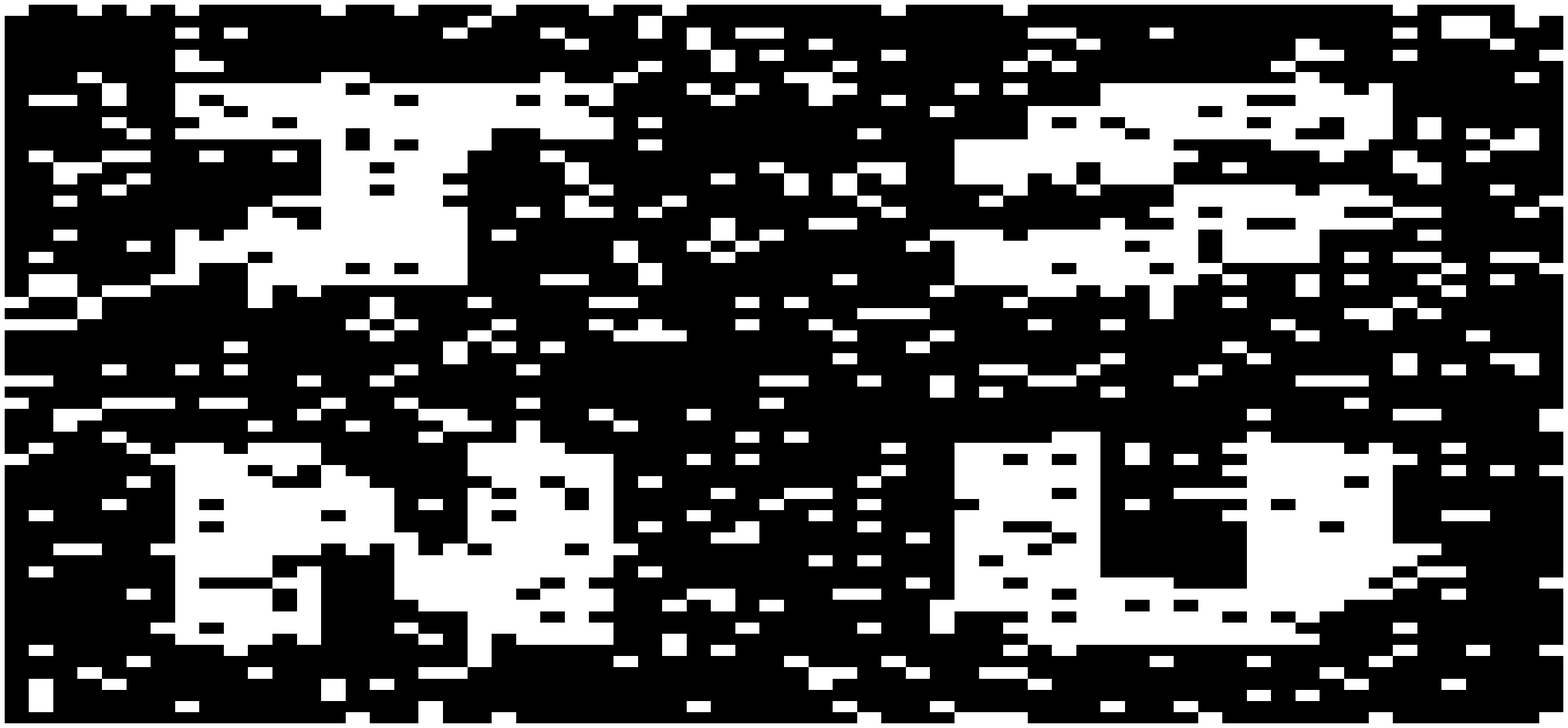} \end{minipage} &
\begin{minipage}{0.07\textwidth} \includegraphics[width=12mm, height=10mm]{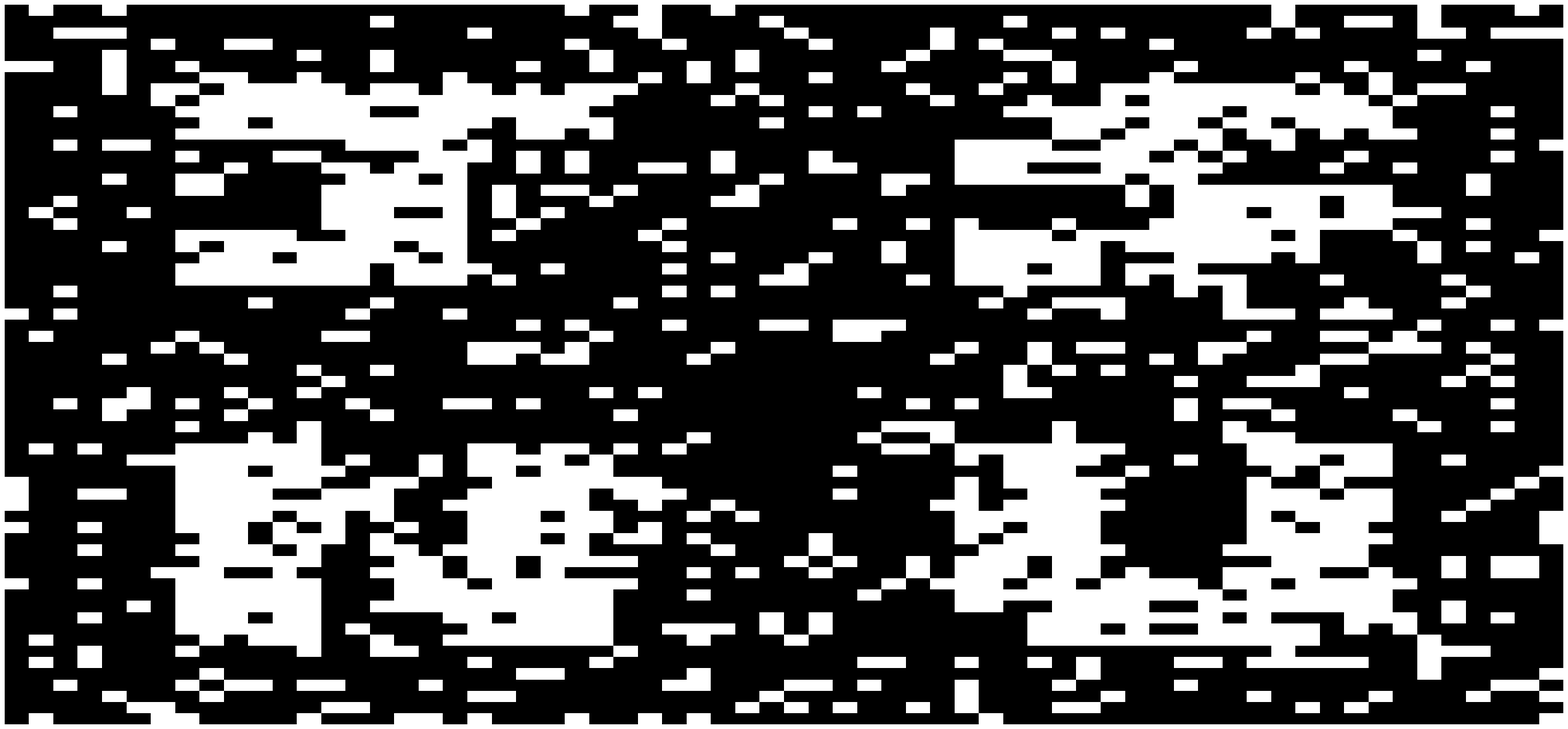} \end{minipage} &
\begin{minipage}{0.07\textwidth} \includegraphics[width=12mm, height=10mm]{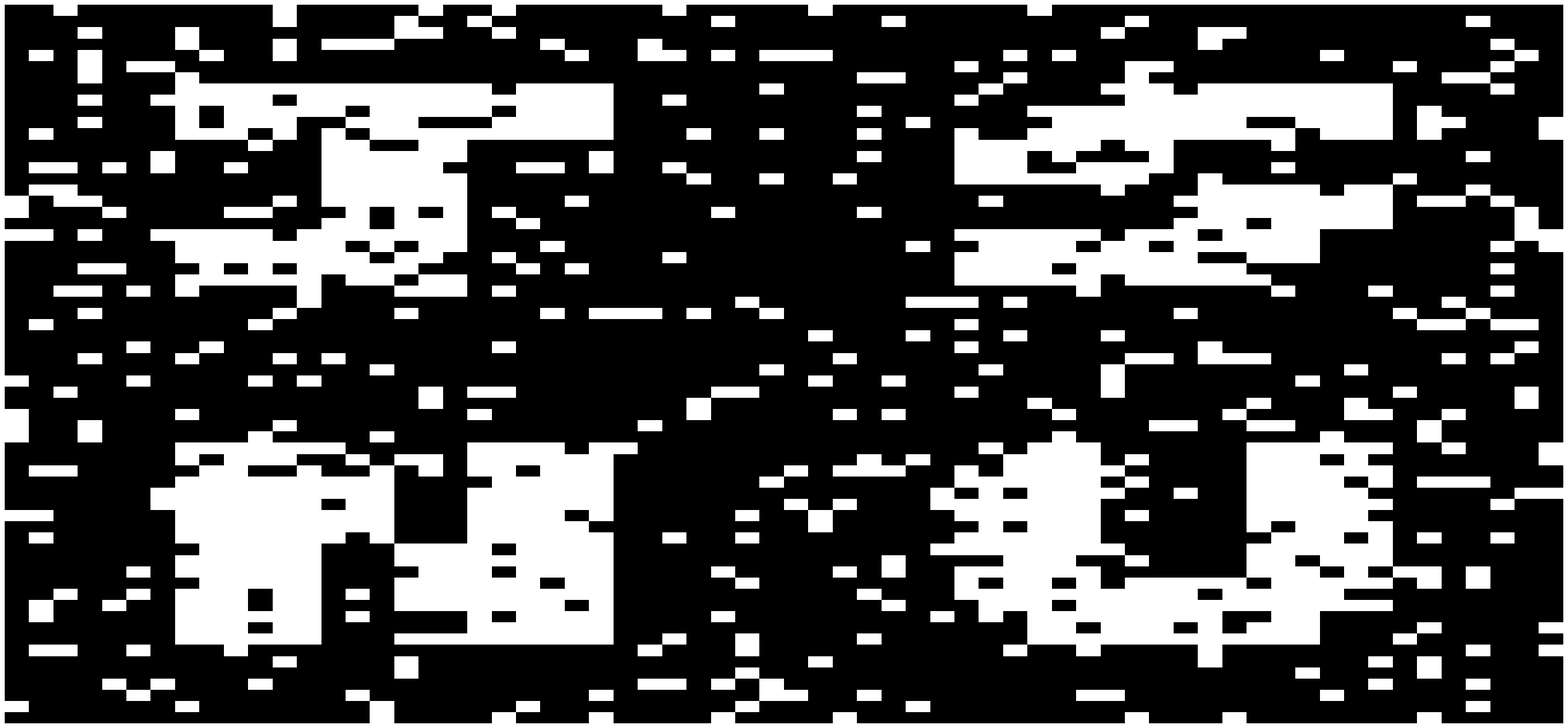} \end{minipage}&
\begin{minipage}{0.07\textwidth} \includegraphics[width=12mm, height=10mm]{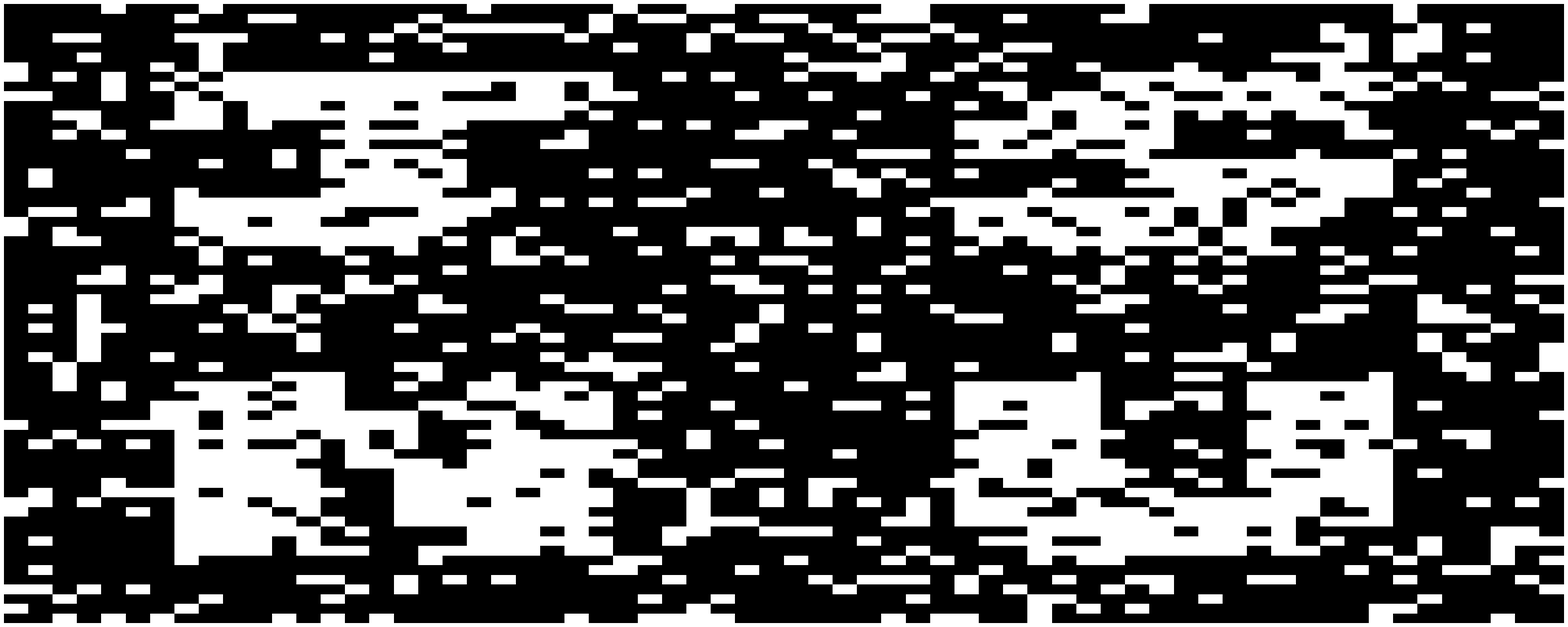} \end{minipage} &
\begin{minipage}{0.07\textwidth} \includegraphics[width=12mm, height=10mm]{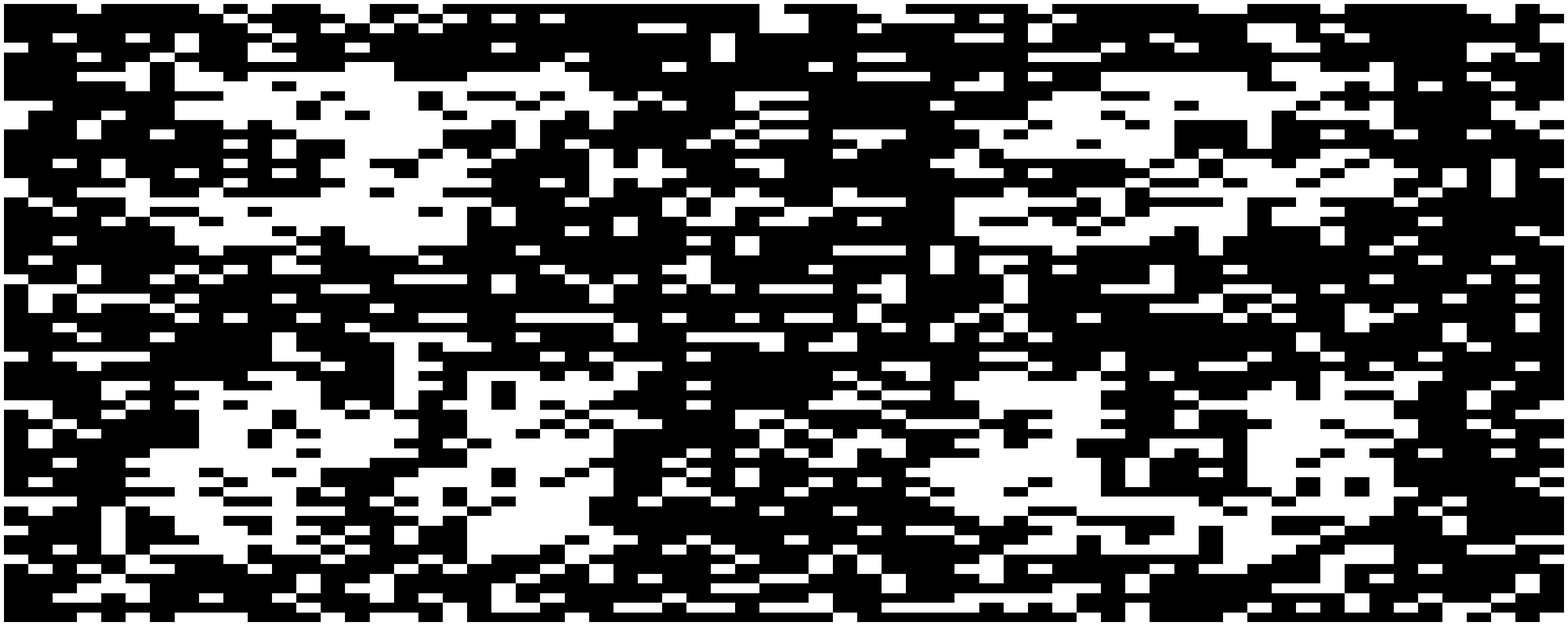} \end{minipage} &
\begin{minipage}{0.07\textwidth} \includegraphics[width=12mm, height=10mm]{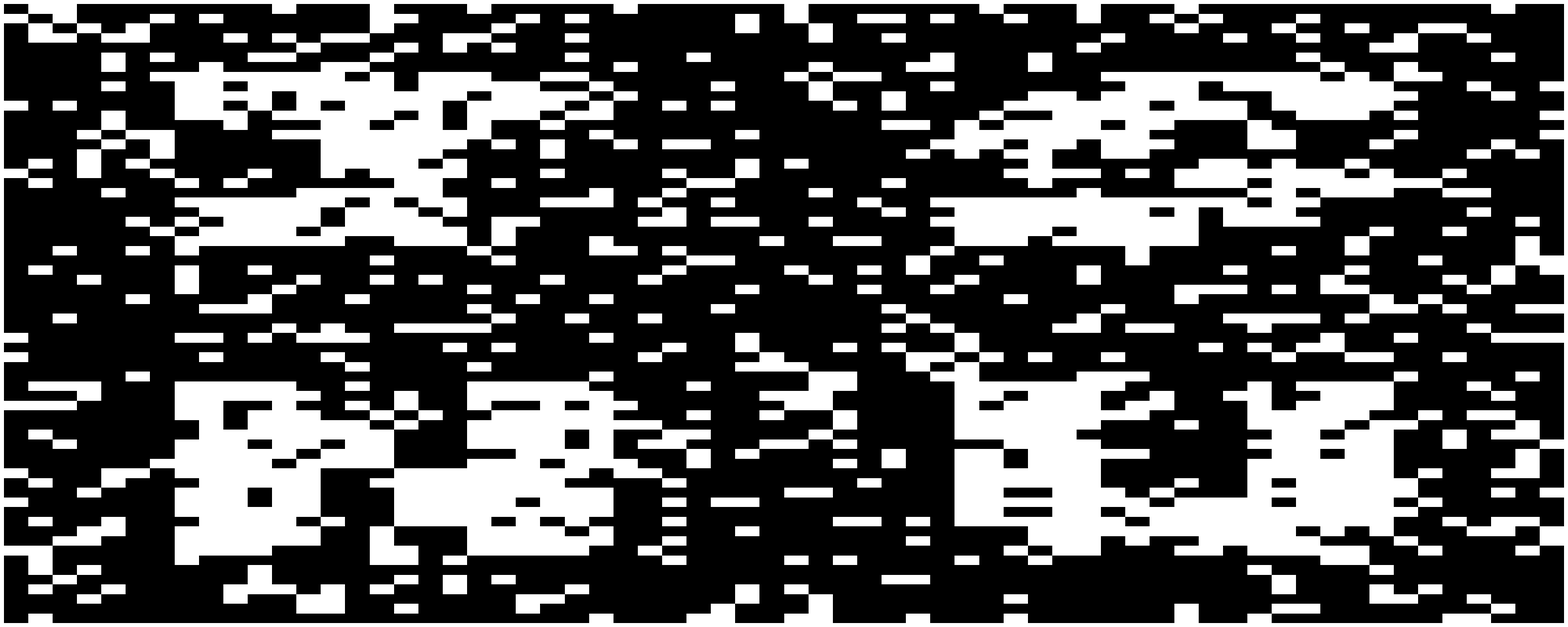} \end{minipage}\\
\cline{2-8}
& BER & 0.1428 & 0.1699 & 0.1265 & 0.1973 & 0.2373 & 0.1848\\
\hline
\end{tabular}
\end{table*}

\textbf{3. The execution time comparison analysis}

Regarding the execution time, it is another important criterion. After many times running, as can be seen from Table \ref{table_6}, the average execution time of the proposed method is much smaller than that of Liu \emph{et al.}\cite{FLCZ18}. This means that the proposed method is efficient. Therefore, it should be noted that the calculation amount of the proposed QSVD algorithm is much less than that of the conventional QSVD algorithm. Li \emph{et al.}'s method \cite{LWZZ16} is based on the built-in function ${\tt svd}$ from Matlab, the execution time is the fastest.

\begin{table}[!htbp]
\caption{The comparison of execution times between methods(Second) }\label{table_6}
\centering
\begin{tabular}{|p{2cm}<{\centering}|p{1.2cm}<{\centering}|p{1cm}<{\centering}|p{1cm}<{\centering}|}
\hline
Method & Embedding time & Extraction time &  Total time\\
\hline
Li \emph{et al.} \cite{LWZZ16}& 0.1247 & 0.0630 & 6.4330\\
\hline
Liu \emph{et al.} \cite{FLCZ18}& 40.8966 & 21.9549 & 69.0130\\
\hline
Proposed method & 1.4293 & 1.2725 & 8.8930\\
\hline
\end{tabular}
\end{table}

\textbf{4. The capacity and security analysis}.

\begin{figure}
\centering
\includegraphics[width=8cm,height=5cm]{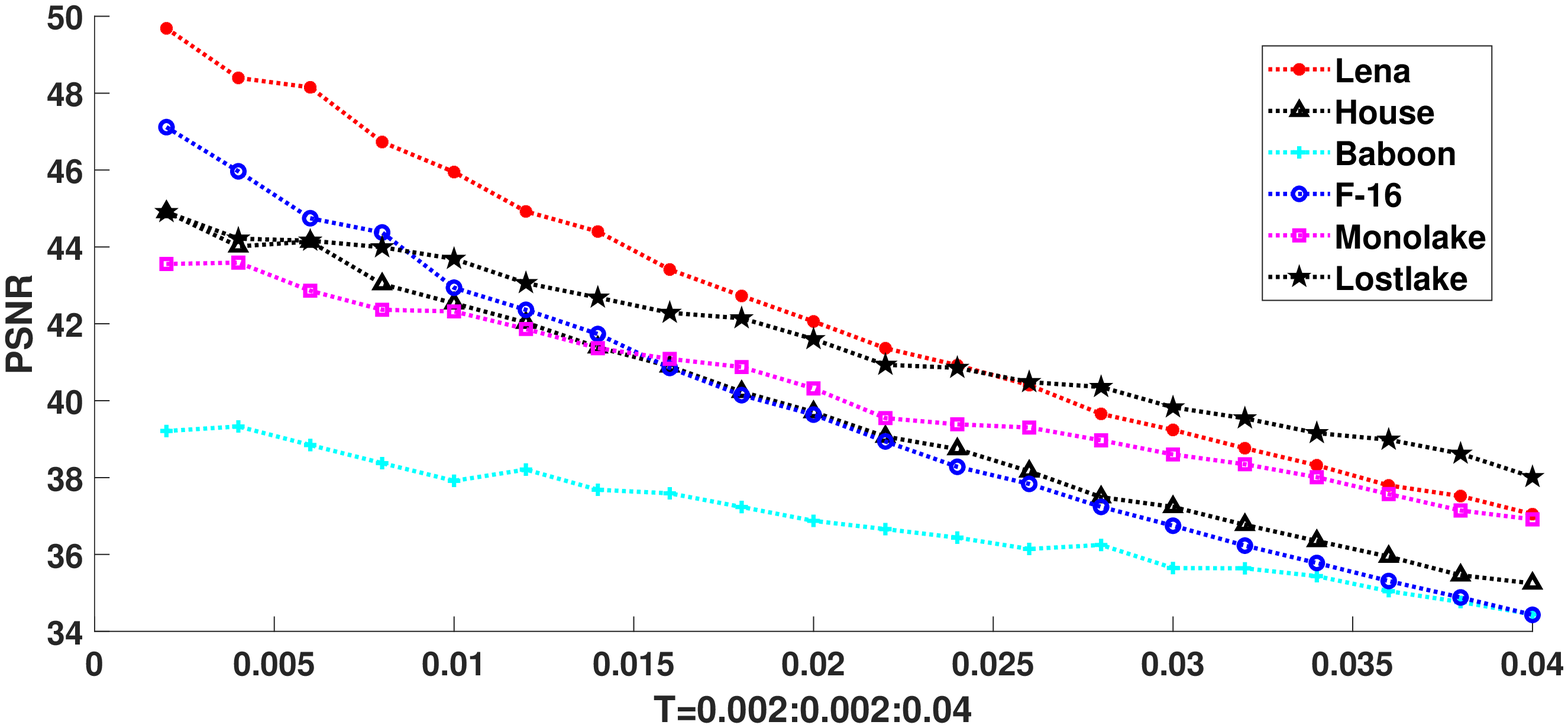}
\caption{Corresponding PSNR for triple watermarks embedded in different host images at appropriate interval $T$.}\label{PSNR_3}
\end{figure}

The watermark capacity is a key indicator used to measure the performance of a watermarking scheme. In this paper, since the proposed watermarking scheme uses the optimal coefficient pair selection strategy, the maximum capacity is equal to the number of host images divided into a $4 \times 4$ block matrix. This strategy is different from the traditional QSVD watermarking technology. The traditional QSVD uses a statistical method to select a fixed coefficient correlation pair for watermark embedding, which ignores the equivalence of the three imaginary parts used to represent color images and causes a color image layer to be modified too much. In this paper, three highly correlated coefficient pairs are found for the first time, and the proposed strategy is to adaptively select the coefficient pair with the highest correlation in each $4 \times 4$ block matrix. Therefore, for the $512 \times 512$ host image, the maximum payload of the proposed QSVD and the traditional QSVD are both $128 \times 128$. In addition, the actual payload is limited by the distortion of the watermarked image and the quality of the extracted watermark, so the embedded watermark is set to a $64 \times 64$ binary image and the host image is a $512 \times 512$ color image, that is, the embedding rate is $(64\times 64)/(512\times512)=0.015625$(bpp). In particular, our method can further pushes forward the capacity limit of conventional QSVD. It provides a very high capacity in a single embedding pass three imaginary units simultaneously, when $\Delta_{1}=\delta_{11}\mathbf{i}+\delta_{12}\mathbf{j}+\delta_{13}\mathbf{k}$, $\Delta_{2}=\delta_{21}\mathbf{i}+\delta_{22}\mathbf{j}+\delta_{23}\mathbf{k}$, then 3 bits will be carried in each image block. We shows the visual quality of triple watermarks (as shown in Fig.~\ref{watermark}) simultaneously embedded in the host image. For example, for images Lena, House, Baboon, F-16, Monolake and Lostlake, we show their PSNR values in Fig.~\ref{PSNR_3}. All PSNR values are greater than 34 dB, which means that the visual quality of the watermarked image is unnoticeable. Thus the triple watermarking scheme is effective and feasible, but this paper mainly involves adaptive embedding, and the research on the robustness of triple embedding will be carried out in the future work.

A random coordinate sequence based on the private key {\tt Ka} is adopted, which increases the unpredictable choice for embedded blocks and embedded locations. Moreover, the watermark bits are adaptively embedded in the three imaginary units, which also increases security. They greatly increases the security of the entire watermarking process.

\section{Conclusion}\label{sec6}
In this paper, we propose a new robust watermarking scheme based on efficient QSVD, coefficient pair selection, and adaptive embedding. Our method focuses on efficient algorithms and highly correlated coefficient pairs. Compared with conventional QSVD, the structure-preserving method frees up more cumbersome calculations for better performance. Moreover, compared with other state-of-the-art structure-preserving methods, the CPU runtime of our method is about half of that of the algorithms in \cite{LWZZ67} and \cite{LWZZ96}. Highly correlated coefficient pairs are determined by the NC method and are adaptively selected. These make the watermarked image more fidelity and the extracted watermark has better visual quality against common watermark attacks. Experimental results demonstrate that the proposed method successfully makes the watermark imperceptible and robust against various geometric attacks and signal processing operations. Compared with other state-of-the-art watermarking schemes, the new proposed method has better performance to protect the copyright of color images that will be transmitted over the Internet. This paper also proposes a high-capacity mechanism and verifies its feasibility. In the future, a high-capacity and more robust watermarking scheme is expected.

\ifCLASSOPTIONcaptionsoff
  \newpage
\fi


\begin{thebibliography}{1}

\bibitem{TR07} D. M. Thodi and J. J. Rodriguez, ``Expansion Embedding Techniques for Reversible Watermarking," \emph{IEEE Trans. Image Process.}, vol. 16, no. 3, pp.721-730, 2007.

\bibitem{ZWWX19} H. Zheng, C. Wang, J. Wang, and S. Xiang, ``A new reversible watermarking scheme using the content-adaptive block size for prediction," \emph{Signal Process.}, vol. 164, pp. 74-83, 2019.

\bibitem{CSTS05} M. U. Celik, G. Sharma, A. M. Tekalp, and E. Saber, ``Lossless Generalized-LSB Data Embedding," \emph{IEEE Trans. Image Process.}, vol. 14, no. 2, pp. 253-266,2005.

\bibitem{LSZLT12} M. Liu, H. S. Seah, C. Zhu, W. Lin, and F. Tian, ``Reducing location map in prediction-based difference expansion for reversible image data embedding," \emph{Signal Process.}, vol.92, no.3, pp. 819-828, 2012.

\bibitem{DC14} I.-C. Dragoi and D. Coltuc, ``Local-prediction-based difference expansion reversible watermarking," \emph{IEEE Trans. Image Process.}, vol. 23, no.4, pp.1779-1790, 2014.

\bibitem{DC15} I.-C. Dragoi and D. Coltuc, ``On local prediction based reversible watermarking," \emph{IEEE Trans. Image Process.}, vol. 24, no.4, pp.1244-1246, 2015.

\bibitem{LLYZ13} X. Li, B. Li, B. Yang, and T. Zeng, ``General framework to histogram-shifting-based reversible data hiding," \emph{IEEE Trans. Image Process.}, vol. 22, no.6, pp.2181-2191, 2013.

\bibitem{WYWW18} W. Wang, J. Ye, T. Wang, and W. Wang, ``A high capacity reversible data hiding scheme based on right-left shift," \emph{Signal Process.}, vol. 150, pp. 102-115, 2018.

\bibitem{JYZL19} Y. Jia, Z. Yin, X. Zhang, and Y. Luo, ``Reversible data hiding based on reducing invalid shifting of pixels in histogram shifting,"  \emph{Signal Process.}, vol.163, pp. 238-246, 2019.

\bibitem{LYZ11} X. Li, B. Yang, and T. Zeng, ``Efficient reversible watermarking based on adaptive prediction-error expansion and pixel selection," \emph{IEEE Trans. Image Process.}, vol. 20,  no.12, pp. 3524-3533, 2011.

\bibitem{OLZN15} B. Ou, X. Li, Y. Zhao, and R. Ni, ``Efficient color image reversible data hiding based on channel-dependent payload partition and adaptive embedding," \emph{Signal Process.}, vol. 108, pp. 642-657, 2015.

\bibitem{OLZN13} B. Ou, X. Li, Y. Zhao, R. Ni, and Y. Q. Shi, ``Pairwise prediction-error expansion for efficient reversible data hiding," \emph{IEEE Trans. Image Process.}, vol. 22, no. 12, pp. 5010-5021, 2013.

\bibitem{C12} D. Coltuc, ``Low Distortion Transform for Reversible Watermarking," \emph{IEEE Trans. Image Process.}, vol. 21, no. 1, pp. 412-417, 2012.


\bibitem{ZPP10} X. Zeng, L. Ping, and X. Pan,``A lossless robust data hiding scheme," \emph{Pattern Recogn.}, vol. 43, no.4, pp. 1656-1667, 2010.

\bibitem{AGLTDL12} L. An, X. Gao, X. Li, D. Tao, C. Deng, and J. Li, ``Robust Reversible Watermarking via Clustering and Enhanced Pixel-Wise Masking," \emph{IEEE Trans. Image Process.}, vol. 21, no. 8, pp. 3598-3611, 2012.

\bibitem{SNZPL13} Q. Su, Y. Niu, Y. Zhao, S. Pang, and X. Liu, ``A dual color images watermarking scheme based on the optimized compensation of singular value decomposition," \emph{AEU-Int. J. Electron. and Commun.}, vol. 67, no. 8, pp. 652-664, 2013.

\bibitem{SNZL13} Q. Su, Y. Niu, H. Zou, and X. Liu, ``A blind dual color images watermarking based on singular value decomposition," \emph{Appl. Math. Comput.}, vol. 219, no. 16, pp. 8455-8466, 2013.

\bibitem{FLCZ18} F. Liu, L. Ma, C. Liu, and Z. Lu. ``Optimal blind watermarking for color images based on the U matrix of quaternion singular value decomposition," \emph{Multimed. Tools Appl.}, vol. 77, no. 18, pp. 23483-23500, 2018.

\bibitem{CS17} C. Chang and J. Shen, ``Features Classification Forest: A Novel Development that is Adaptable to Robust Blind Watermarking Techniques," \emph{IEEE Trans. Image Process.}, vol. 26, no. 8, pp. 3921-3935, 2017.

\bibitem{BL13} B. Abdelhamid and L. Lamri, ``New images watermarking scheme based on singular value decomposition," \emph{J. Inf. Hiding and Multimedia Signal Process.}, vol. 4, no. 1, pp. 1-10, 2013.

\bibitem{MAS08} A. A. Mohammad, A. Alhaj, and S. Shaltaf, ``An improved SVD-based watermarking scheme for protecting rightful ownership," \emph{Signal Process.}, vol. 88, no. 9, pp. 2158-2180, 2008.

\bibitem{LWZZ16} Y. Li, M. Wei, F. Zhang, and J. Zhao, ``A New Double Color Image Watermarking Algorithm Based on the SVD and Arnold Scrambling," \emph{J. Appl. Math.}, vol. 2497379, pp. 1-9, 2016.

\bibitem{BM05} P. Bao and X. Ma. ``Image adaptive watermarking using wavelet domain singular value decomposition," \emph{IEEE Trans. Circuits Syst. Video Technol.}, vol. 15, no.1, pp. 96-102, 2005.

\bibitem{MY15} M. Natarajan and Y. Govindarajan, ``A study of DWT-SVD based multiple watermarking scheme for medical Images," \emph{Int. J. Netw. Secur.} vol. 17, no. 5, pp. 558-568, 2015.

\bibitem{WR11} W. R. Hamilton, \emph{The Mathematical Papers of Sir William Rowan Hamilton}, Cambridge: Cambridge University Press, 1967.

\bibitem{DJD21} D. Finkelstein, J. M. Jauch, and D. Speiser, ``Notes on quaternion quantum mechanics," \emph{CERN Yellow Reports: Monographs}, vol. 59, no. 9, pp. 1-36, 1959.

\bibitem{DJSD20} D. Finkelstein, J. M. Jauch, S. Schiminovich, and D. Speiser, ``Foundations of quaternion quantum mechanics," \emph{J. Math. Phys.}, vol. 3, pp. 207-220, 1962.

\bibitem{NS12} N. L. Bihan and S. J. Sangwine, ``Quaternion principal component analysis of color images," \emph{IEEE Int. Conf. Image Process.}, vol. 1, pp. 809-812, 2003.

\bibitem{JNW19} Z. Jia, M. K. Ng, and W. Wang, ``Color image restoration by saturation-value total variation,'' \emph{SIAM J. Imaging Sci.}, vol. 12, no. 2, 972-1000, 2019.

\bibitem{JNS19} Z. Jia, M. K. Ng, and G. Song, ``Robust quaternion matrix completion with applications to image inpainting,'' \emph{Numerical Linear Algebra with Applications}, vol. 26, no. 4, e2245, 2019.

\bibitem{HD18} K. M. Hosny and M. M. Darwish, ``Robust Color Image Watermarking Using Invariant Quaternion Legendre-Fourier Moments," \emph{Multimed. Tools Appl.}, vol. 77, no. 19, pp. 24727-24750, 2018.

\bibitem{HD19} K. M. Hosny and M. M. Darwish, ``Resilient Color Image Watermarking Using Quaternion Radial Substituted Chebychev Moments," \emph{ACM Trans. Multim. Comput.}, vol. 15, no. 2, 2019.

\bibitem{MCWLWS20} B. Ma, L. Chang, C. Wang, J. Li, X. Wang, and Y. Shi, ``Robust image watermarking using invariant accurate polar harmonic Fourier moments and chaotic mapping," \emph{Signal Process.}, vol. 172, 2020.

\bibitem{GC20} G. H. Golub and C. Reinsch, ``Singular value decomposition and least squares solutions," \emph{Numer. Math.}, vol. 14, no. 5, pp. 403-420, 1970.
\bibitem{GC96} G. H. Golub and C. F. Van Loan, \emph{Matrix Computations}, 4th Edition, The Johns Hopkins University Press, 2013.

\bibitem{ZF57} F.~Zhang, ``Quaternions and matrices of quaternions," \emph{Linear Algebra Appl.}, vol. 251, no. 15, pp. 21-57, 1997.

\bibitem{SN} S. J. Sangwine and N. L. Bihan, Quaternion toolbox for matlab. http://qtfm.sourceforge.net/.
\bibitem{JWL24} Z. Jia, M. Wei, and S. Ling, ``A new structure-preserving method for quaternion Hermitian eigenvalue problems," \emph{J. Comput. Appl. Math.}, vol. 239, pp. 12-24, 2013.
\bibitem{LWZZ67} Y. Li, M. Wei, F. Zhang, and J. Zhao, ``A fast structure-preserving method for computing the singular value decomposition of quaternion matrix," \emph{Appl. Math. Comput.}, vol. 235, pp. 157-167, 2014.
\bibitem{LWZZ96} Y. Li, M. Wei, F. Zhang, and J. Zhao, ``Real structure-preserving algorithms of Householder based transformations for quaternion matrices," \emph{J. Comput. Appl. Math.}, vol. 305, pp. 82-91, 2016.
\bibitem{SN06} S. J. Sangwine and N. L. Bihan, ``Quaternion singular value decomposition based on bidiagonalization to a real or complex matrix using quaternion Householder transformations," \emph{Appl. Math. Comput.}, vol. 182, pp. 727-738, 2006.

\bibitem{JWZC18} Z. Jia, M. Wei, M. Zhao, and Y. Chen, ``A new real structure-preserving quaternion QR algorithm," \emph{J. Comput. Appl. Math.}, vol. 343, pp. 26-48, 2018.

\bibitem{GBM89} A. Bunse-Gerstner, R. Byers, and V. Mehrmann, ``A quaternion QR algorithm," \emph{Numer. Math.}, vol. 55, pp. 83-95, 1989.

\bibitem{UoG} Unversity of Granada, Computer Vision Group. CVG-UGR image Data base. [2012-10-22]. http://decsai.ugr.es/cvg/dbimagenes/c512.php

\end{thebibliography}
\end{document}